\documentclass{article}

\usepackage[preprint]{neurips_2025}

\usepackage[utf8]{inputenc} %
\usepackage[T1]{fontenc}    %
\usepackage{hyperref}       %
\usepackage{url}            %
\usepackage{booktabs}       %
\usepackage{amsfonts}       %
\usepackage{nicefrac}       %
\usepackage{microtype}      %
\usepackage{xcolor}         %
\usepackage{graphicx}
\usepackage{xspace}
\usepackage{amsmath} 
\usepackage{tikz}
\usepackage{pgfplots}
\pgfplotsset{compat=newest}
\usepgfplotslibrary{groupplots}
\usepackage{enumitem}
\usepackage{multirow}
\usepackage{caption}
\usepackage{wrapfig}

\definecolor{classa}{HTML}{d53e4f}
\definecolor{classb}{HTML}{f46d43}
\definecolor{classc}{HTML}{fee08b}
\definecolor{classd}{HTML}{e6f598}
\definecolor{classe}{HTML}{abdda4}
\definecolor{classf}{HTML}{66c2a5}
\definecolor{classg}{HTML}{3288bd}

\definecolor{darkblue}{HTML}{2832C2}
\newcommand{\bluebold}[1]{\textcolor{darkblue}{\textbf{#1}}}

\newcommand{\ourdata}{\textsc{Kitten}\xspace}
\newcommand{\custompara}[1]{{\noindent\textbf{#1}}}
\newcommand{\nlp}[1]{\texttt{\footnotesize #1}}

\usepackage[capitalize]{cleveref}

\crefname{section}{Sec.}{Secs.}
\Crefname{section}{Section}{Sections}
\Crefname{table}{Table}{Tables}
\crefname{table}{Tab.}{Tabs.}

\definecolor{iccvblue}{rgb}{0.21,0.49,0.74}
\hypersetup{
	colorlinks,
	linkcolor={iccvblue},
	citecolor={iccvblue},
	urlcolor={iccvblue}
}

\newcommand{\BackboneMarker}{\raisebox{0.5pt}{\tikz\fill[classa] (0,0) rectangle (1ex,1ex);}}
\newcommand{\RetrievalMarker}{\raisebox{0.5pt}{\tikz\fill[classg] (0,0) circle (.5ex);}}

\title{\raisebox{-0.25em}{\includegraphics[width=1.25em]{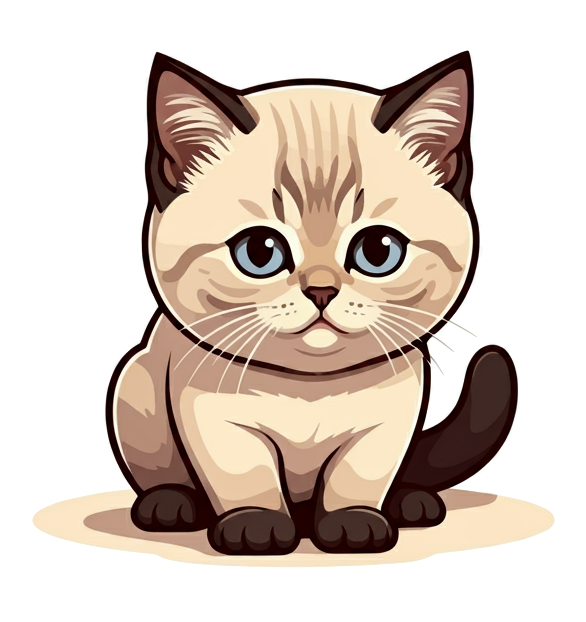}}\ourdata: A Knowledge-Intensive Evaluation of
Image Generation on Visual Entities}
\author{
\textbf{Hsin-Ping Huang}\textsuperscript{\textnormal{*1,2}}\quad
\textbf{Xinyi Wang}\textsuperscript{\textnormal{*1}} \quad
\textbf{Yonatan Bitton}\textsuperscript{\textnormal{1}} \quad
\textbf{Hagai Taitelbaum}\textsuperscript{\textnormal{1}} \\
\textbf{Gaurav Singh Tomar}\textsuperscript{1} \quad
\textbf{Ming-Wei Chang}\textsuperscript{1} \quad
\textbf{Xuhui Jia}\textsuperscript{1} \quad
\textbf{Kelvin C.K. Chan}\textsuperscript{1} \\
\textbf{Hexiang Hu}\textsuperscript{1} \quad
\textbf{Yu-Chuan Su}\textsuperscript{1} \quad
\textbf{Ming-Hsuan Yang}\textsuperscript{\textnormal{1,2}} \\
\\
\textsuperscript{1}Google DeepMind \quad
\textsuperscript{2}University of California, Merced
}

\begin{document}

\renewcommand{\thefootnote}{}
\footnotetext{\textsuperscript{*}Equal contribution}

\maketitle

\begin{abstract}
Recent advances in text-to-image generation have improved the quality of synthesized images, but evaluations mainly focus on aesthetics or alignment with text prompts. Thus, it remains unclear whether these models can accurately represent a wide variety of realistic visual entities. To bridge this gap, we propose \ourdata, a benchmark for \textbf{K}nowledge-\textbf{\textsc{i}}n\textbf{\textsc{t}}ensive image genera\textbf{\textsc{t}}ion on real-world \textbf{\textsc{en}}tities. Using \ourdata, we conduct a systematic study of the latest text-to-image models and retrieval-augmented models, focusing on their ability to generate real-world visual entities, such as landmarks and animals. Analysis using carefully designed human evaluations, automatic metrics, and MLLM evaluations show that even advanced text-to-image models fail to generate accurate visual details of entities. While retrieval-augmented models improve entity fidelity by incorporating reference images, they tend to over-rely on them and struggle to create novel configurations of the entity in creative text prompts.

\end{abstract}

\section{Introduction}
Recent advances in generative AI have revolutionized multimedia content creation. Large Language Models (LLMs) excel at knowledge-intensive tasks like question-answering and summarization. Cutting-edge image generation models, such as Imagen~\cite{saharia2022photorealistic,baldridge2024imagen,hu2024instructimagen}, DALL-E~\cite{ramesh2021dalle,ramesh2022dalle}, and Stable Diffusion~\cite{rombach2022stablediffusion}, produce photorealistic and creative images from text. However, as these models become more capable and popular, assessing their reliability is crucial. Research on LLMs shows that even the most advanced models can generate inaccuracies, potentially undermining trust and causing societal harm~\cite{muhlgay2023generating,feng2023factkb}.

Despite increasing attention to factuality in LLMs, the accuracy of image generation models remains underexplored. Existing benchmarks mainly assess alignment with general text descriptions~\cite{lin2015microsoftcococommonobjects}, compliance with image-editing instructions~\cite{ku2024imagenhub}, or adherence to spatial relationships~\cite{gokhale2022spatial}. However, they fall short in evaluating how well models generate images that faithfully reproduce the precise visual details of real-world entities, objects, and scenes grounded in trustworthy knowledge sources (see examples in \cref{fig:image_comparison}).

Recently, HEIM~\cite{lee2024holistic} introduces an evaluation suite for assessing various aspects of image generation, including their ability to generate entities such as historical figures or well-known subjects. However, real-world visual entities are far more diverse than those covered by HEIM, requiring a broader assessment. 
Moreover, HEIM primarily evaluates the alignment between generated images and entity names in text prompts. It fails to capture the fine-grained visual details essential for assessing the reproduction of visual-world knowledge since nuances of real-world entities cannot be conveyed through text alone. Thus, directly evaluating the fidelity in the generated images is essential.

\begin{figure*}
    \centering
    \includegraphics[width=0.8\textwidth]{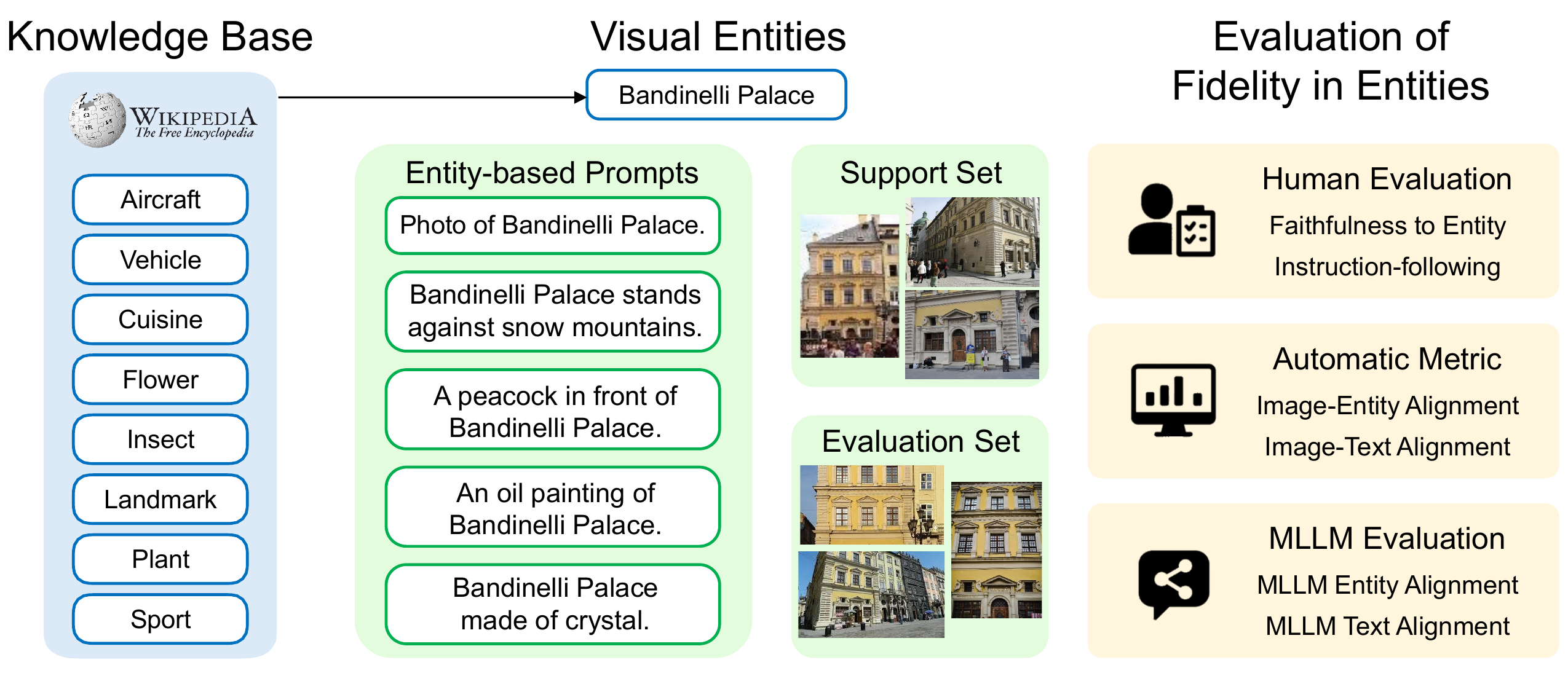}
\caption{\textbf{\ourdata~benchmark} is constructed from real-world entities across eight domains. For each selected entity, we define five evaluation tasks of image-generation prompts incorporating the entity. \ourdata~includes a support set of entity images from the knowledge source for evaluating retrieval-augmented models, and an evaluation set for assessing the fidelity of the generated entities.}
    \label{fig:benchmark}
\end{figure*}

To address the gap in evaluating image generation models' ability to reproduce visual world knowledge, we introduce \ourdata, a benchmark dataset and evaluation suite designed to assess how well models generate visually accurate representations of real-world entities grounded in trustworthy knowledge sources. Unlike prior benchmarks that focus on aesthetics, text alignment, or commonsense reasoning, \ourdata uses prompts derived from visual entities documented in Wikipedia~\cite{Hu_2023_ICCV}, a reliable knowledge base, and evaluates real-world entities across eight visual domains (see \cref{fig:benchmark}). This ensures that generated images are compared with verifiable visual information crowdsourced from the internet. Additionally, we have developed a comprehensive set of human evaluation criteria that focus on the precise visual depiction of entities, capturing subtle but essential details for visual accuracy. By directly assessing entity fidelity in the generated images against established knowledge, \ourdata aims to advance the evaluation of world knowledge in image generation models.

Using \ourdata, we conduct a comprehensive evaluation of various text-to-image models, including standard and customization models fine-tuned or utilizing in-context learning with retrieved reference images~\cite{chen2022reimagen}. Our findings show that even the most advanced models~\cite{baldridge2024imagen,blackforest2024flux} often fail to produce accurate representations, generating images missing critical details essential for visual correctness. While retrieval-augmented models improve visual fidelity by incorporating reference images during testing, they tend to over-rely on these references, limiting their ability to generate novel configurations of entities from creative prompts. These findings highlight a key challenge in current image generation models: balancing entity fidelity with creative flexibility, underscoring the need for techniques that can generate precise visual details without sacrificing the ability to respond to diverse and imaginative user inputs.

\input{figs/teaser.tikz}

\section{Related Work}

\custompara{Existing evaluation for text-to-image generation.}
Evaluating text-to-image models has long been challenging, with many efforts aimed at improving performance measurement. Fréchet Inception Distance (FID)~\citep{heusel2017gans} is a common metric for assessing perceptual quality by measuring the distribution gap between generated and real-world images. CLIP-T scores~\citep{hessel2021clipscore} evaluate text-image alignment by comparing the CLIP feature similarity between generated images and input prompts. These metrics summarize the overall image quality. Several works assess alignment between generated images and text descriptions~\citep{yarom2024you,gordon2023mismatch,hu2023tifa,wiles2024revisiting,cho2023davidsonian}, but they mainly focus on semantic consistency rather than fine-grained visual accuracy of depicted entities and specialized visual-world knowledge.

Recent works aim to evaluate models more thoroughly by decomposing the evaluation into sub-categories, such as attribute binding and numeracy, with corresponding benchmarks. For example, the SR$_\text{2D}$ dataset~\citep{gokhale2022spatial} and the VISOR metric evaluate spatial relationships in text-to-image models, assessing whether objects in the generated image adhere to specified relationships (e.g., an orange \underline{\textit{above}} a giraffe). T2I-CompBench++~\citep{huang2023t2i} contains text prompts from four categories (e.g., attribute binding) along with associated metrics. HRS-Bench~\citep{bakr2023hrs} evaluates model performance across five major groups (i.e., bias, fairness, generalization, accuracy, and robustness). TIFA v1.0~\citep{hu2023tifa} is a benchmark across 12 categories, paired with an automatic evaluation metric that measures image faithfulness via visual question answering. 
GenAI-Bench~\citep{li2024genaibench} and ConceptMix~\citep{wu2024conceptmix} focus on the evaluation of compositional text prompts, e.g., objects with specific colors, shapes, or spatial relationships.
Additionally, ImagenHub~\citep{ku2024imagenhub} evaluates models across different tasks by measuring semantic consistency and perceptual quality.

\custompara{Fidelity of entities in text-to-image generation.} 
While text-to-image models enable the generation of creative images from text descriptions, challenges arise when visual-world knowledge is needed, i.e., generating accurate visual details of entities. Existing works have identified this issue and proposed solutions to mitigate hallucination~\citep{lim2024addressing}. However, no clear methodology exists to systematically assess these models' limitations, which is crucial for improvement.
In this work, we propose \ourdata, a benchmark addressing a novel problem of evaluating image generation models' ability to generate fine-grained details of specific visual entities. Using~\ourdata, we systematically assess the latest text-to-image and retrieval-augmented models with carefully designed human evaluations, automatic metrics, and MLLM evaluations.

\section{\ourdata Benchmark}

We introduce the \ourdata~benchmark to evaluate the reliability of text-to-image models in generating knowledge-intensive concepts.

\subsection{Design Desiderata of \ourdata}

The key to creating the benchmark is constructing a set of image-generation prompts that require grounding in visual-world knowledge. Two specific properties differentiate our benchmark from prior evaluation frameworks of image generation. First, while existing benchmarks aim to test the common-sense knowledge of image generation models such as spatial or physical relationships~\citep{gokhale2022spatial,huang2023t2i}, we would like to stress-test the image generation models by focusing on generating entities from specific domains. Therefore, we create the benchmark using image concepts from Wikipedia, a rich knowledge-intensive data source, which contains several domain-specific entities and their corresponding images. Second, while most existing benchmarks focus on evaluating the instruction-following capability of the models, we would like to understand how well these models are at faithfully representing real-world concepts grounded in visual knowledge sources. Therefore, we design a specific set of evaluations targeted at capturing the visual fidelity of generated entities.     
Guided by the above principles, next, we clarify the details of the \ourdata~benchmark.

\begin{figure*}
    \centering
    \includegraphics[width=0.8\textwidth]{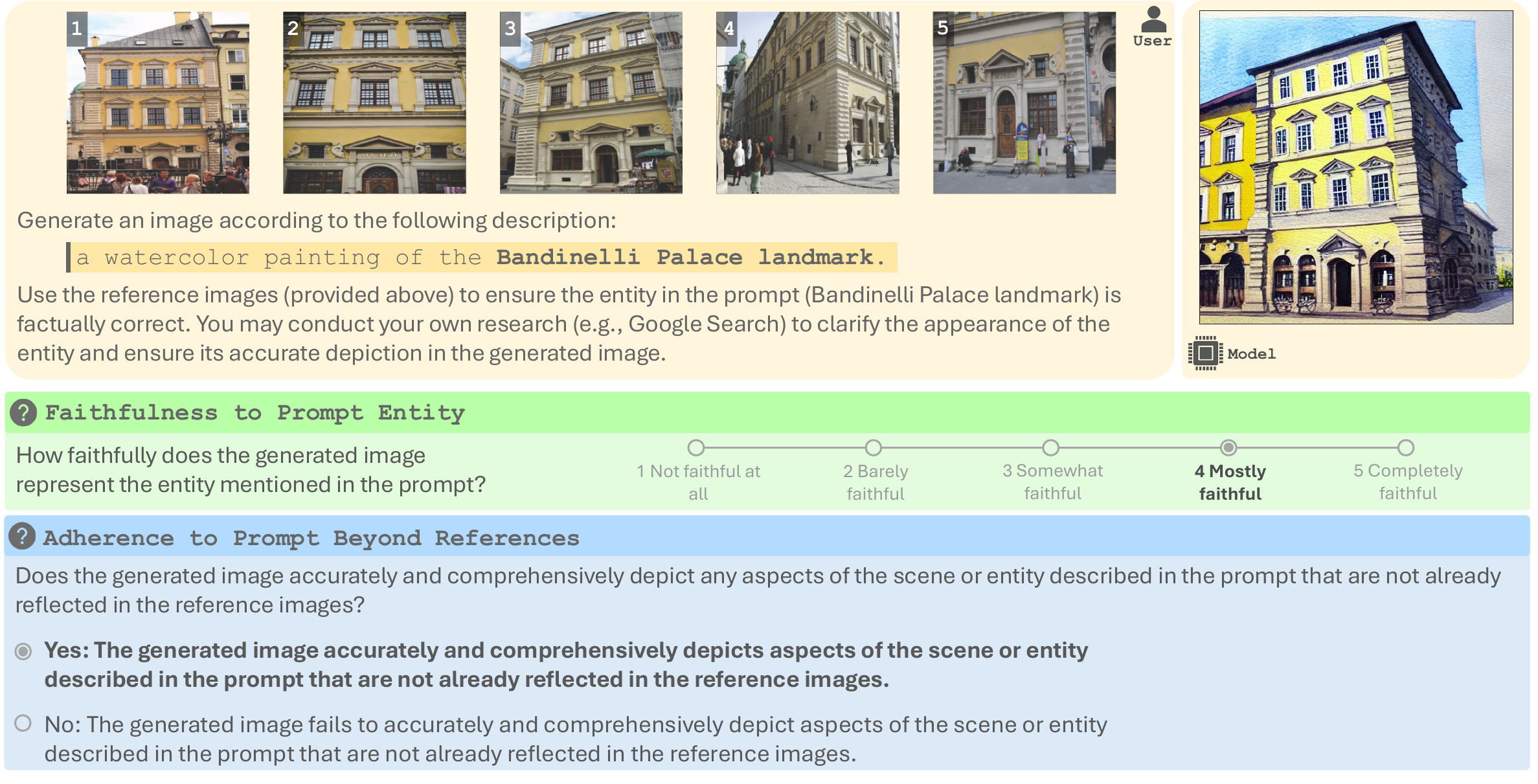}
    \caption{\textbf{Annotation interface.} Raters are asked to: (1) rate the image's faithfulness to the prompt entity on a 1–5 scale, and (2) indicate whether the image follows the prompt with a yes or no response. We calculate the percentage of responses marked as ``yes.''}
    \label{fig:annotation_interface}
\end{figure*}

\subsection{Creating Entity-based Prompts}
\label{sec:data_details}

\cref{fig:benchmark} shows the benchmark creation process.  
To generate a diverse set of prompts focused on faithfulness to knowledge-grounded concepts, we first select entity domains from the OVEN-Wiki dataset~\citep{Hu_2023_ICCV}, the most comprehensive open-domain image recognition dataset.  
We select 8 domains encompassing 322 entities, covering human-made objects, natural species, and human activities. This selection offers broader coverage than existing benchmarks~\cite{lee2024holistic}.  
For each entity, we collect an evaluation set of entity images from Wikipedia for human evaluation, and a support set of images to evaluate retrieval-augmented models that leverage external knowledge sources for image generation~(see evaluated models in \cref{sec:eval_models}).  

After selecting the entities, we design five evaluation tasks of image-generation prompts. 
\begin{itemize}[leftmargin=*, itemsep=0pt, topsep=0pt]
\item {Basic prompt (4.58\%):} \nlp{Photo of Bandinelli Palace.}
\item {Entity in a specified location (30.57\%):} \nlp{Bandinelli Palace stands against snow mountains.}
\item {Composition with other objects (22.78\%):} \nlp{A peacock in front of Bandinelli Palace.}
\item {Entity in specific styles (21.20\%):} \nlp{An oil painting of Bandinelli Palace.}
\item {Entity made of specific materials (20.87\%):} \nlp{Bandinelli Palace made of crystal.}
\end{itemize}
The evaluation tasks are designed to cover key scenarios in customized image generation~\cite{ruiz2023dreamboothfinetuningtexttoimage,kumari2022customdiffusion}, ensuring relevance to both researchers and real-world applications.  
For each task, ChatGPT~\cite{gpt} is instructed to propose prompts for different entity domains, which are then tailored to evaluate knowledge-entity generation by incorporating entity names directly into the prompts.
This process resulted in a set of 6,440 prompts to assess the models' ability to handle diverse and imaginative user inputs. The prompts range from 4 to 24 words, with an average length of 9.91 and a standard deviation of 2.86.

\subsection{Human Evaluation}
Since no established metrics exist for evaluating the generation of visual entities, human evaluation plays a critical role in reliably assessing model performance.
We design human evaluations focusing on the visual fidelity of the target entity in the generated image, decomposing the evaluation into two aspects: 1) faithfulness to the prompt entity, and 2) adherence to prompt instructions beyond the entity. This design allows raters to focus on distinct criteria, enabling more informative comparisons of models, as our results often show trade-offs between these aspects. Raters are shown reference images of the prompt entity and encouraged to verify the entity's faithfulness through their own research. The final evaluation score is the average of five raters per image to ensure robust assessments. The human annotation interface is shown in~\cref{fig:annotation_interface}. %

\begin{itemize}[leftmargin=*, itemsep=0pt, topsep=0pt]
\item \textbf{Faithfulness to Entity.} We use a scale from 1 to 5, where 5 means completely faithful to the prompt entity, and 1 means the generated image has no similarity to the prompt entity.
\item \textbf{Instruction-following.} We use a yes/no question to evaluate whether the generated image adheres to the prompt instructions beyond the entity. We calculate the percentage of answers ``yes.''
\end{itemize}

\subsection{Automatic Metrics}
We gather the results of popular automatic metrics for image generation models, which primarily measure the similarity between the generated images and the references or prompts. While these metrics are not specialized for capturing the visual fidelity of the generated entity, we include their results to provide a comprehensive analysis of their alignment with human evaluation. %
\begin{itemize}[leftmargin=*, itemsep=0pt, topsep=0pt]
    \item \textbf{Image-Text Alignment.} We measure the cosine similarity between the generated image and the text prompt in CLIP's feature space~\citep{radford2021clip}, i.e., \textit{CLIP-T Score}~\citep{hessel2021clipscore}.
    \item \textbf{Image-Entity Alignment.} We measure the average pairwise cosine similarity between the generated image and reference images of the target entity in the evaluation set using DINO's feature space~\citep{dinov2}.
    This serves as a proxy for how closely fine-grained details match.
\end{itemize}

\subsection{MLLM Evaluation}
Multimodal large language models (MLLMs) have recently demonstrated impressive progress in multimodal understanding. We investigate the use of MLLMs as automatic evaluators to reduce human effort and address the limitations of traditional metrics in assessing visual fidelity. Specifically, we prompt GPT-4o-mini~\cite{gpt} using the same criteria as our human evaluation. %
\begin{itemize}[leftmargin=*, itemsep=0pt, topsep=0pt]
\item \textbf{MLLM Text Alignment.} Given the text prompt, we use an MLLM to evaluate whether the generated image follows the prompt instructions on a 1–5 scale.
\item \textbf{MLLM Entity Alignment.} Given reference images of the target entity, we use an MLLM to assess how well the generated image resembles the target entity on a 1–5 scale.
\end{itemize}

\section{Evaluated Models}
\label{sec:eval_models}
We present a comprehensive analysis using \ourdata to understand the visual-world knowledge in current state-of-the-art models.

\subsection{Text-to-Image Backbone Models}
First, we examine general text-to-image backbone models that directly generate images solely based on text prompts without using additional tools or reference images.

\begin{itemize}[leftmargin=*, itemsep=0pt, topsep=0pt]
    \item Stable Diffusion~\citep{rombach2022stablediffusion} maps images to a latent space where a diffusion model is trained.
    \item Imagen~\citep{saharia2022photorealistic} uses a T5 encoder and cascaded diffusion models for high-resolution image generation.
    \item Imagen-3~\citep{baldridge2024imagen} is a successor to Imagen, notable for its ability to handle long prompts. %
    \item Flux\footnote{We use Flux.1-dev.}~\citep{blackforest2024flux} is a successor to Stable Diffusion, integrating parallel diffusion transformer blocks. %
\end{itemize}

\subsection{Retrieval-augmented Text-to-Image Models}

The retrieval-augmented method is a family of image generation approaches that use support images (e.g., retrieved by a search engine) to enhance the model through fine-tuning or in-context learning, improving the fidelity of entities in the generated images. Our goal is to evaluate these models and determine whether incorporating such \textit{support} images enhances the fine-grained visual fidelity of the entity in generation. 
Specifically, we provide some ground-truth reference entity images (held out from the entity images used for evaluation) as the support images to the above methods and then generate new images from them following the evaluation text prompts.
We study the following models:

\begin{itemize}[leftmargin=*, itemsep=0pt, topsep=0pt]
    \item DreamBooth~\citep{ruiz2023dreamboothfinetuningtexttoimage} \textbf{fine-tunes} the Stable Diffusion model to learn a special token encoding the target entity. It then generates the entity in new contexts using prompts that include this token.
    \item Custom-Diff~\citep{kumari2022customdiffusion}, similar to DreamBooth, \textbf{fine-tunes} partial weights of the backbone model. %
    \item Instruct-Imagen~\citep{hu2024instructimagen} generates the target entity through \textbf{in-context learning} by encoding reference images into a multi-modal instruction: {\small{\texttt{Generate an image of <entity\_name>, referring to the images <ref\_image\_1>, ..., <ref\_image\_K>, and follow the caption: <prompt>}}}.
\end{itemize}

\input{figs/human_and_autometric.tikz}

\section{Evaluation Results}

\subsection{Human Evaluation}

\custompara{Retrieval-augmented models enhance faithfulness but weaken instruction-following.}
\cref{fig:tradeoff} (top) shows that retrieval-augmented models --- Custom-Diff, DreamBooth, and Instruct-Imagen --- generally produce images more faithful to the entities than their base models, SD and Imagen. This is because these models incorporate reference images during testing, enabling them to generate visual concepts not well-represented in the base models' parameters.
However, retrieval-augmented models tend to have reduced instruction-following capabilities compared to their backbone models, as they often over-rely on reference images and struggle to create novel configurations of the entity as requested in creative text prompts.
Although this trend is consistent across methods, the extent of the impact varies. Notably, Instruct-Imagen shows a significant increase in faithfulness score (2.81 $\rightarrow$ 4.22) but also a substantial drop in instruction-following score (72.2 $\rightarrow$ 46.5).

\custompara{Enhancing backbone models improves both faithfulness and instruction-following.}
Our results show that improvements to base models alone can enhance both instruction-following and faithfulness. For example, Flux outperforms its predecessor SD by 0.23, and Imagen-3 surpasses its predecessor Imagen by 0.35 in faithfulness. 
However, these faithfulness improvements are still minor compared to those achieved by retrieval-augmented methods, such as DreamBooth, which shows a 0.57 improvement over SD.

On the other hand, Imagen-3 achieves the highest instruction-following score (83.6) and a high faithfulness score (3.17), outperforming the retrieval-augmented models DreamBooth (3.08) and Custom-Diff (2.90).
This demonstrates that Imagen-3, as a strong backbone model, can generate specialized entities solely from text prompts. 
However, a notable gap remains in entity fidelity compared to the highest score achieved by Instruct-Imagen (4.22).
These findings show that enhancing the backbone model can improve both instruction-following and entity fidelity, while it is essential to incorporate advanced retrieval-augmented techniques to achieve higher levels of faithfulness.

\custompara{Balancing faithfulness and instruction-following is achievable.}
The retrieval-augmented model DreamBooth improves entity faithfulness compared to its baseline, SD (2.51 $\rightarrow$ 3.08), without compromising SD’s instruction-following score (72.2 $\rightarrow$ 73.8).
This demonstrates that a well-designed retrieval-augmented method can enhance entity fidelity without sacrificing creativity.
These findings also suggest future research directions, emphasizing that combining a strong backbone with an effective retrieval-augmented approach can achieve a balance between faithfulness and instruction-following.

\subsection{Automatic Metrics and MLLM Evaluation}

\custompara{Retrieval-augmented models improve entity alignment but reduce text alignment.}
\cref{fig:tradeoff} (bottom) shows that retrieval models increase the entity alignment score while decreasing the text alignment score compared to their base models in both automatic metrics and MLLM evaluation.
These observations align with the human evaluation, where retrieval-augmented models show improved entity faithfulness but reduced instruction-following.
In addition, this trend is consistent with observations in recent works~\cite {materzynska2023customizing}, which indicate that models incorporating additional inputs, such as reference images, tend to have lower text alignment scores than base models due to a trade-off between aligning with the text and with the images.

\custompara{Alignment between automatic metrics and human evaluation.}
While the overall observations from the automatic metrics align with the human evaluation, there are notable discrepancies.
We observe that improving base models does not necessarily lead to gains in the automatic metrics. For example, Flux (0.329) performs worse than its predecessor SD (0.338) in the image-text metric, and Imagen-3 (0.389) shows only a marginal improvement over Imagen (0.386) in the image-entity score. These findings suggest that automatic metrics have a limited ability to capture meaningful variations between models of similar quality. 
In addition, although DreamBooth achieves a higher instruction-following score compared to its base model SD, it has a lower image-text alignment score.
We hypothesize that the image-text score may not accurately assess the alignment between the generated image and rare entities. 
For example, with the prompt ``The Teufelsmauer landmark shimmers in the sunlight,'' it is unclear whether the image-text similarity for ``Teufelsmauer'' is evaluated correctly. This highlights that the traditional metrics~\cite {hessel2021clipscore,lee2024holistic} might fail to measure true alignment between the unique entity and the generated image. In addition, Imagen-3 ranks higher in faithfulness in the human evaluation, yet DreamBooth outperforms Imagen-3 in image-entity scores, showing a misalignment between human perception and the learned semantic features~\cite{dinov2}. %

\custompara{Alignment between MLLM and human evaluation.} 
The MLLM score generally shows a high correlation with human evaluation, though slight discrepancies exist. Specifically, DreamBooth (3.37) falls behind SD (3.46) and Imagen (3.61) in the MLLM text alignment score. However, in human evaluation, DreamBooth (73.8) outperforms both Imagen and SD (72.2). This suggests that while the MLLM can capture overall trends and identify models with clearly stronger or weaker performance, it may struggle to distinguish between models with similar capabilities. On the other hand, Flux (2.04) receives a lower MLLM entity alignment score than SD (2.39), despite outperforming SD in human evaluation (2.74 vs. 2.51), which suggests that the MLLM may have different preferences or biases compared to human raters.

\subsection{Ablation Study}

\begin{wraptable}{r}{0.23\textwidth}
    \vspace{-4mm}
        \centering
        \scriptsize
        \caption{Correlation with human evaluation.}
        \tabcolsep 1pt
\begin{tabular}{l|cc}
    \toprule
    Metric & Pearson & Spearman \\
    \midrule
    CLIP-T & 0.337 & 0.384 \\
    CLIP-I & 0.239 & 0.340 \\
    DINO   & 0.510 & 0.504 \\
    MLLM-T & 0.618 & 0.589 \\ 
    MLLM-I & 0.703 & 0.695 \\
    \bottomrule
\end{tabular}
        \label{tab:correlation}
\end{wraptable}

\textbf{Correlation of automatic and MLLM metrics with human evaluation.}
While manual evaluation ensures high accuracy, automated methods offer a cost-effective alternative, albeit with slightly lower alignment to human perception. To quantify the consistency between automatic and MLLM metrics with human evaluation, we computed Pearson and Spearman correlations in \cref{tab:correlation}. CLIP-T and CLIP-I show moderate alignment with user evaluations. Compared to CLIP-I metrics, DINO demonstrates stronger alignment with user evaluations of faithfulness.
Surprisingly, MLLM-based evaluation shows a strong correlation with human judgments. This highlights the potential of MLLM metrics to capture subtle visual details when assessing the faithfulness of generated knowledge entities. It suggests that our study can be reliably scaled up using MLLM as an automatic evaluator.

\begin{wraptable}{r}{0.23\textwidth}
    \centering
    \vspace{-6mm}
    \scriptsize
    \tabcolsep 5pt
        \centering
        \tabcolsep 1.5pt
        \caption{Selection of Image-Entity metrics.}
        \begin{tabular}{l|cc}
        \toprule
         Model & CLIP-I & DINO \\
        \midrule
         SD & 0.646 & 0.350\\
         Imagen & 0.646 & 0.386\\
         Flux & 0.639 & 0.380\\
         Imagen-3 & 0.650 & 0.389\\
         \midrule
         Custom-Diff & 0.643 & 0.388\\
         DreamBooth & 0.674 & 0.412\\
         Instruct-Imagen & 0.751 & 0.582\\
        \bottomrule
        \end{tabular}
        \label{tab:ablation}
\end{wraptable}

\textbf{Selection of image-entity alignment metrics.}  
In \cref{tab:ablation}, two popular visual features are tested for calculating cosine similarity scores between reference and generated images as the image-entity alignment metric: \textit{CLIP-I}~\citep{radford2021clip} and \textit{DINO}~\citep{dinov2}. 
We find that DINO scores provide a clearer separation between models compared to CLIP-I scores, making it a more discriminative metric at capturing subtle differences in faithfulness. For example, the difference between Custom-Diff and Instruct-Imagen is much larger when using DINO (0.19) compared to CLIP-I (0.11). This may be due to DINO's focus on primary entities, allowing for a more accurate estimation of similarity between the generated entities and reference images.

\begin{figure}[t]
    \centering
    \begin{tikzpicture}[scale=1]
        \begin{groupplot}[
            group style={group size=1 by 2, vertical sep=5mm},
        ]

\nextgroupplot[
    width=\textwidth,
    height=3.25cm,
    ybar,
    ymin=10,
    ymax=95,
            bar width=1mm,
    symbolic x coords={Aircraft, Vehicle, Cuisine, Flower, Insect, Landmark, Plant, Sport},
    xtick=data,  %
    font=\scriptsize,
    ylabel={\texttt{Image-Entity(\%)}}, %
    xtick pos=left, %
    ytick pos=left, %
            scaled y ticks=false,        %
            nodes near coords,
            nodes near coords style={font=\tiny, rotate=90, anchor=west},
            yticklabel=\pgfkeys{/pgf/number format/.cd,fixed,precision=1,zerofill}\pgfmathprintnumber{\tick},
            axis lines=left,
            enlarge x limits=true,
            legend style={font=\scriptsize, at={(0.5, 1.35)}, anchor=north, draw=none, legend columns=-1},
        ]

            \addplot[fill=classa, area legend] coordinates {(Aircraft, 44.9) (Vehicle, 54.0) (Cuisine, 36.7) (Flower, 38.6) (Insect, 18.1) (Landmark, 30.6) (Plant, 18.9) (Sport, 37.9)}; %
            \addplot[fill=classb, area legend] coordinates {(Aircraft, 61.0) (Vehicle, 55.9) (Cuisine, 35.9) (Flower, 46.9) (Insect, 19.8) (Landmark, 35.3) (Plant, 20.7) (Sport, 33.1)}; %
            \addplot[fill=classc, area legend] coordinates {(Aircraft, 52.4) (Vehicle, 57.6) (Cuisine, 38.2) (Flower, 47.7) (Insect, 15.3) (Landmark, 34.3) (Plant, 21.8) (Sport, 36.9)}; %
            \addplot[fill=classd, area legend] coordinates {(Aircraft, 58.8) (Vehicle, 59.0) (Cuisine, 35.1) (Flower, 43.5) (Insect, 19.4) (Landmark, 36.6) (Plant, 21.5) (Sport, 36.9)}; %
            \addplot[fill=classe, area legend] coordinates {(Aircraft, 55.5) (Vehicle, 57.6) (Cuisine, 28.5) (Flower, 49.0) (Insect, 26.3) (Landmark, 38.2) (Plant, 26.7) (Sport, 28.8)}; %
            \addplot[fill=classf, area legend] coordinates {(Aircraft, 58.0) (Vehicle, 56.7) (Cuisine, 40.6) (Flower, 43.5) (Insect, 28.0) (Landmark, 40.7) (Plant, 25.0) (Sport, 37.1)}; %
            \addplot[fill=classg, area legend] coordinates {(Aircraft, 75.8) (Vehicle, 71.1) (Cuisine, 53.9) (Flower, 64.6) (Insect, 52.6) (Landmark, 54.5) (Plant, 45.0) (Sport, 48.2)}; %
                        \legend{SD, Imagen, Flux, Imagen-3, Custom-Diff, DreamBooth, Instruct-Imagen}
        \end{groupplot}
    \end{tikzpicture}

    \begin{tikzpicture}[scale = 1]
        \begin{groupplot}[
            group style={group size=1 by 2, vertical sep=5mm},
        ]

\nextgroupplot[
    width=0.75\textwidth,
    height=3.25cm,
    ybar,
    ymin=25.0,
    ymax=80.0,
    bar width=1mm,
    symbolic x coords={Basic, Location, Composition, Style, Material},
    xtick=data,
    font=\scriptsize,
    ylabel={\texttt{Image-Entity(\%)}},
    xtick pos=left,
    ytick pos=left,
    nodes near coords,
    nodes near coords style={font=\tiny, rotate=90, anchor=west},
    scaled y ticks=false,
    yticklabel style={/pgf/number format/.cd, fixed, precision=1, zerofill},
    axis lines=left,
    enlarge x limits=true
]
\addplot[fill=classa] coordinates {(Basic, 46.7) (Location, 37.9) (Composition, 29.5) (Style, 31.1) (Material, 34.9)};
\addplot[fill=classb] coordinates {(Basic, 47.2) (Location, 40.9) (Composition, 33.3) (Style, 37.6) (Material, 38.5)};
\addplot[fill=classc] coordinates {(Basic, 46.4) (Location, 39.4) (Composition, 31.6) (Style, 38.3) (Material, 36.8)};
\addplot[fill=classd] coordinates {(Basic, 48.3) (Location, 41.2) (Composition, 31.7) (Style, 40.3) (Material, 39.1)};
\addplot[fill=classe] coordinates {(Basic, 57.8) (Location, 41.1) (Composition, 34.8) (Style, 35.9) (Material, 40.0)};
\addplot[fill=classf] coordinates {(Basic, 59.7) (Location, 44.0) (Composition, 37.6) (Style, 37.1) (Material, 41.7)};
\addplot[fill=classg] coordinates {(Basic, 62.6) (Location, 57.3) (Composition, 56.6) (Style, 59.3) (Material, 60.6)};
        \end{groupplot}
    \end{tikzpicture}

\caption{\textbf{Performance analysis across domains and tasks.} 
\textbf{(Top)} Retrieval models achieve higher Image-Entity scores in the insect, landmark, and plant domains but perform worse in the cuisine and sport domains, possibly due to the varying occurrence of these entities in common image datasets. %
\textbf{(Bottom)} \textit{Location} scoring highest and \textit{Composition} the least. %
}
\label{fig:domain_and_task}
\end{figure}

\subsection{Analysis of Performance Variations}
\custompara{Performance across entity domains.}
\cref{fig:domain_and_task} (top) shows that the performance of each method is domain-dependent.
Retrieval-augmented models generally achieve higher image-entity alignment scores than backbone models in the insect, landmark, and plant domains.
Since these domains contain less frequent terms in common image datasets, these visual concepts are therefore underrepresented in the backbone model’s parameters. The retrieval-augmented models improve performance by incorporating reference images during inference. 
Additionally, the insect and landmark domains have lower average image-entity scores, likely due to the inherent challenges of generating fine-grained details of insects and the many specifications and features of landmarks.

On the other hand, the retrieval-augmented method, Custom-Diff, performs worse than its base model, SD, in the cuisine and sport domains.
These domains contain common terms, such as snowboarding and guacamole, which the SD model has well memorized. The Custom-Diff model's performance degrades, potentially due to fine-tuning on a smaller reference set.
This variability suggests that the effectiveness of a retrieval-augmented method may be influenced by the nature of the domain-specific content, and the optimal choice of retrieval-augmented method remains an open question.

\custompara{Performance across evaluation tasks.}
\cref{fig:domain_and_task} (bottom) shows that image-entity scores across evaluation tasks generally align with the overall ranking. \textit{Location} scores highest (0.431), followed by \textit{Material} (0.417), \textit{Style} (0.399), and \textit{Composition} (0.364), highlighting the challenge of maintaining entity fidelity when prompts involve complex compositions.

We observe that \textit{Style} prompts show a distinct score distribution. Retrieval-augmented methods, DreamBooth and Custom-Diff, along with their base model SD, receive lower image-entity scores (0.371, 0.359, and 0.311), indicating that models based on SD struggle to generate faithful entities when changing their styles. However, SD achieves the highest image-text score (0.346), followed by DreamBooth (0.341) and Custom-Diff (0.335), suggesting these models are strong in generating accurate styles but may sacrifice entity fidelity.

\begin{figure*}[t]
    \centering
    \scriptsize
    \setlength{\tabcolsep}{2pt} %
    \begin{tabular}{c@{\;\;}c@{\;\;}c@{\;\;}c}
        \toprule %
        \bluebold{Real Photo} & \bluebold{Custom-Diff} / 0.38 / 0.35 & \bluebold{DreamBooth} / 0.56 / 0.36 & \bluebold{Instruct-Imagen} / 0.53 / 0.40 \\
        \includegraphics[width=0.20\textwidth, height=2.5cm]{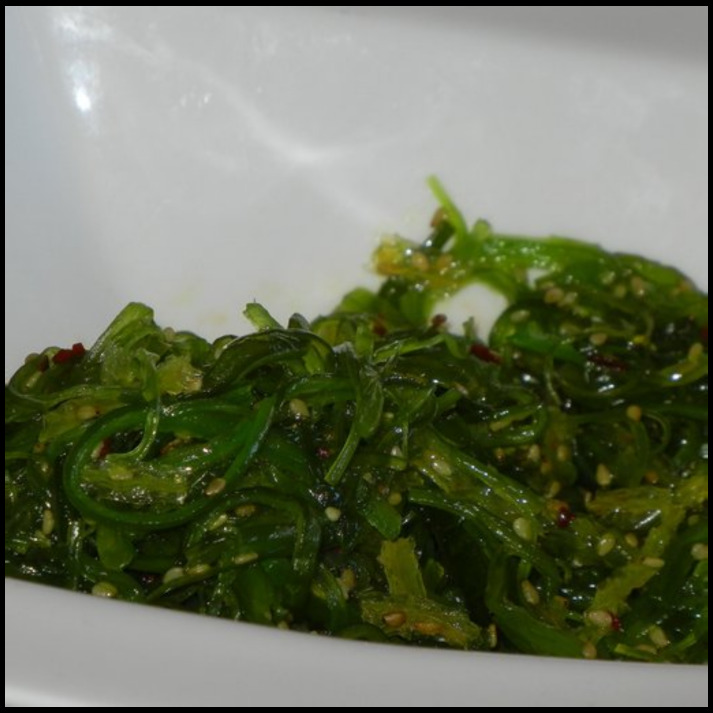} & 
        \includegraphics[width=0.20\textwidth, height=2.5cm]{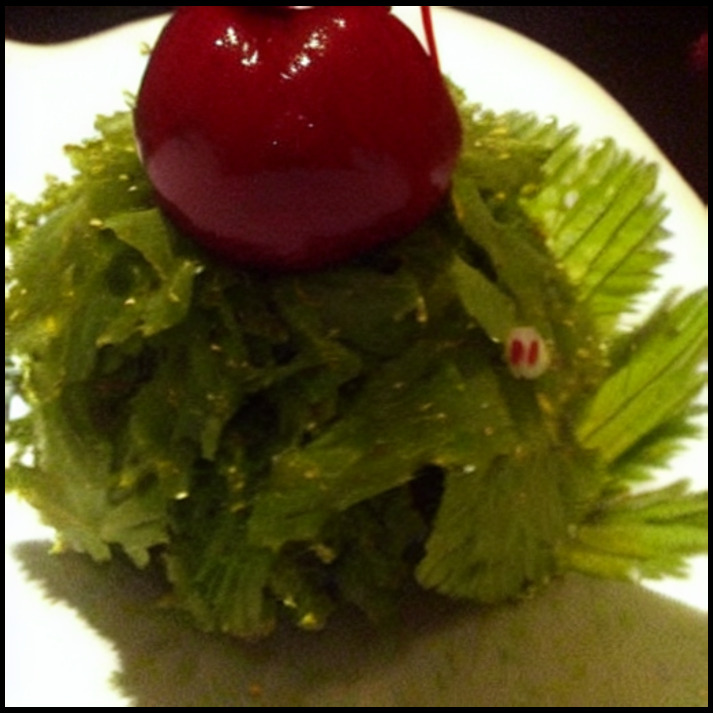} & 
        \includegraphics[width=0.20\textwidth, height=2.5cm]{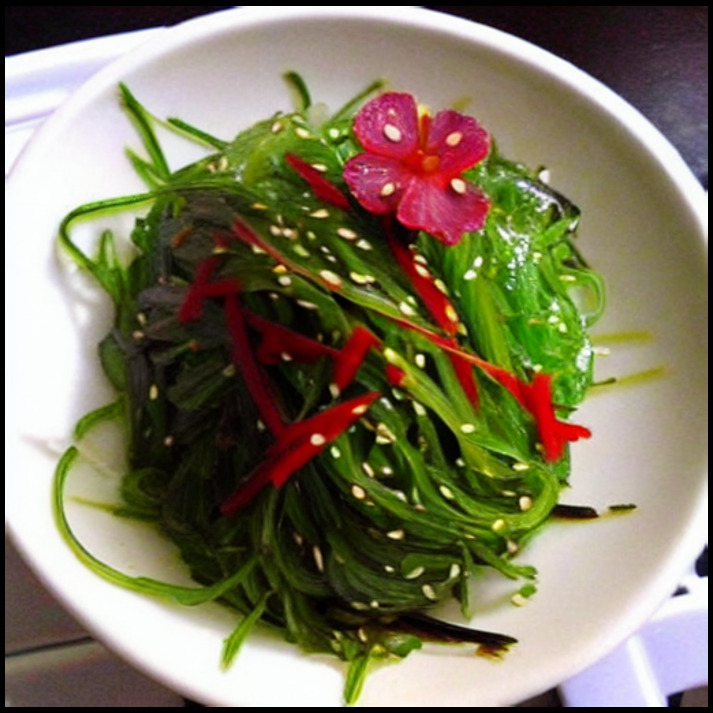} & 
        \includegraphics[width=0.20\textwidth, height=2.5cm]{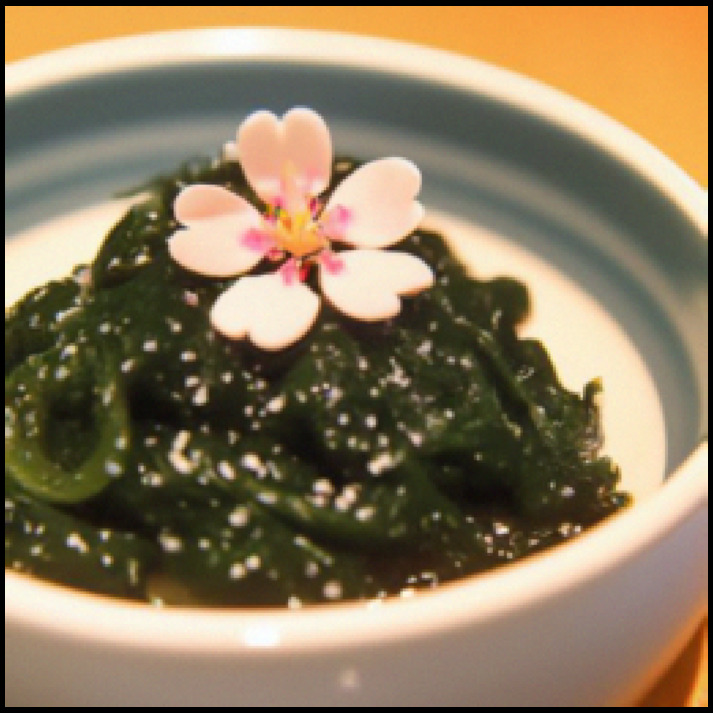} \\
        \bluebold{SD} / 0.49 / 0.39 & \bluebold{Imagen} / 0.29 / 0.37 & \bluebold{Flux} / 0.41 / 0.34 & \bluebold{Imagen-3} / 0.38 / 0.36 \\
        \includegraphics[width=0.20\textwidth, height=2.5cm]{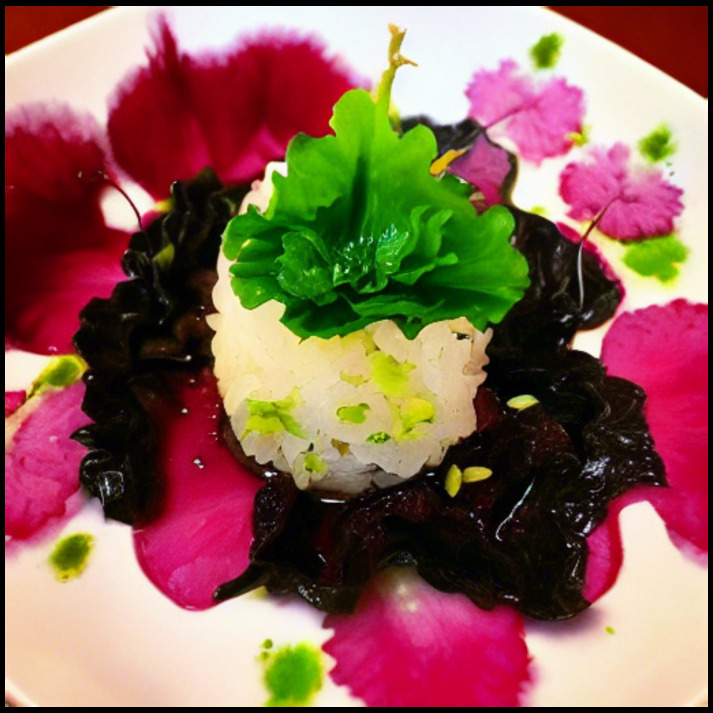} & 
        \includegraphics[width=0.20\textwidth, height=2.5cm]{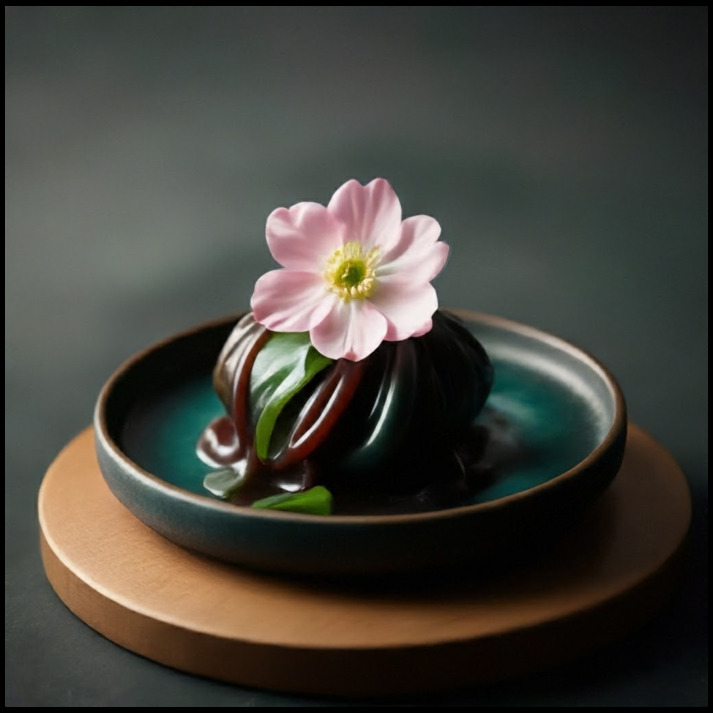} & 
        \includegraphics[width=0.20\textwidth, height=2.5cm]{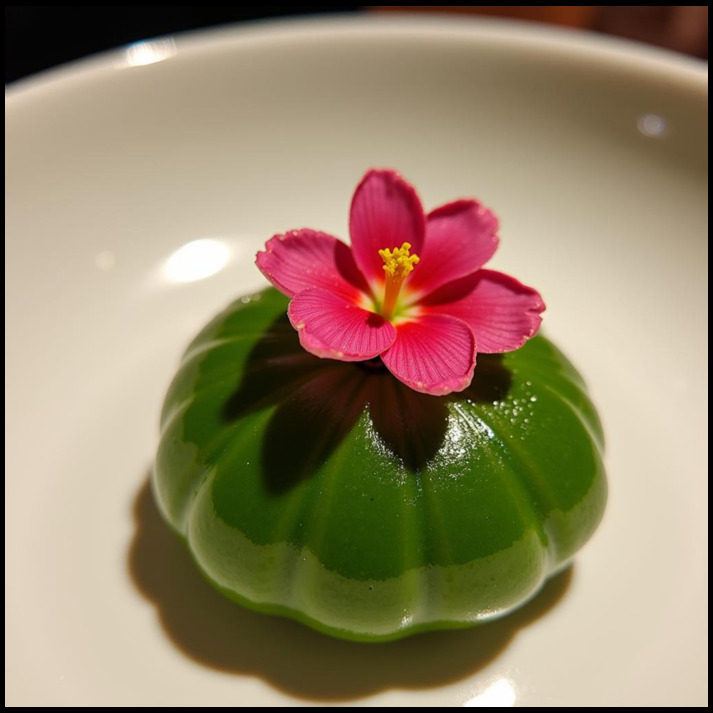} & 
        \includegraphics[width=0.20\textwidth, height=2.5cm]{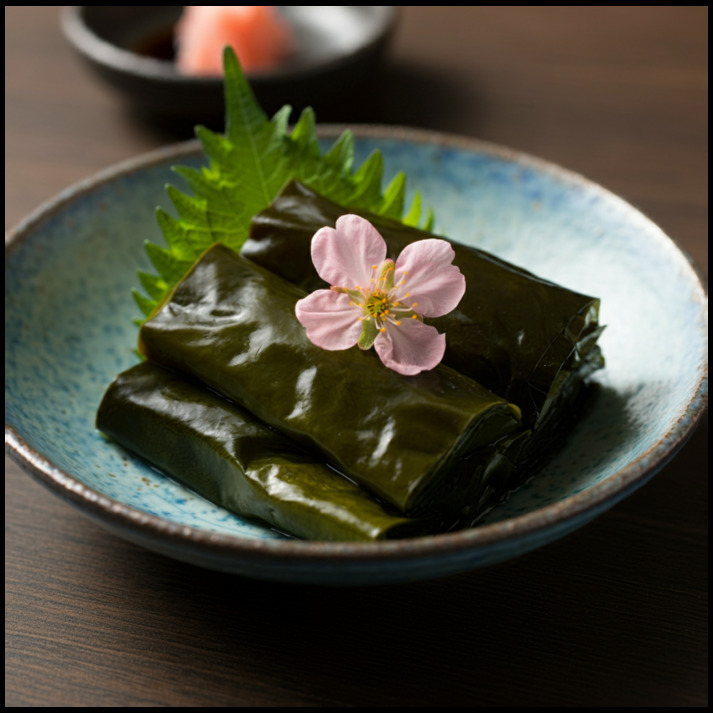} \\
        \multicolumn{4}{c}{\textit{A Wakame dish with a cherry flower on top of it.}} \\
        \midrule
        \bluebold{Real Photo} & \bluebold{Custom-Diff} / 0.61 / 0.39 & \bluebold{DreamBooth} / 0.38 / 0.40 & \bluebold{Instruct-Imagen} / 0.59 / 0.41 \\
        \includegraphics[width=0.20\textwidth, height=2.5cm]{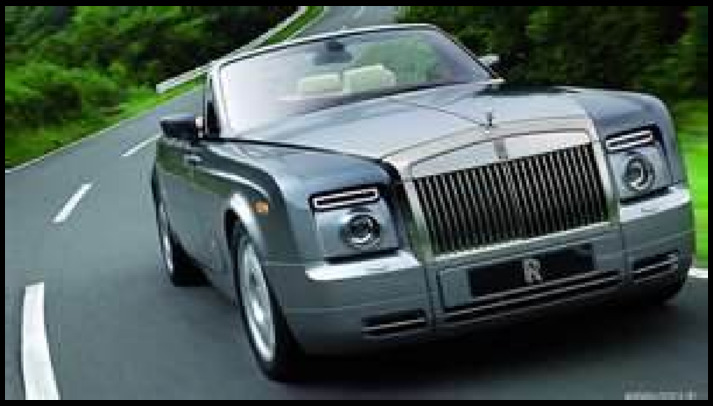} & 
        \includegraphics[width=0.20\textwidth, height=2.5cm]{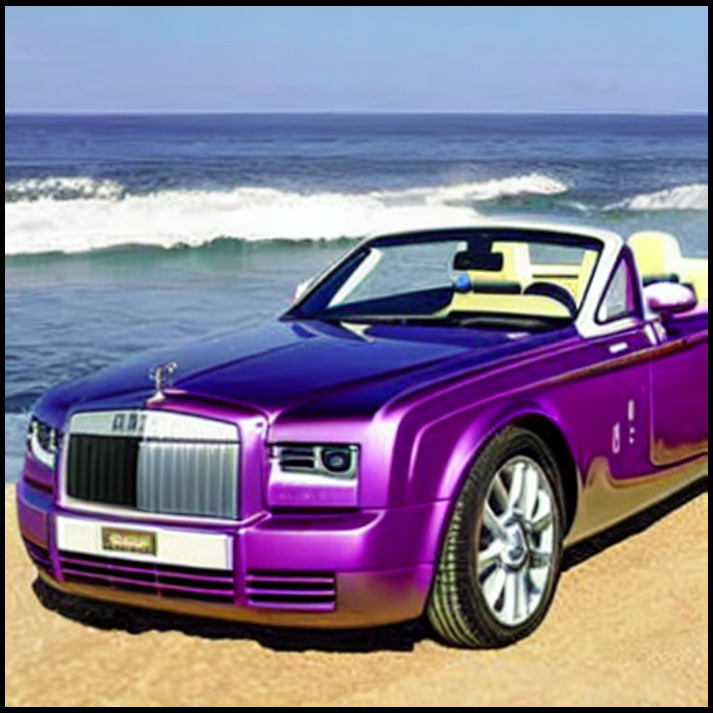} & 
        \includegraphics[width=0.20\textwidth, height=2.5cm]{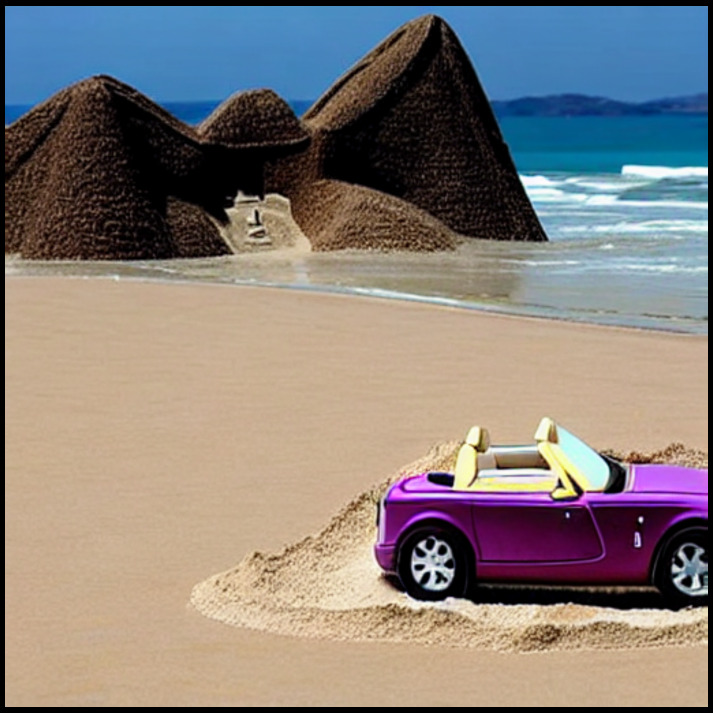} & 
        \includegraphics[width=0.20\textwidth, height=2.5cm]{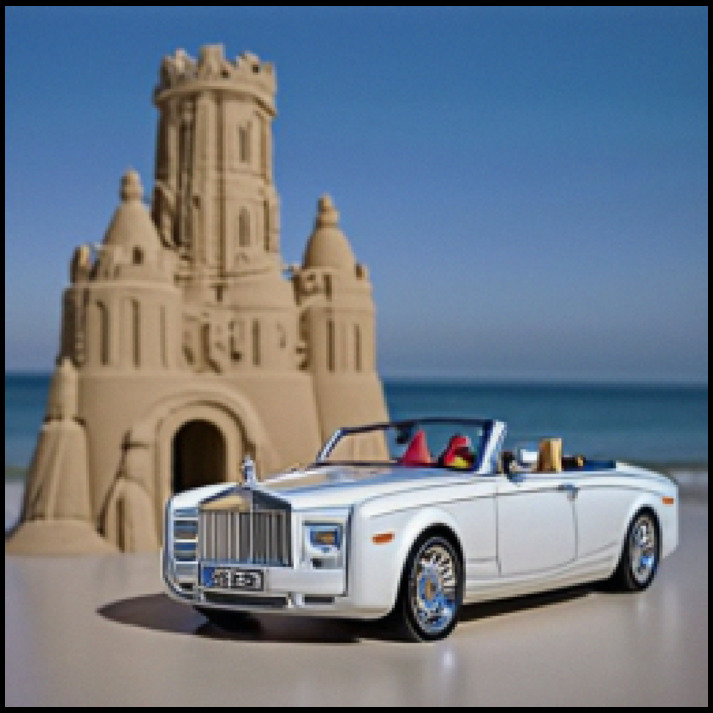} \\
        \bluebold{SD} / 0.53 / 0.41 & \bluebold{Imagen} / 0.43 / 0.42 & \bluebold{Flux} / 0.35 / 0.43 & \bluebold{Imagen-3} / 0.47 / 0.45 \\
        \includegraphics[width=0.20\textwidth, height=2.5cm]{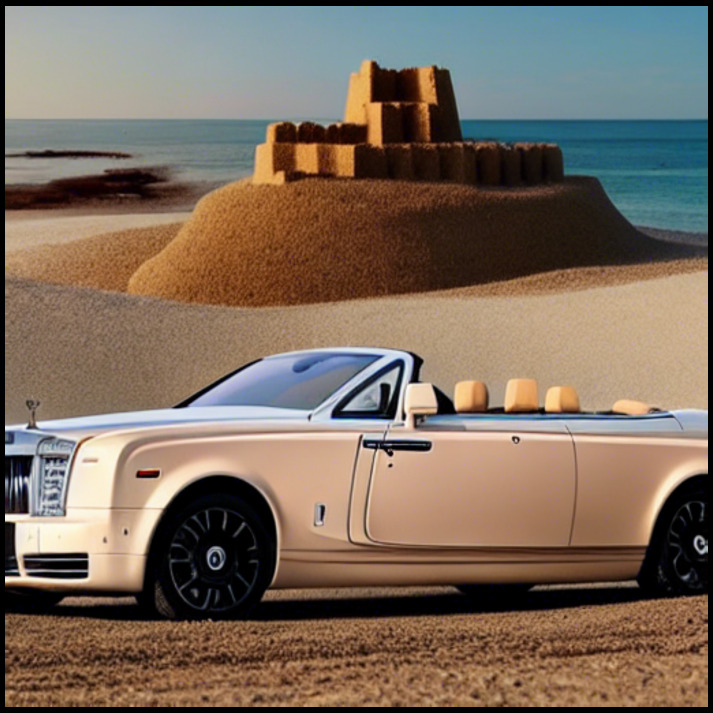} & 
        \includegraphics[width=0.20\textwidth, height=2.5cm]{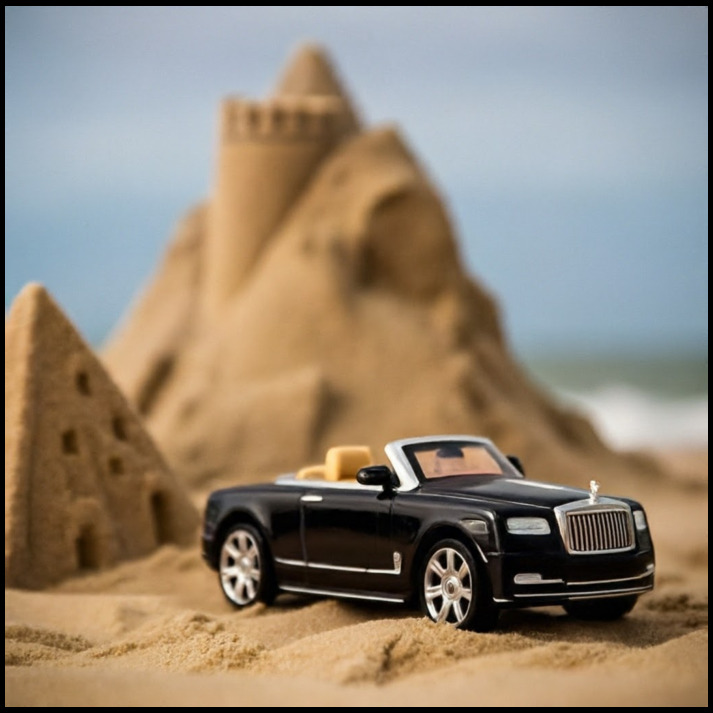} & 
        \includegraphics[width=0.20\textwidth, height=2.5cm]{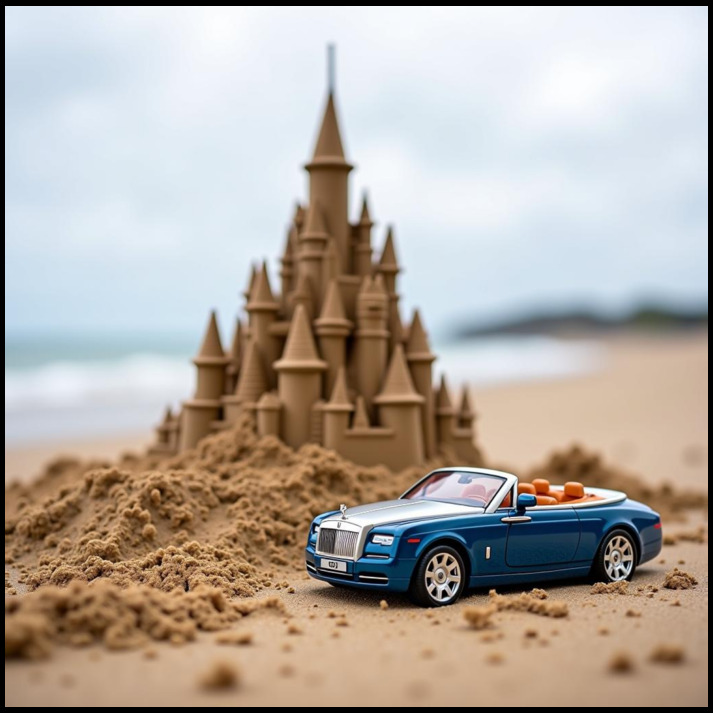} & 
        \includegraphics[width=0.20\textwidth, height=2.5cm]{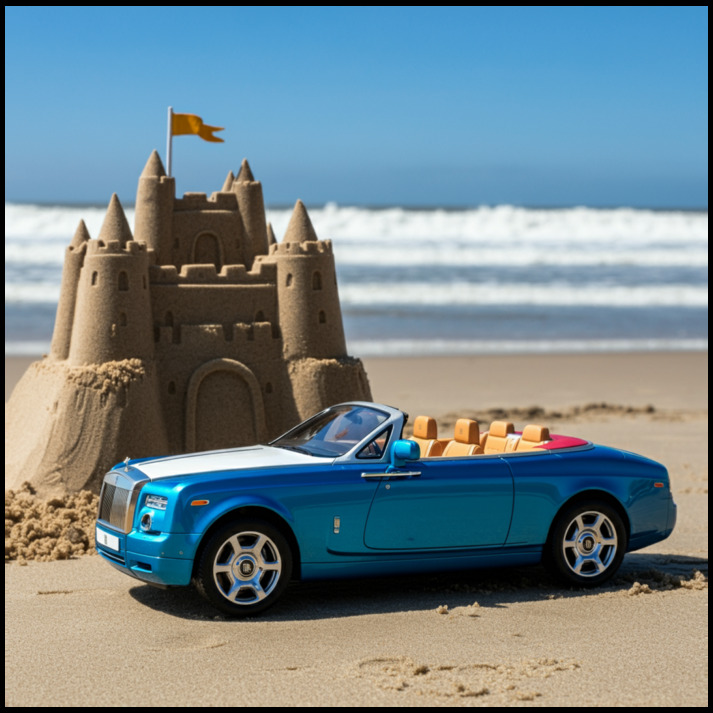} \\
        \multicolumn{4}{c}{\textit{A toy Rolls-Royce Phantom Drophead Coupé next to a giant sandcastle on the beach.}} \\

        \bottomrule
    \end{tabular}
    \caption{\textbf{Qualitative results.}
\textbf{(Top)} The backbone models show lower Faithfulness to Entity, with Wakame’s appearance differing from the reference image.
\textbf{(Bottom)} The retrieval models struggle with Instruction-following, over-relying on references, and failing to create compositions of the entity and the giant sandcastle.
We show the Image-Entity and Image-Text scores with the generated results.
}
    \label{fig:image_comparison3}
\end{figure*}

\subsection{Qualitative Results}
We present the visual results in~\cref{fig:image_comparison3}.
In the above example, the backbone models (second row) show lower faithfulness to the entity, with Wakame’s appearance differing from the reference image. In contrast, retrieval-augmented models (first row), using reference images during testing, achieve better visual alignment with the target. Instruct-Imagen demonstrates a balance between entity fidelity and creative flexibility in the generated images, supported by the highest alignment scores across both metrics.
These findings highlight future research directions, showing that enhancing the backbone model can improve both instruction-following capability and entity fidelity. Furthermore, combining a strong backbone with an advanced retrieval-augmented method enables the coexistence of these two aspects.

In the example below, Custom-Diff and DreamBooth show reduced instruction-following compared to their backbone model, SD. In particular, Custom-Diff struggles to create novel compositions of the entity and the giant sandcastle.
In contrast, the backbone models excel in both instruction-following and entity faithfulness, likely because ``Rolls-Royce Phantom Drophead Coupé'' is well-represented in their training data. The retrieval-augmented models underperform due to over-relying on the reference images and experiencing knowledge forgetting during fine-tuning on small sets of reference images.
These results suggest that the success of retrieval-augmented methods heavily depends on the entity domain and the customization approach.
More results can be found in the supplementary materials.

 \section{Conclusion}
We propose \ourdata, a benchmark for evaluating entity fidelity in text-to-image generation, focusing on visual concepts that require specialized knowledge. We design prompts based on Wikipedia entities and introduce a human evaluation framework to assess visual faithfulness. Extensive analysis reveals that while backbone models can generate specialized entities, retrieval-augmented models achieve higher faithfulness. However, these methods often struggle with creative prompts, highlighting the need for techniques that enhance entity fidelity without compromising instruction-following ability.

\custompara{Limitations.}
\ourdata does not include prompts requiring knowledge reasoning (e.g., the tallest building in Manhattan), as such prompts can be rewritten as entity-based ones by language preprocessing.

\newpage
\counterwithin{figure}{section}
\counterwithin{table}{section}
\renewcommand\thefigure{\thesection.\arabic{figure}}
\renewcommand\thetable{\thesection.\arabic{table}}

\appendix

\section{Additional Qualitative Examples}
\label{app:examples}
\begin{figure}[t]
    \centering
    \scriptsize
    \setlength{\tabcolsep}{2pt} %
    \begin{tabular}{c@{\;\;}c@{\;\;}c@{\;\;}c}
        \toprule %
        \textbf{Real Photo} & \textbf{Custom-Diff} / 0.52 / 0.36 & \textbf{DreamBooth} / 0.53 / 0.35 & \textbf{Instruct-Imagen} / 0.78 / 0.32\\
        \includegraphics[width=0.23\textwidth, height=2.8cm]{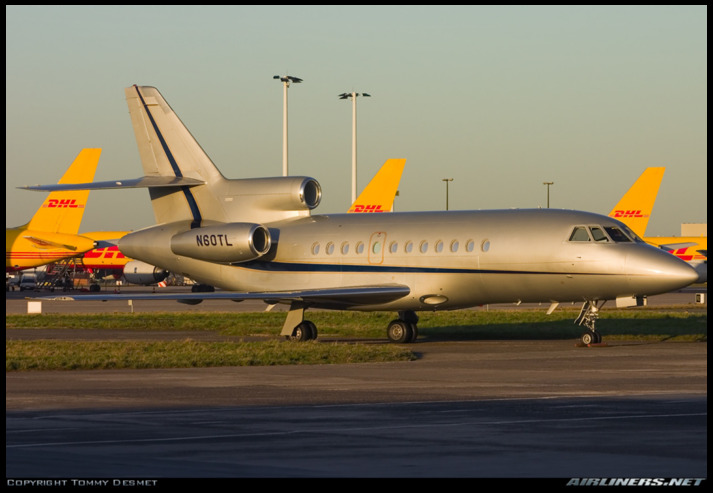} & 
        \includegraphics[width=0.23\textwidth, height=2.8cm]{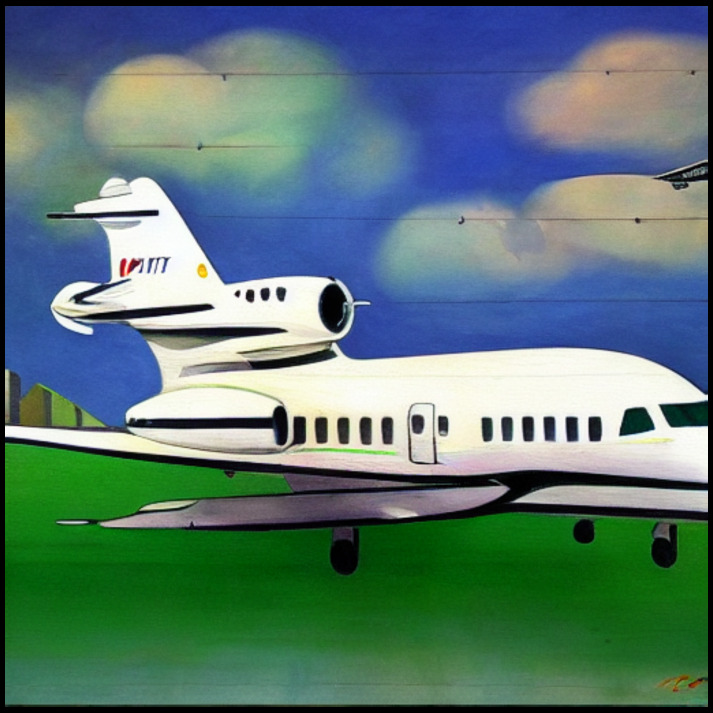} & 
        \includegraphics[width=0.23\textwidth, height=2.8cm]{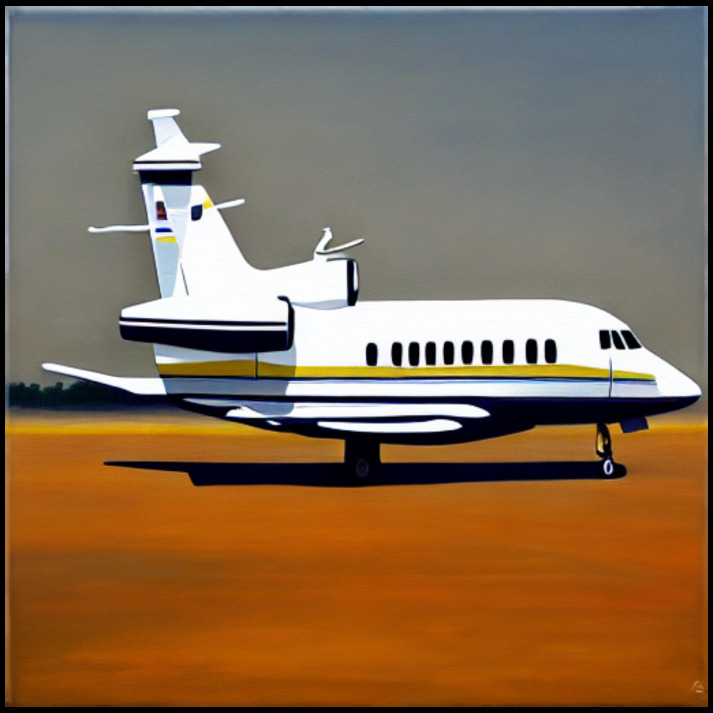} & 
        \includegraphics[width=0.23\textwidth, height=2.8cm]{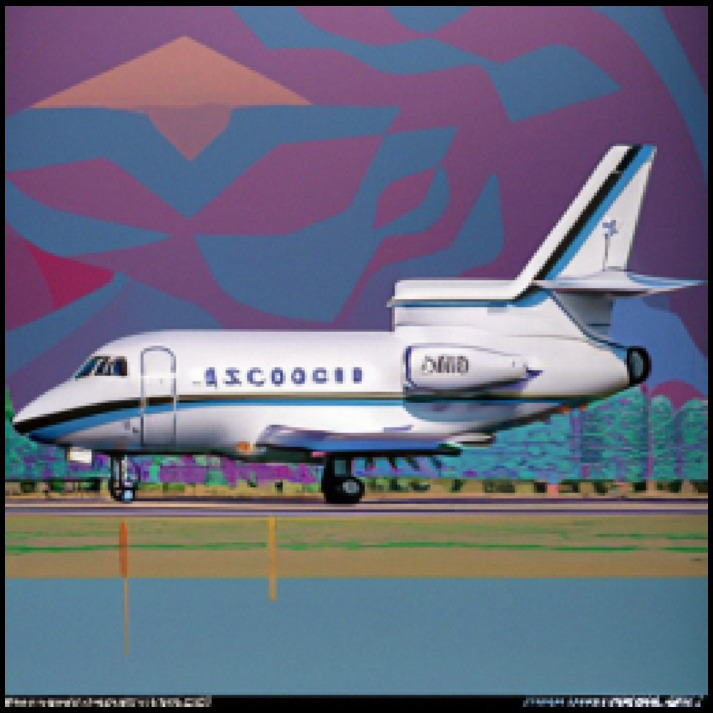} \\
        \textbf{SD} / 0.49 / 0.33 & \textbf{Imagen} / 0.71 / 0.33 & \textbf{Flux} / 0.69 / 0.30 & \textbf{Imagen-3} / 0.74 / 0.33\\
        \includegraphics[width=0.23\textwidth, height=2.8cm]{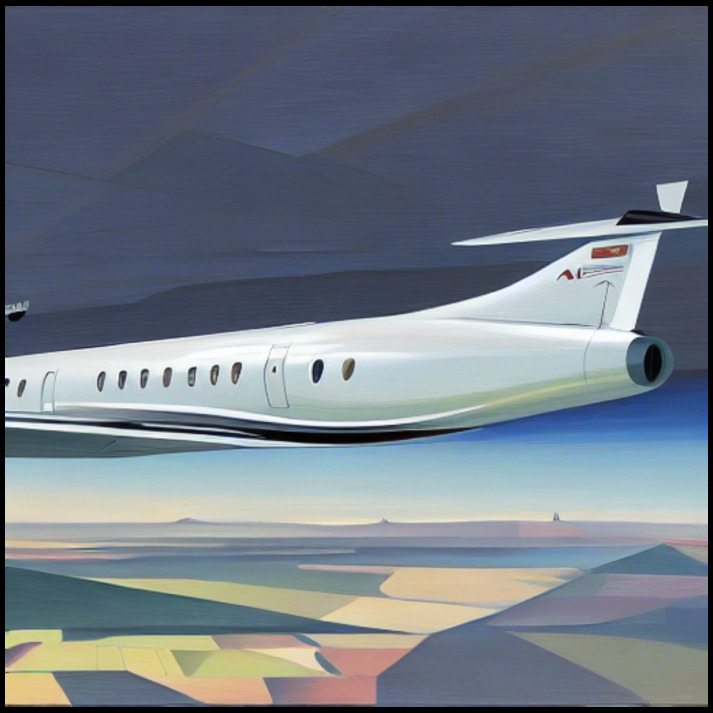} & 
        \includegraphics[width=0.23\textwidth, height=2.8cm]{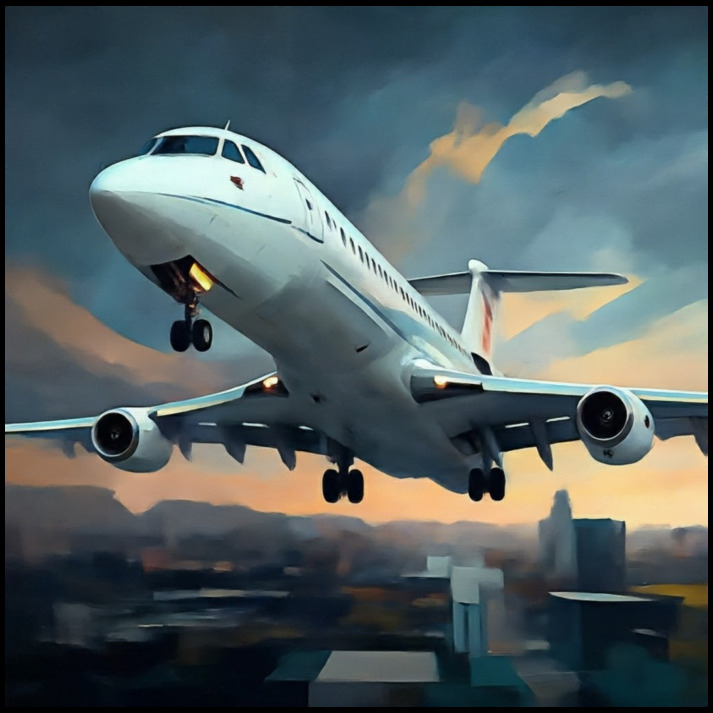} & 
        \includegraphics[width=0.23\textwidth, height=2.8cm]{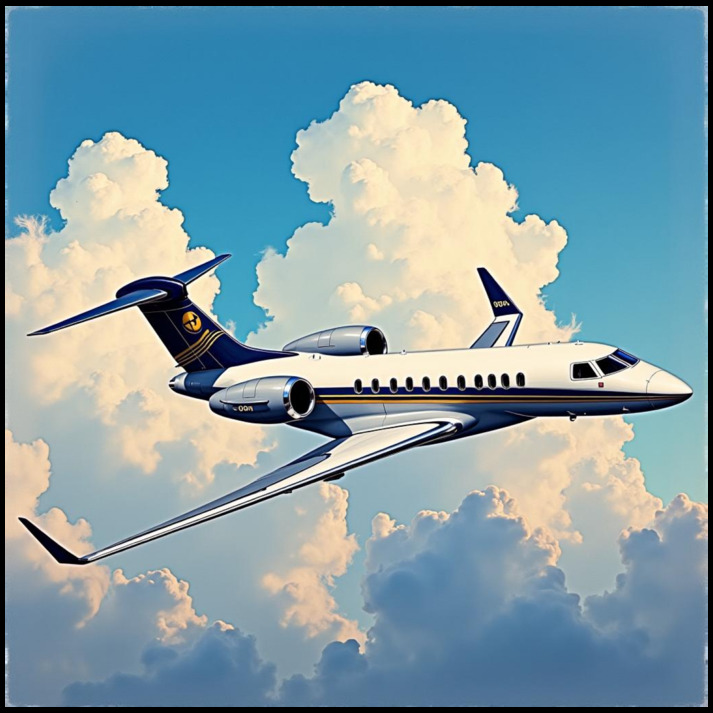} & 
        \includegraphics[width=0.23\textwidth, height=2.8cm]{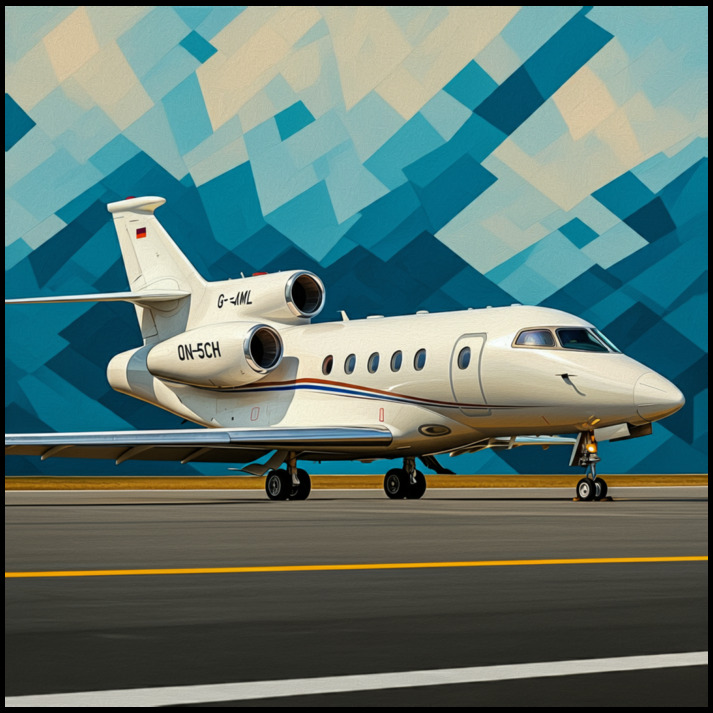} \\
        \multicolumn{4}{c}{\textit{A painting of Dassault Falcon 900 in a cubist style, parked in a surreal landscape.}} \\
        \midrule
        \textbf{Real Photo} & \textbf{Custom-Diff} / 0.50 / 0.38 & \textbf{DreamBooth} / 0.60 / 0.33 & \textbf{Instruct-Imagen} / 0.83 / 0.27 \\
        \includegraphics[width=0.23\textwidth, height=2.8cm]{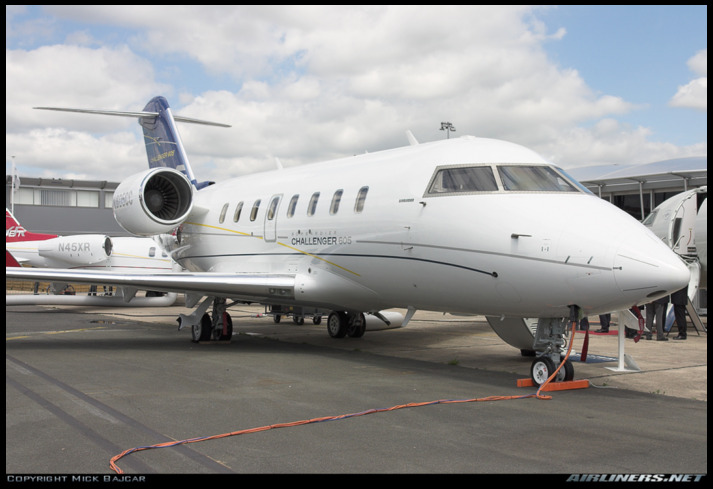} & 
        \includegraphics[width=0.23\textwidth, height=2.8cm]{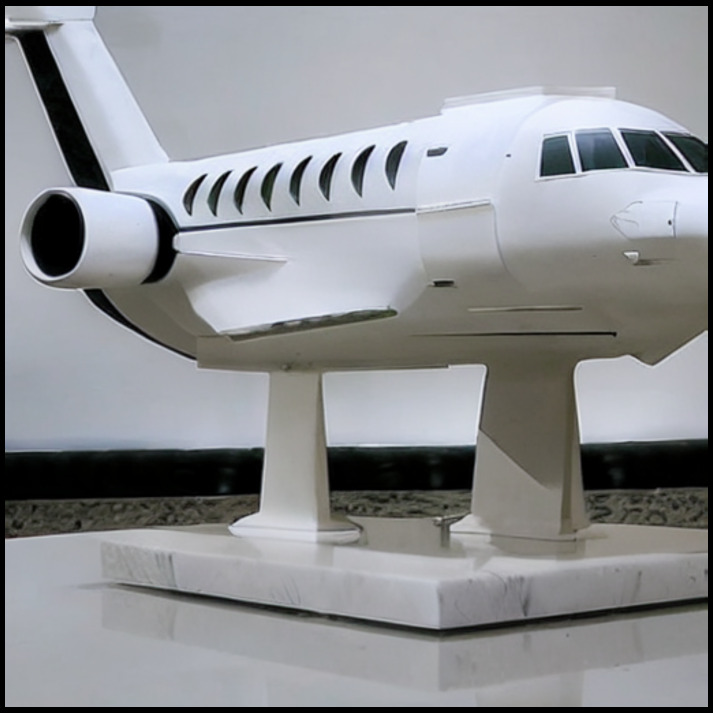} & 
        \includegraphics[width=0.23\textwidth, height=2.8cm]{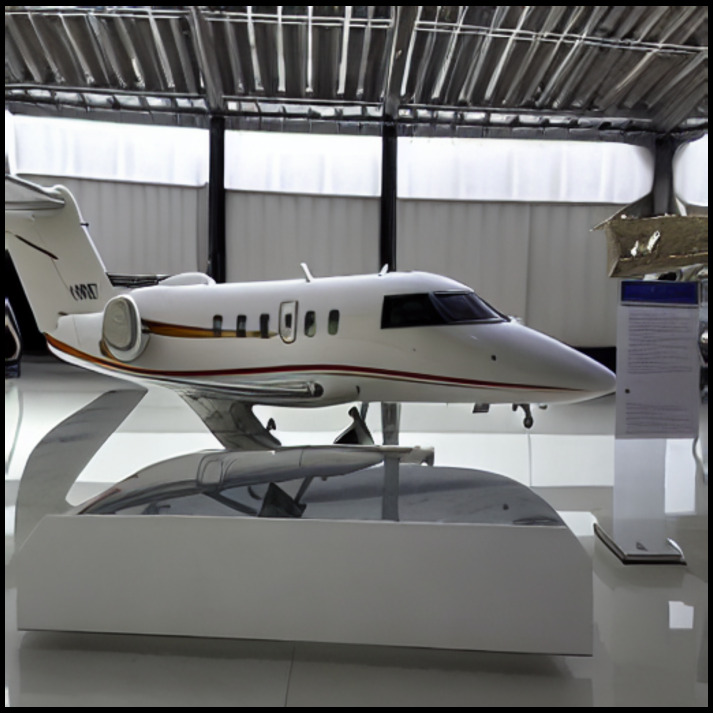} & 
        \includegraphics[width=0.23\textwidth, height=2.8cm]{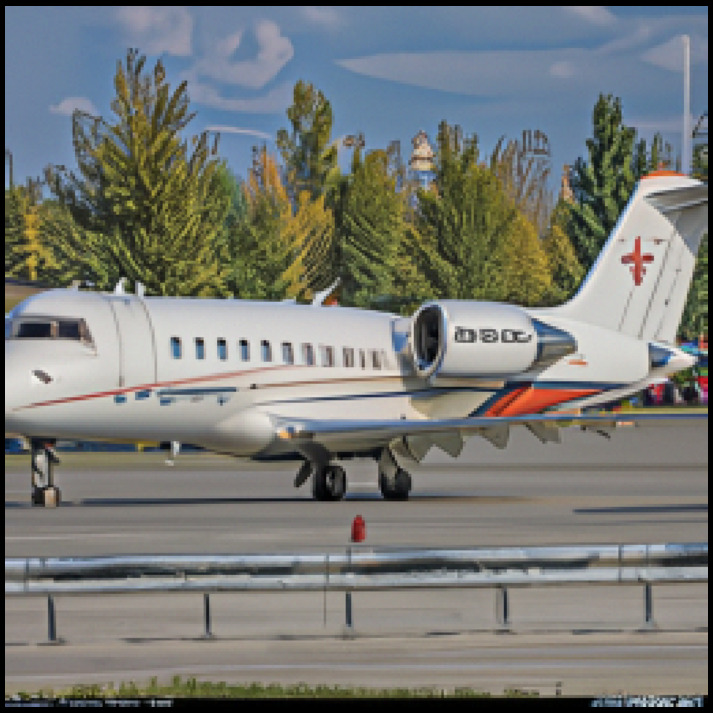} \\
        \textbf{SD} / 0.62 / 0.35 & \textbf{Imagen} / 0.69 / 0.33 & \textbf{Flux} / 0.65 / 0.37 & \textbf{Imagen-3} / 0.58 / 0.43 \\
        \includegraphics[width=0.23\textwidth, height=2.8cm]{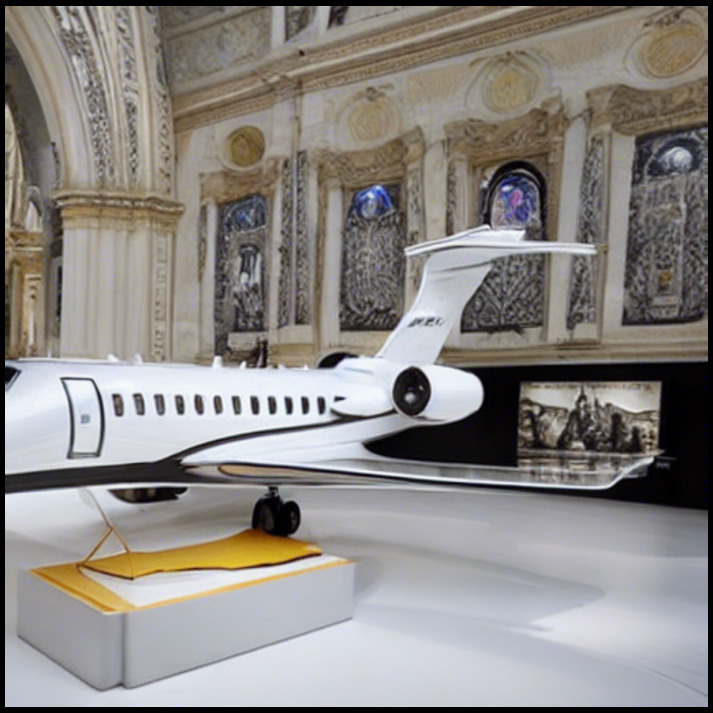} & 
        \includegraphics[width=0.23\textwidth, height=2.8cm]{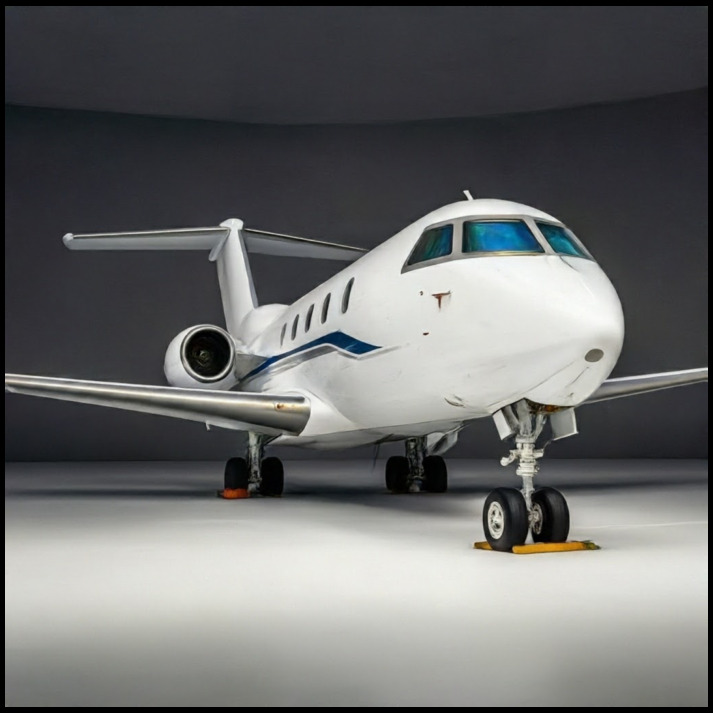} & 
        \includegraphics[width=0.23\textwidth, height=2.8cm]{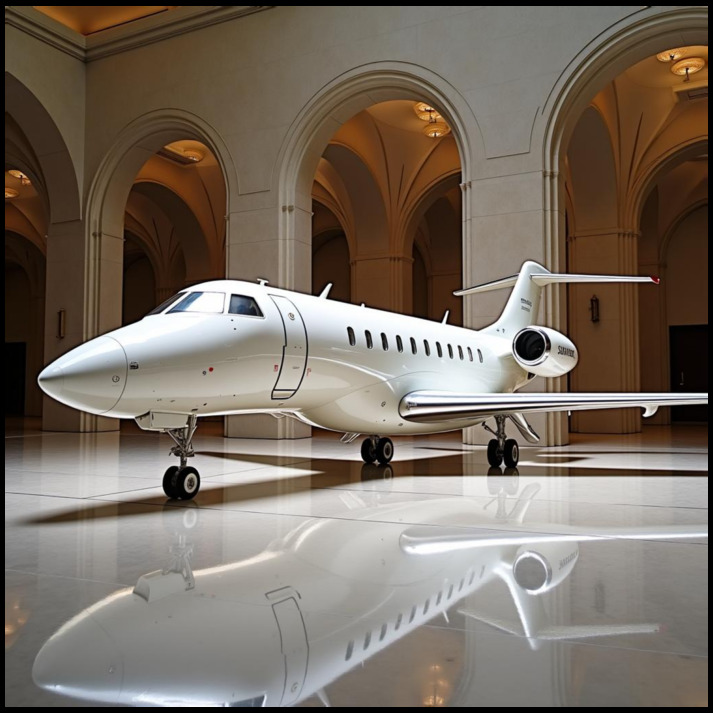} & 
        \includegraphics[width=0.23\textwidth, height=2.8cm]{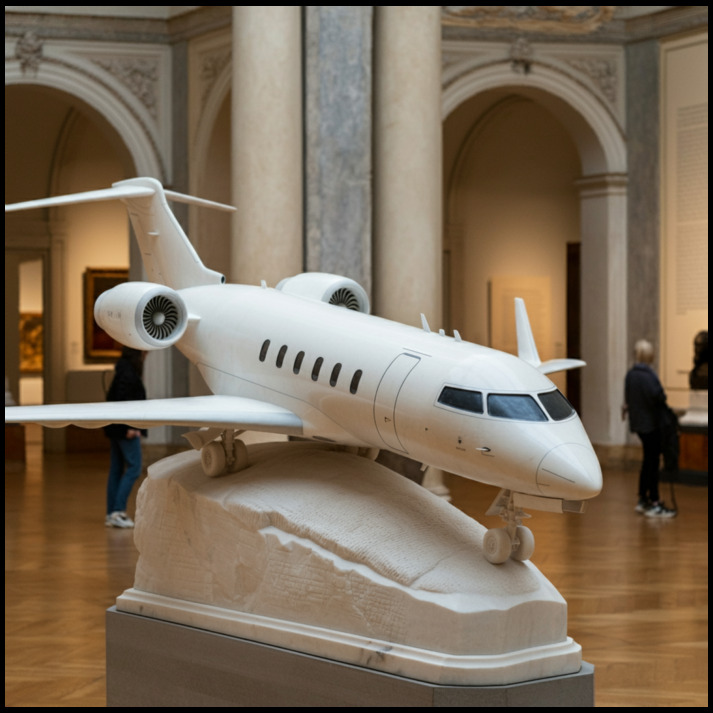} \\
        \multicolumn{4}{c}{\textit{A Bombardier Challenger 600 series aircraft sculpture made from marble in a grand art gallery.}} \\
        \midrule
        \textbf{Real Photo} & \textbf{Custom-Diff} / 0.81 / 0.29 & \textbf{DreamBooth} / 0.78 / 0.29 & \textbf{Instruct-Imagen} / 0.87 / 0.31 \\
        \includegraphics[width=0.23\textwidth, height=2.8cm]{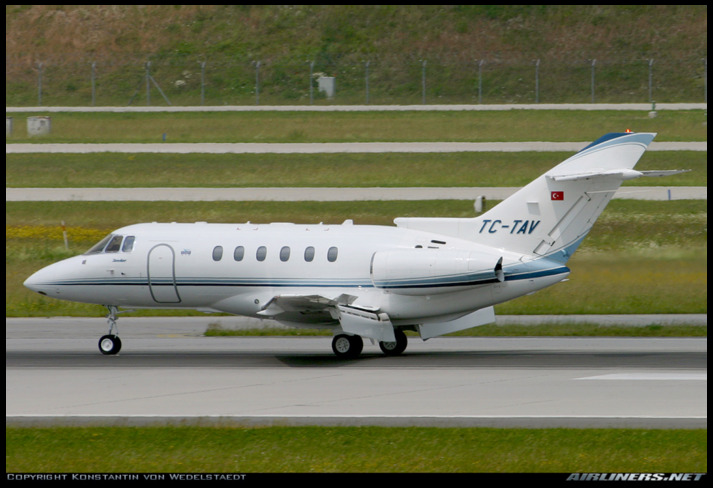} & 
        \includegraphics[width=0.23\textwidth, height=2.8cm]{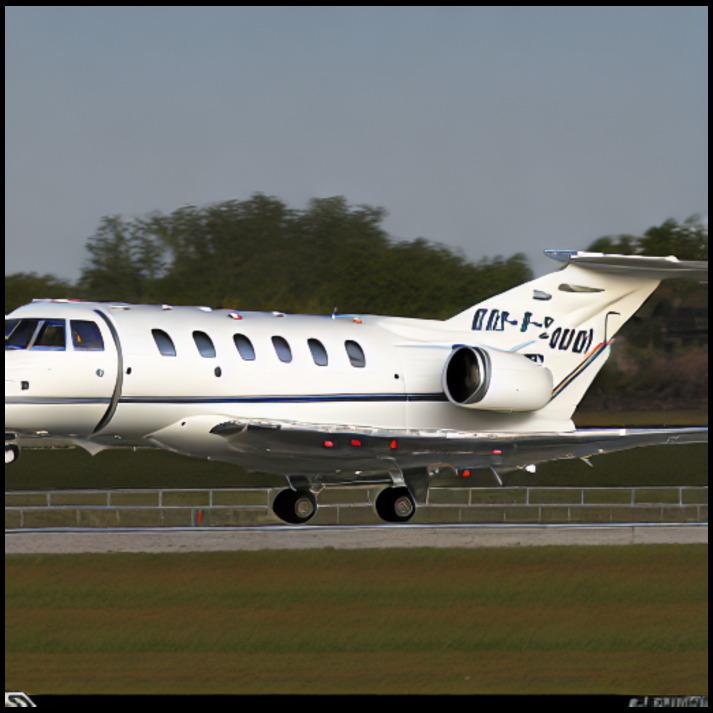} & 
        \includegraphics[width=0.23\textwidth, height=2.8cm]{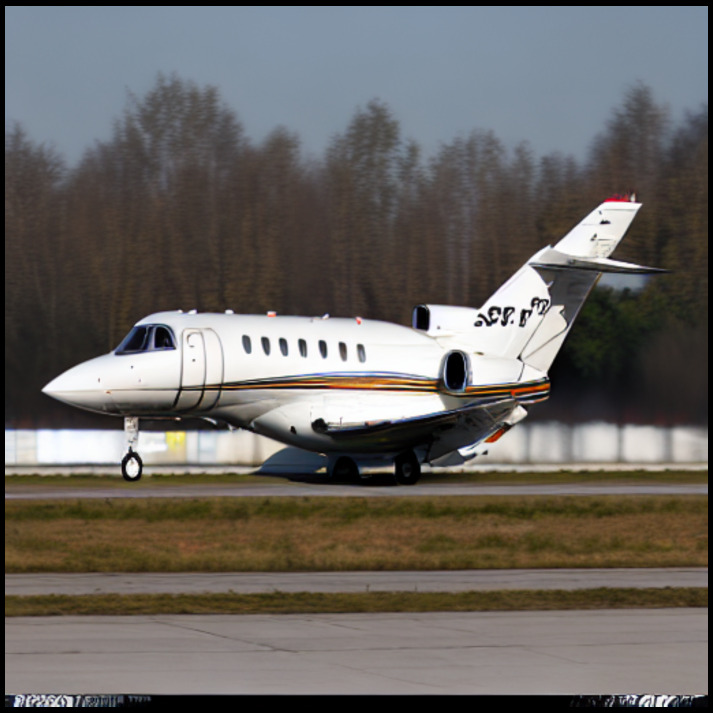} & 
        \includegraphics[width=0.23\textwidth, height=2.8cm]{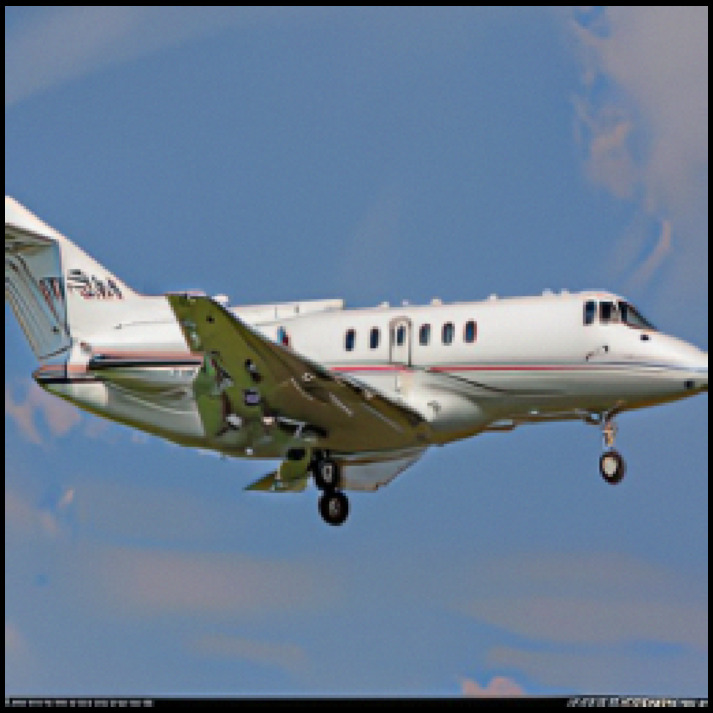} \\
        \textbf{SD} / 0.69 / 0.31 & \textbf{Imagen} / 0.77 / 0.29 & \textbf{Flux} / 0.71 / 0.27 & \textbf{Imagen-3} / 0.76 / 0.30 \\
        \includegraphics[width=0.23\textwidth, height=2.8cm]{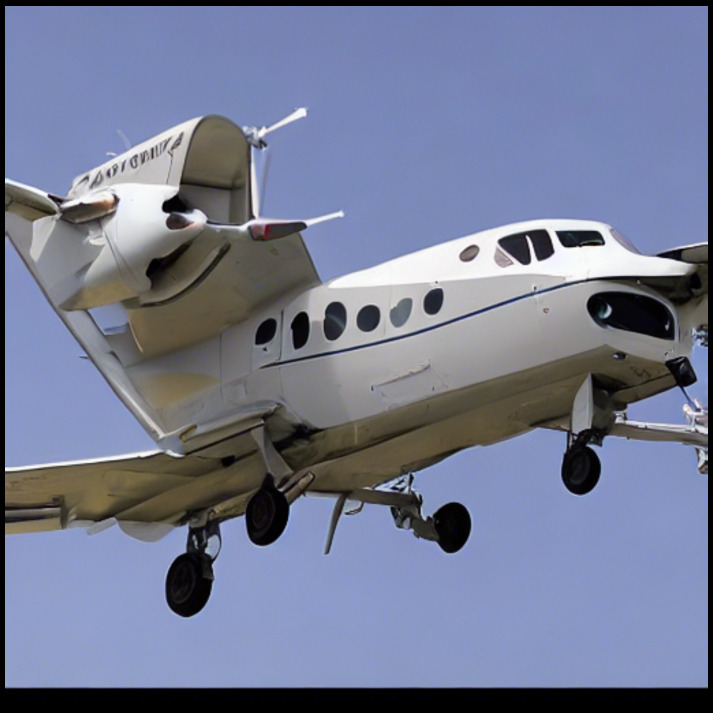} & 
        \includegraphics[width=0.23\textwidth, height=2.8cm]{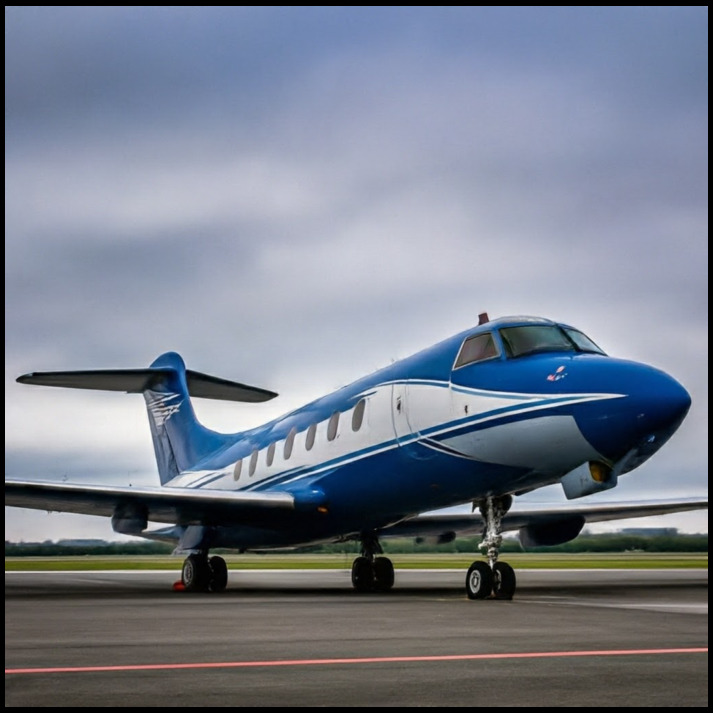} & 
        \includegraphics[width=0.23\textwidth, height=2.8cm]{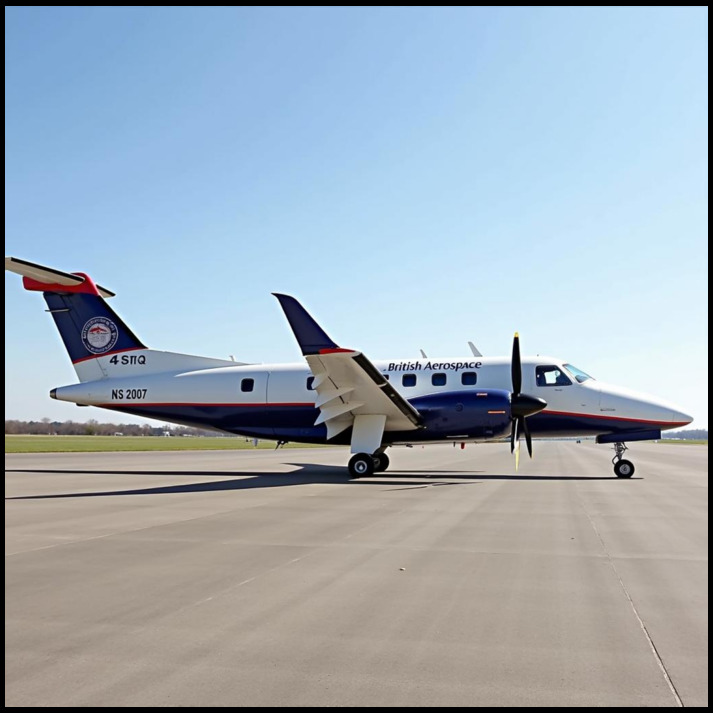} & 
        \includegraphics[width=0.23\textwidth, height=2.8cm]{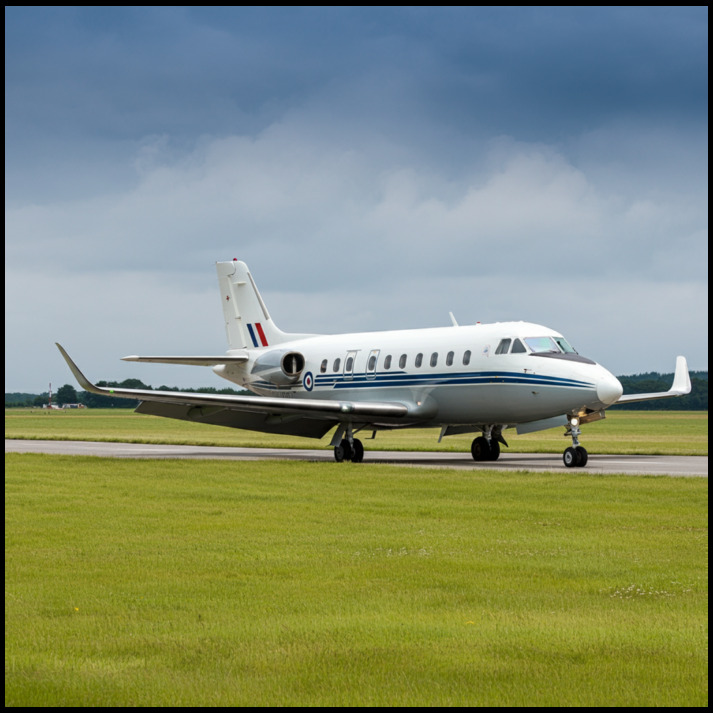} \\
        \multicolumn{4}{c}{\textit{photo of a British Aerospace 125 aircraft.}} \\
        \bottomrule
    \end{tabular}
    \caption{\textbf{Qualitative results} for the aircraft domain, including the DINO and CLIP-T scores.}
    \label{fig:image_comparison_1}
\end{figure}

\begin{figure}[t]
    \centering
    \scriptsize
    \setlength{\tabcolsep}{2pt} %
    \begin{tabular}{c@{\;\;}c@{\;\;}c@{\;\;}c}
        \toprule %
        \textbf{Real Photo} & \textbf{Custom-Diff} / 0.59 / 0.43 & \textbf{DreamBooth} / 0.59 / 0.41 & \textbf{Instruct-Imagen} / 0.72 / 0.40 \\
        \includegraphics[width=0.23\textwidth, height=2.8cm]{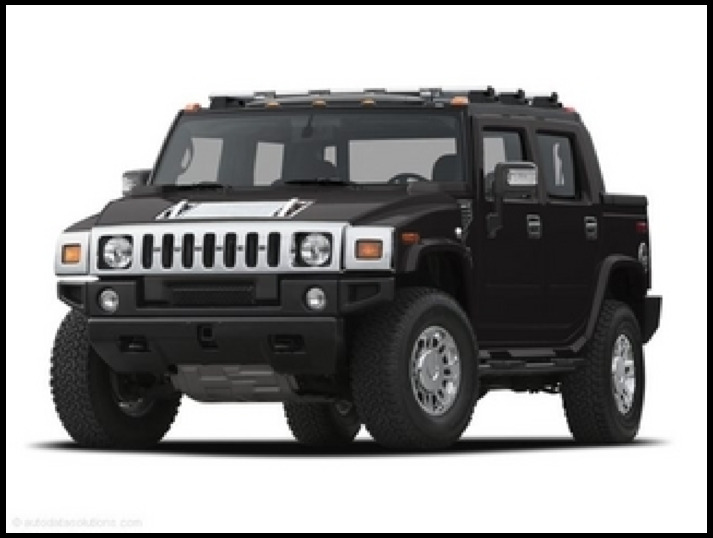} & 
        \includegraphics[width=0.23\textwidth, height=2.8cm]{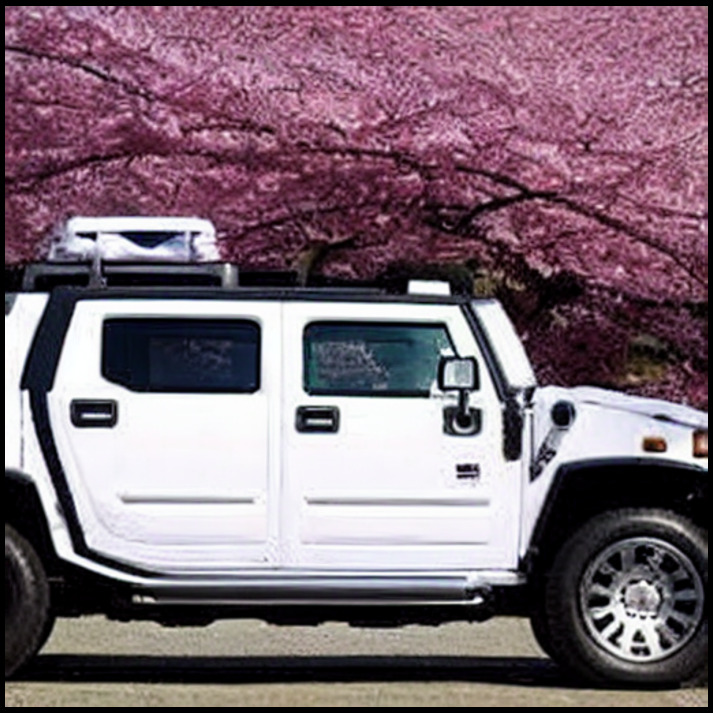} & 
        \includegraphics[width=0.23\textwidth, height=2.8cm]{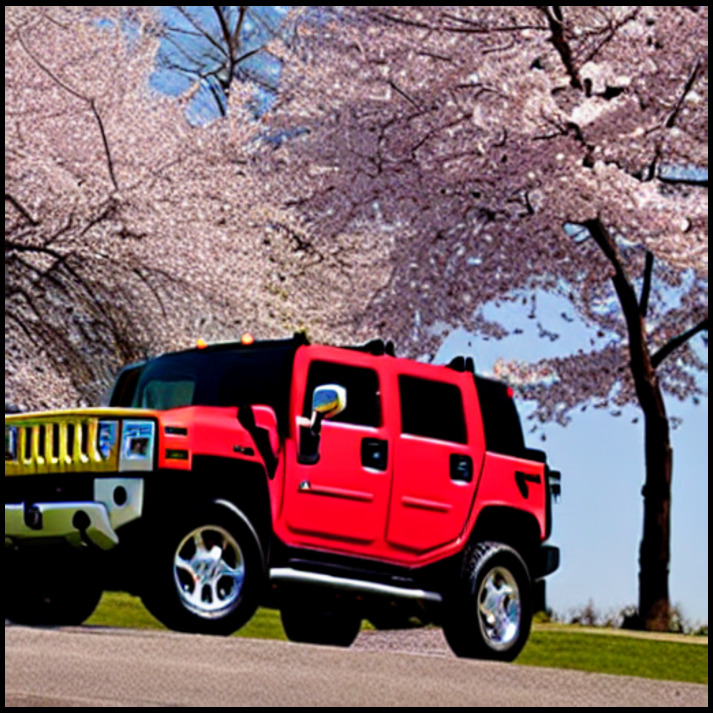} & 
        \includegraphics[width=0.23\textwidth, height=2.8cm]{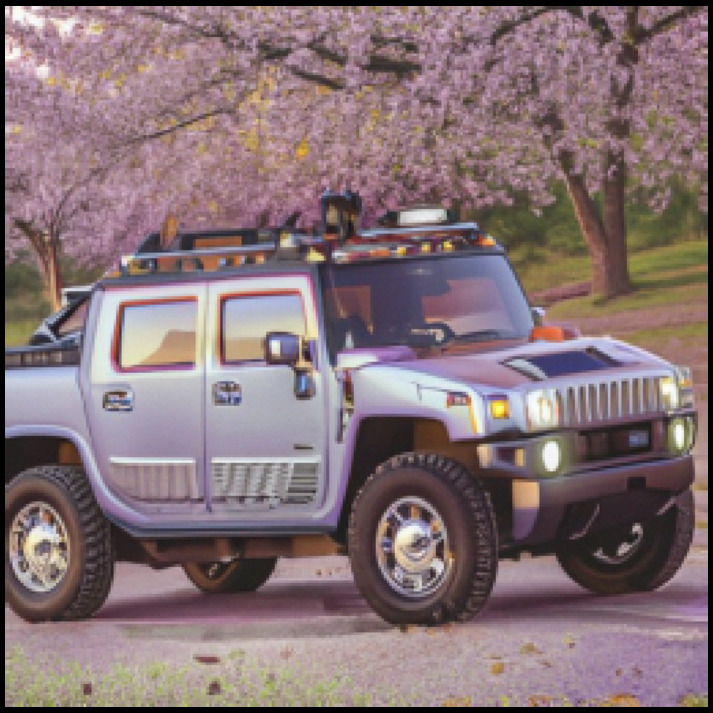} \\
        \textbf{SD} / 0.53 / 0.41 & \textbf{Imagen} / 0.62 / 0.42 & \textbf{Flux} / 0.65 / 0.41 & \textbf{Imagen-3} / 0.69 / 0.43 \\
        \includegraphics[width=0.23\textwidth, height=2.8cm]{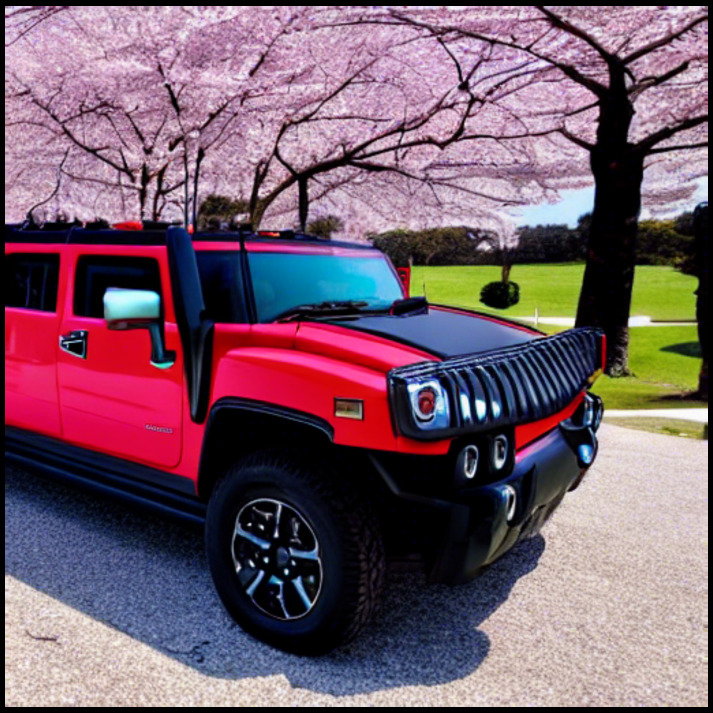} & 
        \includegraphics[width=0.23\textwidth, height=2.8cm]{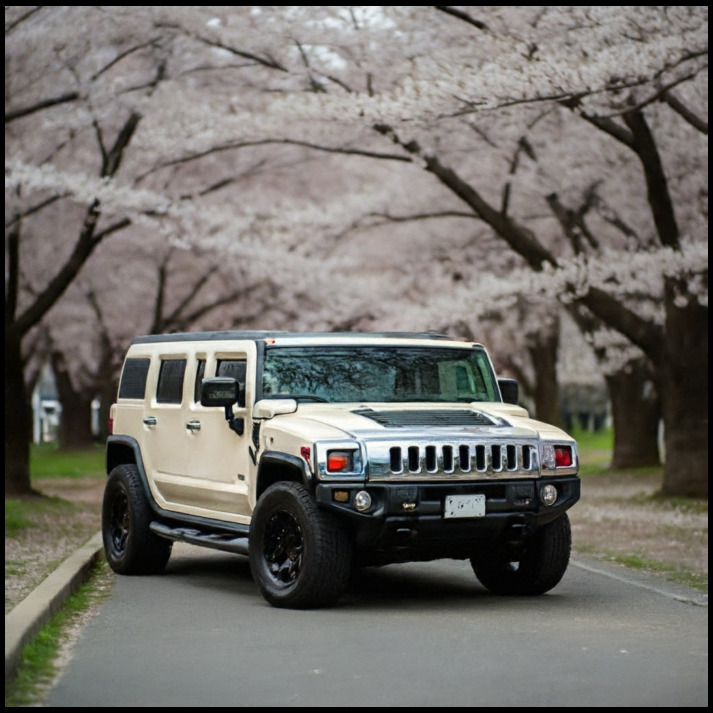} & 
        \includegraphics[width=0.23\textwidth, height=2.8cm]{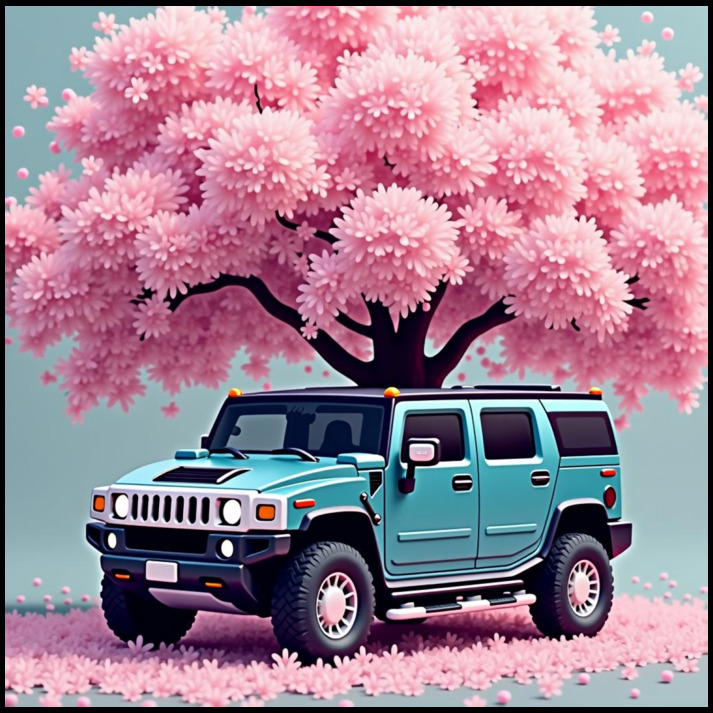} & 
        \includegraphics[width=0.23\textwidth, height=2.8cm]{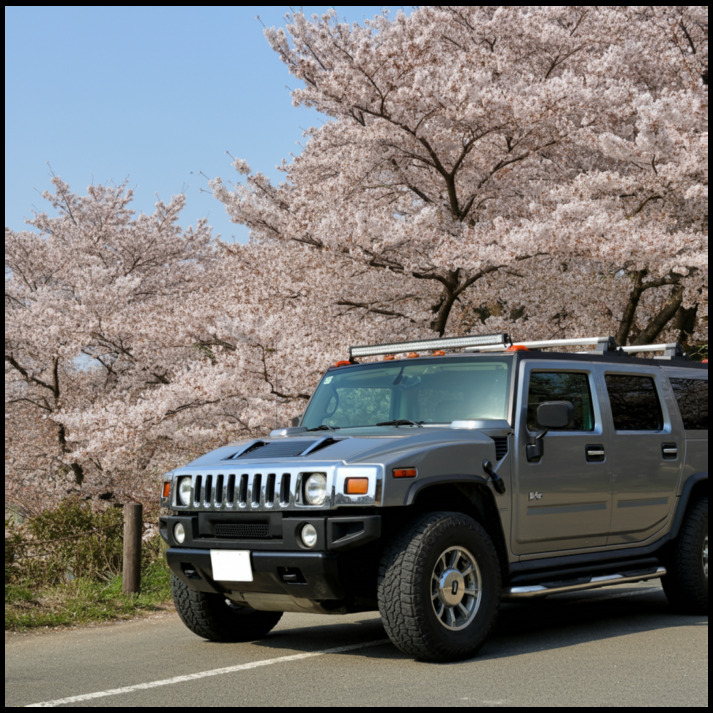} \\
        \multicolumn{4}{c}{\textit{A Hummer H2 car resting beneath the cherry blossoms in full bloom.}} \\
        \midrule
        \textbf{Real Photo} & \textbf{Custom-Diff} / 0.61 / 0.39 & \textbf{DreamBooth} / 0.38 / 0.40 & \textbf{Instruct-Imagen} / 0.59 / 0.41 \\
        \includegraphics[width=0.23\textwidth, height=2.8cm]{images_lowres/car_35_10_photo.jpg} & 
        \includegraphics[width=0.23\textwidth, height=2.8cm]{images_lowres/car_35_10_custom.jpg} & 
        \includegraphics[width=0.23\textwidth, height=2.8cm]{images_lowres/car_35_10_db.jpg} & 
        \includegraphics[width=0.23\textwidth, height=2.8cm]{images_lowres/car_35_10_instruct.jpg} \\
        \textbf{SD} / 0.53 / 0.41 & \textbf{Imagen} / 0.43 / 0.42 & \textbf{Flux} / 0.35 / 0.43 & \textbf{Imagen-3} / 0.47 / 0.45 \\
        \includegraphics[width=0.23\textwidth, height=2.8cm]{images_lowres/car_35_10_sd.jpg} & 
        \includegraphics[width=0.23\textwidth, height=2.8cm]{images_lowres/car_35_10_imagen.jpg} & 
        \includegraphics[width=0.23\textwidth, height=2.8cm]{images_lowres/car_35_10_flux.jpg} & 
        \includegraphics[width=0.23\textwidth, height=2.8cm]{images_lowres/car_35_10_juno.jpg} \\
        \multicolumn{4}{c}{\textit{A toy Rolls-Royce Phantom Drophead Coupé next to a giant sandcastle on the beach.}} \\
        \midrule
        \textbf{Real Photo} & \textbf{Custom-Diff} / 0.75 / 0.34 & \textbf{DreamBooth} / 0.74 / 0.34 & \textbf{Instruct-Imagen} / 0.80 / 0.34 \\
        \includegraphics[width=0.23\textwidth, height=2.8cm]{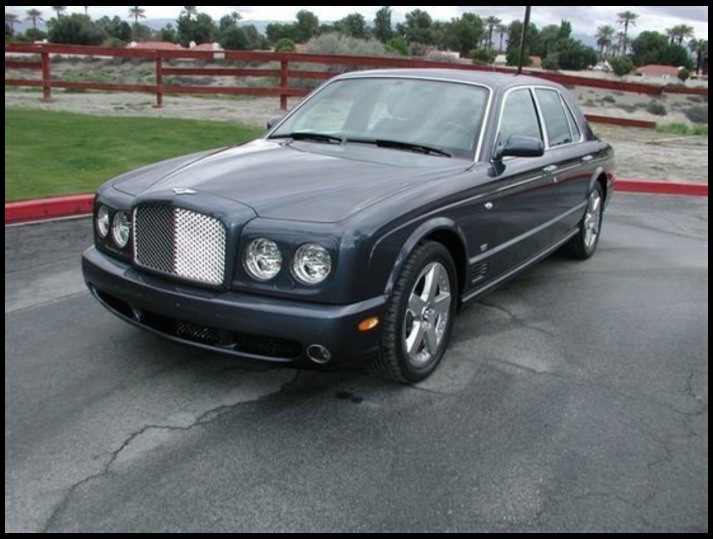} & 
        \includegraphics[width=0.23\textwidth, height=2.8cm]{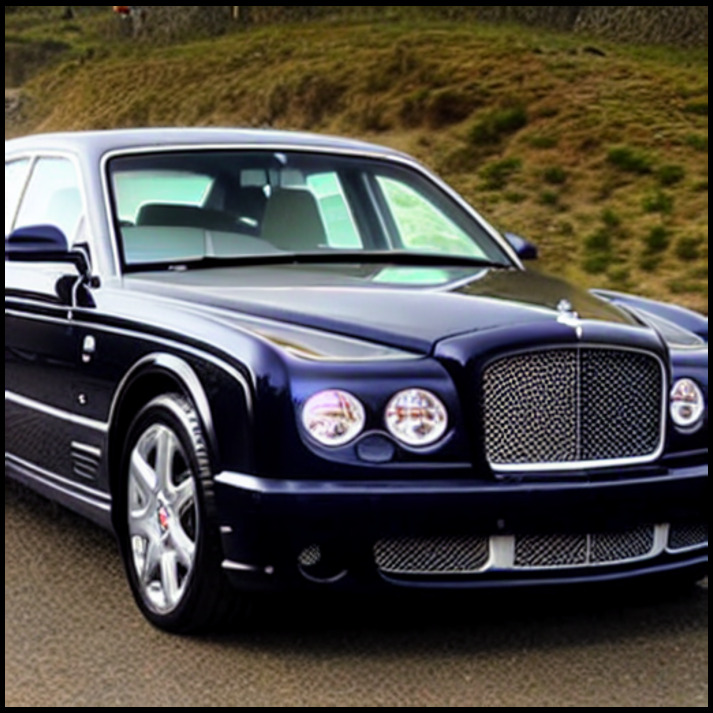} & 
        \includegraphics[width=0.23\textwidth, height=2.8cm]{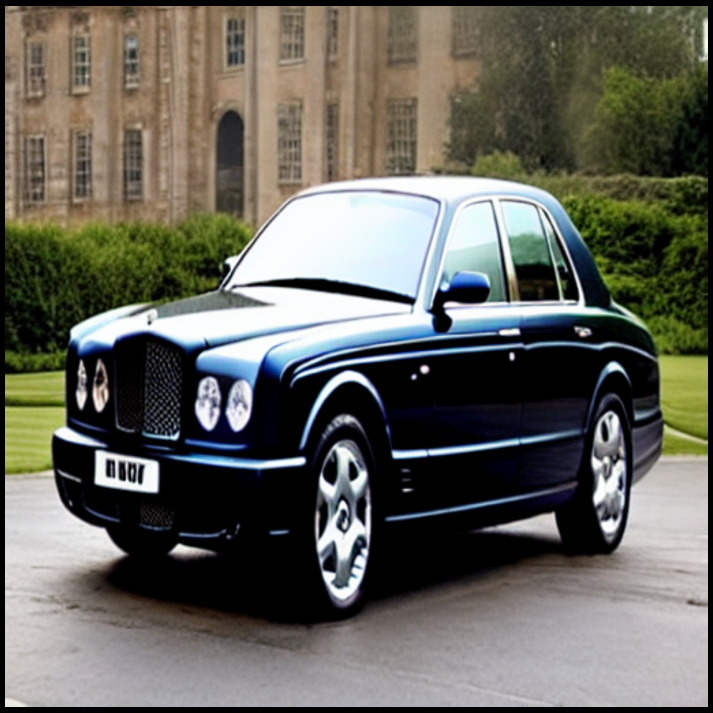} & 
        \includegraphics[width=0.23\textwidth, height=2.8cm]{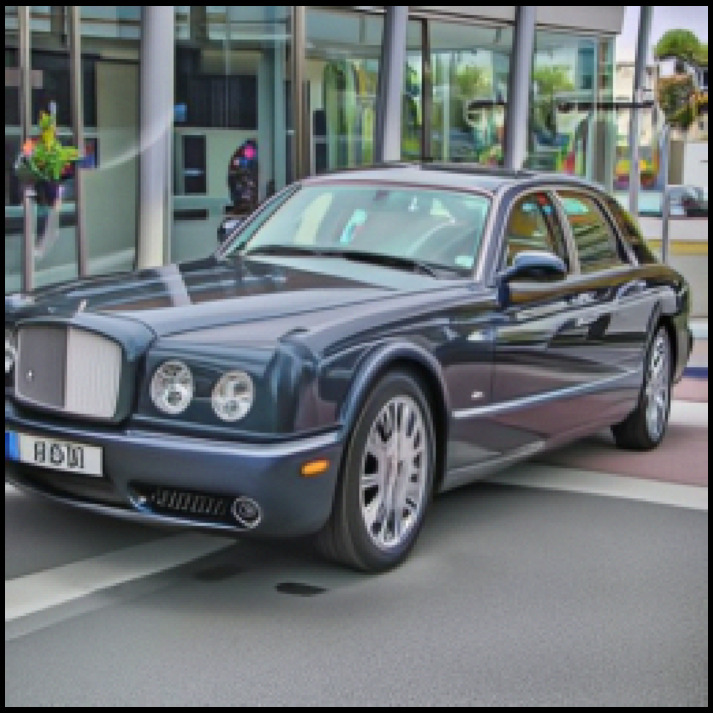} \\
        \textbf{SD} / 0.71 / 0.34 & \textbf{Imagen} / 0.68 / 0.31 & \textbf{Flux} / 0.74 / 0.33 & \textbf{Imagen-3} / 0.75 / 0.32 \\
        \includegraphics[width=0.23\textwidth, height=2.8cm]{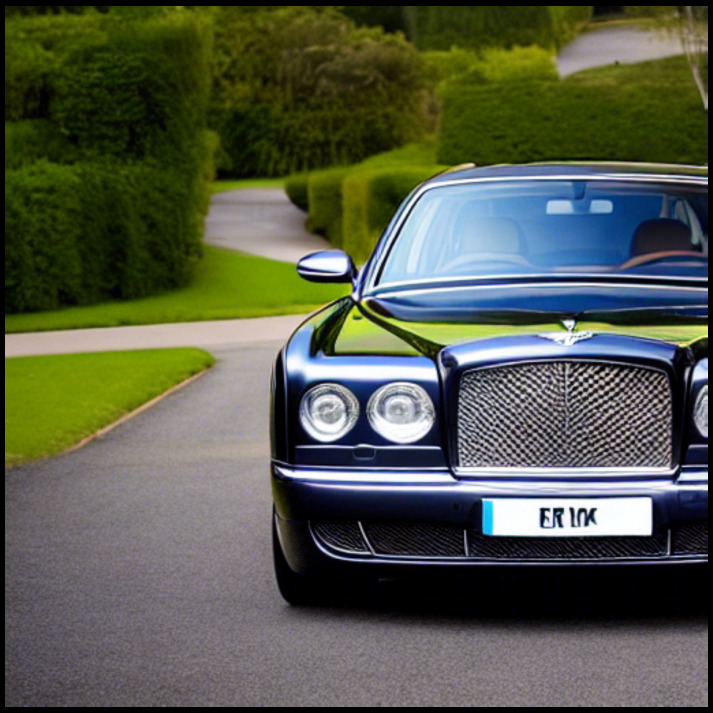} & 
        \includegraphics[width=0.23\textwidth, height=2.8cm]{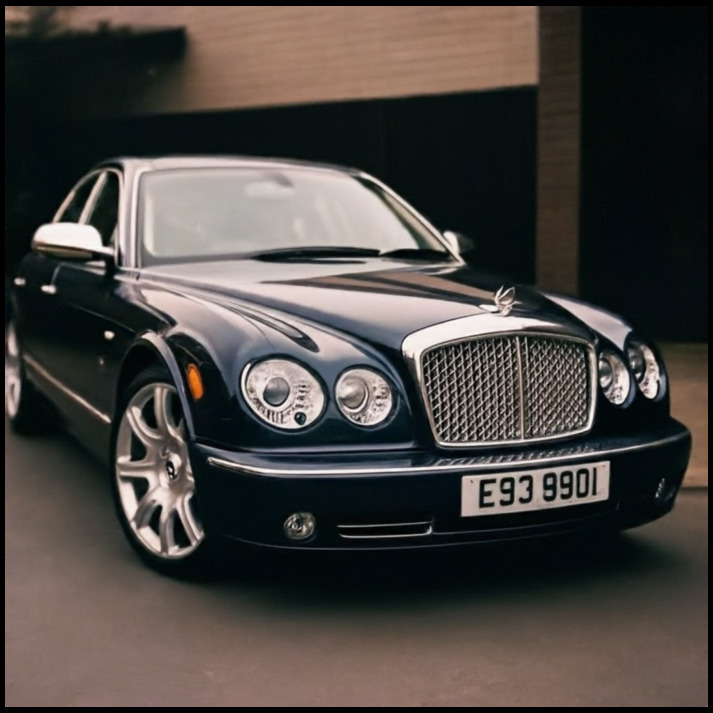} & 
        \includegraphics[width=0.23\textwidth, height=2.8cm]{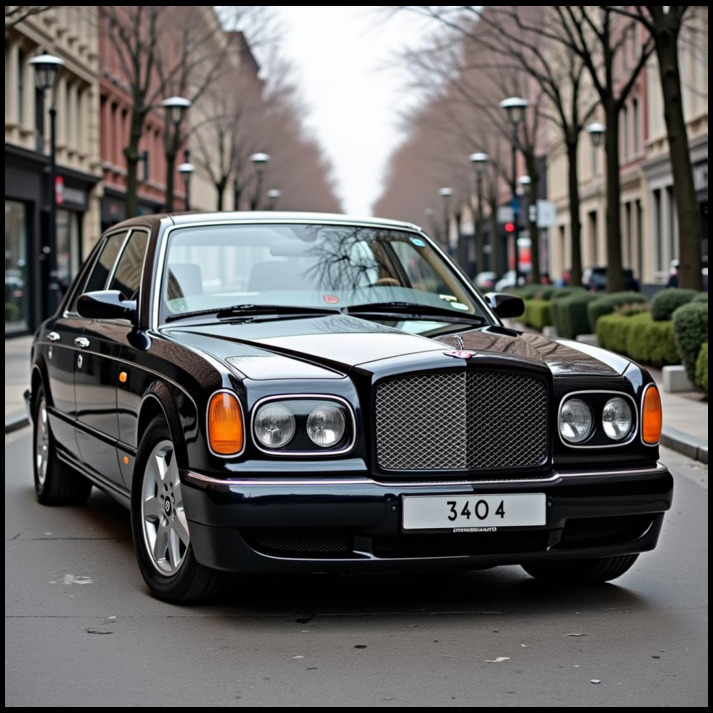} & 
        \includegraphics[width=0.23\textwidth, height=2.8cm]{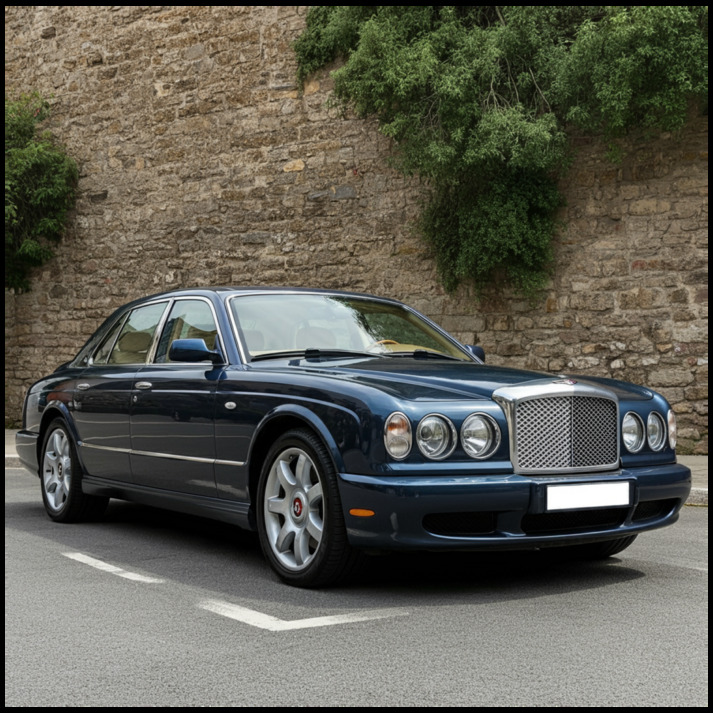} \\
        \multicolumn{4}{c}{\textit{photo of a Bentley Arnage.}} \\
        \bottomrule
    \end{tabular}
    \caption{\textbf{Qualitative results} for the vehicle domain, including the DINO and CLIP-T scores.}
    \label{fig:image_comparison_2}
\end{figure}

\begin{figure}[t]
    \centering
    \scriptsize
    \setlength{\tabcolsep}{2pt} %
    \begin{tabular}{c@{\;\;}c@{\;\;}c@{\;\;}c}
        \toprule %
        \textbf{Real Photo} & \textbf{Custom-Diff} / 0.38 / 0.35 & \textbf{DreamBooth} / 0.56 / 0.36 & \textbf{Instruct-Imagen} / 0.53 / 0.40 \\
        \includegraphics[width=0.23\textwidth, height=2.8cm]{images_lowres/dish_3_12_photo.jpg} & 
        \includegraphics[width=0.23\textwidth, height=2.8cm]{images_lowres/dish_3_12_custom.jpg} & 
        \includegraphics[width=0.23\textwidth, height=2.8cm]{images_lowres/dish_3_12_db.jpg} & 
        \includegraphics[width=0.23\textwidth, height=2.8cm]{images_lowres/dish_3_12_instruct.jpg} \\
        \textbf{SD} / 0.49 / 0.39 & \textbf{Imagen} / 0.29 / 0.37 & \textbf{Flux} / 0.41 / 0.34 & \textbf{Imagen-3} / 0.38 / 0.36 \\
        \includegraphics[width=0.23\textwidth, height=2.8cm]{images_lowres/dish_3_12_sd.jpg} & 
        \includegraphics[width=0.23\textwidth, height=2.8cm]{images_lowres/dish_3_12_imagen.jpg} & 
        \includegraphics[width=0.23\textwidth, height=2.8cm]{images_lowres/dish_3_12_flux.jpg} & 
        \includegraphics[width=0.23\textwidth, height=2.8cm]{images_lowres/dish_3_12_juno.jpg} \\
        \multicolumn{4}{c}{\textit{A Wakame dish with a cherry flower on top of it.}} \\
        \midrule
        \textbf{Real Photo} & \textbf{Custom-Diff} / 0.36 / 0.35 & \textbf{DreamBooth} / 0.18 / 0.33 & \textbf{Instruct-Imagen} / 0.49 / 0.40 \\
        \includegraphics[width=0.23\textwidth, height=2.8cm]{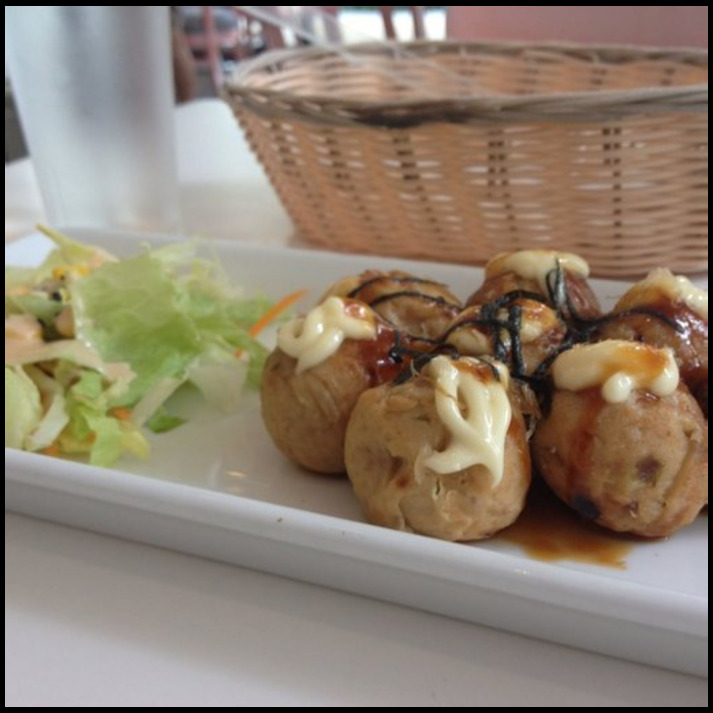} & 
        \includegraphics[width=0.23\textwidth, height=2.8cm]{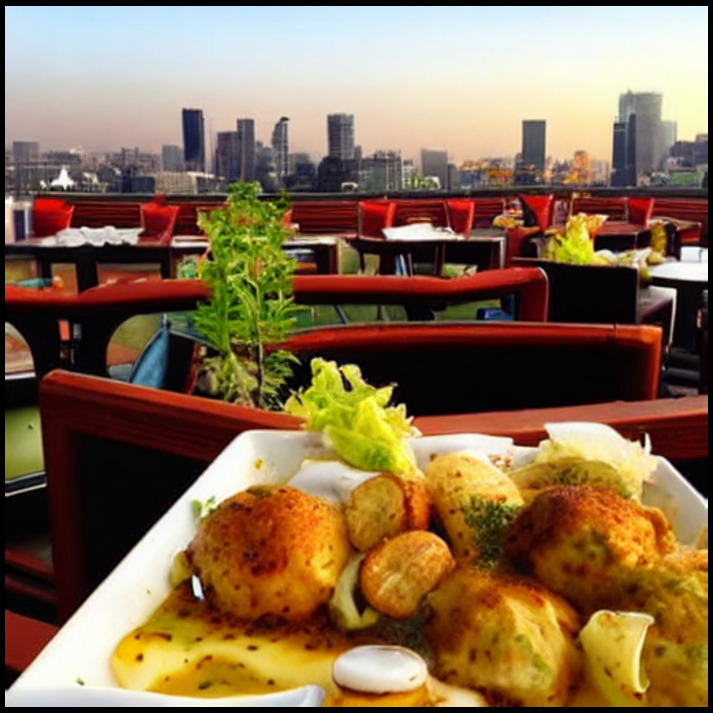} & 
        \includegraphics[width=0.23\textwidth, height=2.8cm]{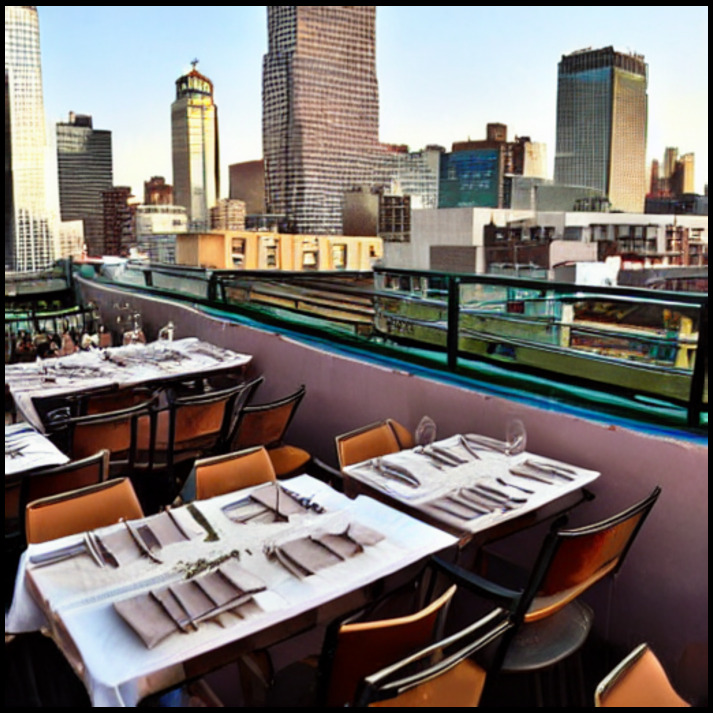} & 
        \includegraphics[width=0.23\textwidth, height=2.8cm]{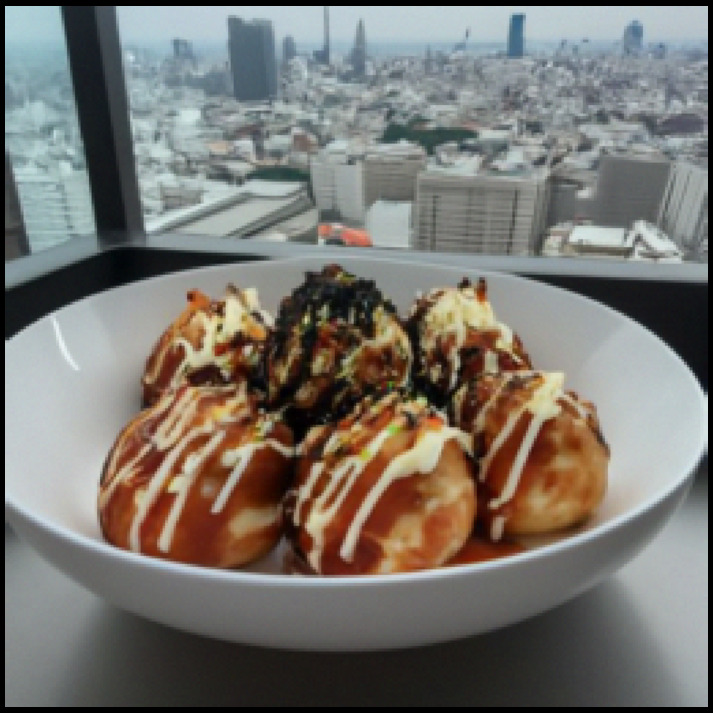} \\
        \textbf{SD} / 0.32 / 0.38 & \textbf{Imagen} / 0.39 / 0.37 & \textbf{Flux} / 0.33 / 0.34 & \textbf{Imagen-3} / 0.35 / 0.41 \\
        \includegraphics[width=0.23\textwidth, height=2.8cm]{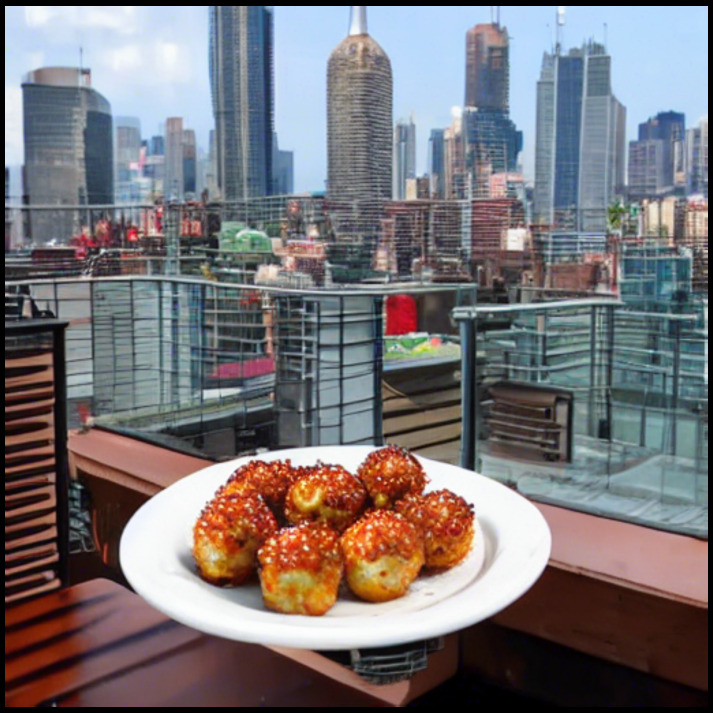} & 
        \includegraphics[width=0.23\textwidth, height=2.8cm]{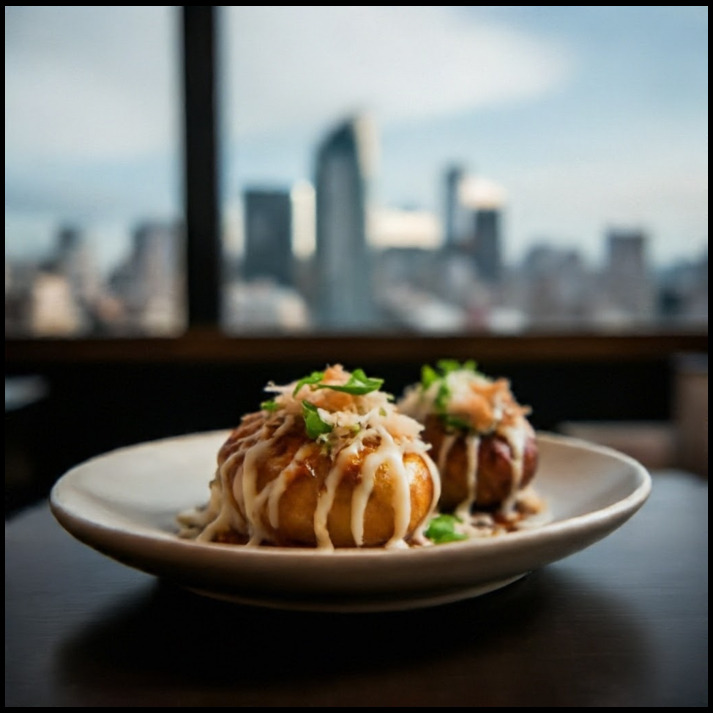} & 
        \includegraphics[width=0.23\textwidth, height=2.8cm]{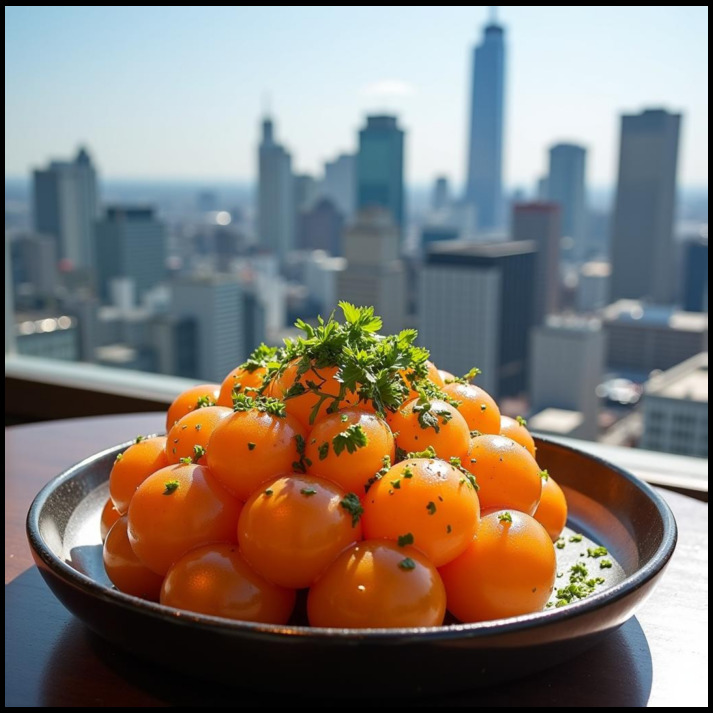} & 
        \includegraphics[width=0.23\textwidth, height=2.8cm]{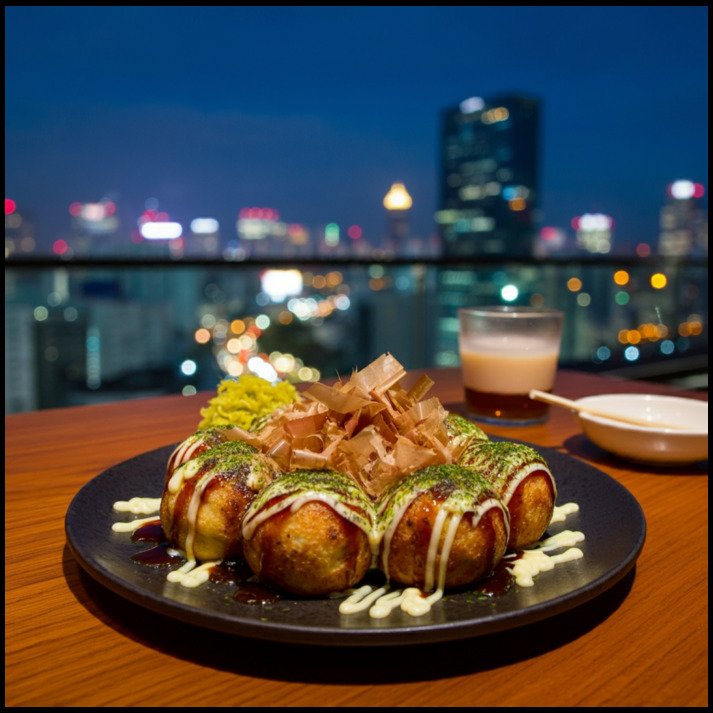} \\
        \multicolumn{4}{c}{\textit{A Takoyaki on a rooftop restaurant with a skyline view.}} \\
        \midrule
        \textbf{Real Photo} & \textbf{Custom-Diff} / 0.72 / 0.35 & \textbf{DreamBooth} / 0.76 / 0.33 & \textbf{Instruct-Imagen} / 0.77 / 0.32 \\
        \includegraphics[width=0.23\textwidth, height=2.8cm]{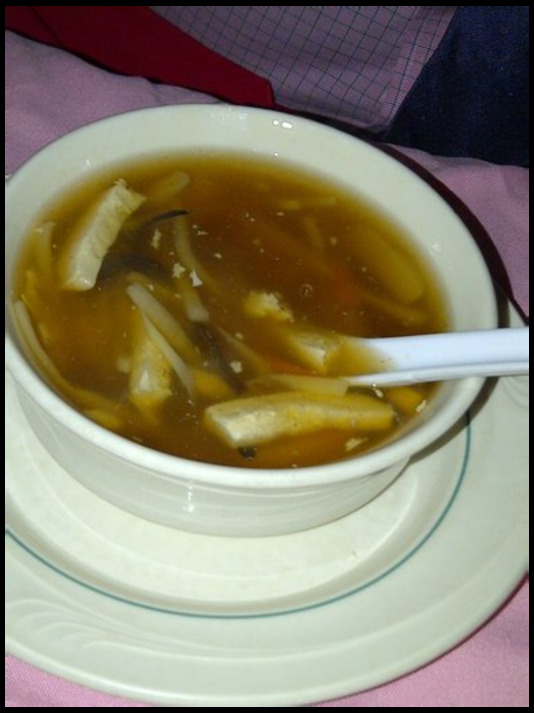} & 
        \includegraphics[width=0.23\textwidth, height=2.8cm]{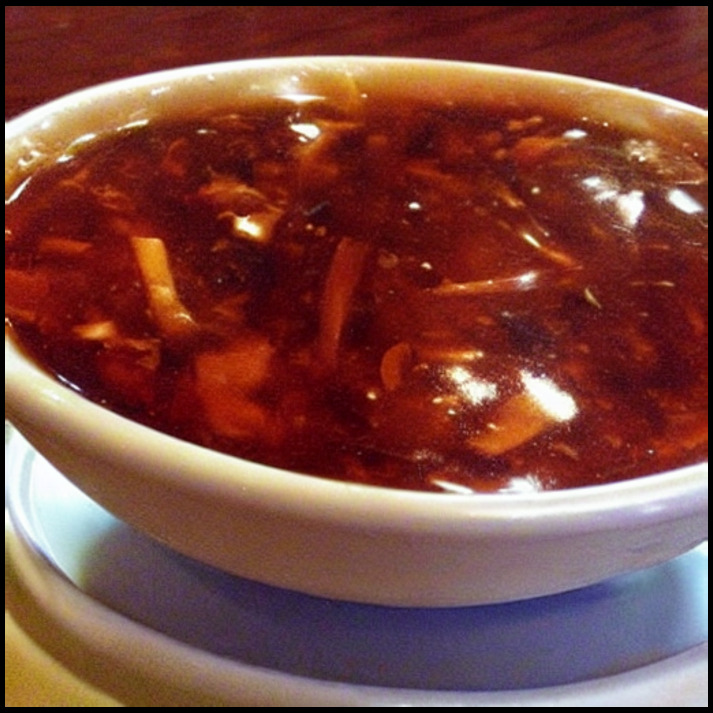} & 
        \includegraphics[width=0.23\textwidth, height=2.8cm]{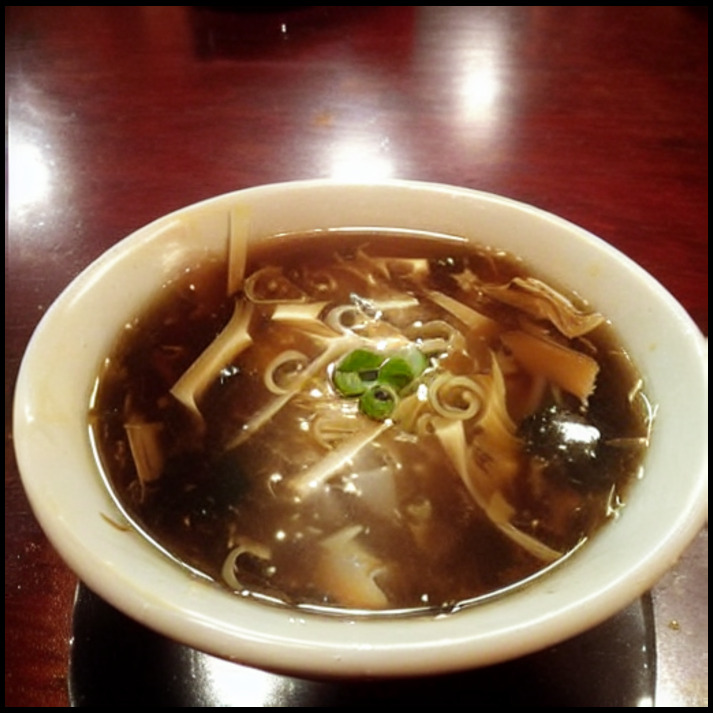} & 
        \includegraphics[width=0.23\textwidth, height=2.8cm]{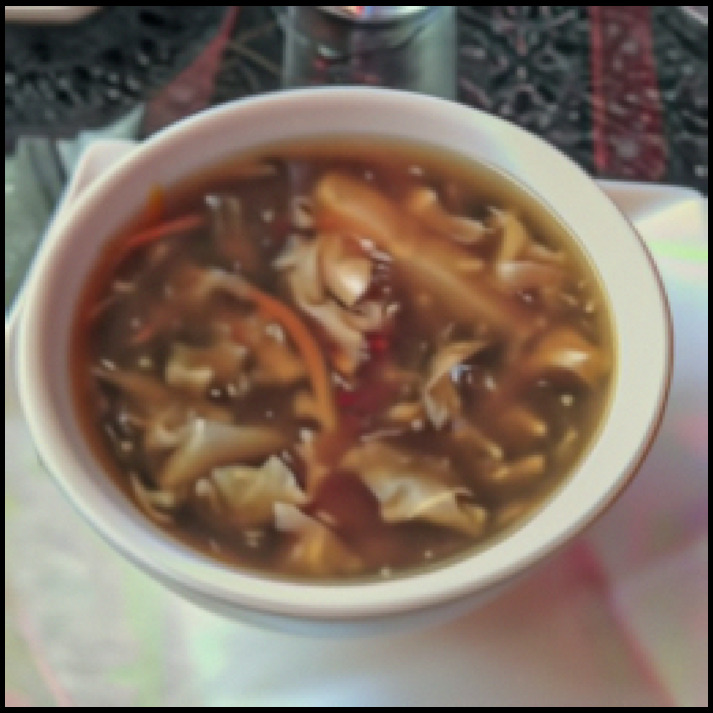} \\
        \textbf{SD} / 0.63 / 0.31 & \textbf{Imagen} / 0.71 / 0.32 & \textbf{Flux} / 0.66 / 0.30 & \textbf{Imagen-3} / 0.60 / 0.30 \\
        \includegraphics[width=0.23\textwidth, height=2.8cm]{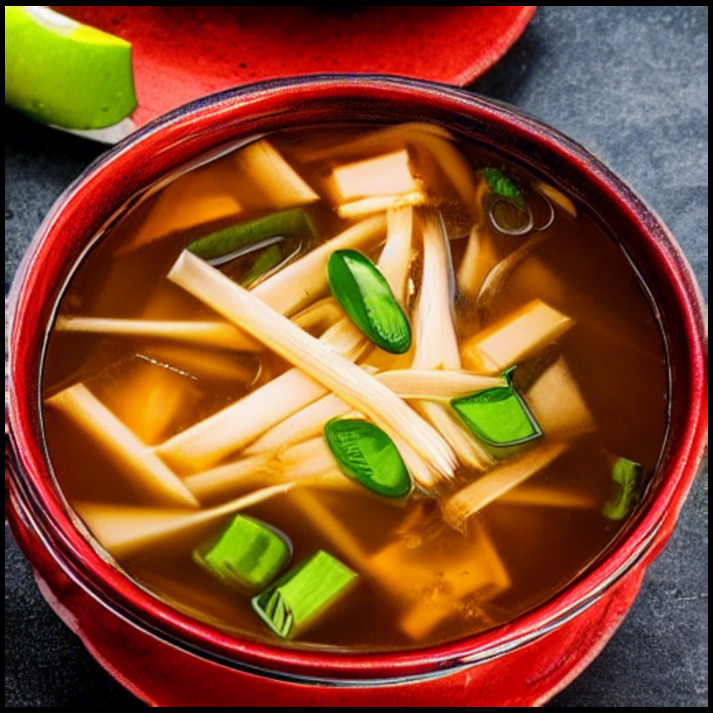} & 
        \includegraphics[width=0.23\textwidth, height=2.8cm]{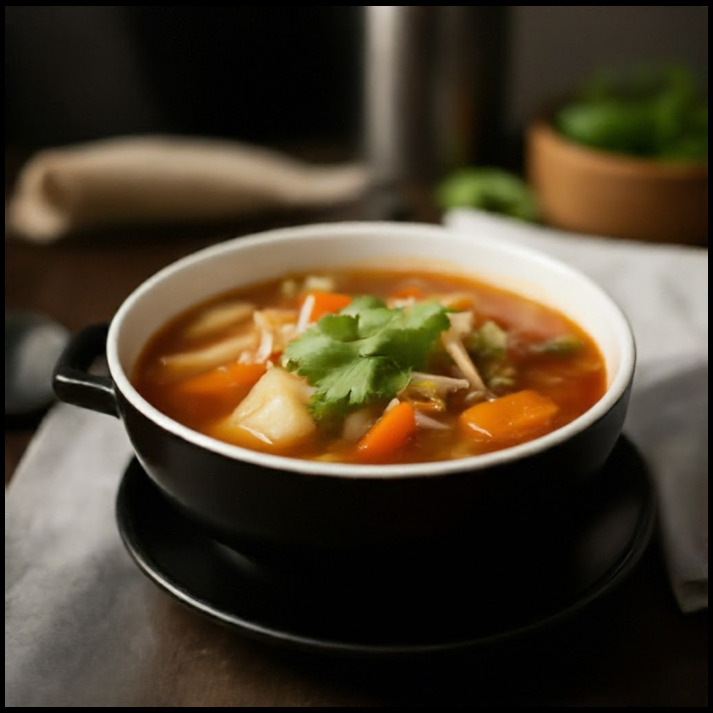} & 
        \includegraphics[width=0.23\textwidth, height=2.8cm]{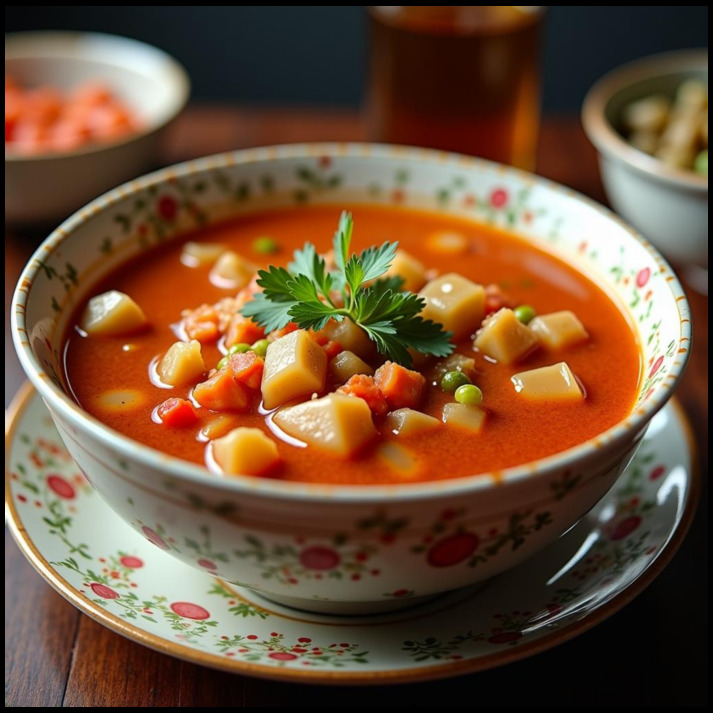} & 
        \includegraphics[width=0.23\textwidth, height=2.8cm]{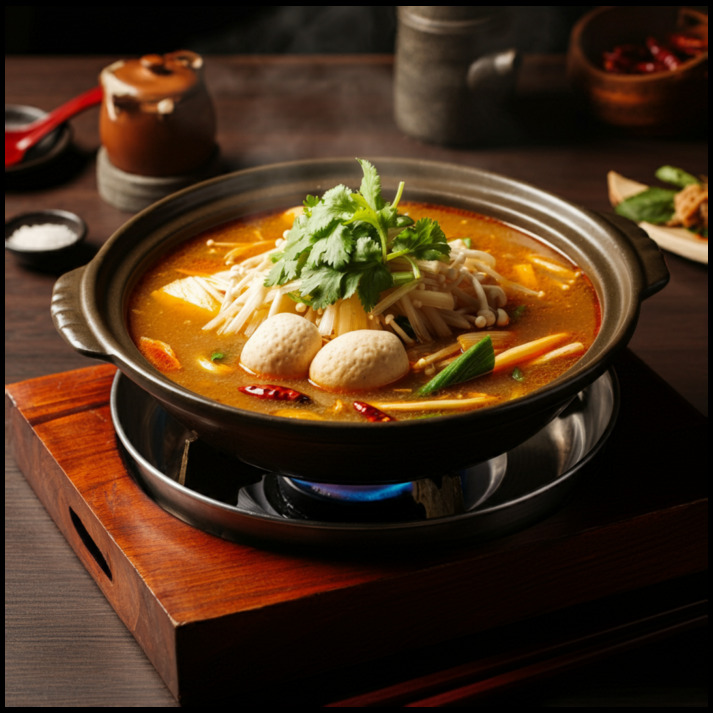} \\
        \multicolumn{4}{c}{\textit{photo of a Hot and sour soup.}} \\
        \bottomrule
    \end{tabular}
    \caption{\textbf{Qualitative results} for the cuisine domain, including the DINO and CLIP-T scores.}
    \label{fig:image_comparison_3}
\end{figure}

\begin{figure}[t]
    \centering
    \scriptsize
    \setlength{\tabcolsep}{2pt} %
    \begin{tabular}{c@{\;\;}c@{\;\;}c@{\;\;}c}
        \toprule %
\textbf{Real Photo} & \textbf{Custom-Diff} / 0.38 / 0.35 & \textbf{DreamBooth} / 0.56 / 0.36 & \textbf{Instruct-Imagen} / 0.53 / 0.40 \\
\includegraphics[width=0.23\textwidth, height=2.8cm]{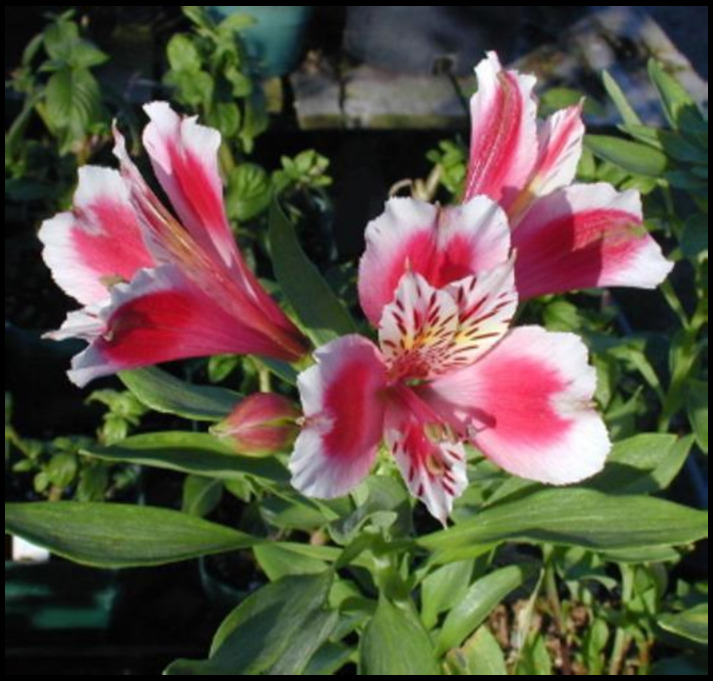} & 
\includegraphics[width=0.23\textwidth, height=2.8cm]{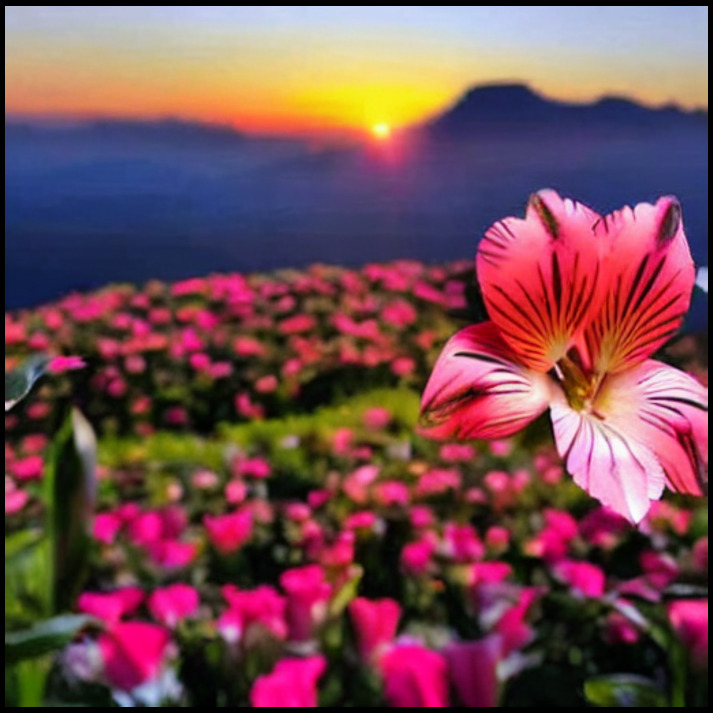} & 
\includegraphics[width=0.23\textwidth, height=2.8cm]{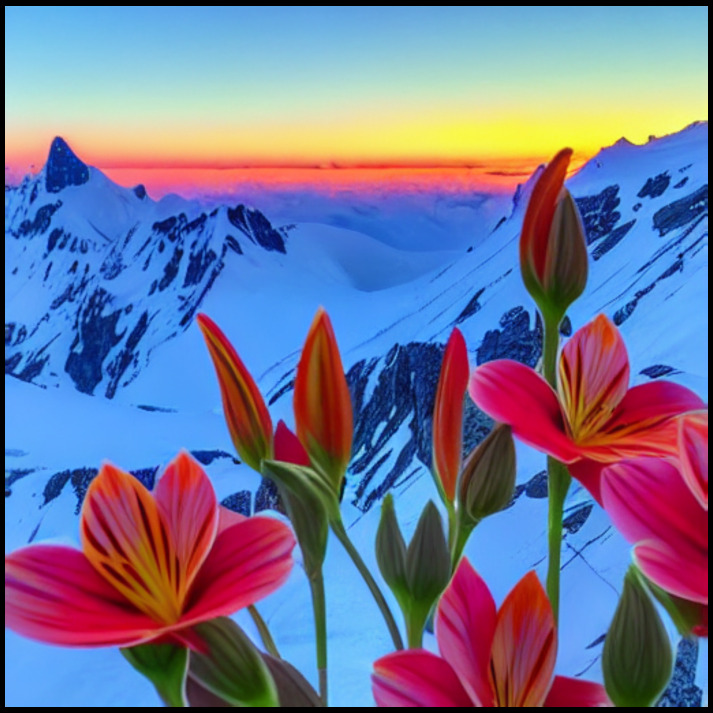} & 
\includegraphics[width=0.23\textwidth, height=2.8cm]{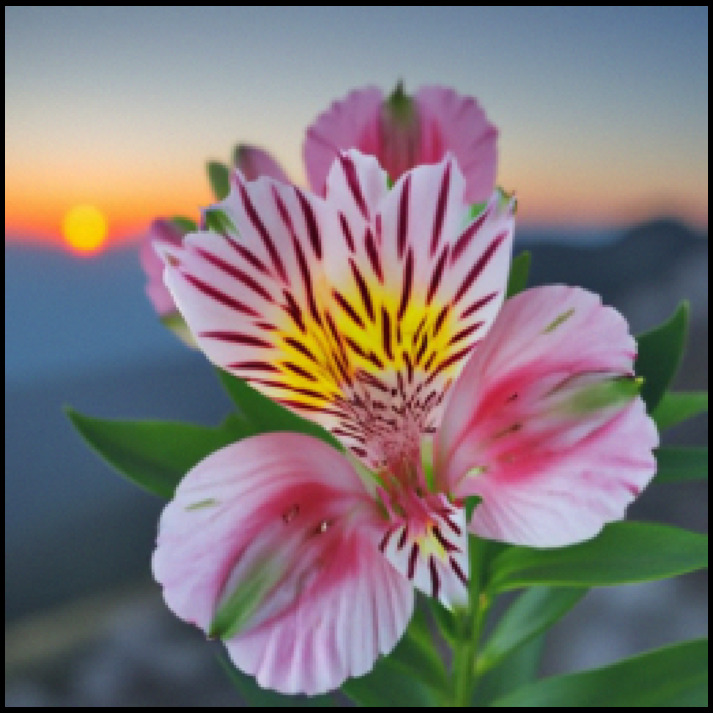} \\
\textbf{SD} / 0.49 / 0.39 & \textbf{Imagen} / 0.29 / 0.37 & \textbf{Flux} / 0.41 / 0.34 & \textbf{Imagen-3} / 0.38 / 0.36 \\
\includegraphics[width=0.23\textwidth, height=2.8cm]{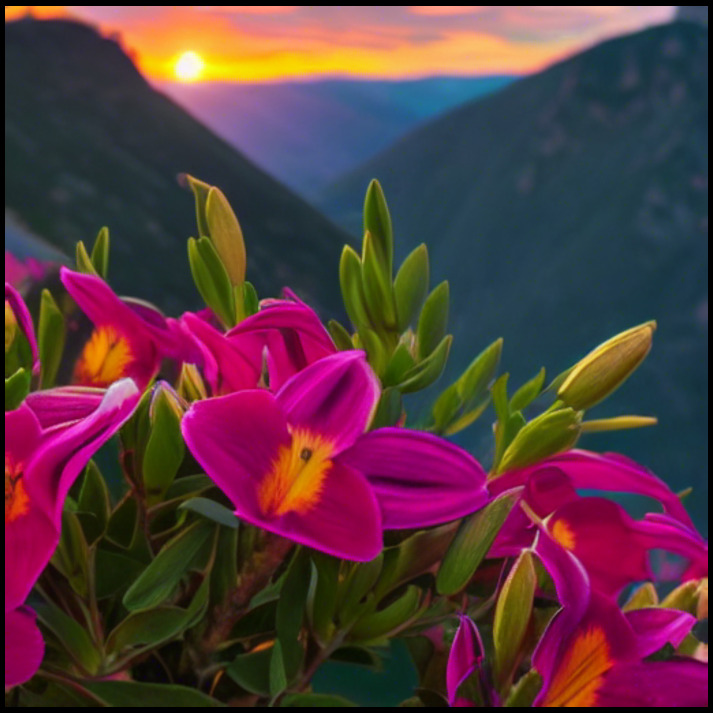} & 
\includegraphics[width=0.23\textwidth, height=2.8cm]{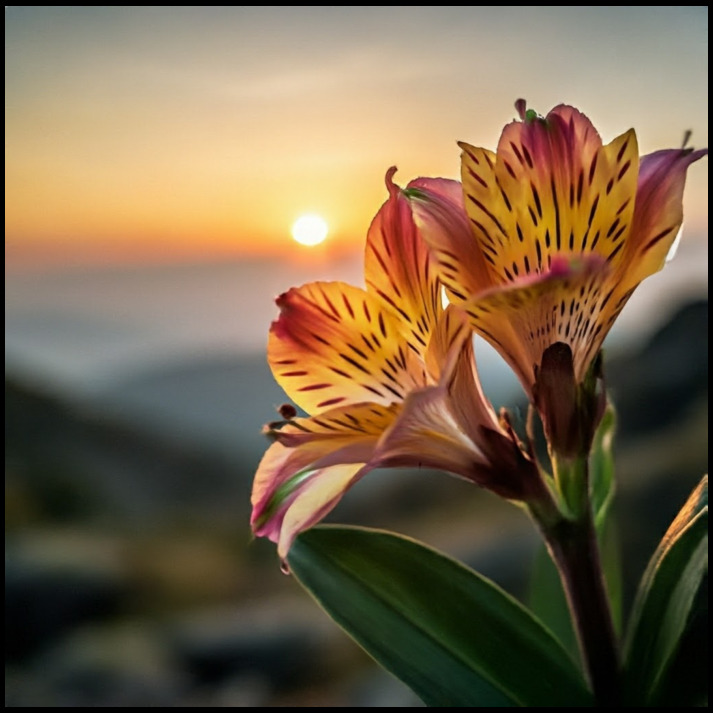} & 
\includegraphics[width=0.23\textwidth, height=2.8cm]{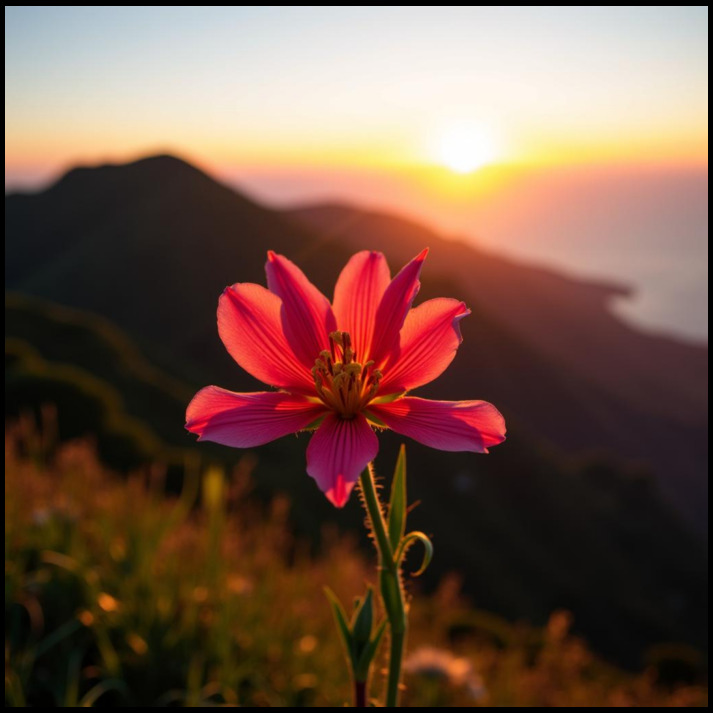} & 
\includegraphics[width=0.23\textwidth, height=2.8cm]{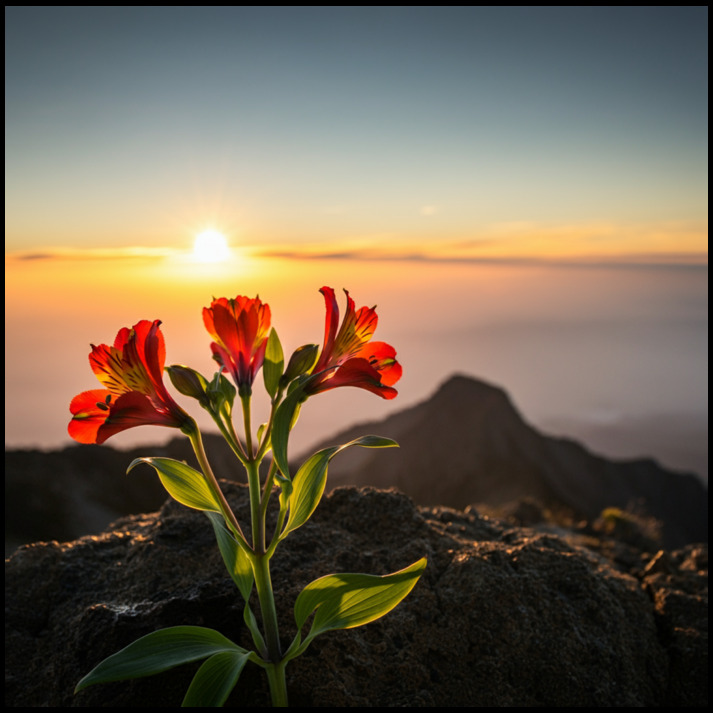} \\

        \multicolumn{4}{c}{\textit{Alstroemeria on top of a mountain with sunrise in the background.}} \\
        \midrule
\textbf{Real Photo} & \textbf{Custom-Diff} / 0.53 / 0.37 & \textbf{DreamBooth} / 0.48 / 0.37 & \textbf{Instruct-Imagen} / 0.65 / 0.41 \\
\includegraphics[width=0.23\textwidth, height=2.8cm]{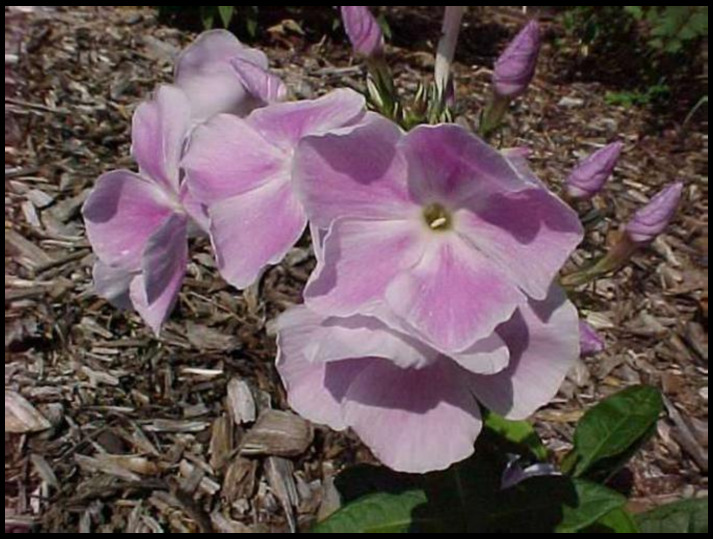} & 
\includegraphics[width=0.23\textwidth, height=2.8cm]{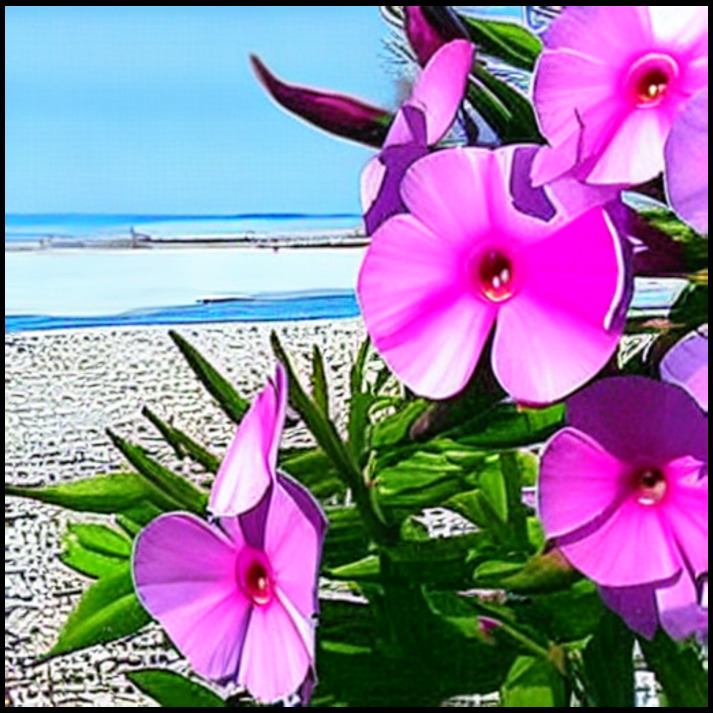} & 
\includegraphics[width=0.23\textwidth, height=2.8cm]{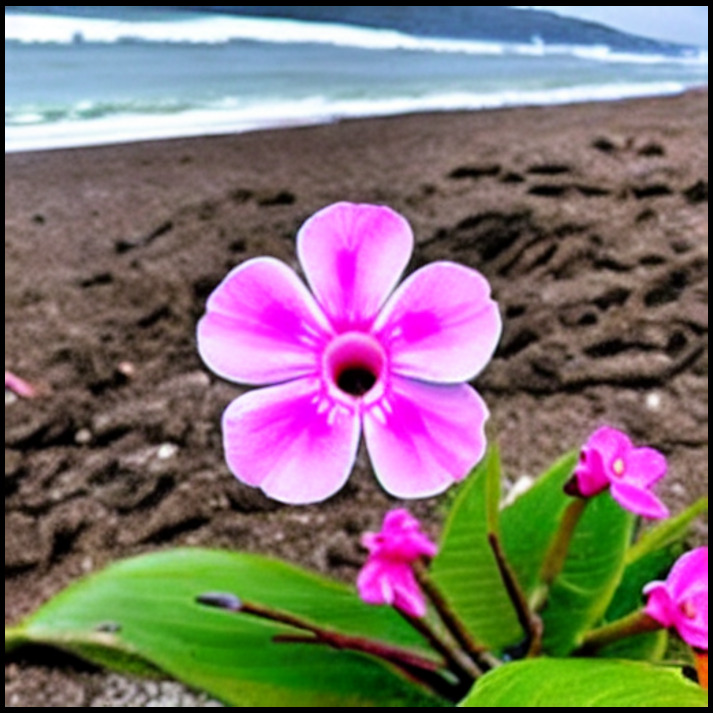} & 
\includegraphics[width=0.23\textwidth, height=2.8cm]{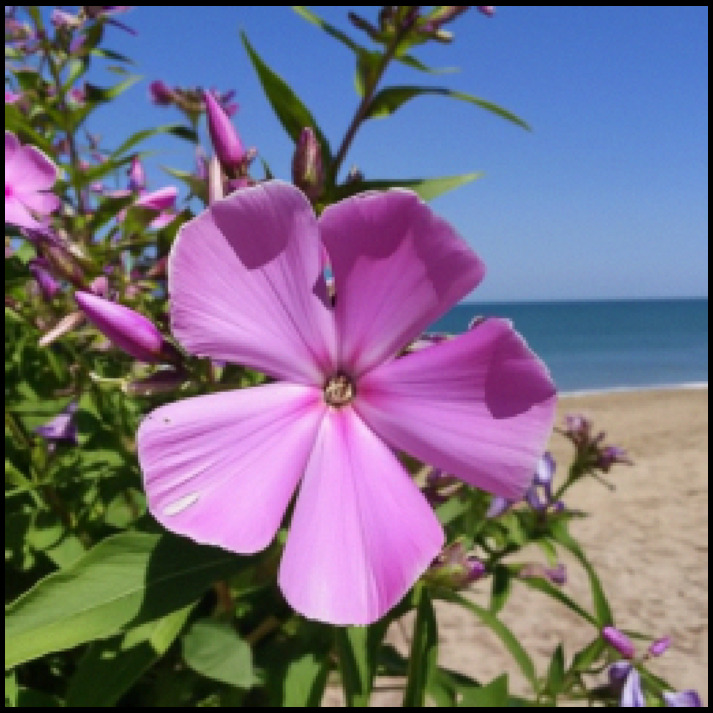} \\
\textbf{SD} / 0.30 / 0.38 & \textbf{Imagen} / 0.49 / 0.40 & \textbf{Flux} / 0.58 / 0.39 & \textbf{Imagen-3} / 0.48 / 0.41 \\
\includegraphics[width=0.23\textwidth, height=2.8cm]{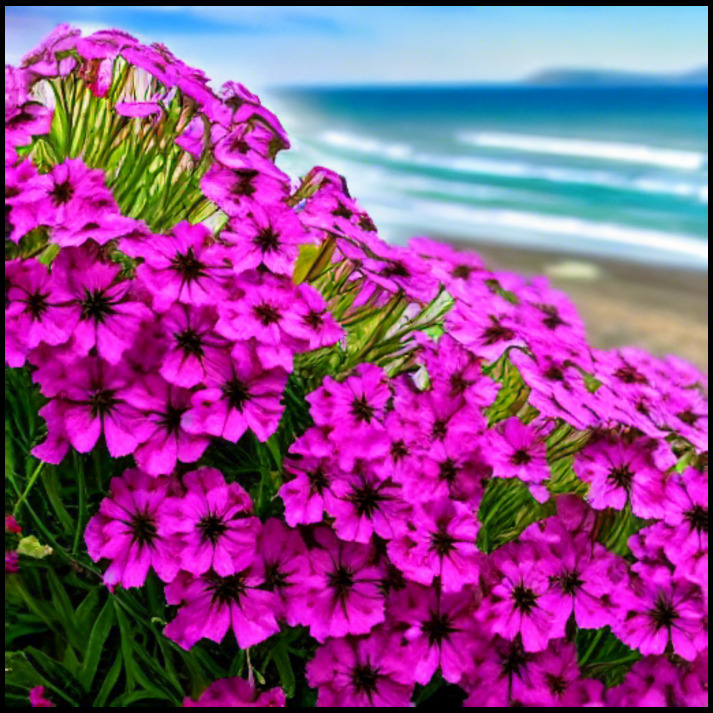} & 
\includegraphics[width=0.23\textwidth, height=2.8cm]{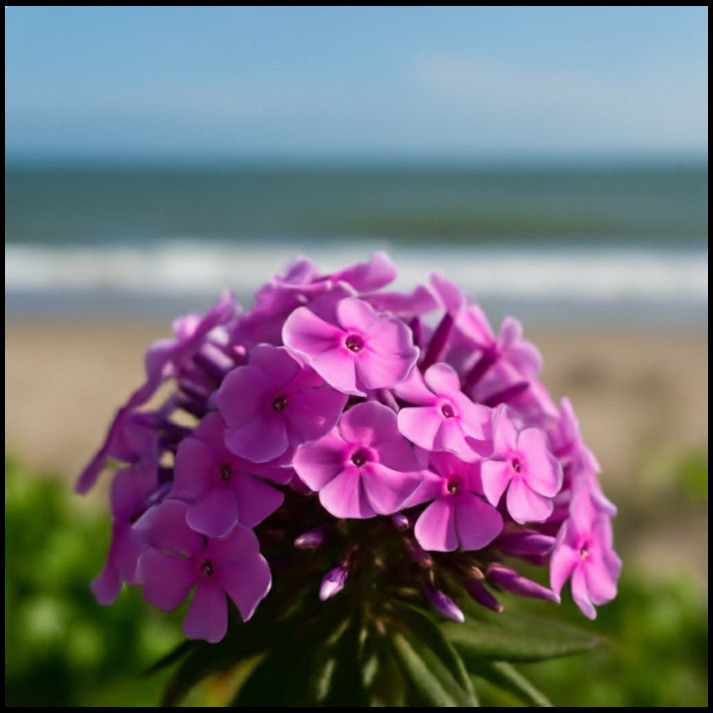} & 
\includegraphics[width=0.23\textwidth, height=2.8cm]{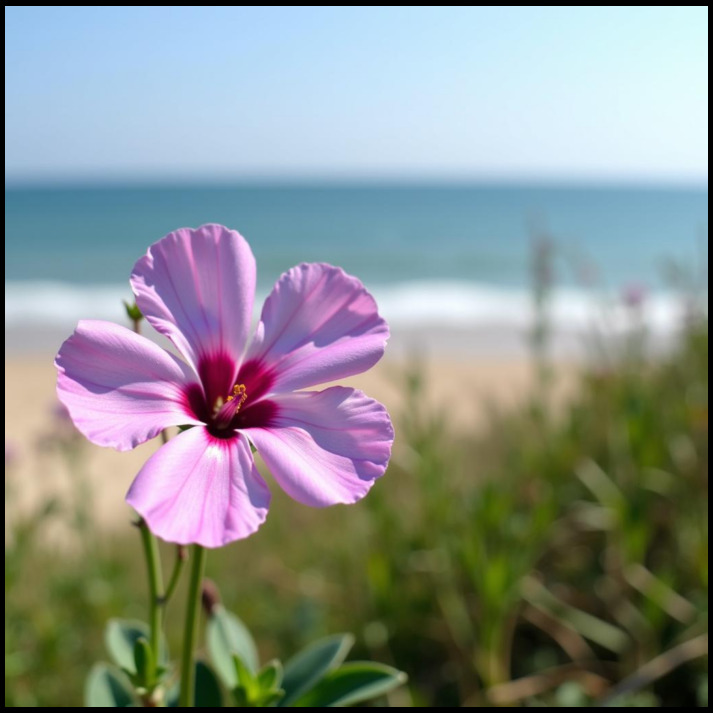} & 
\includegraphics[width=0.23\textwidth, height=2.8cm]{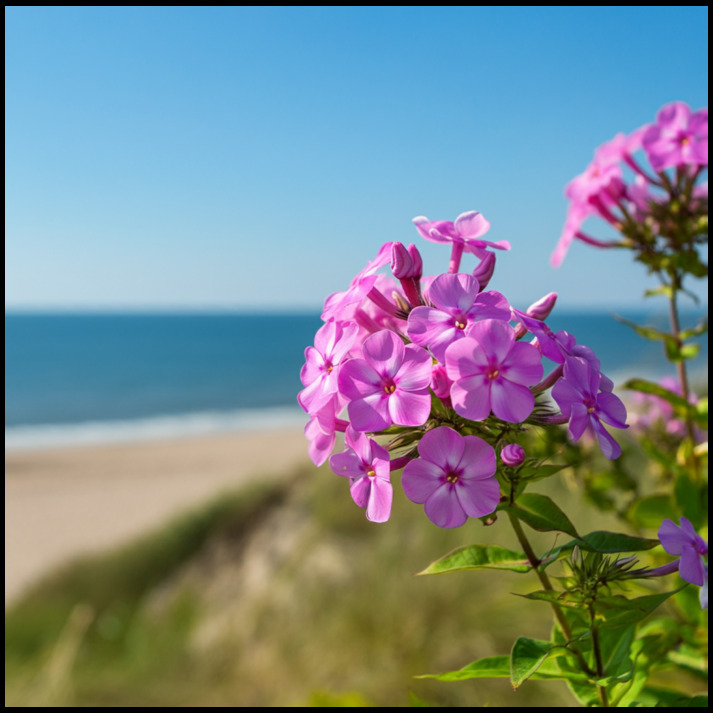} \\

        \multicolumn{4}{c}{\textit{Phlox paniculata at a beach with a view of the seashore.}} \\
        \midrule
\textbf{Real Photo} & \textbf{Custom-Diff} / 0.65 / 0.38 & \textbf{DreamBooth} / 0.69 / 0.37 & \textbf{Imagen} / 0.66 / 0.34 \\
\includegraphics[width=0.23\textwidth, height=2.8cm]{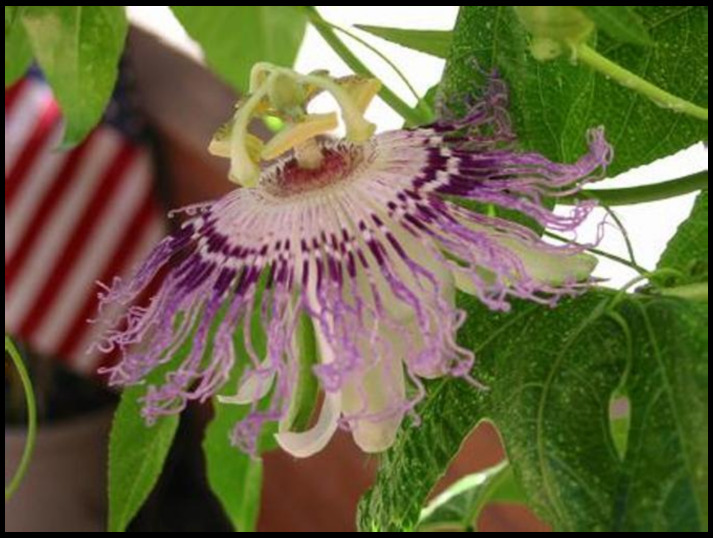} & 
\includegraphics[width=0.23\textwidth, height=2.8cm]{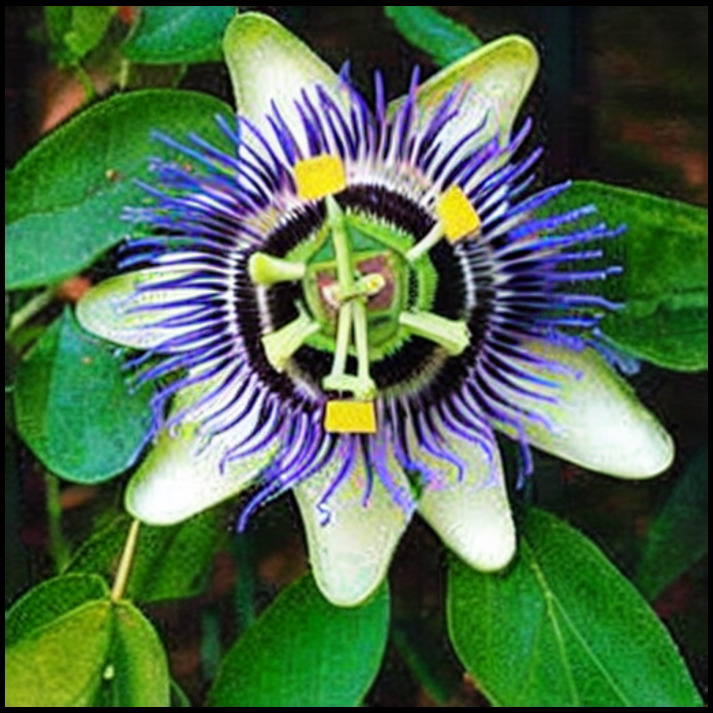} & 
\includegraphics[width=0.23\textwidth, height=2.8cm]{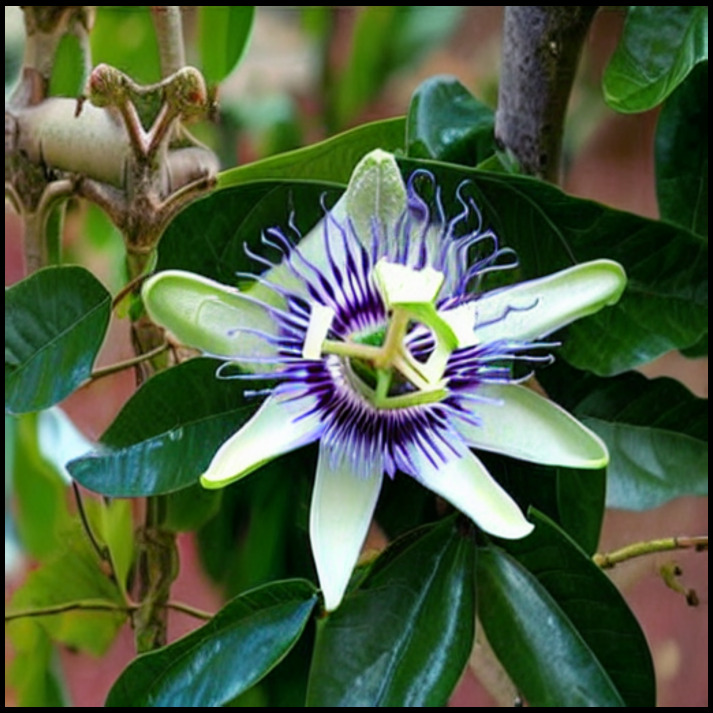} & 
\includegraphics[width=0.23\textwidth, height=2.8cm]{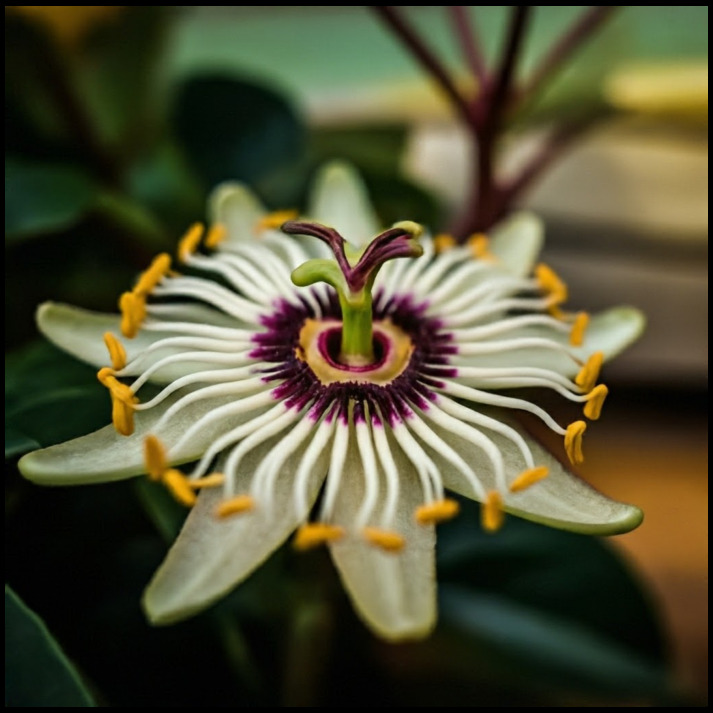} \\
\textbf{SD} / 0.56 / 0.36 & \textbf{Instruct-Imagen} / 0.70 / 0.33 & \textbf{Flux} / 0.65 / 0.36 & \textbf{Imagen-3} / 0.71 / 0.36 \\
\includegraphics[width=0.23\textwidth, height=2.8cm]{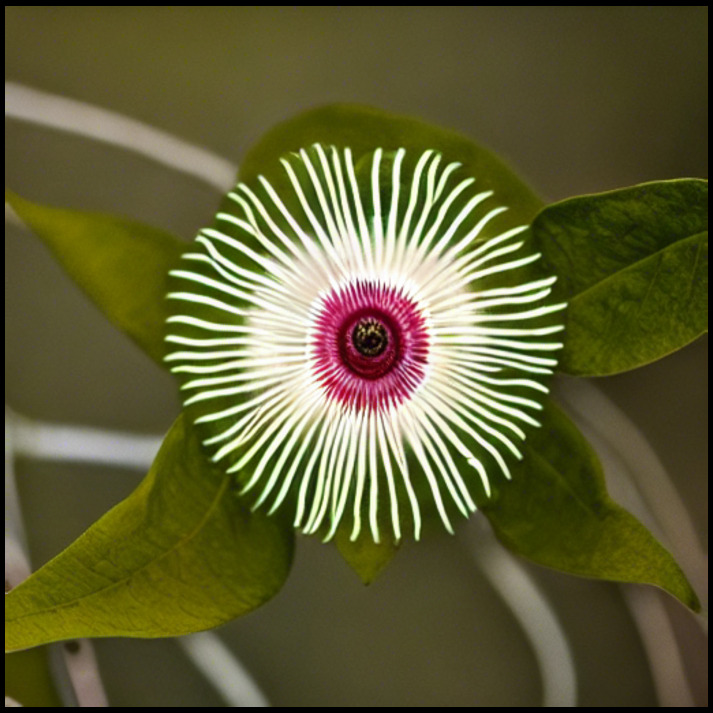} & 
\includegraphics[width=0.23\textwidth, height=2.8cm]{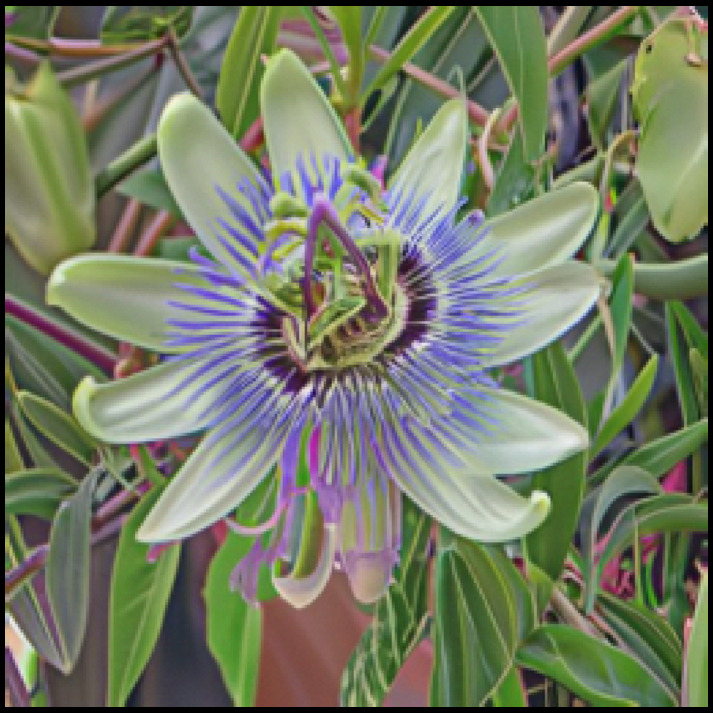} & 
\includegraphics[width=0.23\textwidth, height=2.8cm]{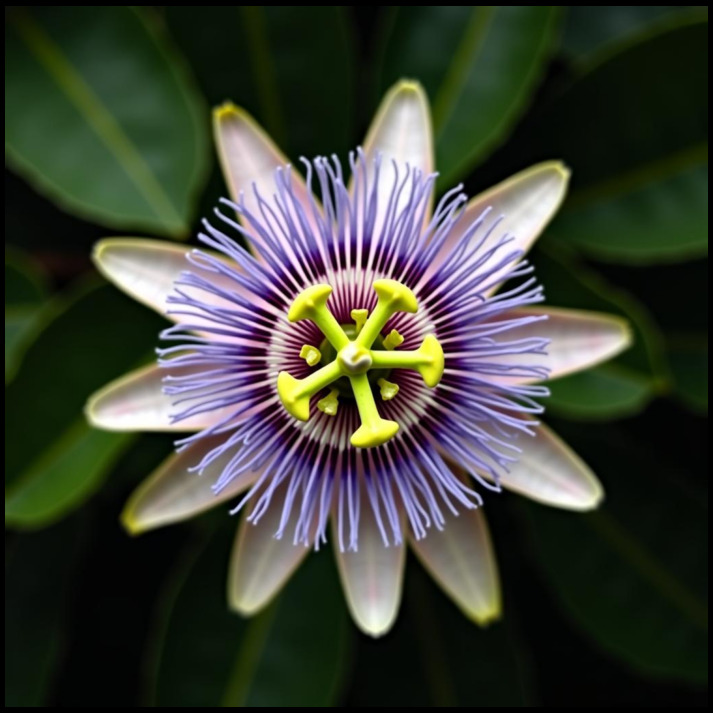} & 
\includegraphics[width=0.23\textwidth, height=2.8cm]{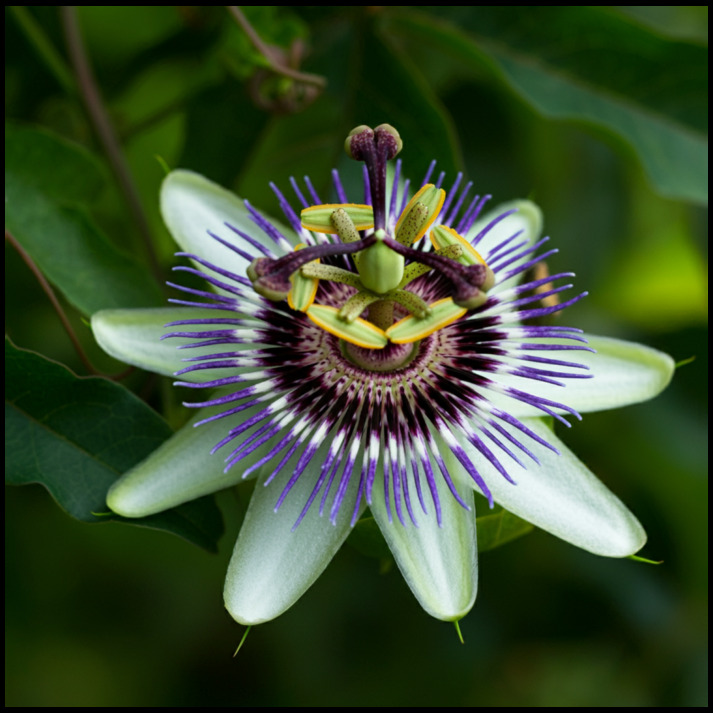} \\

        \multicolumn{4}{c}{\textit{Photo of a Passiflora flower.}} \\
        \bottomrule
    \end{tabular}
    \caption{\textbf{Qualitative results} for the flower domain, including the DINO and CLIP-T scores.}
    \label{fig:image_comparison_4}
\end{figure}

\begin{figure}[t]
    \centering
    \scriptsize
    \setlength{\tabcolsep}{2pt} %
    \begin{tabular}{c@{\;\;}c@{\;\;}c@{\;\;}c}
        \toprule %
        \textbf{Real Photo} & \textbf{Custom-Diff} / 0.36 / 0.32 & \textbf{DreamBooth} / 0.42 / 0.37 & \textbf{Instruct-Imagen} / 0.78 / 0.36 \\
        \includegraphics[width=0.23\textwidth, height=2.8cm]{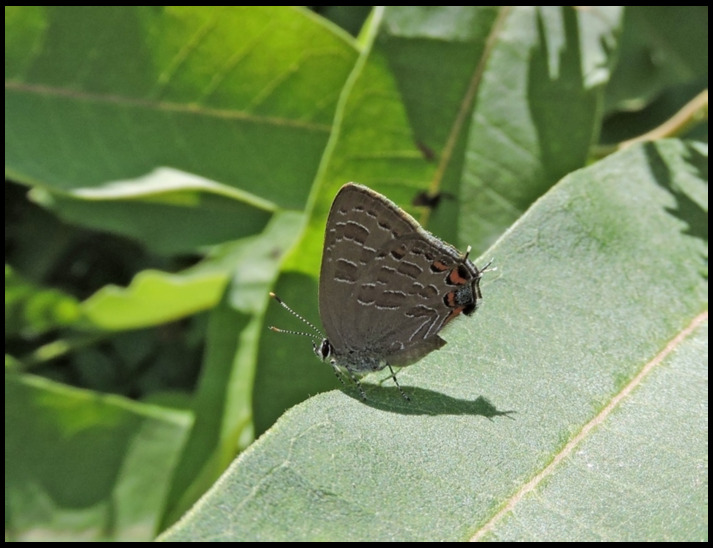} & 
        \includegraphics[width=0.23\textwidth, height=2.8cm]{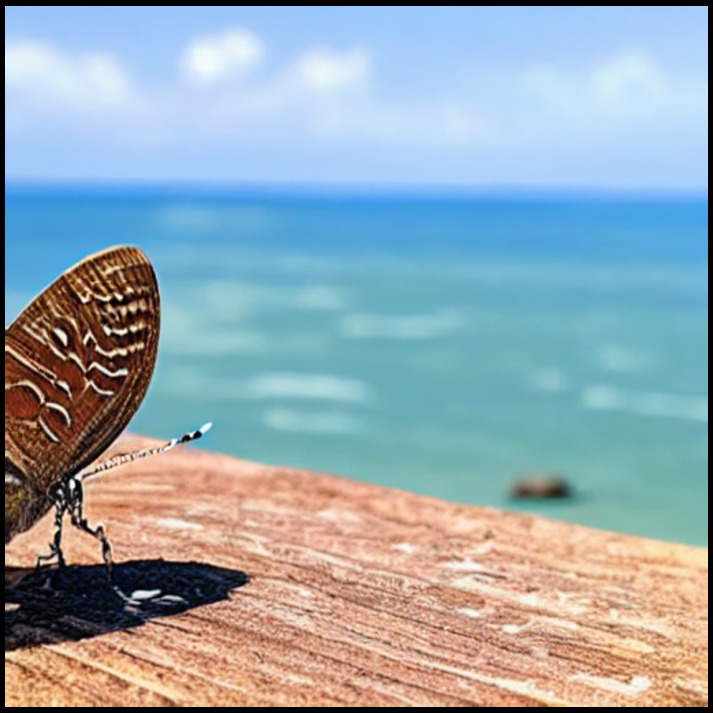} & 
        \includegraphics[width=0.23\textwidth, height=2.8cm]{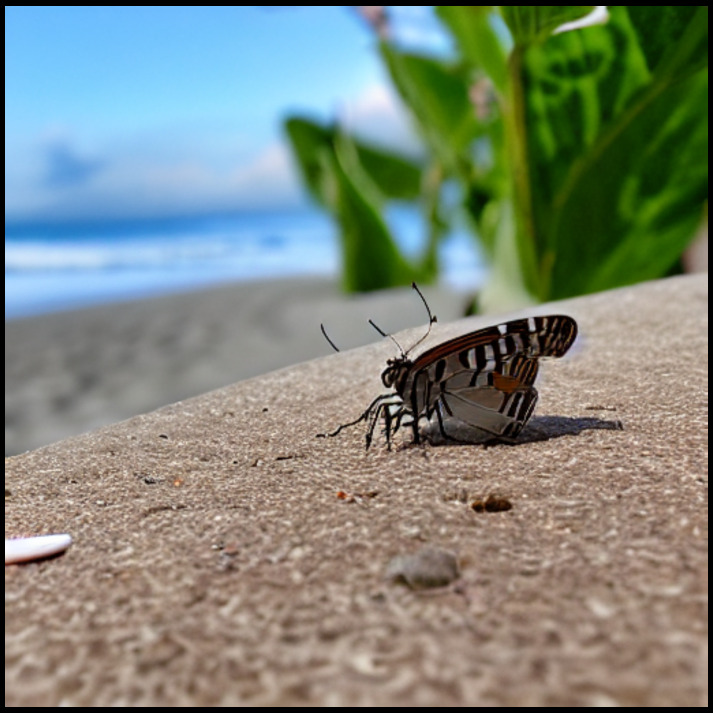} & 
        \includegraphics[width=0.23\textwidth, height=2.8cm]{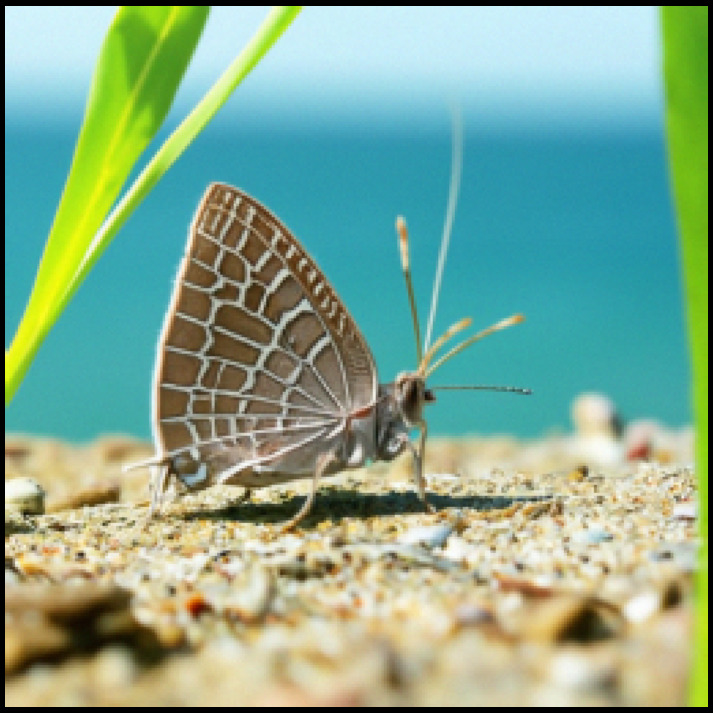} \\
        \textbf{SD} / 0.29 / 0.31 & \textbf{Imagen} / 0.23 / 0.35 & \textbf{Flux} / 0.17 / 0.31 & \textbf{Imagen-3} / 0.18 / 0.34 \\
        \includegraphics[width=0.23\textwidth, height=2.8cm]{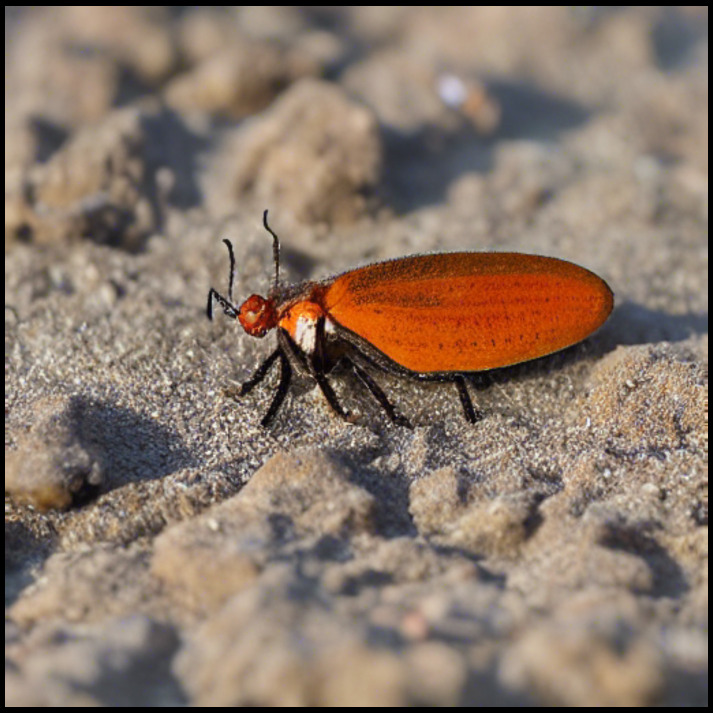} & 
        \includegraphics[width=0.23\textwidth, height=2.8cm]{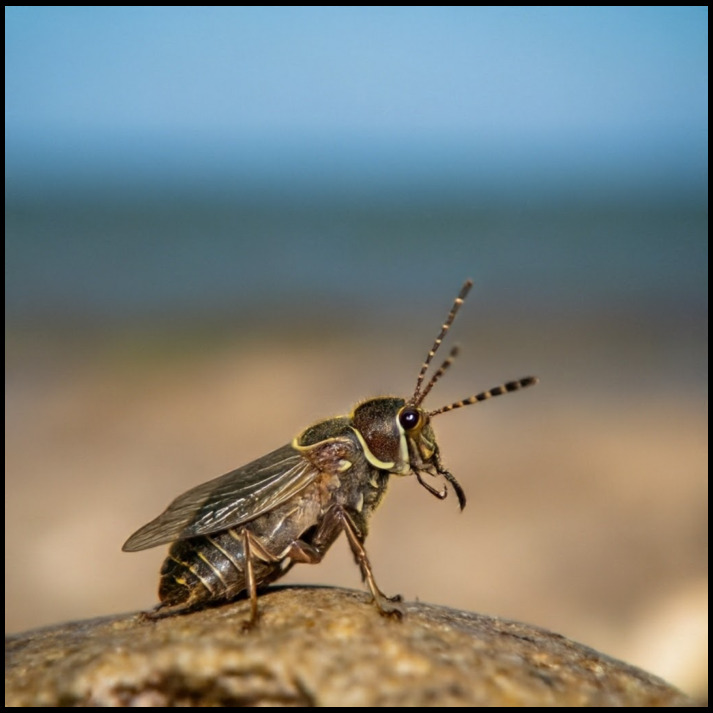} & 
        \includegraphics[width=0.23\textwidth, height=2.8cm]{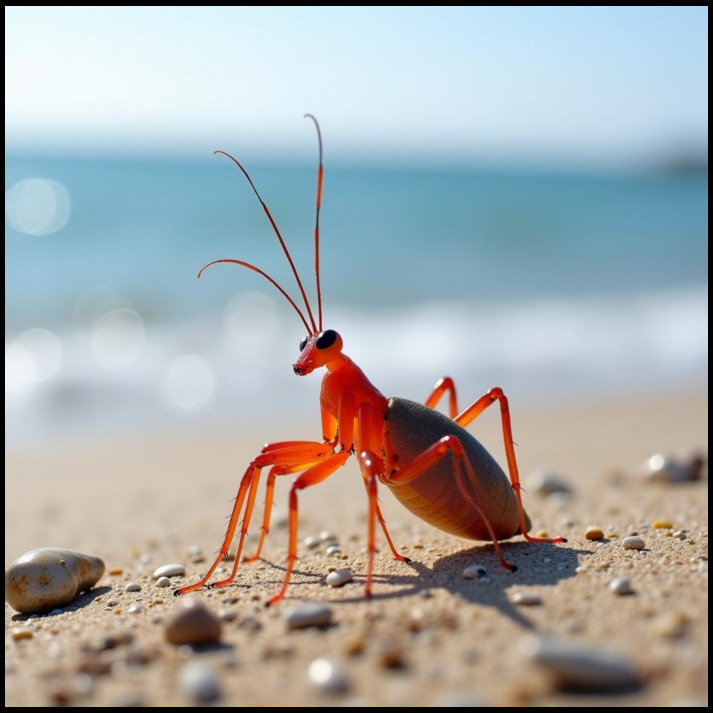} & 
        \includegraphics[width=0.23\textwidth, height=2.8cm]{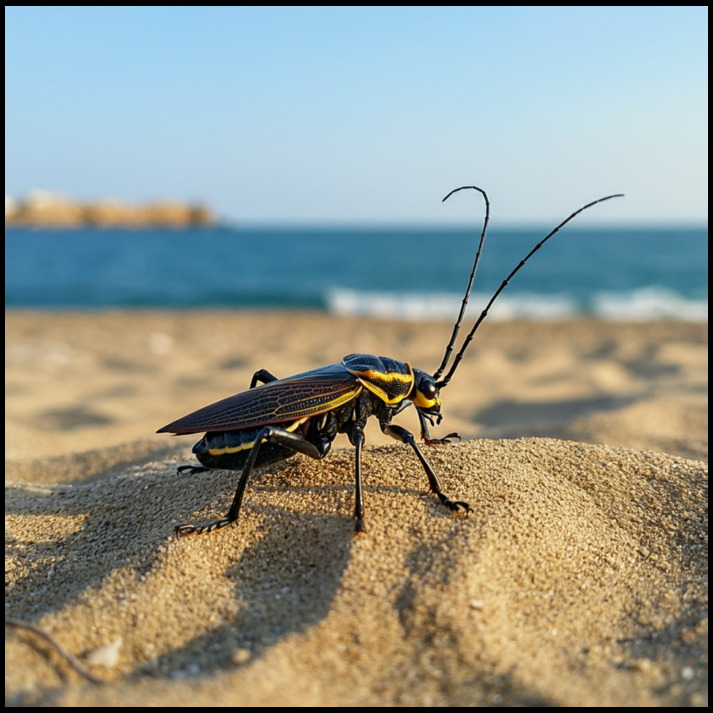} \\
        \multicolumn{4}{c}{\textit{Satyrium liparops sitting at the beach with a view of the sea.}} \\
        \midrule
        \textbf{Real Photo} & \textbf{Custom-Diff} / 0.52 / 0.30 & \textbf{DreamBooth} / 0.34 / 0.31 & \textbf{Instruct-Imagen} / 0.65 / 0.29 \\
        \includegraphics[width=0.23\textwidth, height=2.8cm]{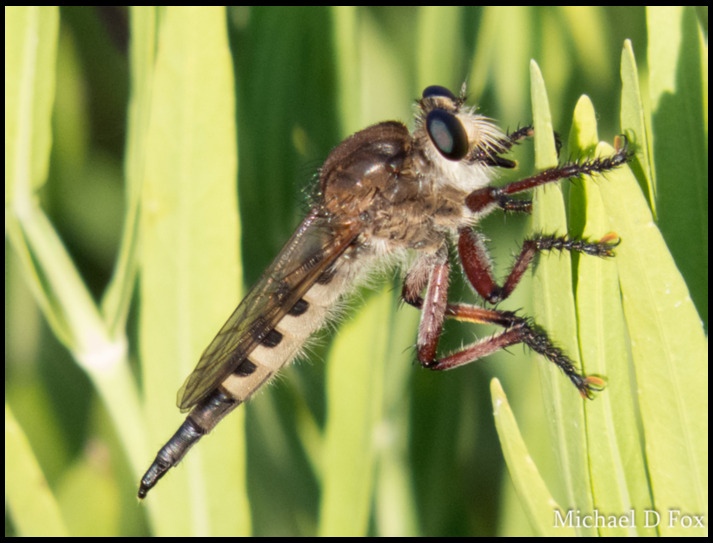} & 
        \includegraphics[width=0.23\textwidth, height=2.8cm]{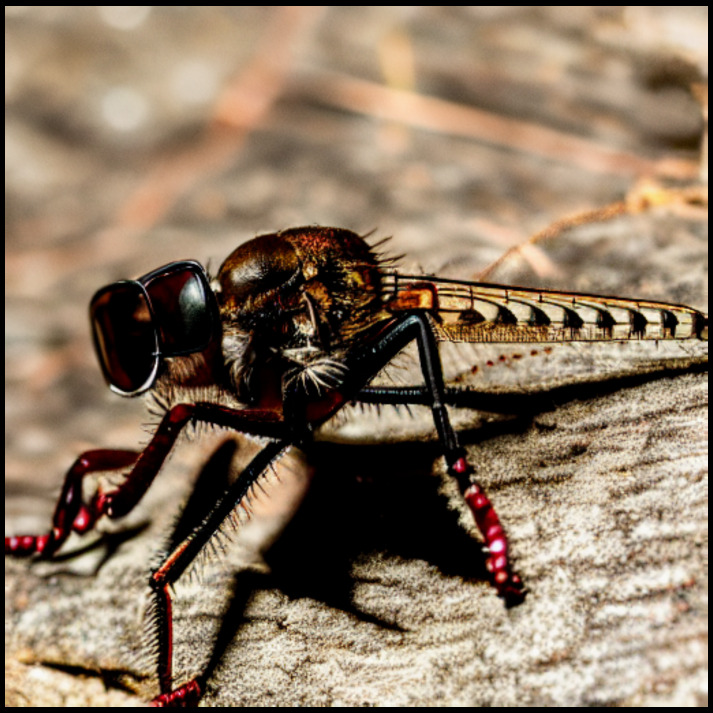} & 
        \includegraphics[width=0.23\textwidth, height=2.8cm]{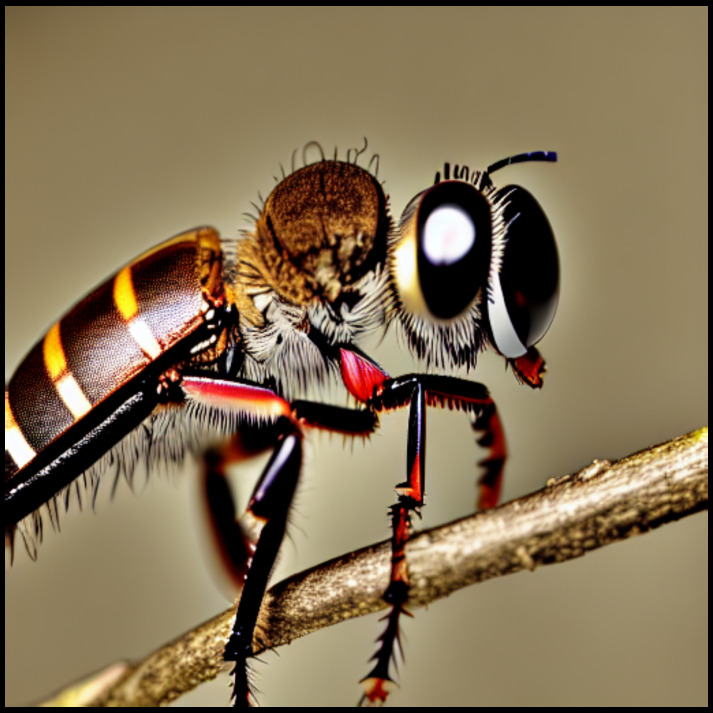} & 
        \includegraphics[width=0.23\textwidth, height=2.8cm]{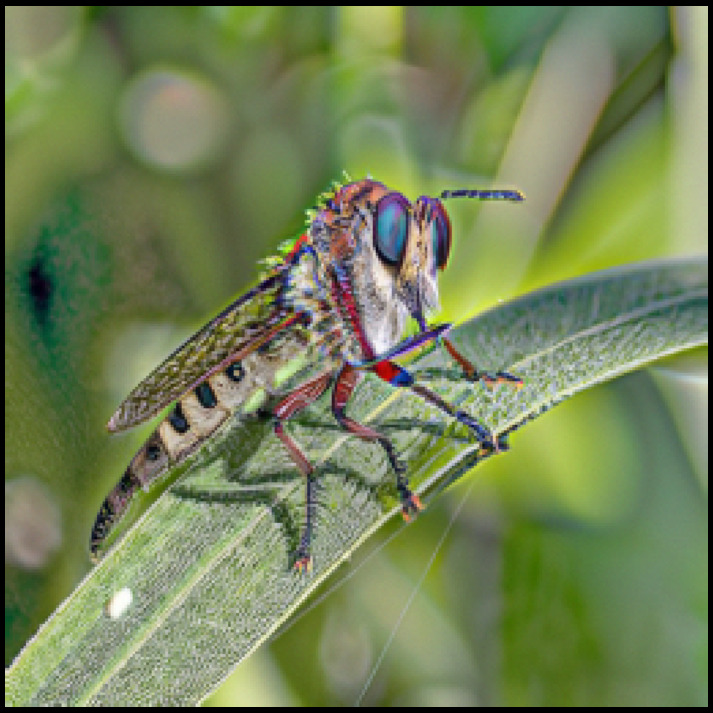} \\
        \textbf{SD} / 0.28 / 0.30 & \textbf{Imagen} / 0.35 / 0.34 & \textbf{Flux} / 0.50 / 0.31 & \textbf{Imagen-3} / 0.58 / 0.31 \\
        \includegraphics[width=0.23\textwidth, height=2.8cm]{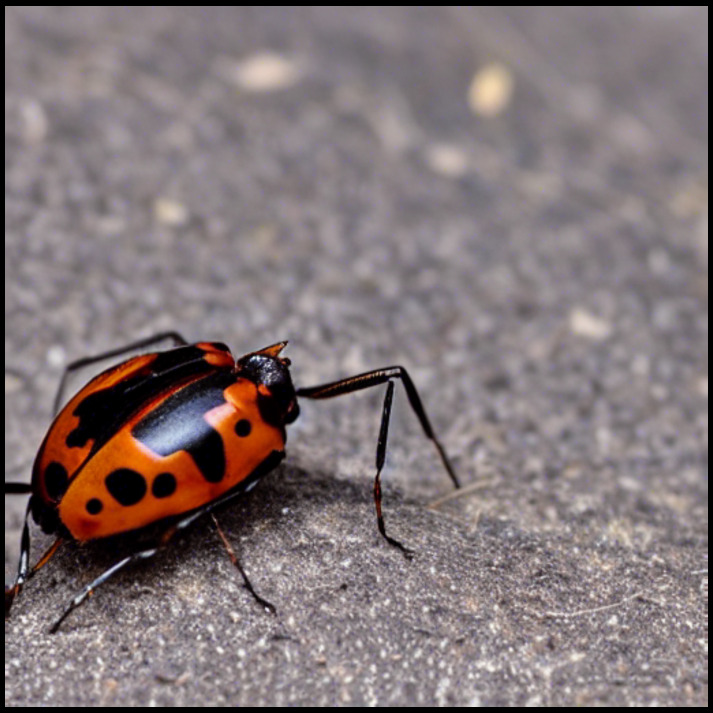} & 
        \includegraphics[width=0.23\textwidth, height=2.8cm]{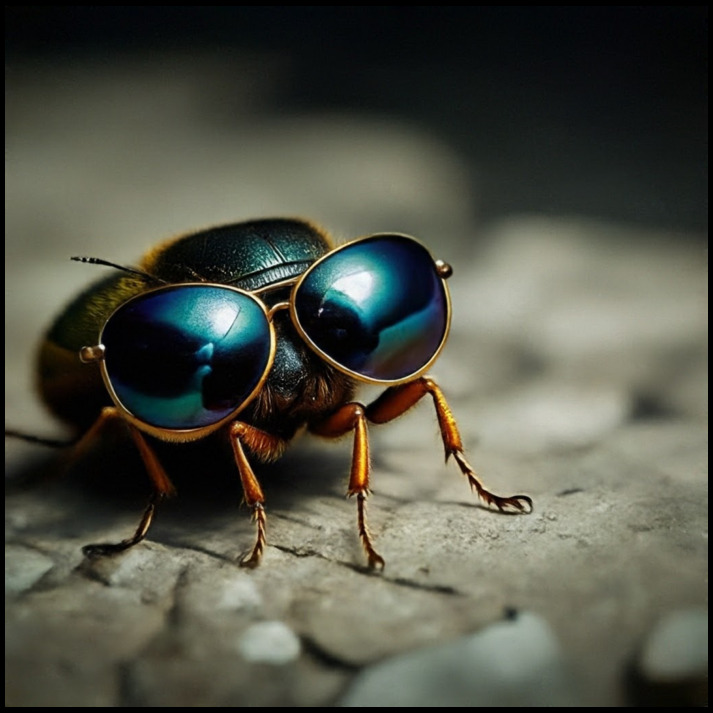} & 
        \includegraphics[width=0.23\textwidth, height=2.8cm]{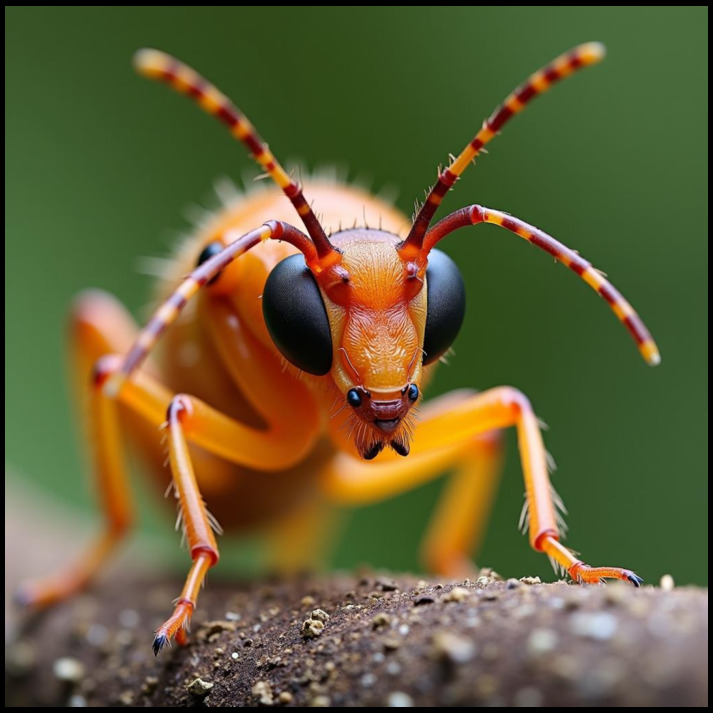} & 
        \includegraphics[width=0.23\textwidth, height=2.8cm]{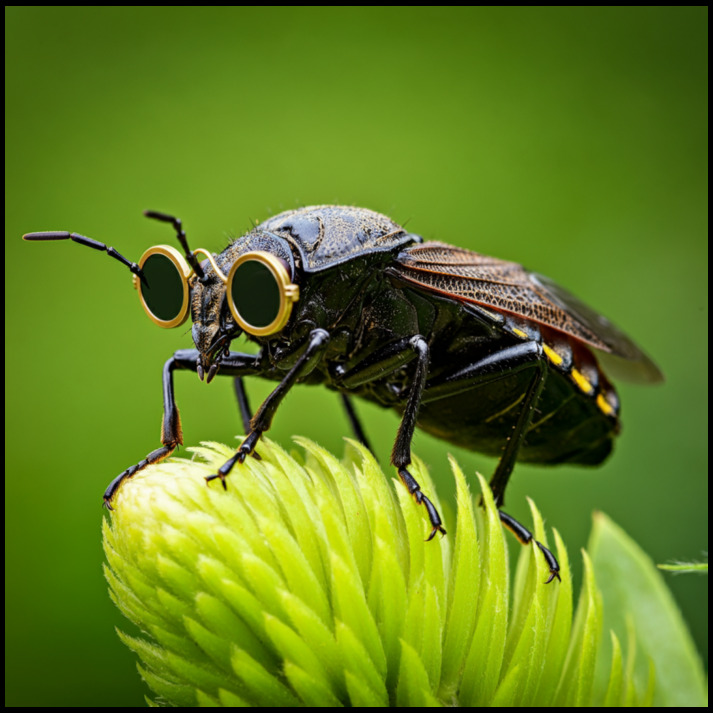} \\
        \multicolumn{4}{c}{\textit{Promachus hinei wearing sunglasses.}} \\
        \midrule
        \textbf{Real Photo} & \textbf{Custom-Diff} / 0.44 / 0.30 & \textbf{DreamBooth} / 0.44 / 0.30 & \textbf{Instruct-Imagen} / 0.48 / 0.30 \\
        \includegraphics[width=0.23\textwidth, height=2.8cm]{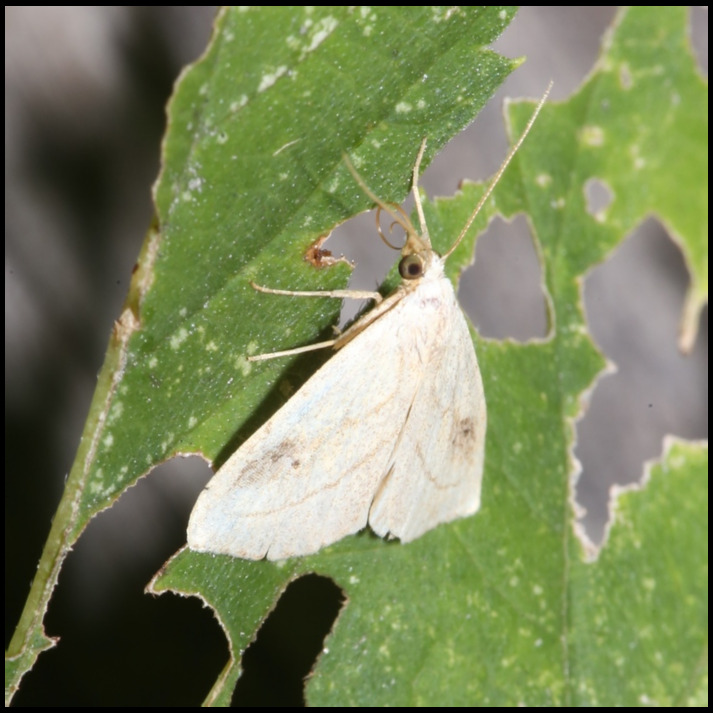} & 
        \includegraphics[width=0.23\textwidth, height=2.8cm]{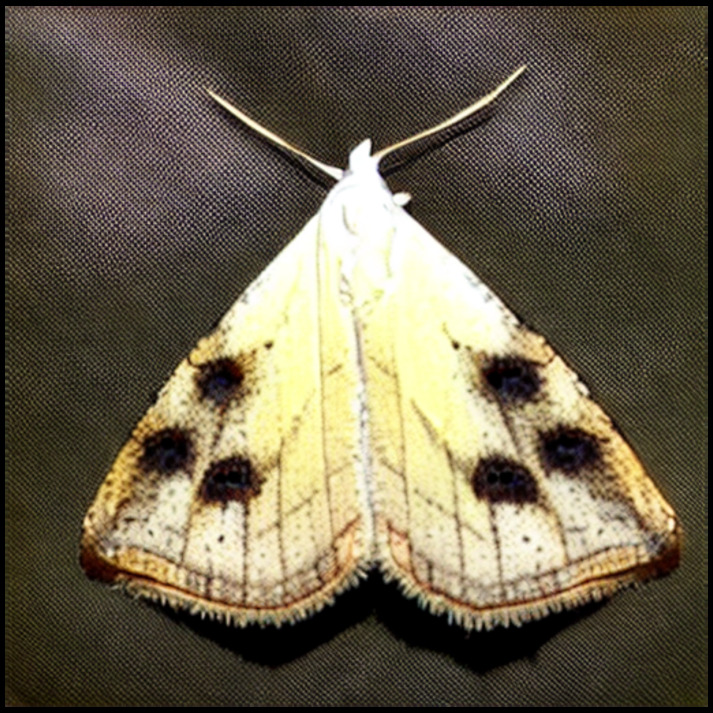} & 
        \includegraphics[width=0.23\textwidth, height=2.8cm]{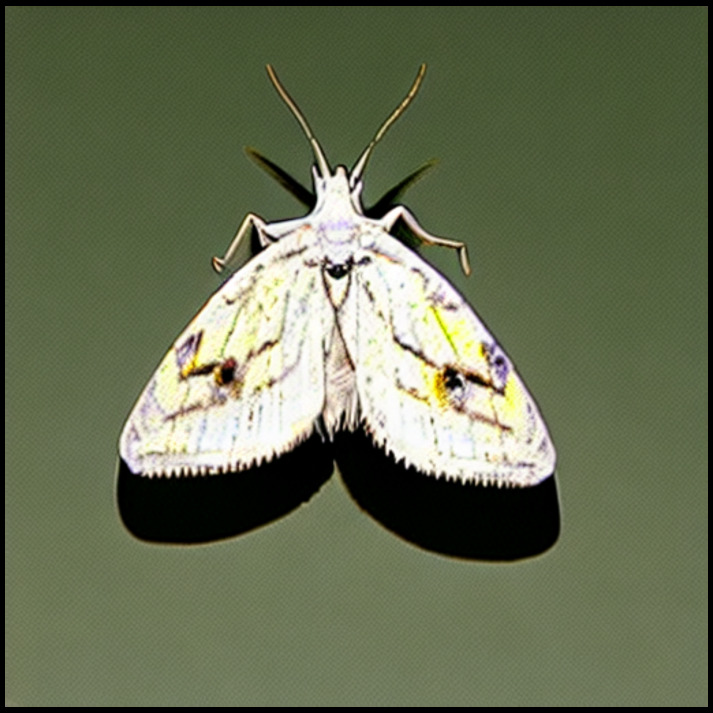} & 
        \includegraphics[width=0.23\textwidth, height=2.8cm]{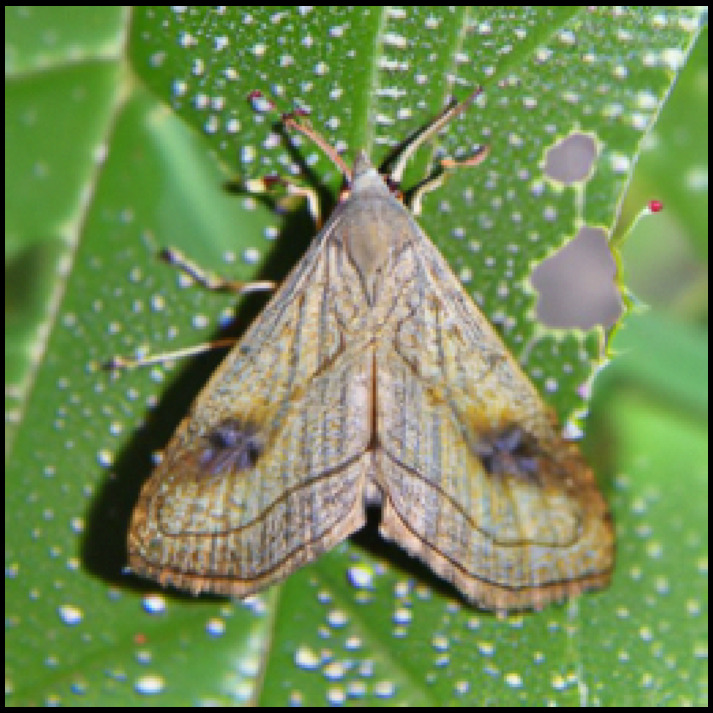} \\
        \textbf{SD} / 0.23 / 0.30 & \textbf{Imagen} / 0.22 / 0.30 & \textbf{Flux} / 0.23 / 0.29 & \textbf{Imagen-3} / 0.23 / 0.30 \\
        \includegraphics[width=0.23\textwidth, height=2.8cm]{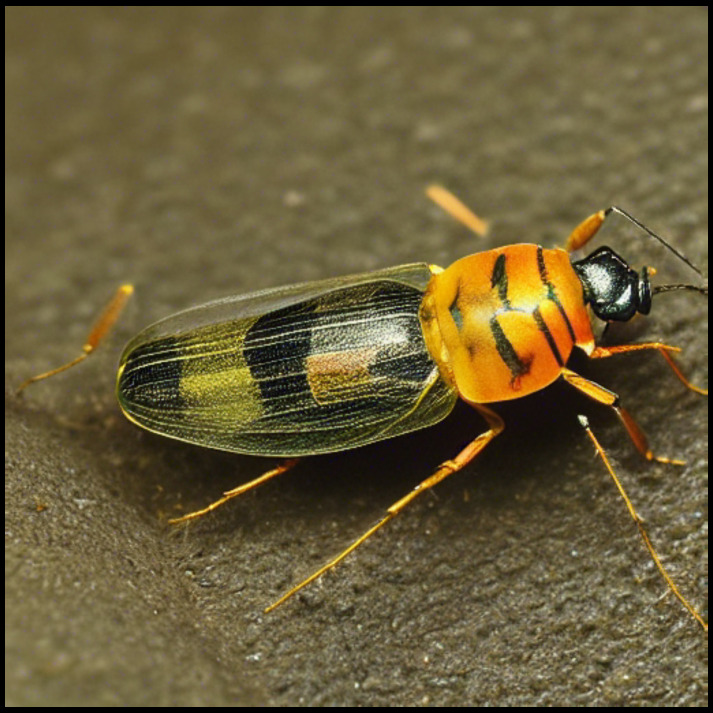} & 
        \includegraphics[width=0.23\textwidth, height=2.8cm]{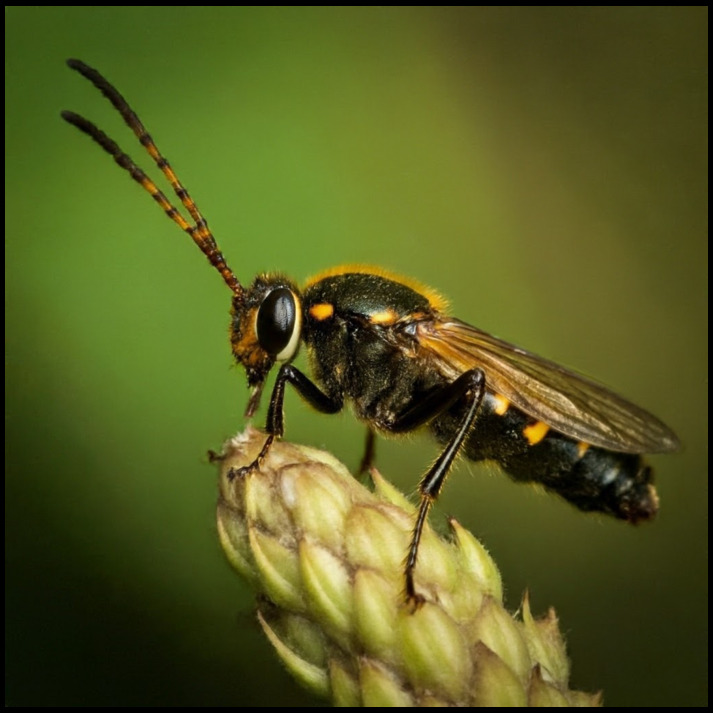} & 
        \includegraphics[width=0.23\textwidth, height=2.8cm]{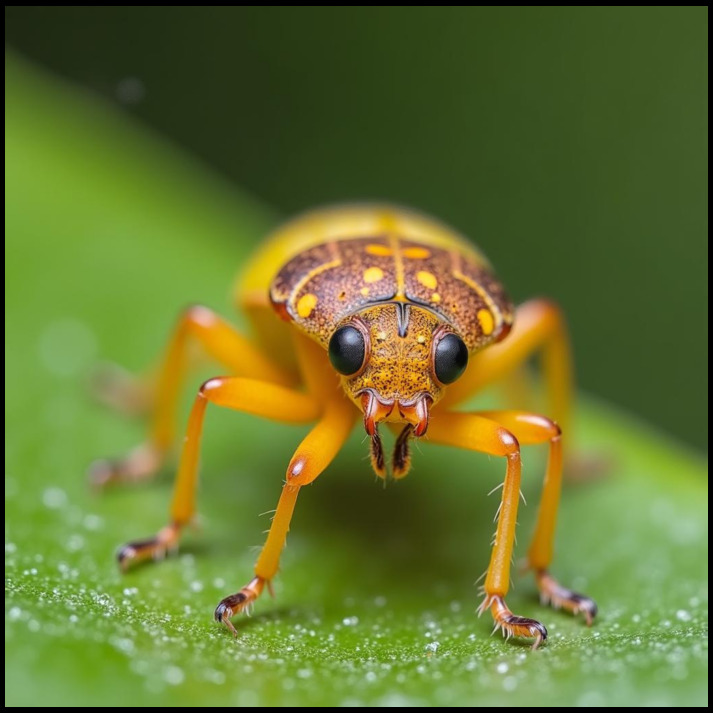} & 
        \includegraphics[width=0.23\textwidth, height=2.8cm]{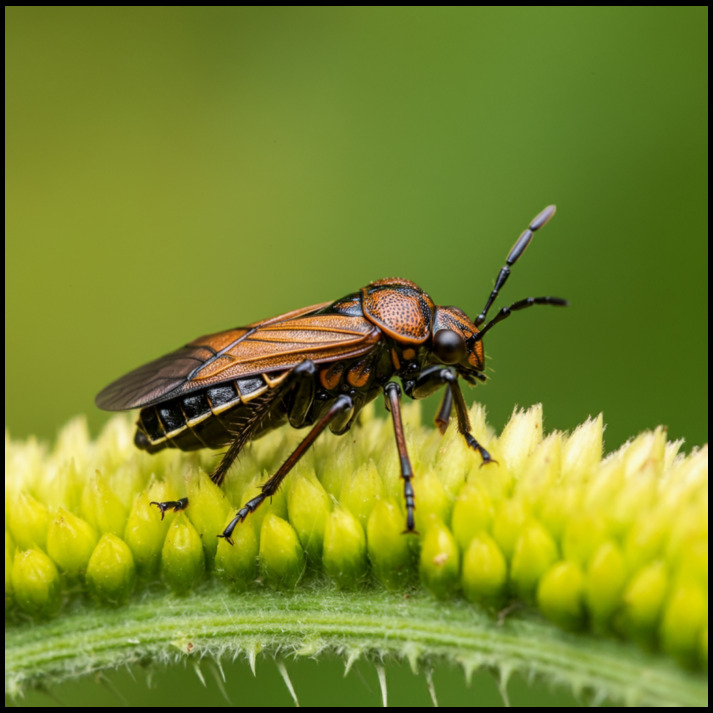} \\
        \multicolumn{4}{c}{\textit{Photo of a Rivula propinqualis.}} \\
        \bottomrule
    \end{tabular}
    \caption{\textbf{Qualitative results} for the insect domain, including the DINO and CLIP-T scores.}
    \label{fig:image_comparison_5}
\end{figure}

\begin{figure}[t]
    \centering
    \scriptsize
    \setlength{\tabcolsep}{2pt} %
    \begin{tabular}{c@{\;\;}c@{\;\;}c@{\;\;}c}
        \toprule %
        \textbf{Real Photo} & \textbf{Custom-Diff} / 0.39 / 0.37 & \textbf{DreamBooth} / 0.33 / 0.38 & \textbf{Instruct-Imagen} / 0.76 / 0.30 \\
        \includegraphics[width=0.23\textwidth, height=2.8cm]{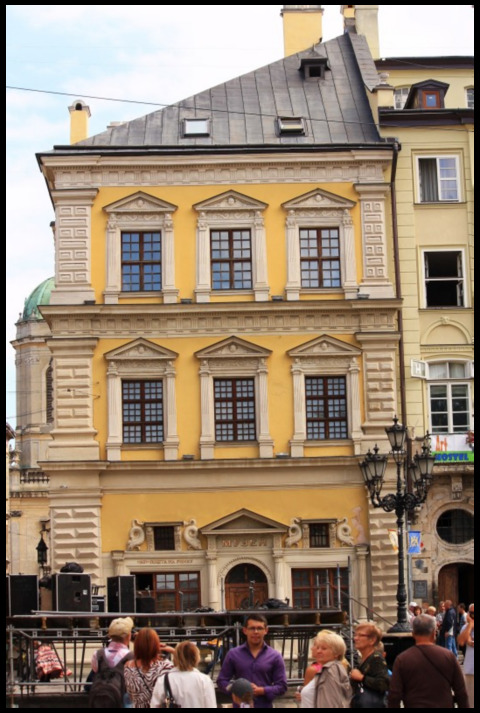} & 
        \includegraphics[width=0.23\textwidth, height=2.8cm]{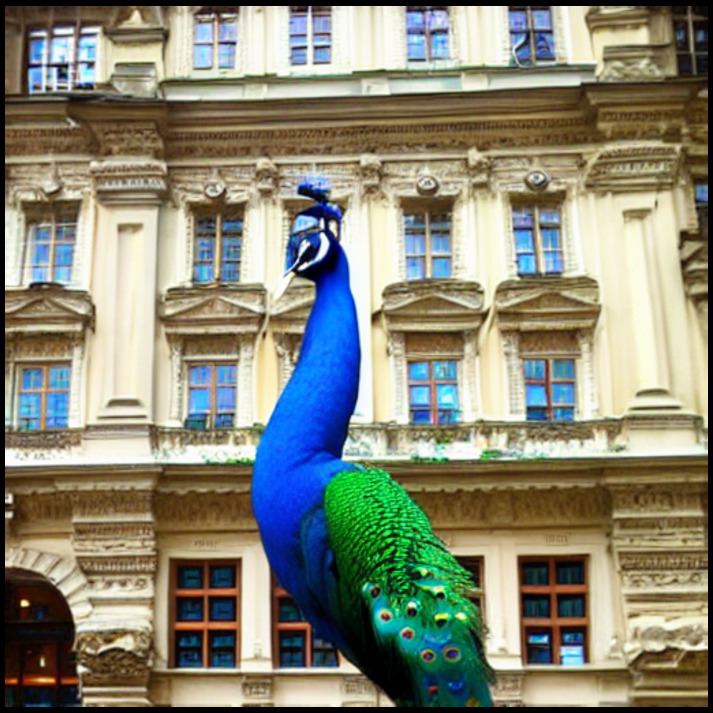} & 
        \includegraphics[width=0.23\textwidth, height=2.8cm]{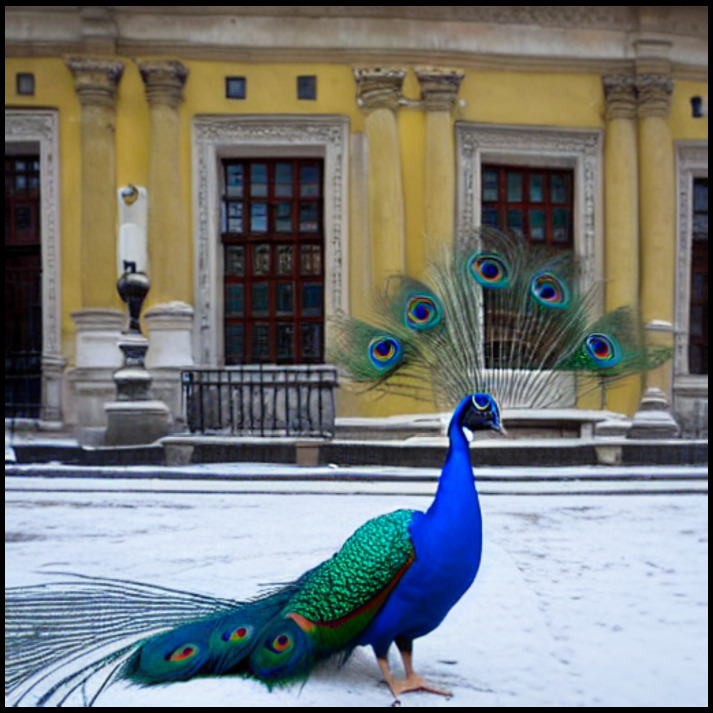} & 
        \includegraphics[width=0.23\textwidth, height=2.8cm]{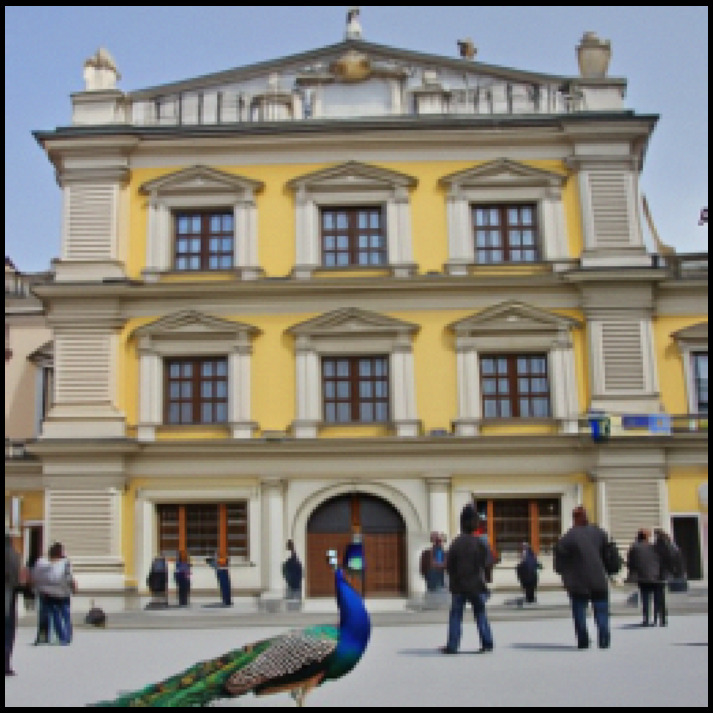} \\
        \textbf{SD} / 0.42 / 0.39 & \textbf{Imagen} / 0.15 / 0.35 & \textbf{Flux} / 0.22 / 0.37 & \textbf{Imagen-3} / 0.10 / 0.34 \\
        \includegraphics[width=0.23\textwidth, height=2.8cm]{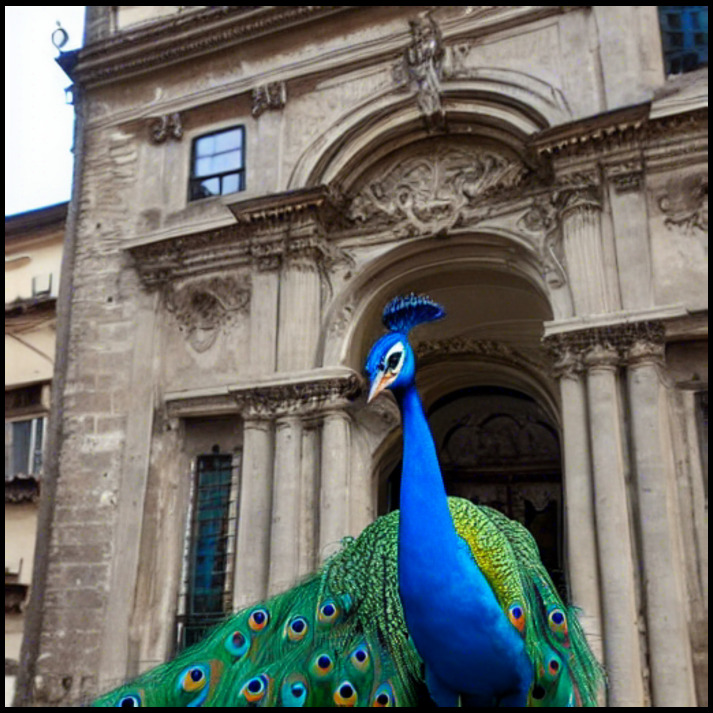} & 
        \includegraphics[width=0.23\textwidth, height=2.8cm]{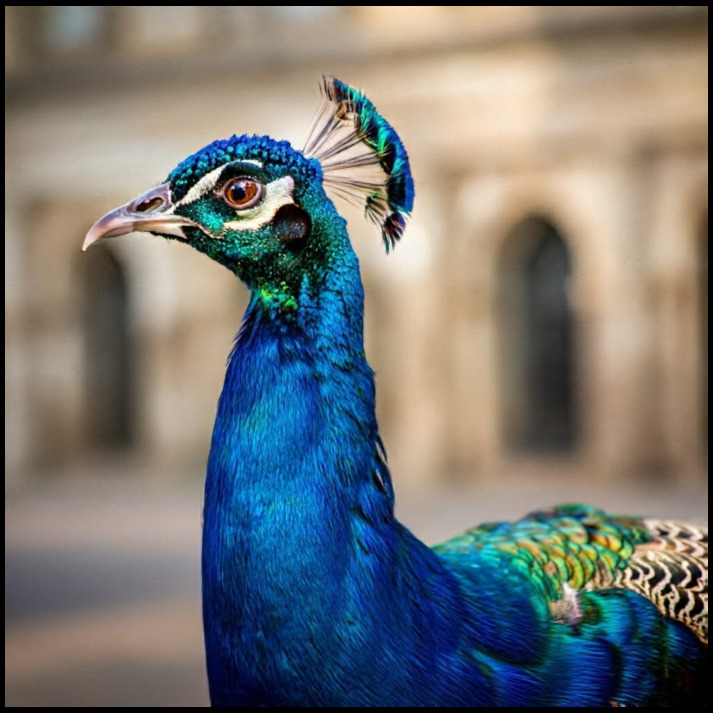} & 
        \includegraphics[width=0.23\textwidth, height=2.8cm]{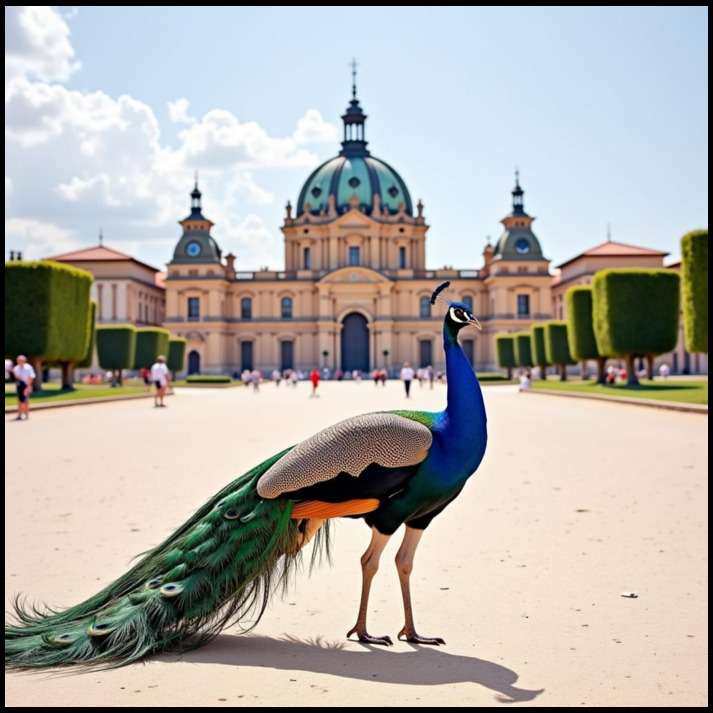} & 
        \includegraphics[width=0.23\textwidth, height=2.8cm]{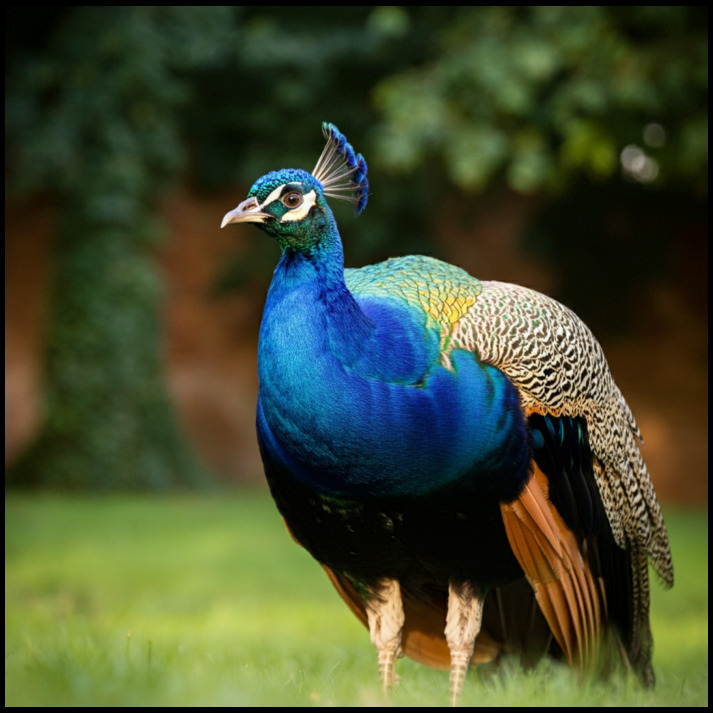} \\
        \multicolumn{4}{c}{\textit{A peacock in front of the Bandinelli Palace.}} \\
        \midrule
        \textbf{Real Photo} & \textbf{Custom-Diff} / 0.59 / 0.29 & \textbf{DreamBooth} / 0.62 / 0.28 & \textbf{Instruct-Imagen} / 0.72 / 0.29 \\
        \includegraphics[width=0.23\textwidth, height=2.8cm]{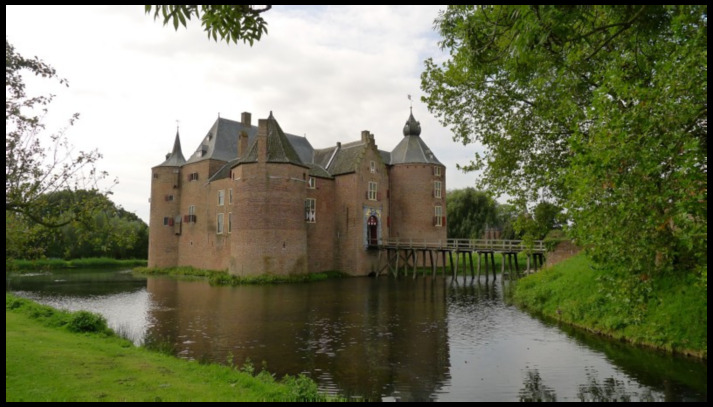} & 
        \includegraphics[width=0.23\textwidth, height=2.8cm]{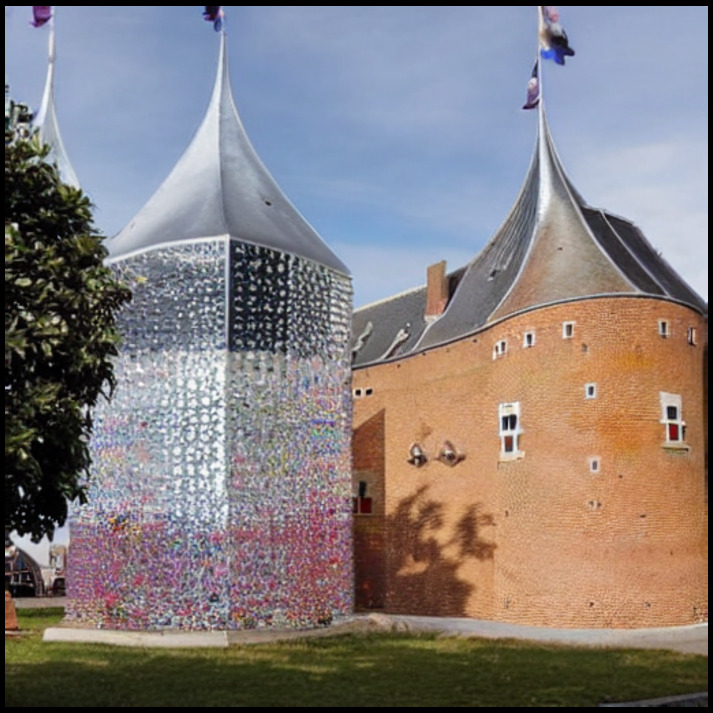} & 
        \includegraphics[width=0.23\textwidth, height=2.8cm]{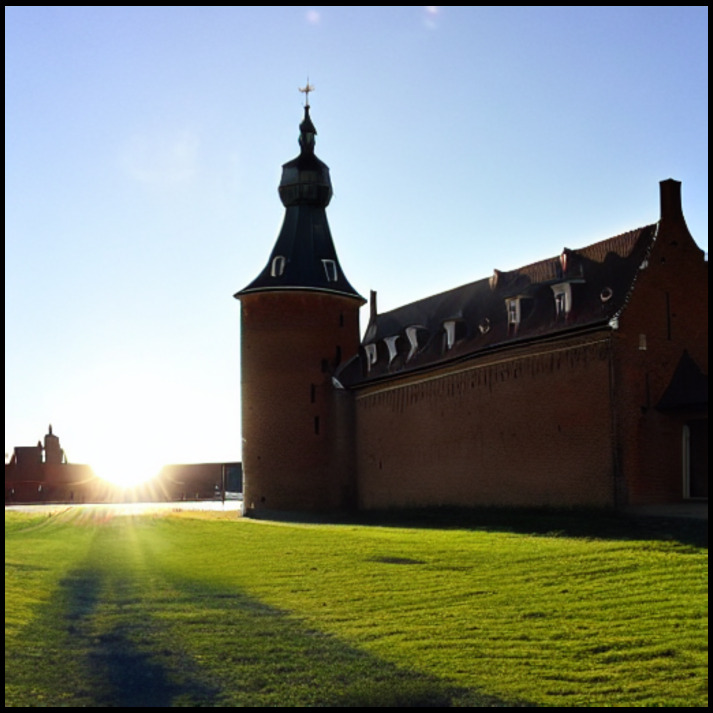} & 
        \includegraphics[width=0.23\textwidth, height=2.8cm]{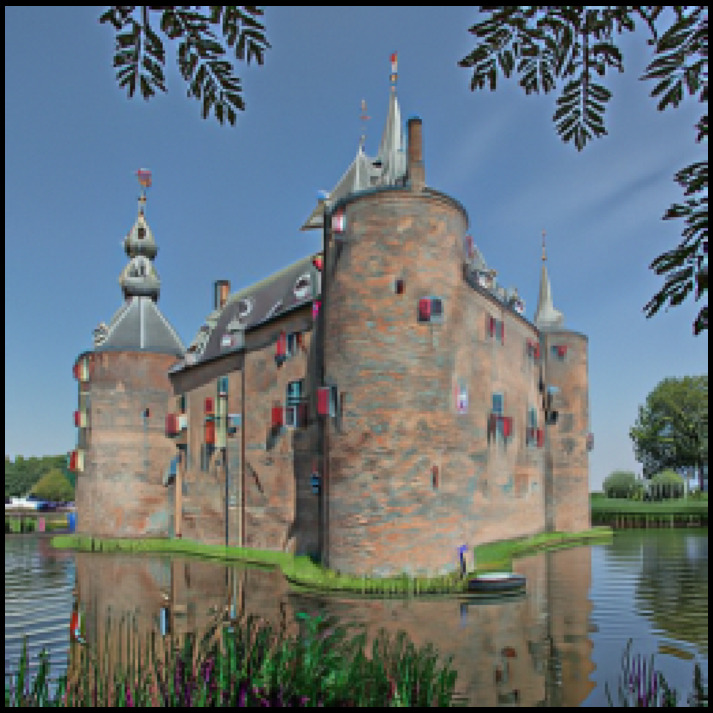} \\
        \textbf{SD} / 0.60 / 0.30 & \textbf{Imagen} / 0.59 / 0.29 & \textbf{Flux} / 0.61 / 0.27 & \textbf{Imagen-3} / 0.59 / 0.28 \\
        \includegraphics[width=0.23\textwidth, height=2.8cm]{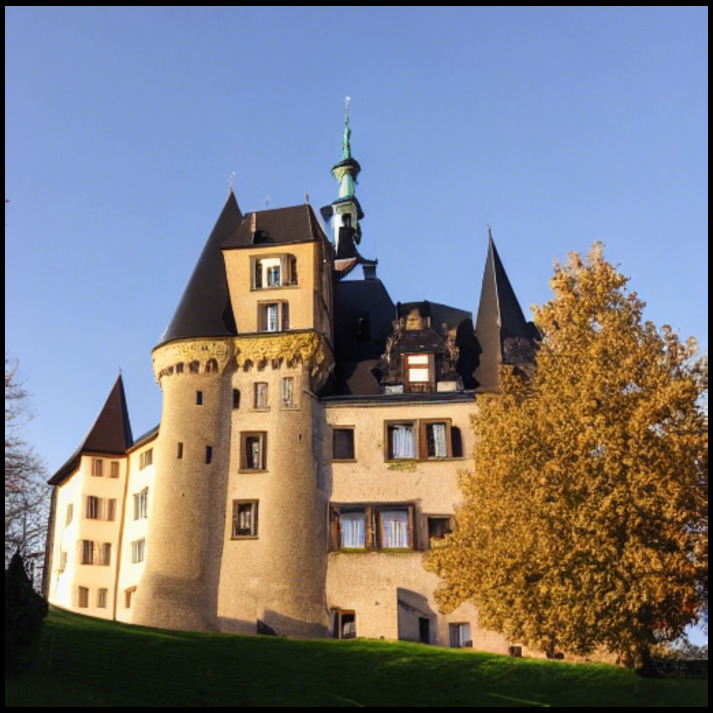} & 
        \includegraphics[width=0.23\textwidth, height=2.8cm]{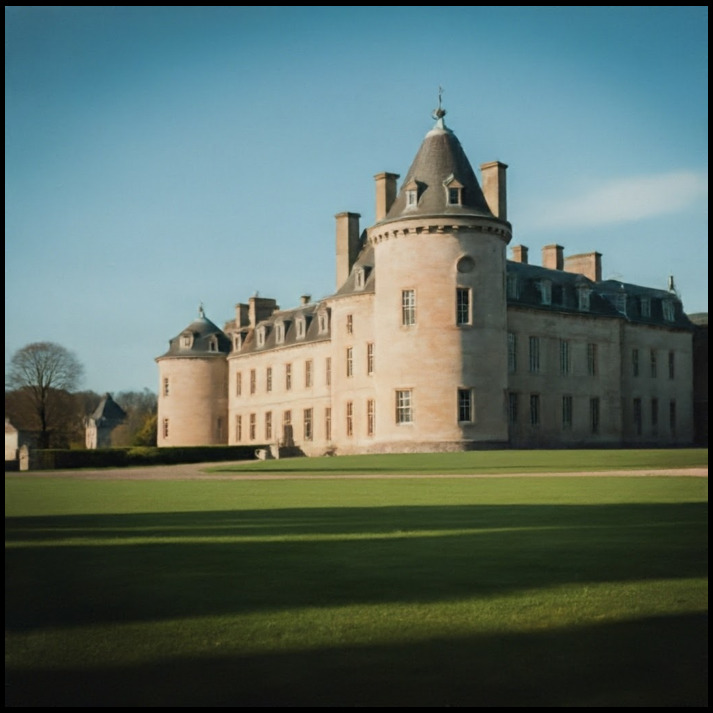} & 
        \includegraphics[width=0.23\textwidth, height=2.8cm]{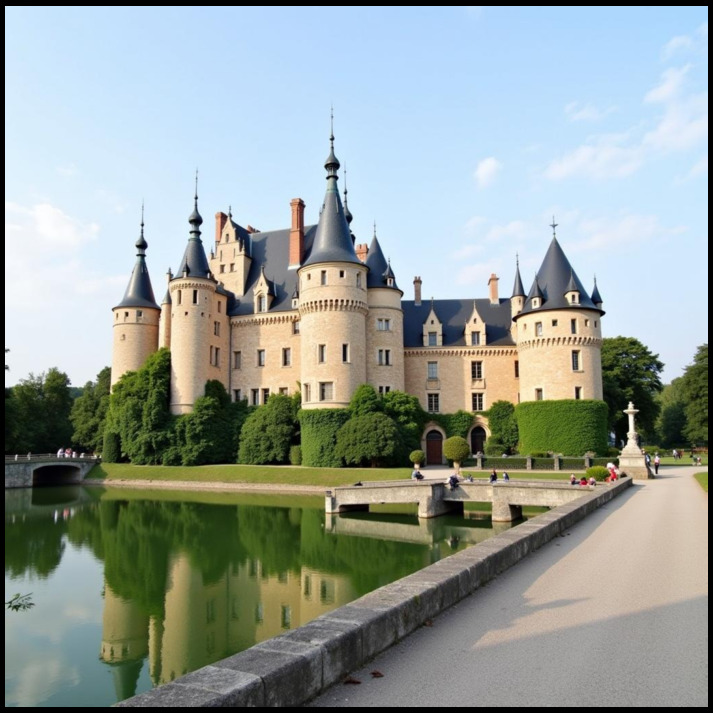} & 
        \includegraphics[width=0.23\textwidth, height=2.8cm]{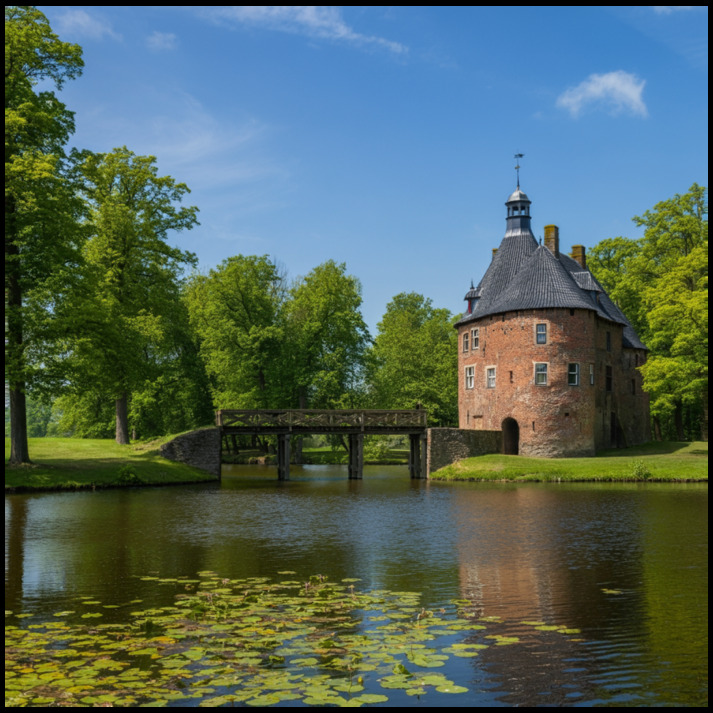} \\
        \multicolumn{4}{c}{\textit{The Ammersoyen Castle, made of crystal, shimmers in the sunlight.}} \\
        \midrule
        \textbf{Real Photo} & \textbf{Custom-Diff} / 0.66 / 0.32 & \textbf{DreamBooth} / 0.58 / 0.33 & \textbf{Instruct-Imagen} / 0.58 / 0.31 \\
        \includegraphics[width=0.23\textwidth, height=2.8cm]{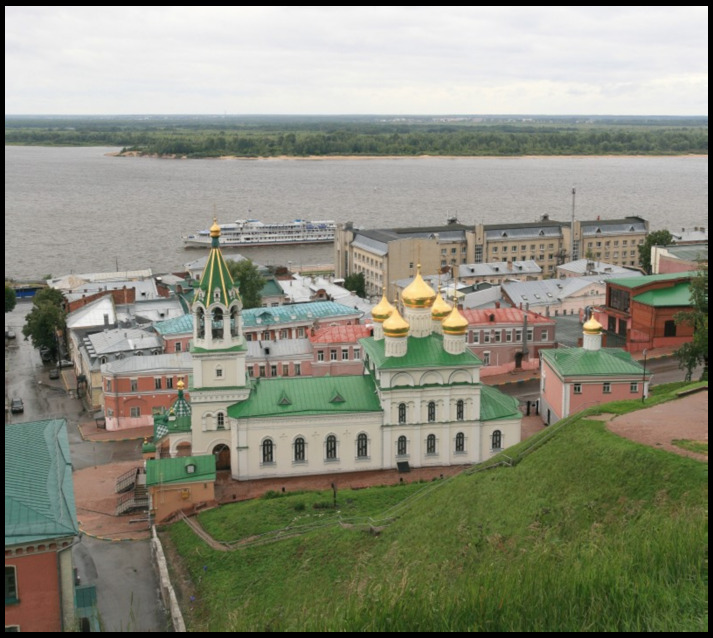} & 
        \includegraphics[width=0.23\textwidth, height=2.8cm]{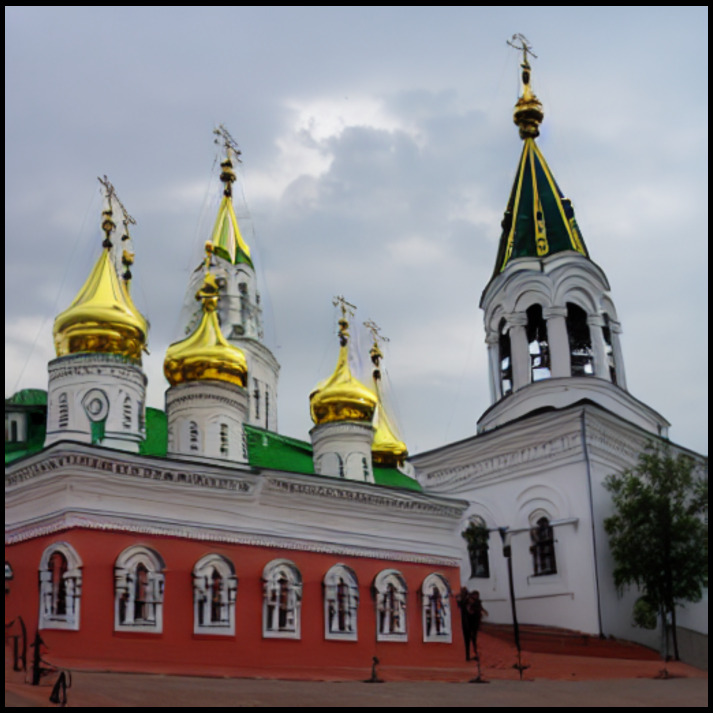} & 
        \includegraphics[width=0.23\textwidth, height=2.8cm]{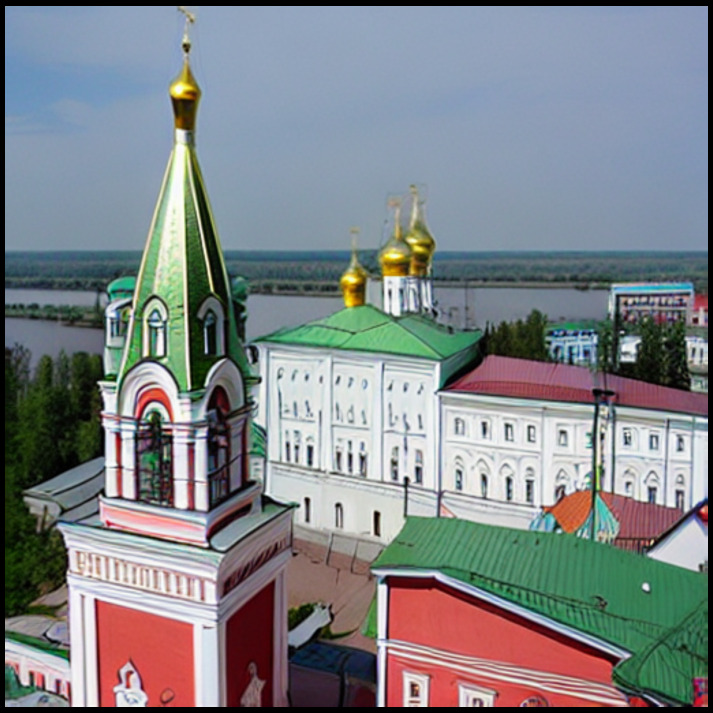} & 
        \includegraphics[width=0.23\textwidth, height=2.8cm]{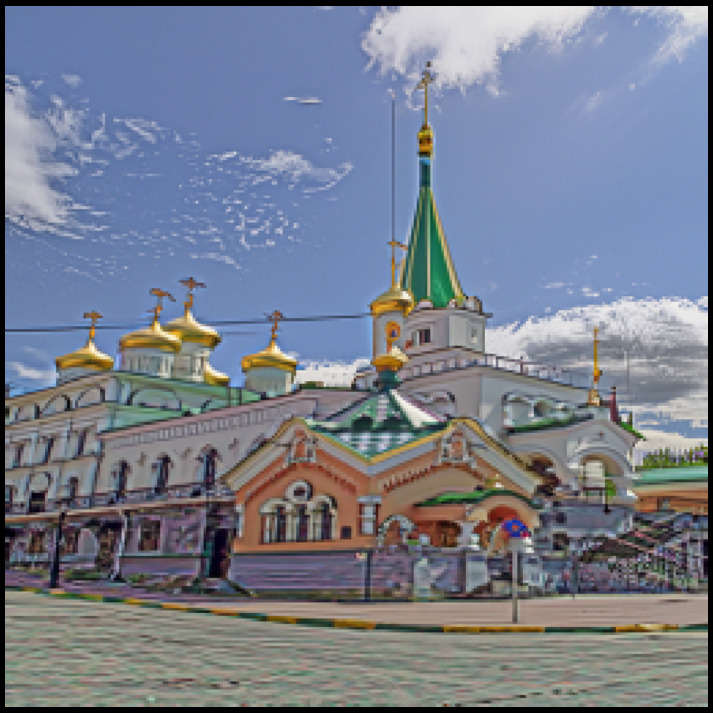} \\
        \textbf{SD} / 0.55 / 0.32 & \textbf{Imagen} / 0.60 / 0.32 & \textbf{Flux} / 0.61 / 0.31 & \textbf{Imagen-3} / 0.61 / 0.32 \\
        \includegraphics[width=0.23\textwidth, height=2.8cm]{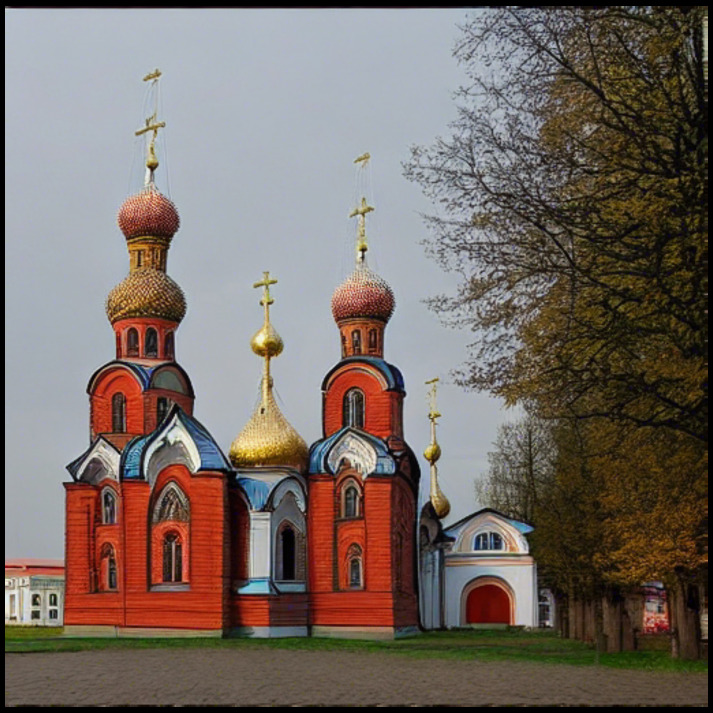} & 
        \includegraphics[width=0.23\textwidth, height=2.8cm]{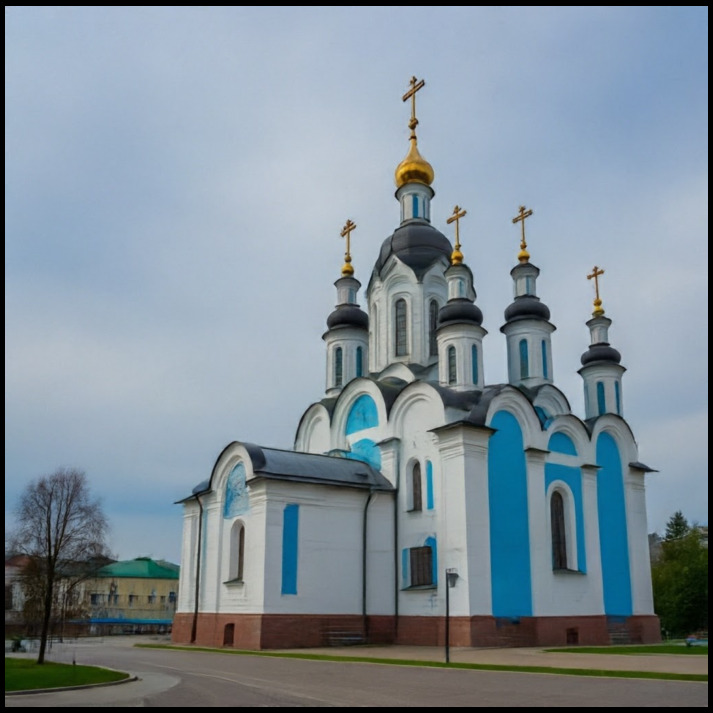} & 
        \includegraphics[width=0.23\textwidth, height=2.8cm]{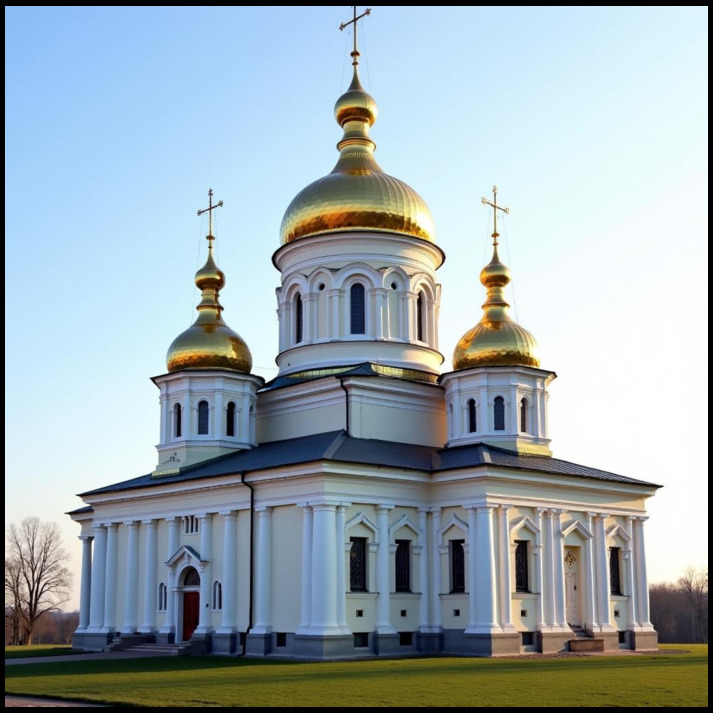} & 
        \includegraphics[width=0.23\textwidth, height=2.8cm]{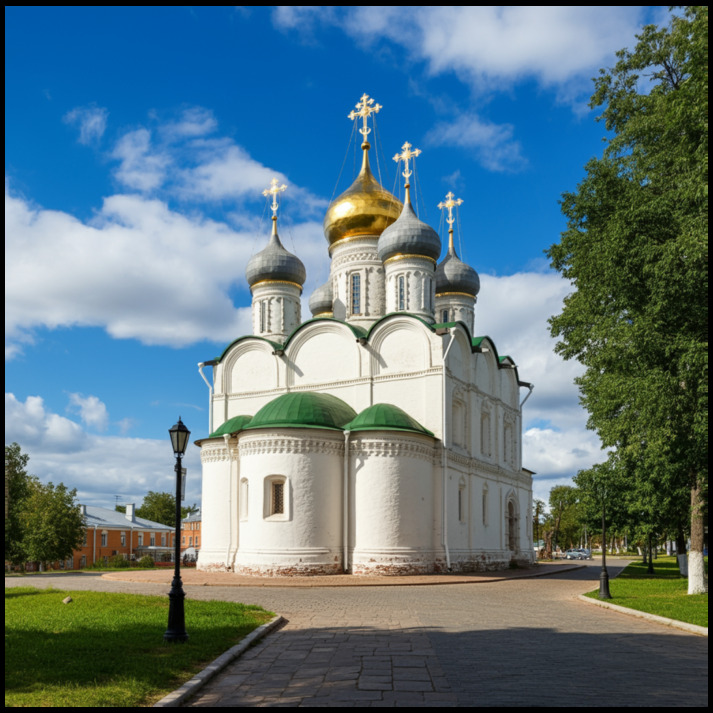} \\
        \multicolumn{4}{c}{\textit{Photo of a Church of St. John the Baptist (Nizhny Novgorod).}} \\
        \bottomrule
    \end{tabular}
    \caption{\textbf{Qualitative results} for the landmark domain, including the DINO and CLIP-T scores.}
    \label{fig:image_comparison_6}
\end{figure}

\begin{figure}[t]
    \centering
    \scriptsize
    \setlength{\tabcolsep}{2pt} %
    \begin{tabular}{c@{\;\;}c@{\;\;}c@{\;\;}c}
        \toprule %
        \textbf{Real Photo} & \textbf{Custom-Diff} / 0.55 / 0.35 & \textbf{DreamBooth} / 0.44 / 0.35 & \textbf{Instruct-Imagen} / 0.74 / 0.30 \\
        \includegraphics[width=0.23\textwidth, height=2.8cm]{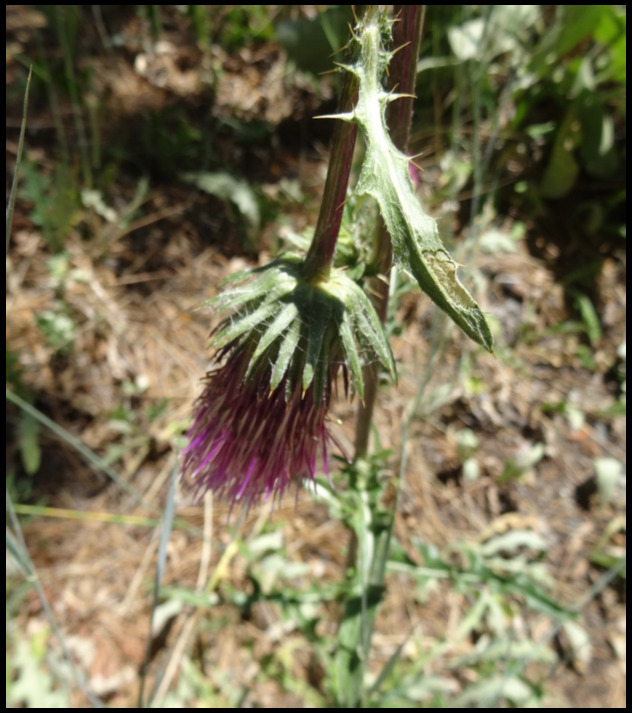} & 
        \includegraphics[width=0.23\textwidth, height=2.8cm]{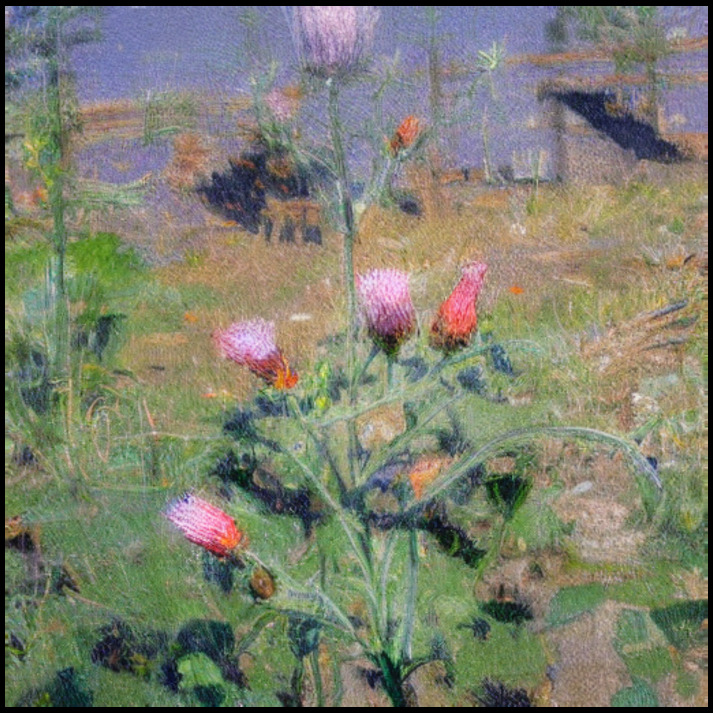} & 
        \includegraphics[width=0.23\textwidth, height=2.8cm]{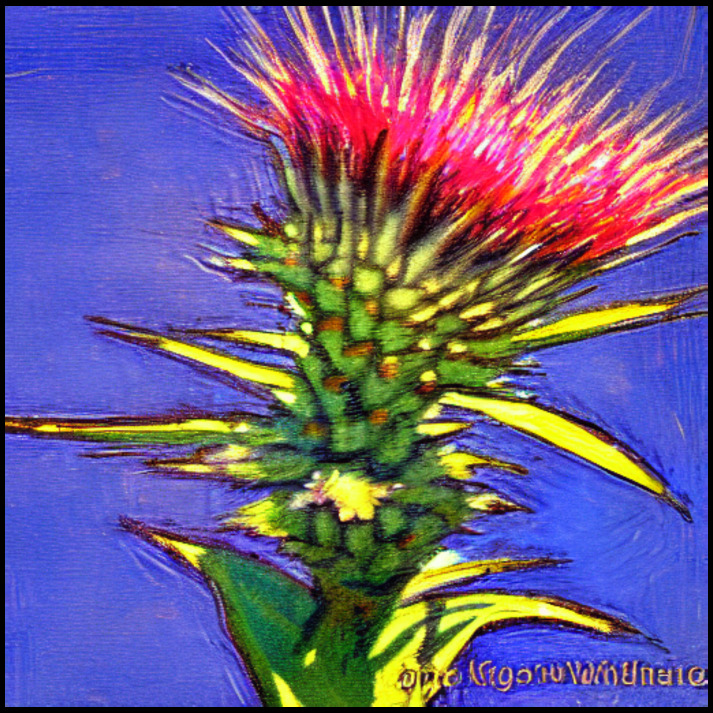} & 
        \includegraphics[width=0.23\textwidth, height=2.8cm]{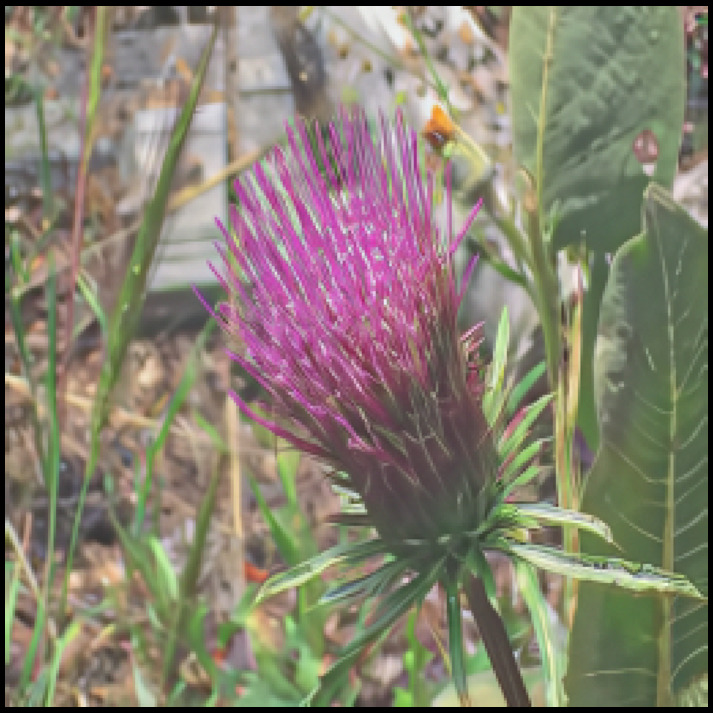} \\
        \textbf{SD} / 0.51 / 0.32 & \textbf{Imagen} / 0.43 / 0.34 & \textbf{Flux} / 0.29 / 0.29 & \textbf{Imagen-3} / 0.61 / 0.32 \\
        \includegraphics[width=0.23\textwidth, height=2.8cm]{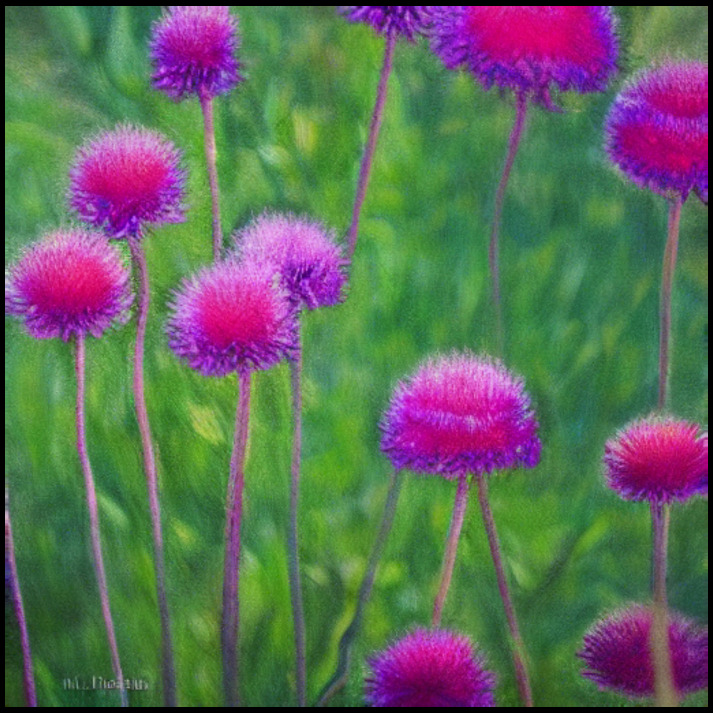} & 
        \includegraphics[width=0.23\textwidth, height=2.8cm]{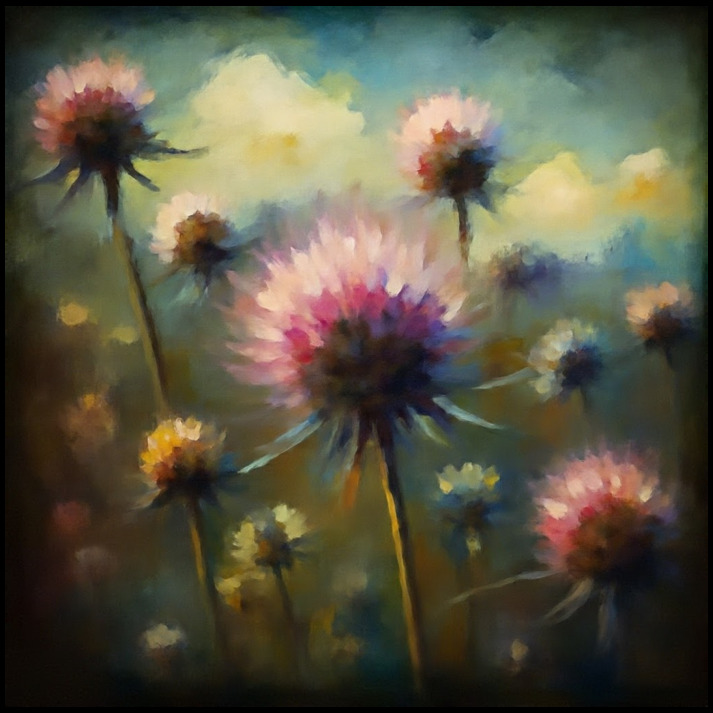} & 
        \includegraphics[width=0.23\textwidth, height=2.8cm]{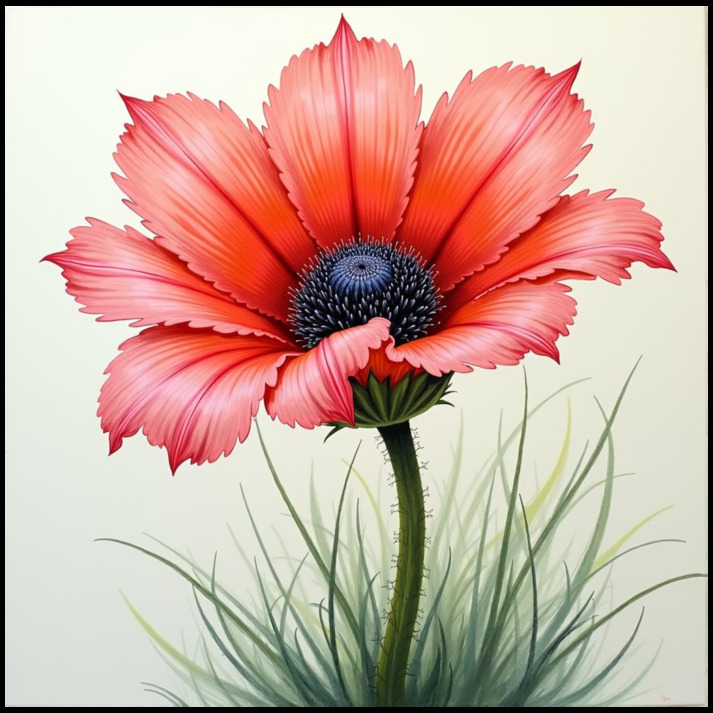} & 
        \includegraphics[width=0.23\textwidth, height=2.8cm]{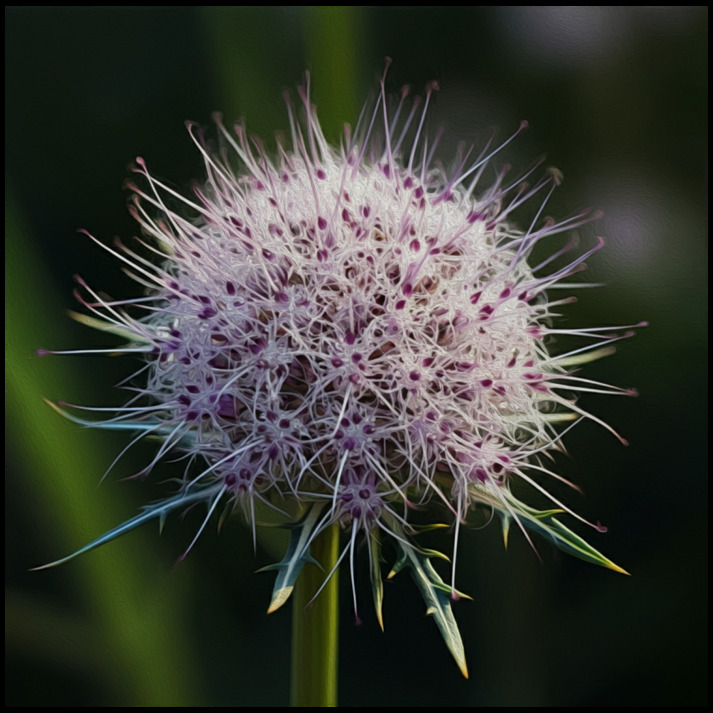} \\
        \multicolumn{4}{c}{\textit{An impressionistic painting of Cirsium andersonii.}} \\
        \midrule
        \textbf{Real Photo} & \textbf{Custom-Diff} / 0.46 / 0.30 & \textbf{DreamBooth} / 0.34 / 0.32 & \textbf{Instruct-Imagen} / 0.67 / 0.31 \\
        \includegraphics[width=0.23\textwidth, height=2.8cm]{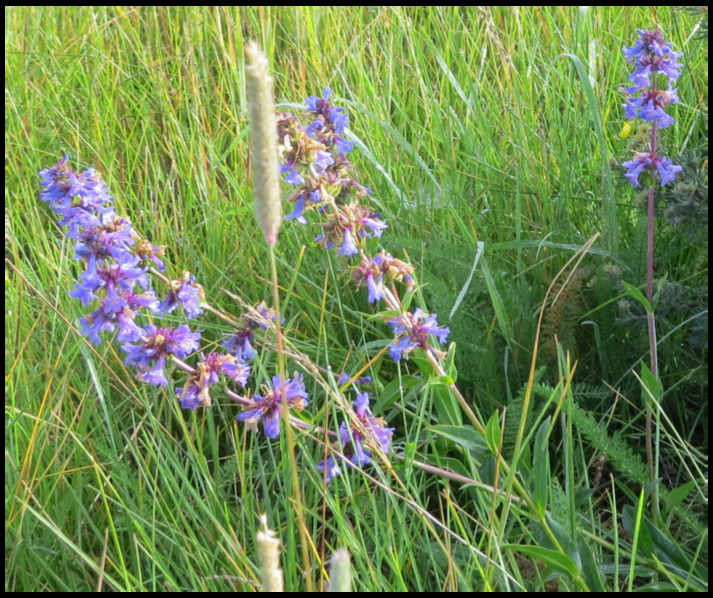} & 
        \includegraphics[width=0.23\textwidth, height=2.8cm]{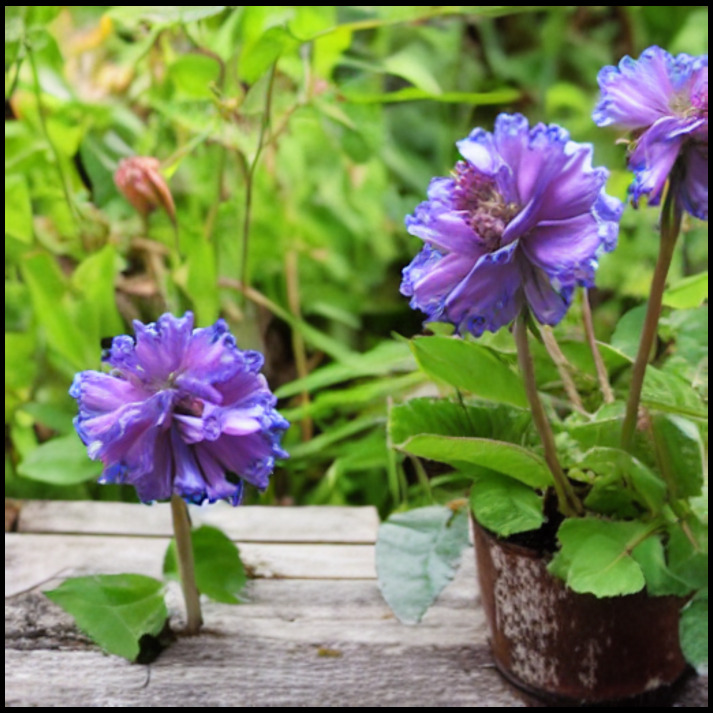} & 
        \includegraphics[width=0.23\textwidth, height=2.8cm]{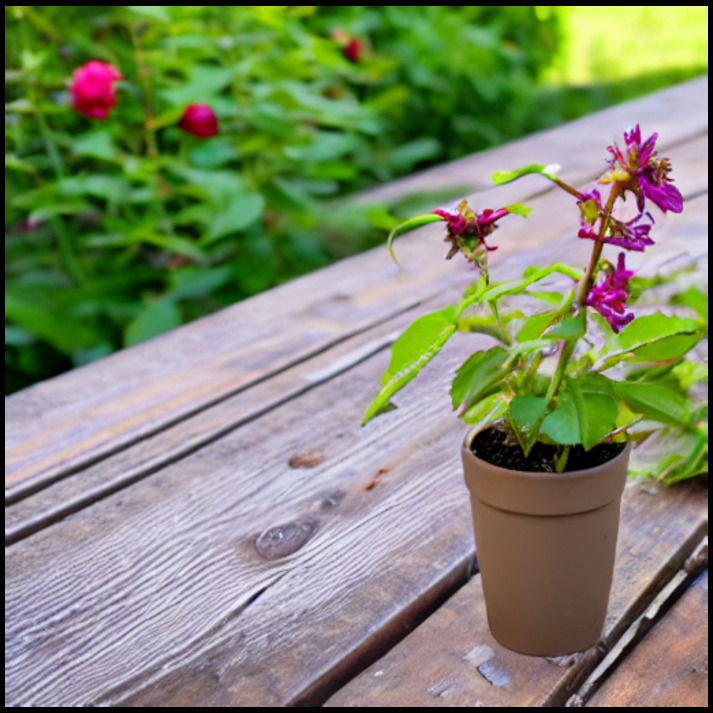} & 
        \includegraphics[width=0.23\textwidth, height=2.8cm]{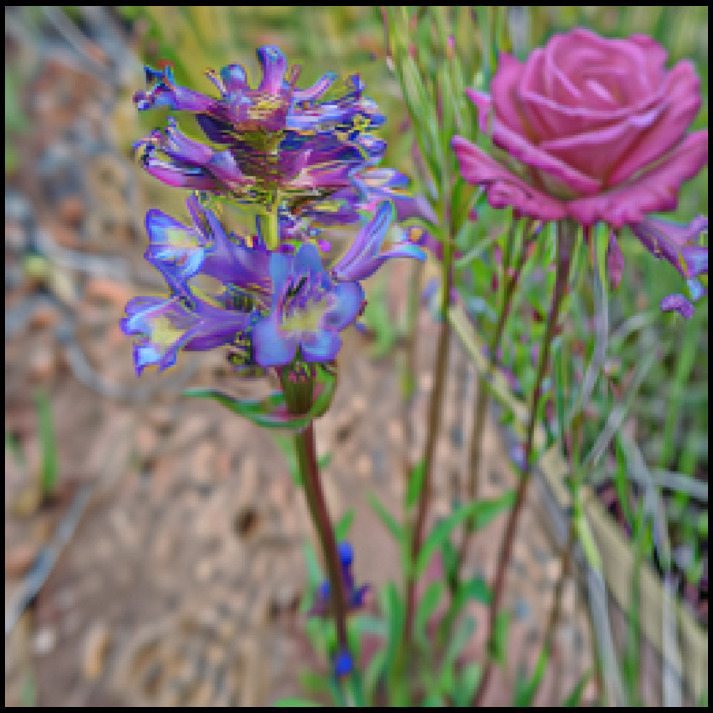} \\
        \textbf{SD} / 0.36 / 0.30 & \textbf{Imagen} / 0.32 / 0.31 & \textbf{Flux} / 0.34 / 0.30 & \textbf{Imagen-3} / 0.31 / 0.33 \\
        \includegraphics[width=0.23\textwidth, height=2.8cm]{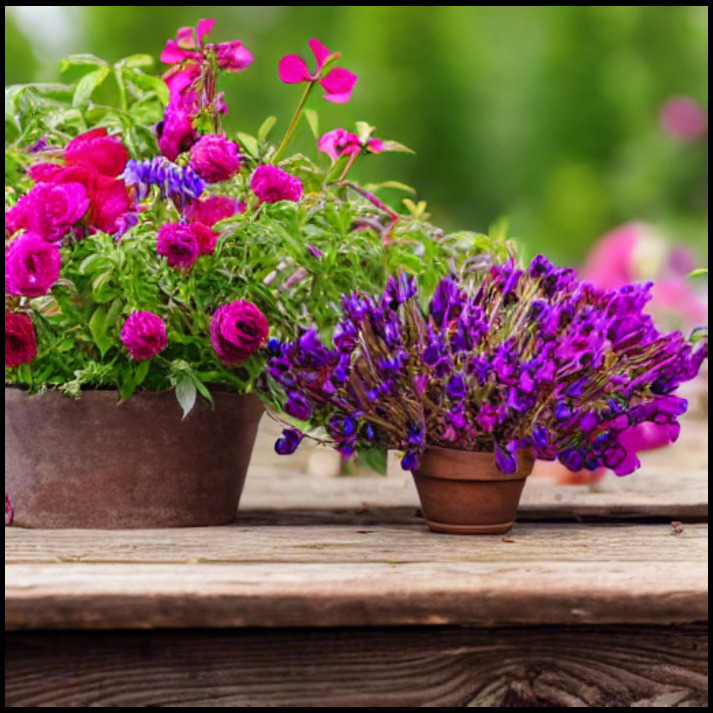} & 
        \includegraphics[width=0.23\textwidth, height=2.8cm]{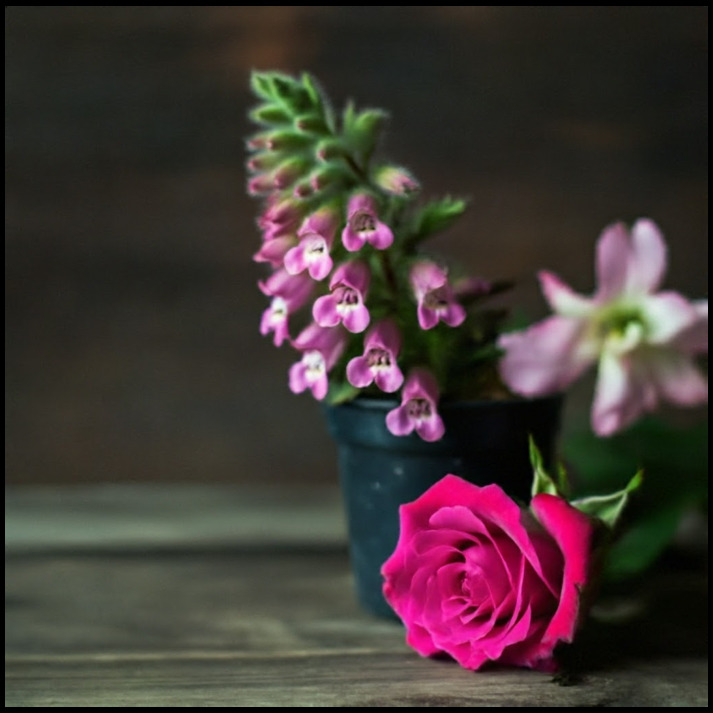} & 
        \includegraphics[width=0.23\textwidth, height=2.8cm]{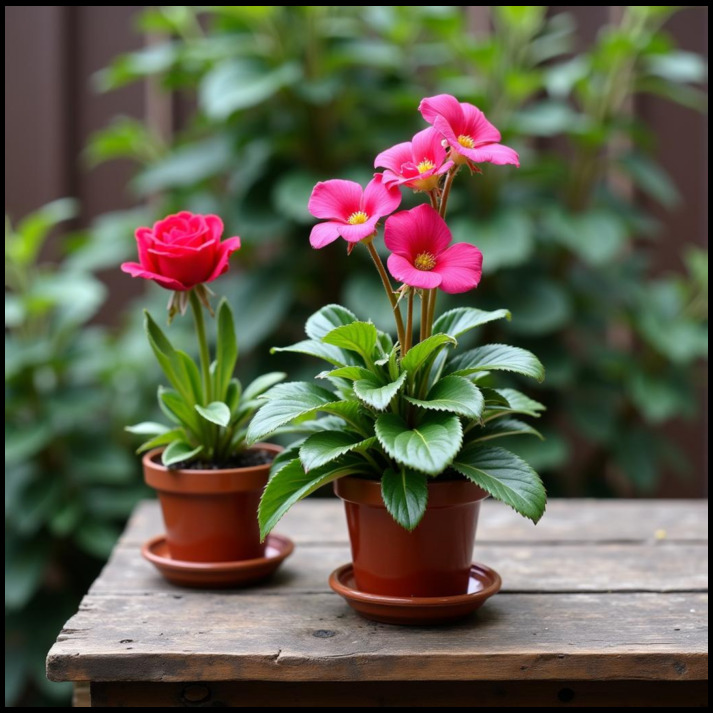} & 
        \includegraphics[width=0.23\textwidth, height=2.8cm]{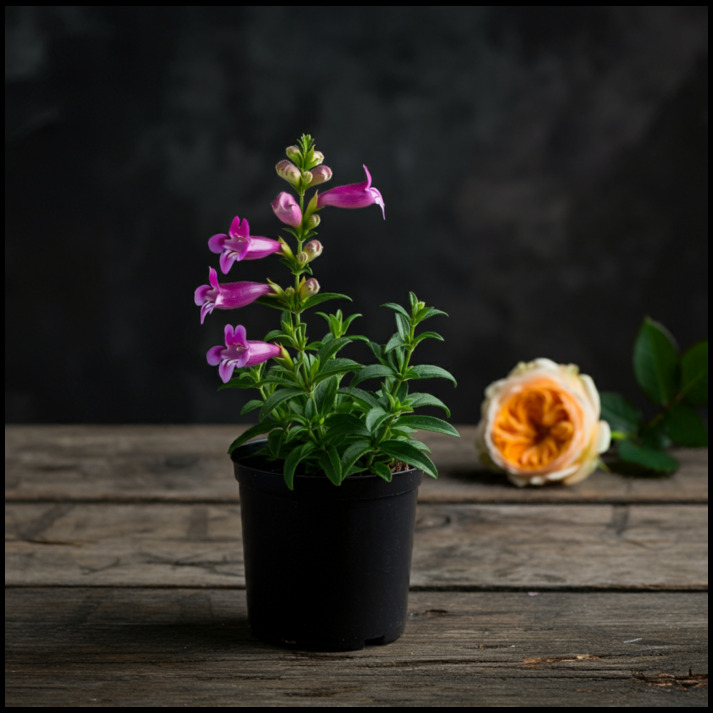} \\
        \multicolumn{4}{c}{\textit{Penstemon rydbergii on a rustic wooden table next to a rose plant.}} \\
        \midrule
        \textbf{Real Photo} & \textbf{Custom-Diff} / 0.64 / 0.29 & \textbf{DreamBooth} / 0.68 / 0.28 & \textbf{Instruct-Imagen} / 0.78 / 0.30 \\
        \includegraphics[width=0.23\textwidth, height=2.8cm]{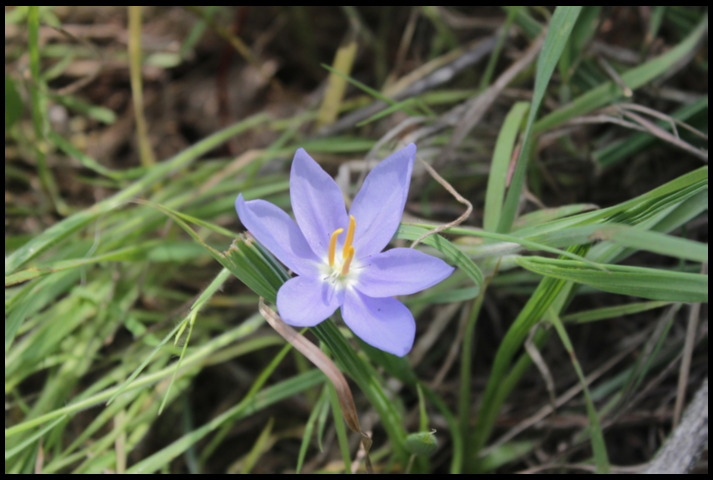} & 
        \includegraphics[width=0.23\textwidth, height=2.8cm]{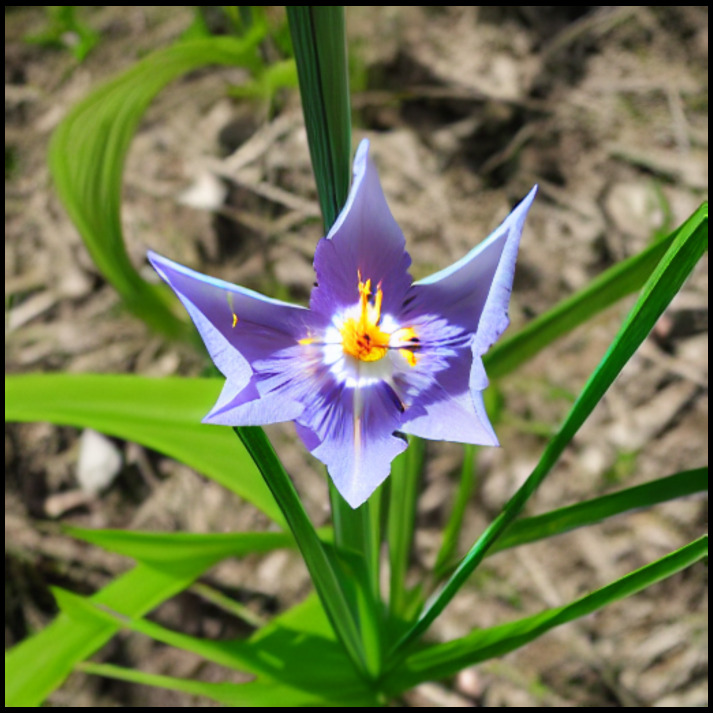} & 
        \includegraphics[width=0.23\textwidth, height=2.8cm]{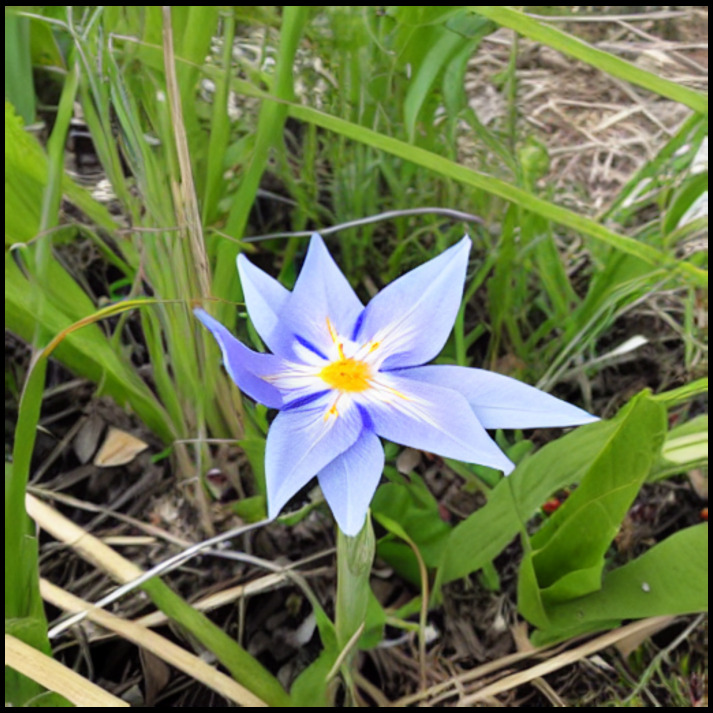} & 
        \includegraphics[width=0.23\textwidth, height=2.8cm]{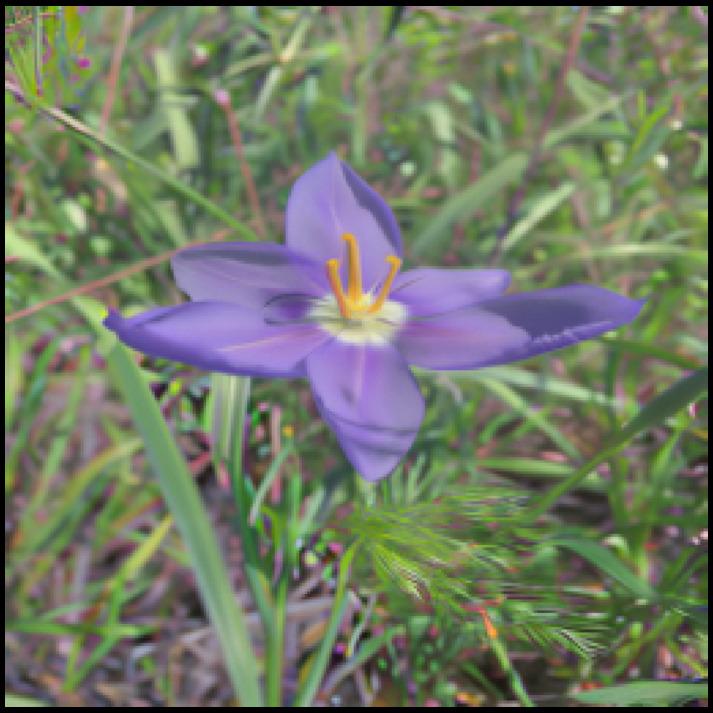} \\
        \textbf{SD} / 0.23 / 0.25 & \textbf{Imagen} / 0.44 / 0.28 & \textbf{Flux} / 0.58 / 0.30 & \textbf{Imagen-3} / 0.46 / 0.28 \\
        \includegraphics[width=0.23\textwidth, height=2.8cm]{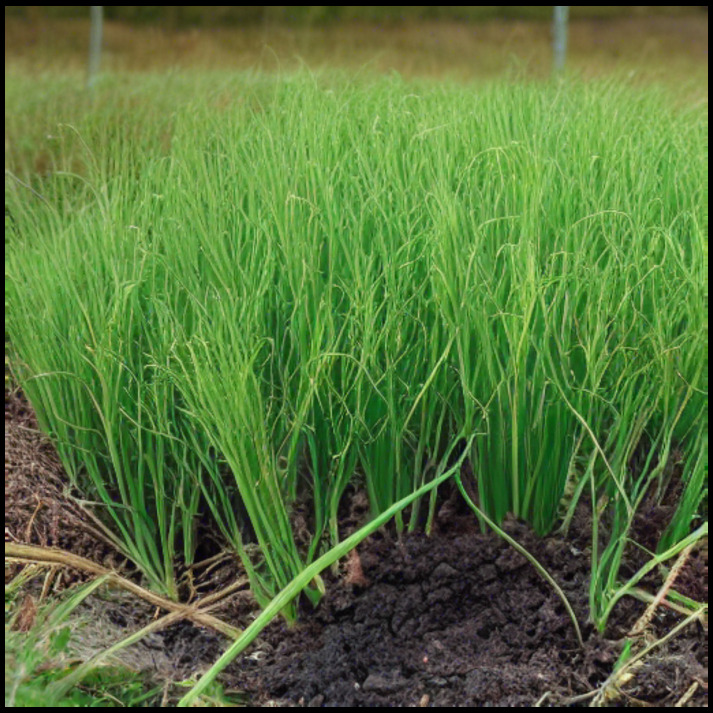} & 
        \includegraphics[width=0.23\textwidth, height=2.8cm]{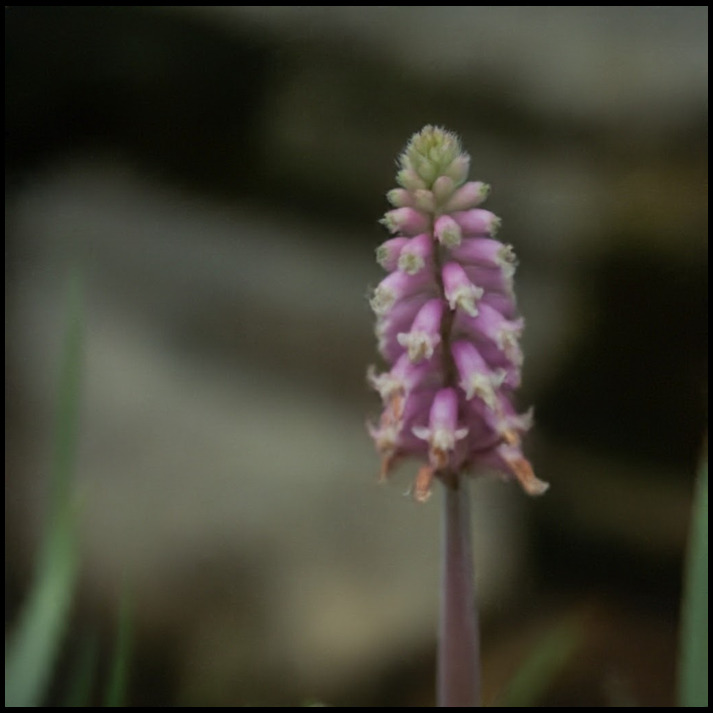} & 
        \includegraphics[width=0.23\textwidth, height=2.8cm]{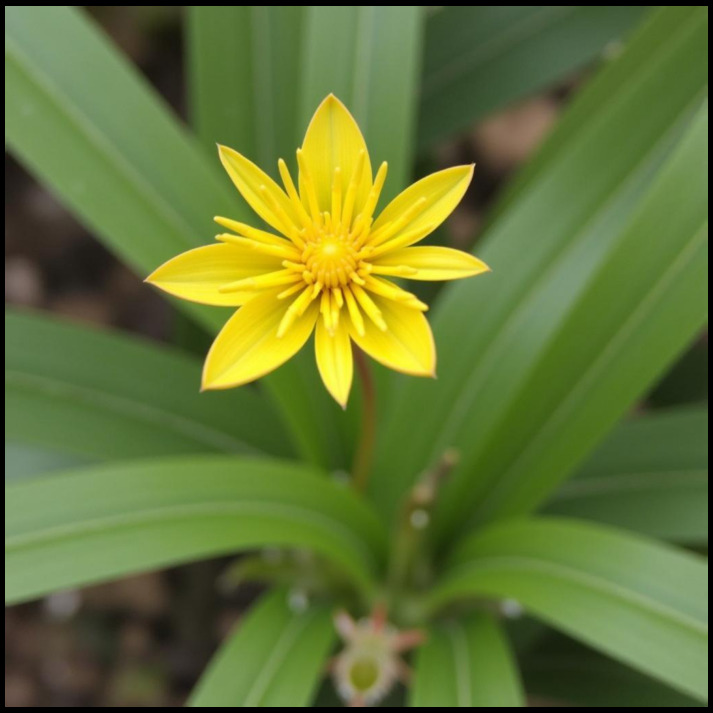} & 
        \includegraphics[width=0.23\textwidth, height=2.8cm]{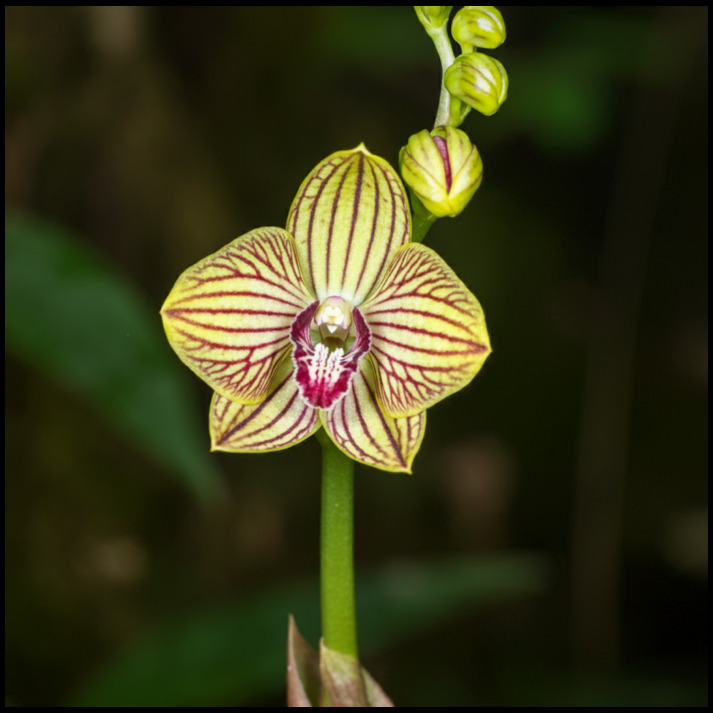} \\
        \multicolumn{4}{c}{\textit{Photo of Nemastylis geminiflora.}} \\
        \bottomrule
    \end{tabular}
    \caption{\textbf{Qualitative results} for the plant domain, including the DINO and CLIP-T scores.}
    \label{fig:image_comparison_7}
\end{figure}

\begin{figure}[t]
    \centering
    \scriptsize
    \setlength{\tabcolsep}{2pt} %
    \begin{tabular}{c@{\;\;}c@{\;\;}c@{\;\;}c}
        \toprule %
        \textbf{Real Photo} & \textbf{Custom-Diff} / 0.28 / 0.28 & \textbf{DreamBooth} / 0.56 / 0.36 & \textbf{Instruct-Imagen} / 0.70 / 0.32 \\
        \includegraphics[width=0.23\textwidth, height=2.8cm]{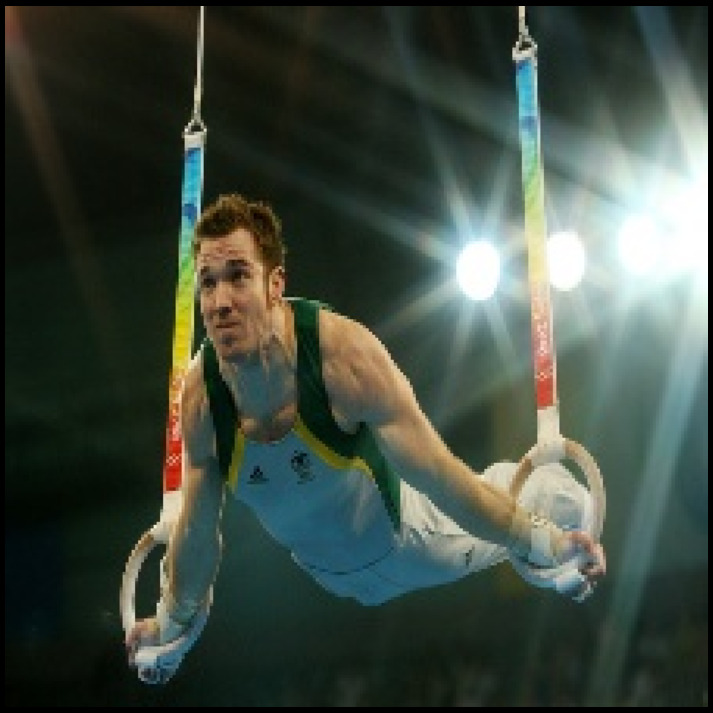} & 
        \includegraphics[width=0.23\textwidth, height=2.8cm]{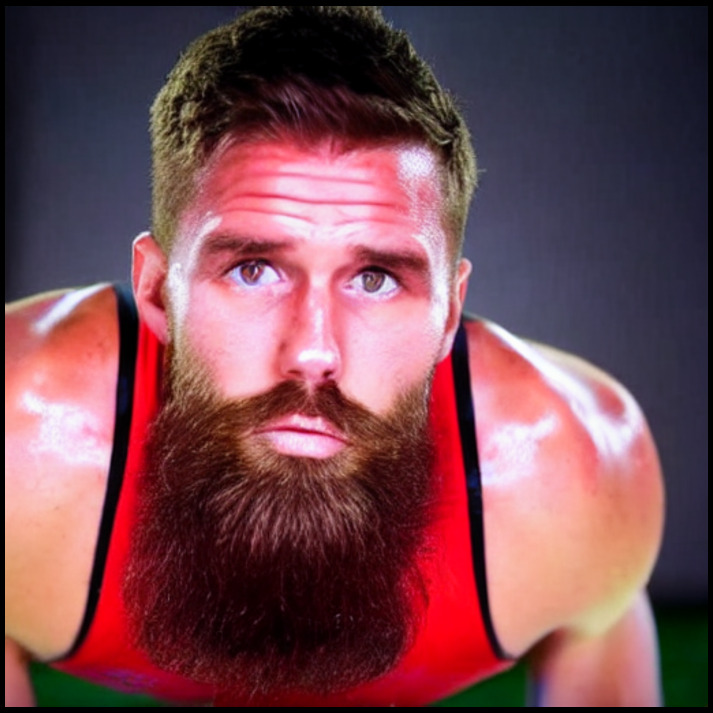} & 
        \includegraphics[width=0.23\textwidth, height=2.8cm]{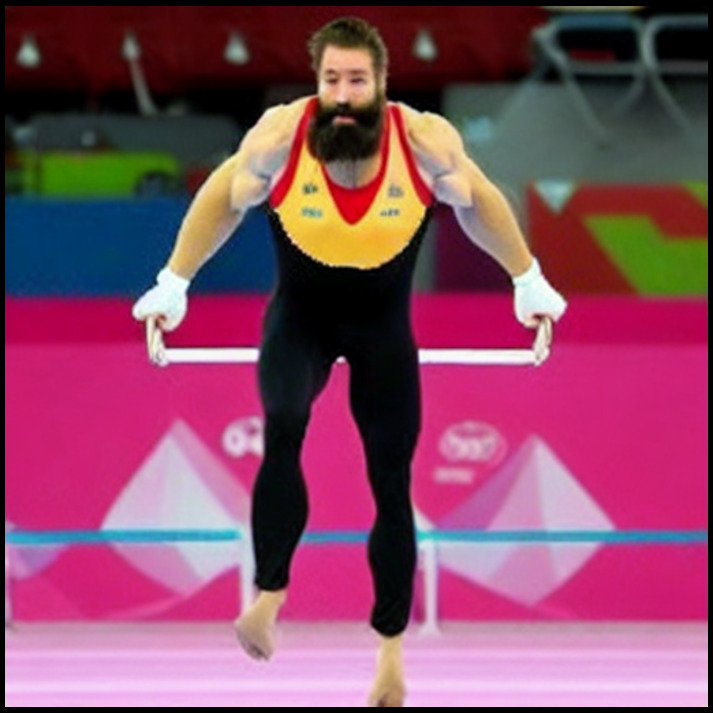} & 
        \includegraphics[width=0.23\textwidth, height=2.8cm]{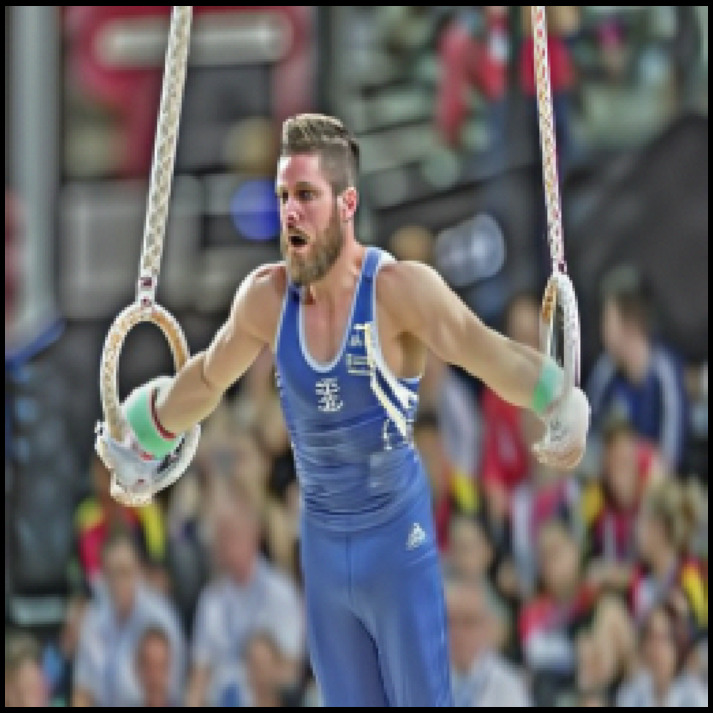} \\
        \textbf{SD} / 0.40 / 0.33 & \textbf{Imagen} / 0.37 / 0.32 & \textbf{Flux} / 0.46 / 0.33 & \textbf{Imagen-3} / 0.69 / 0.37 \\
        \includegraphics[width=0.23\textwidth, height=2.8cm]{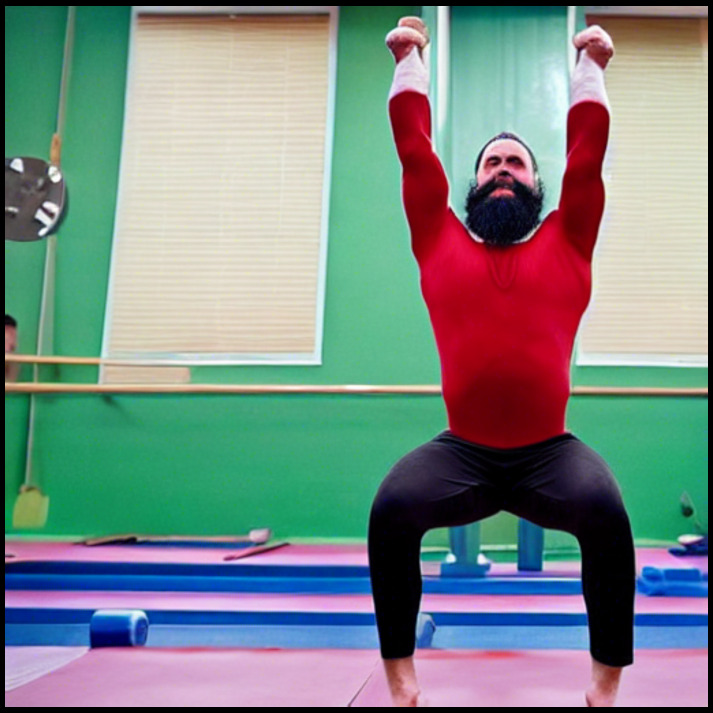} & 
        \includegraphics[width=0.23\textwidth, height=2.8cm]{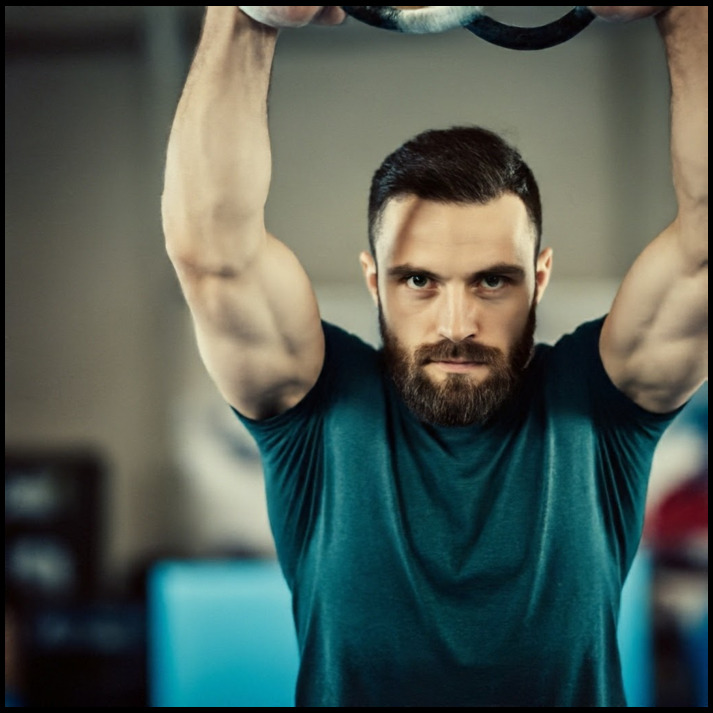} & 
        \includegraphics[width=0.23\textwidth, height=2.8cm]{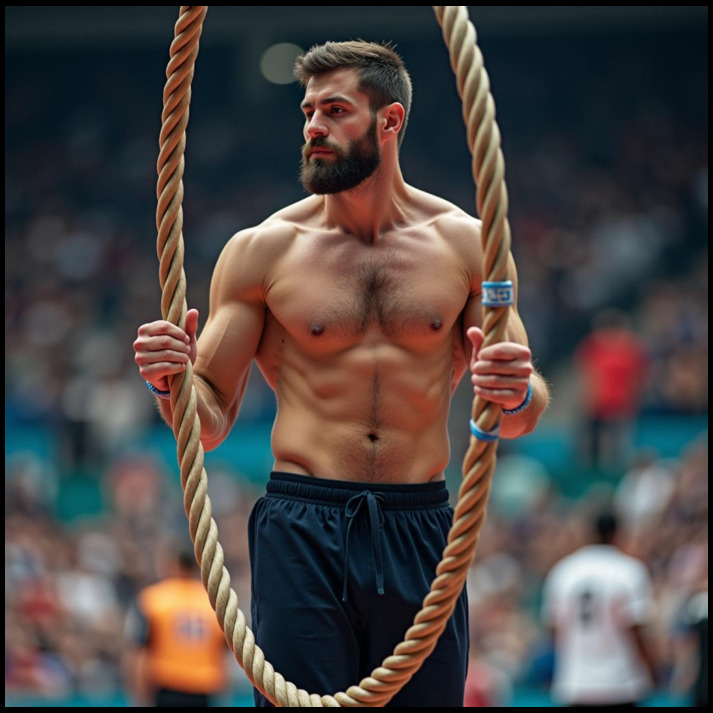} & 
        \includegraphics[width=0.23\textwidth, height=2.8cm]{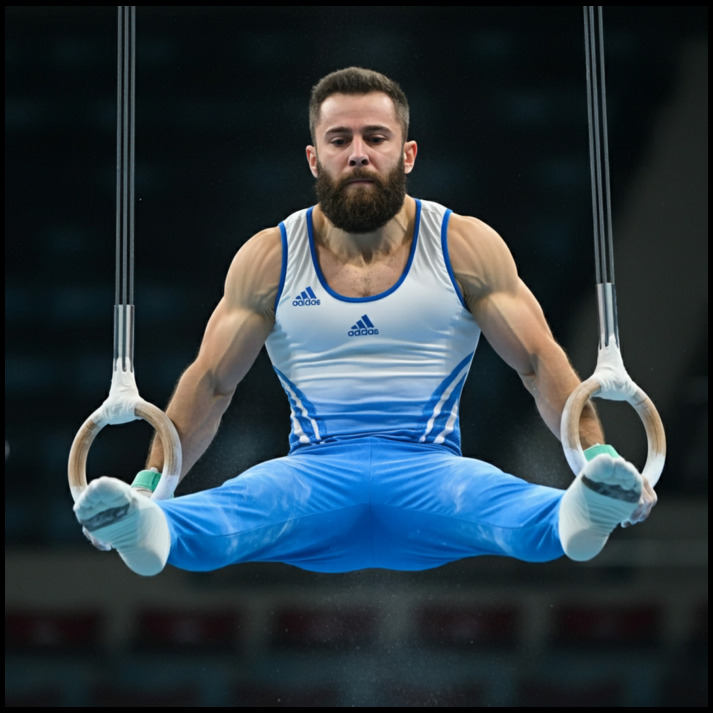} \\
        \multicolumn{4}{c}{\textit{A man with a beard doing the Rings gymnastics.}} \\
        \midrule
        \textbf{Real Photo} & \textbf{Custom-Diff} / 0.23 / 0.23 & \textbf{DreamBooth} / 0.65 / 0.35 & \textbf{Instruct-Imagen} / 0.63 / 0.35 \\
        \includegraphics[width=0.23\textwidth, height=2.8cm]{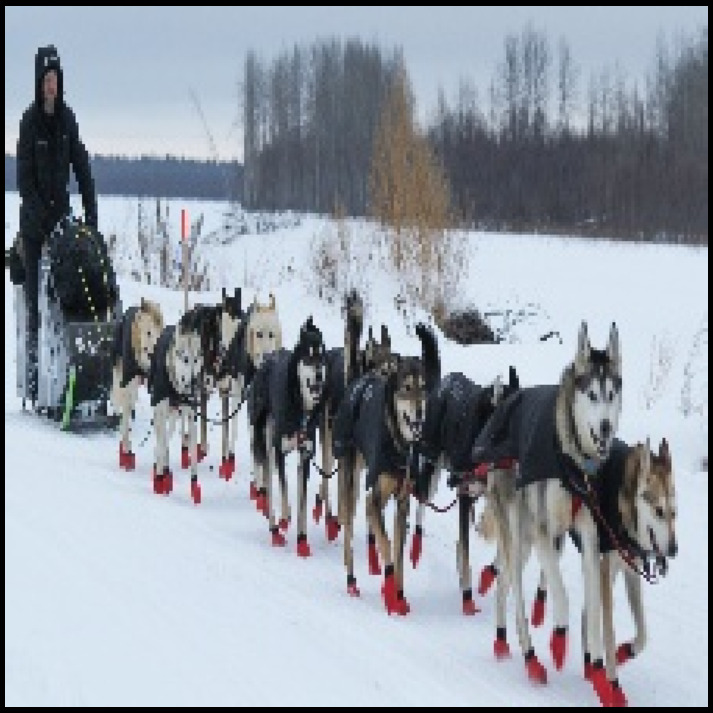} & 
        \includegraphics[width=0.23\textwidth, height=2.8cm]{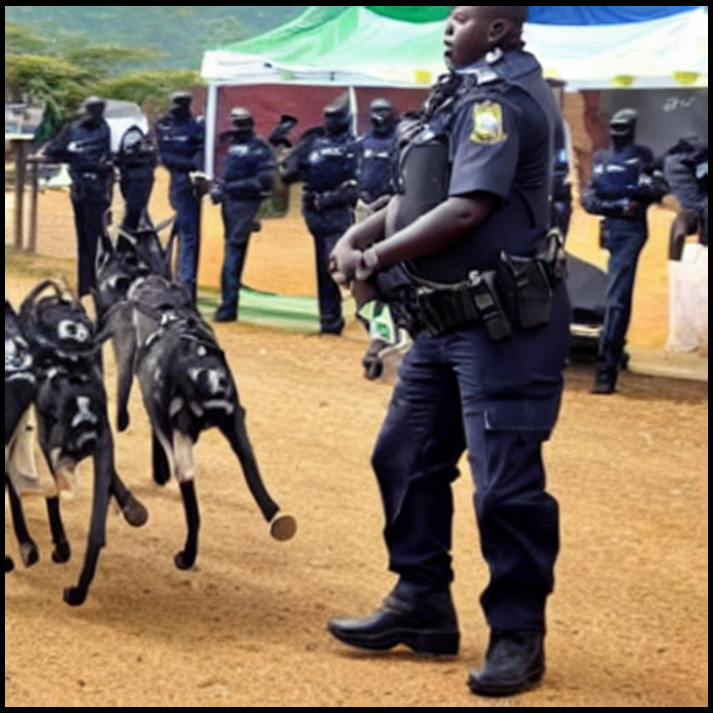} & 
        \includegraphics[width=0.23\textwidth, height=2.8cm]{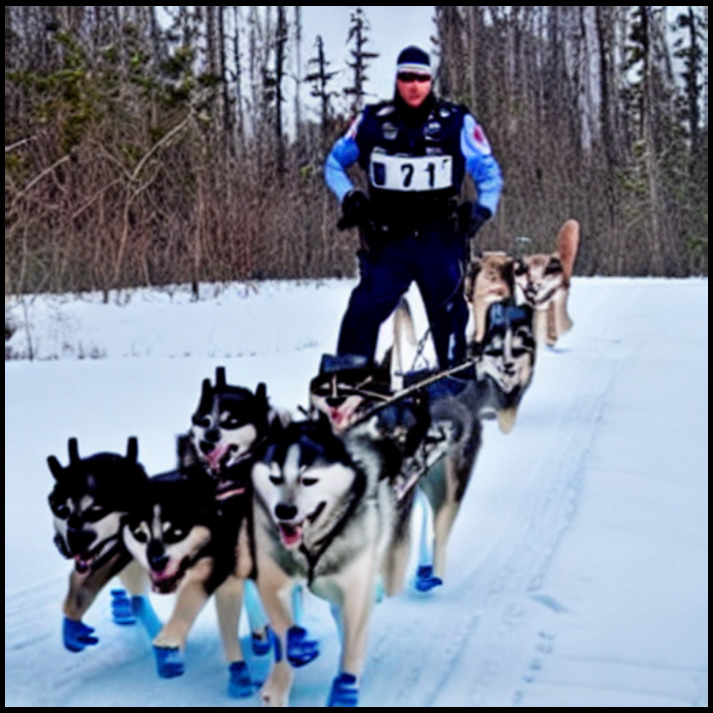} & 
        \includegraphics[width=0.23\textwidth, height=2.8cm]{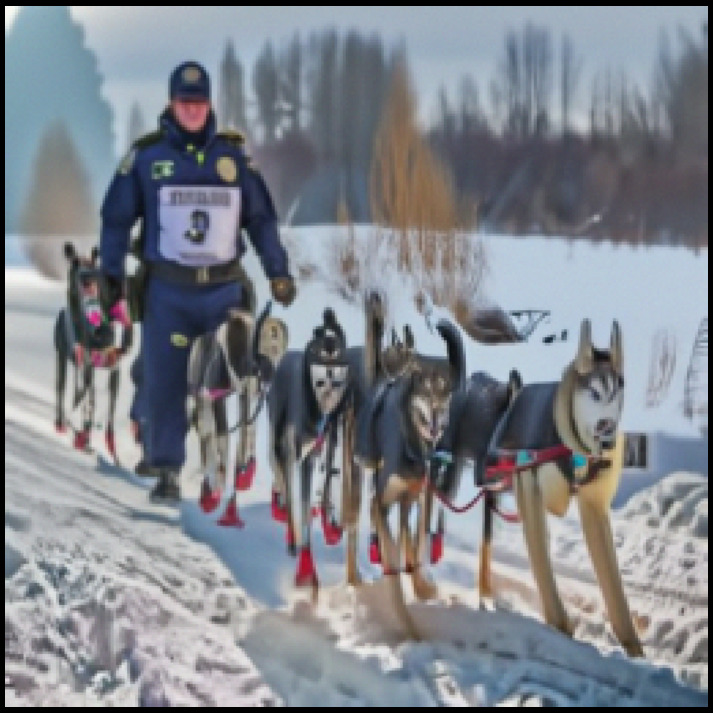} \\
        \textbf{SD} / 0.65 / 0.36 & \textbf{Imagen} / 0.63 / 0.37 & \textbf{Flux} / 0.25 / 0.29 & \textbf{Imagen-3} / 0.61 / 0.37 \\
        \includegraphics[width=0.23\textwidth, height=2.8cm]{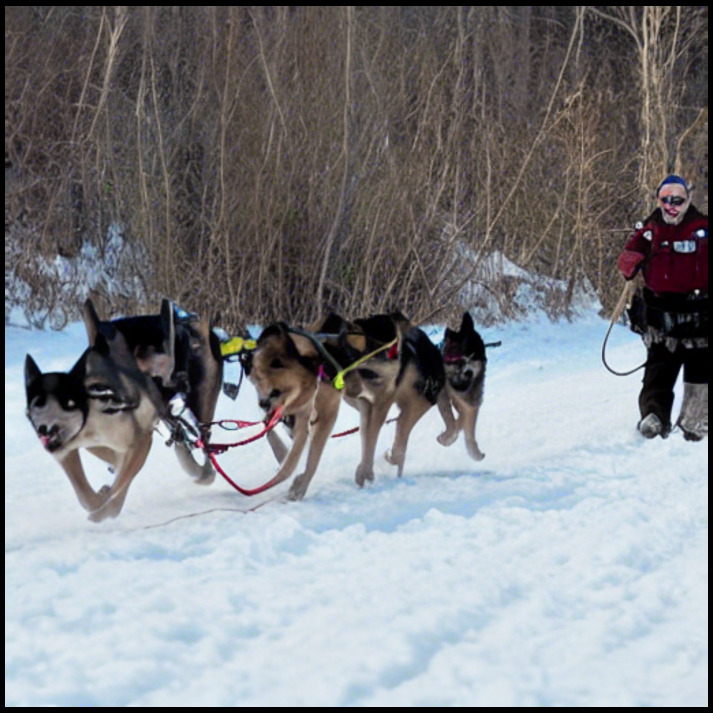} & 
        \includegraphics[width=0.23\textwidth, height=2.8cm]{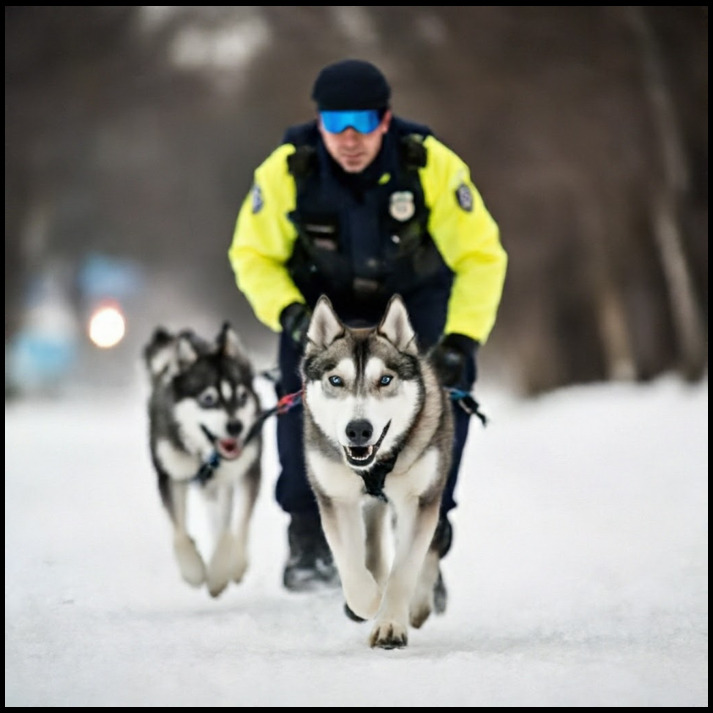} & 
        \includegraphics[width=0.23\textwidth, height=2.8cm]{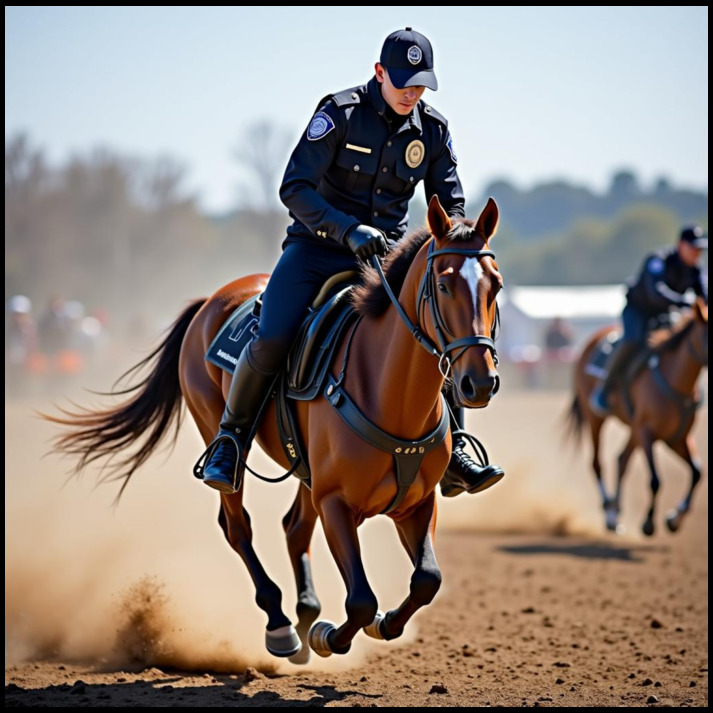} & 
        \includegraphics[width=0.23\textwidth, height=2.8cm]{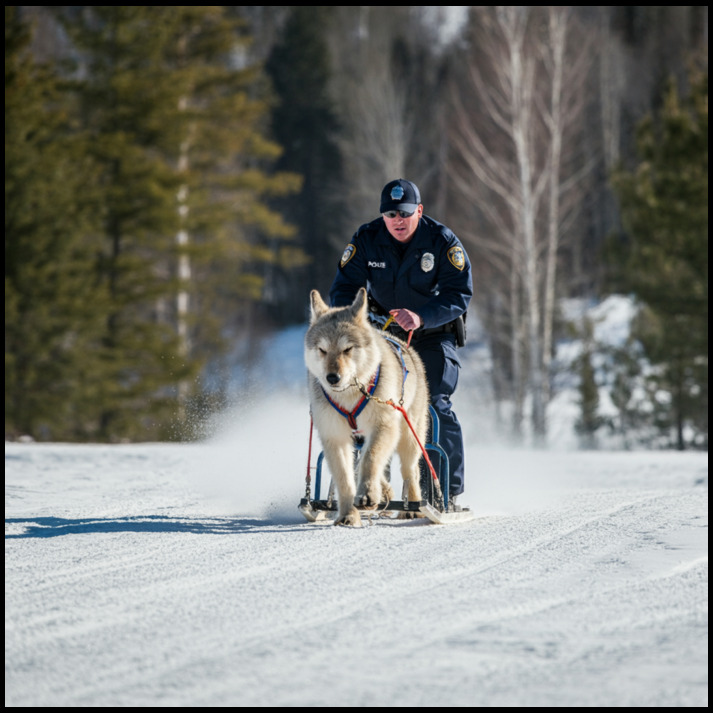} \\
        \multicolumn{4}{c}{\textit{A police officer doing Mushing.}} \\
        \midrule
        \textbf{Real Photo} & \textbf{Custom-Diff} / 0.55 / 0.35 & \textbf{DreamBooth} / 0.56 / 0.35 & \textbf{Instruct-Imagen} / 0.55 / 0.36 \\
        \includegraphics[width=0.23\textwidth, height=2.8cm]{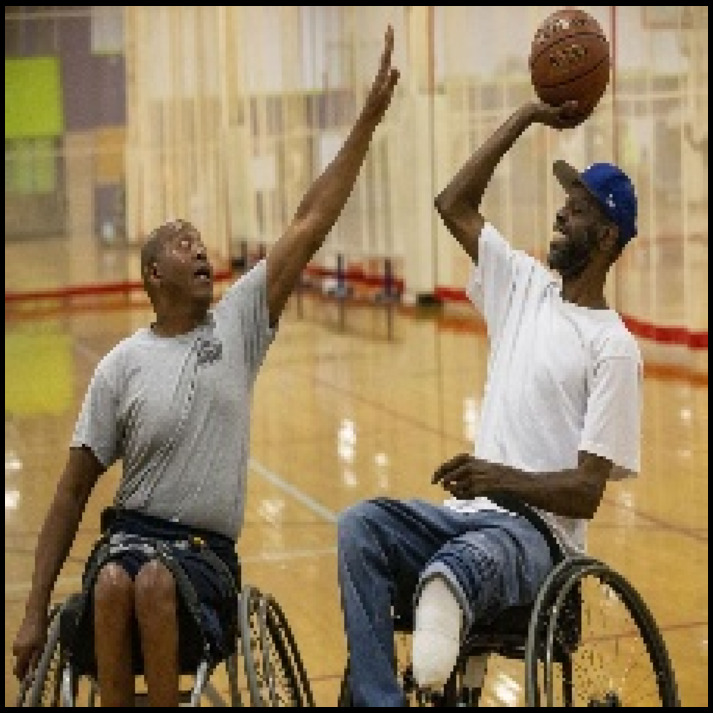} & 
        \includegraphics[width=0.23\textwidth, height=2.8cm]{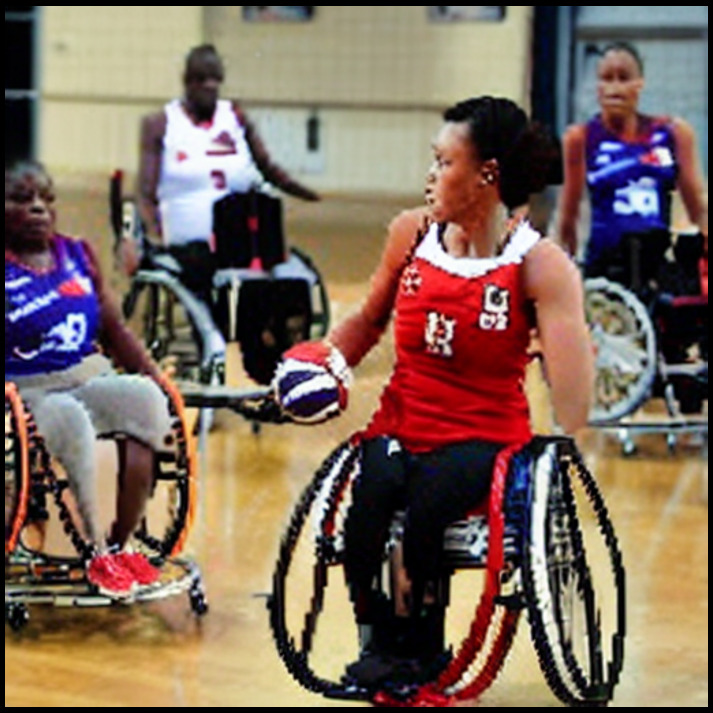} & 
        \includegraphics[width=0.23\textwidth, height=2.8cm]{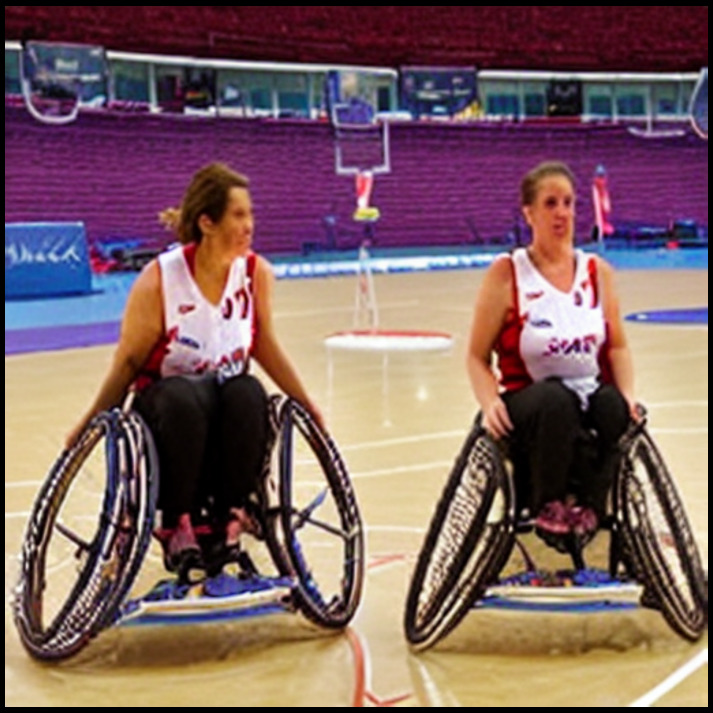} & 
        \includegraphics[width=0.23\textwidth, height=2.8cm]{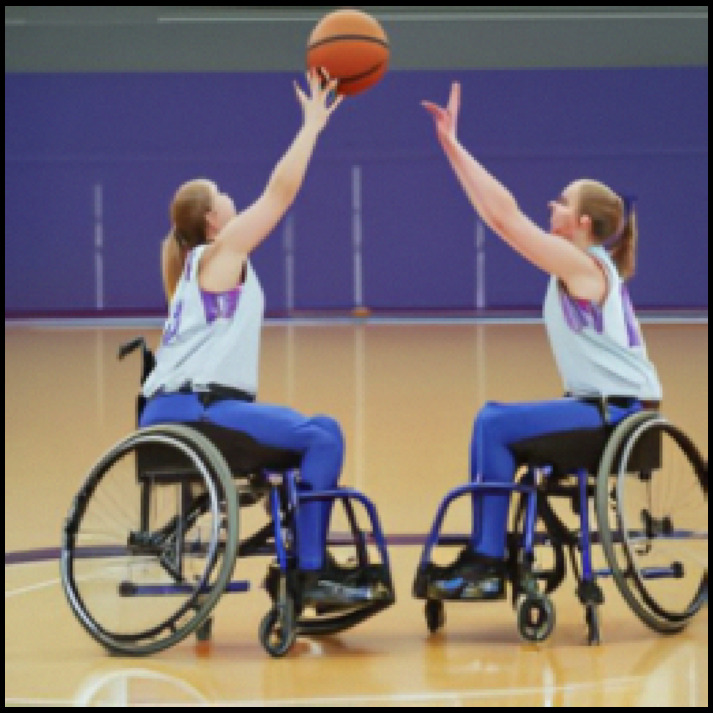} \\
        \textbf{SD} / 0.48 / 0.37 & \textbf{Imagen} / 0.46 / 0.37 & \textbf{Flux} / 0.56 / 0.38 & \textbf{Imagen-3} / 0.51 / 0.38 \\
        \includegraphics[width=0.23\textwidth, height=2.8cm]{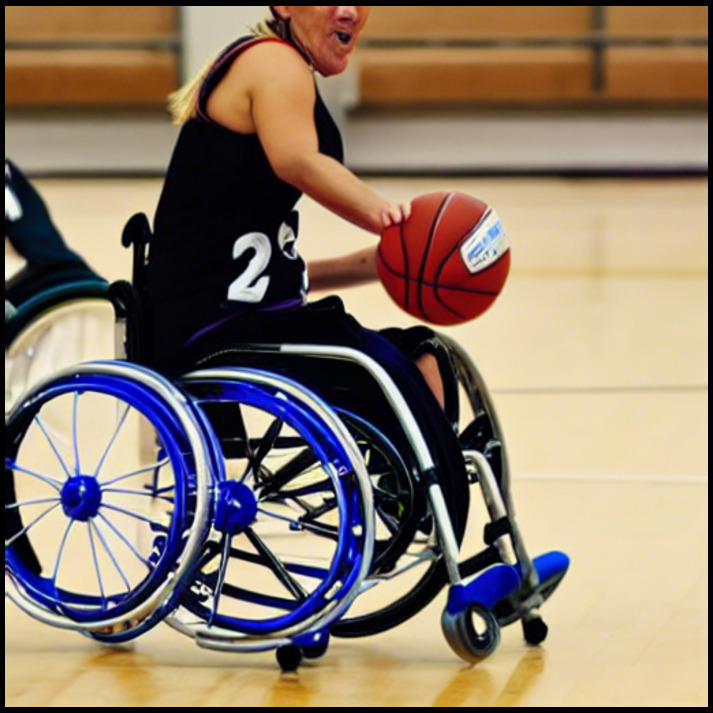} & 
        \includegraphics[width=0.23\textwidth, height=2.8cm]{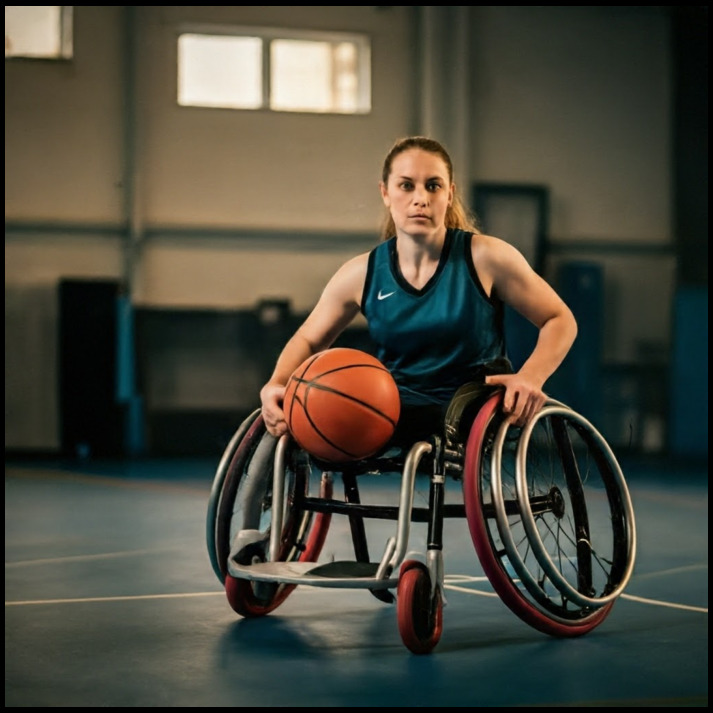} & 
        \includegraphics[width=0.23\textwidth, height=2.8cm]{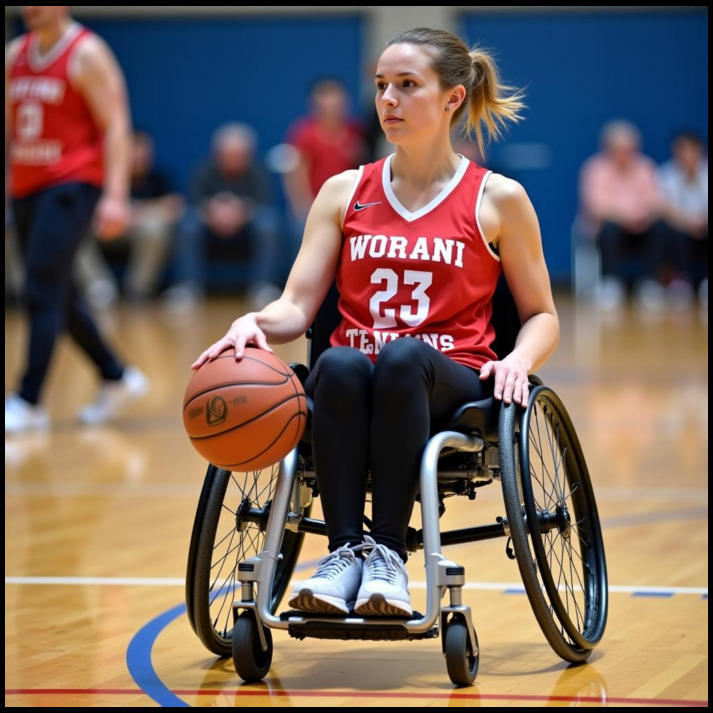} & 
        \includegraphics[width=0.23\textwidth, height=2.8cm]{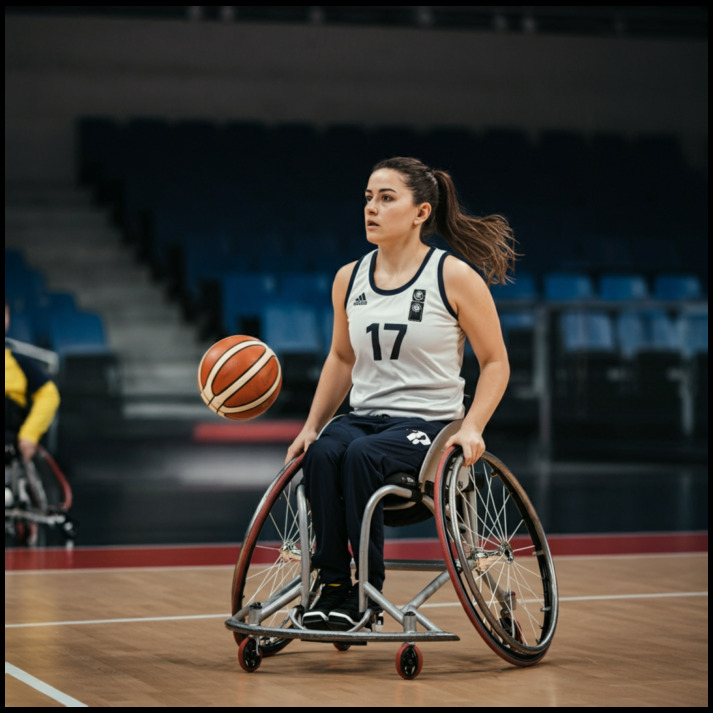} \\
        \multicolumn{4}{c}{\textit{A woman doing Wheelchair basketball.}} \\
        \bottomrule
    \end{tabular}
    \caption{\textbf{Qualitative results} for the sport domain, including the DINO and CLIP-T scores.}
    \label{fig:image_comparison_8}
\end{figure}

We present additional qualitative examples evaluating the backbone models---SD, Imagen, Flux, and Imagen-3---as well as the retrieval models---Custom-Diff, DreamBooth, and Instruct-Imagen---across various domains in the \ourdata benchmark. The results for the aircraft, vehicle, cuisine, flower, insect, landmark, plant, and sport domains are shown in~\cref{fig:image_comparison_1},~\cref{fig:image_comparison_2},~\cref{fig:image_comparison_3},~\cref{fig:image_comparison_4},~\cref{fig:image_comparison_5},~\cref{fig:image_comparison_6},~\cref{fig:image_comparison_7},~\cref{fig:image_comparison_8}, respectively.

\cref{fig:image_comparison_3} presents the results for the cuisine domain. For the prompt, ``A Wakame dish with a cherry flower on top of it,'' retrieval-based models such as Custom-Diff, DreamBooth, and Instruct-Imagen successfully generate the entity Wakame. However, Custom-Diff fails to capture the composition of the cherry flower. In contrast, the backbone models—Imagen, Flux, and Imagen-3—correctly generate the cherry flower, but the representation of Wakame does not align with its real-world appearance. In the example prompt, ``photo of a Hot and sour soup,'' the backbone models introduce incorrect ingredients that significantly deviate from a real-world hot and sour soup, demonstrating how these models hallucinate entities instead of reproducing real-world knowledge accurately.

\cref{fig:image_comparison_5} shows the results for the insect domain. For the prompt, ``Satyrium liparops sitting at the beach with a view of the sea,'' all models correctly generate the beach scene as instructed. However, the backbone models hallucinate the insect's appearance, generating a completely different insect, while the retrieval models accurately depict the visual details of the Satyrium liparops. For the prompt, ``Promachus hinei wearing sunglasses,'' the backbone models demonstrate strong instruction-following ability, with Imagen and Imagen-3 correctly composing the sunglasses on the insect. In contrast, the retrieval models fail to generate this composition. However, the backbone models do not generate an accurate representation of the insect itself, while the retrieval models correctly generate the entity.

\cref{fig:image_comparison_7} presents the results for the plant domain. In the example, ``An impressionistic painting of Cirsium andersonii,'' none of the models effectively follow the prompt in generating the "impressionistic painting," highlighting the challenge of rendering specific artistic styles. However, the retrieval models generate the entity more accurately than the backbone models. In another example, ``Penstemon rydbergii on a rustic wooden table next to a rose plant,'' although the generated entity lacks the fine details of the reference, the backbone models successfully capture the composition of "next to a rose," demonstrating their stronger instruction-following ability.

These observations align with the quantitative results, further demonstrating that while advanced backbone models can generate specialized entities, there is a notable gap in the fidelity of entities compared to retrieval-augmented models. Although retrieval-augmented methods improve faithfulness to entities, they often struggle with creative prompts, highlighting the need for techniques that enhance entity accuracy without compromising instruction-following abilities.

In addition, we have provided examples illustrating a balance between entity fidelity and creative flexibility in the generated images. Specifically, Instruct-Imagen's results for prompts such as "A Wakame dish with a cherry flower on top of it" in~\cref{fig:image_comparison_3}, "Alstroemeria on top of a mountain with sunrise in the background" in~\cref{fig:image_comparison_4}, and "Satyrium liparops sitting at the beach with a view of the sea" in~\cref{fig:image_comparison_5} demonstrate enhanced entity fidelity while maintaining creative flexibility. These examples are supported by the highest alignment scores across both metrics when compared to other models.

\section{Details of the Human Evaluation Scores Across Domains}
\label{app:human_eval}
We have presented the human evaluation results for backbone and retrieval-augmented text-to-image models, comparing their performance in terms of entity faithfulness and instruction-following accuracy.
Here, we provide a detailed breakdown of the human evaluation results across eight domains of the~\ourdata benchmark in~\cref{tab:instruction_he}.

We observe that the performance of each method varies across domains. For instance, while the overall faithfulness scores of Custom-Diff and DreamBooth are lower than the backbone model Imagen-3, these retrieval models consistently perform better in the insect, landmark, and plant domains.
When comparing backbone models, Flux outperforms SD in the overall score; however, SD demonstrates higher faithfulness in the insect, landmark, and plant domains, suggesting that Flux struggles to generate less frequent entities.
Additionally, the retrieval models, DreamBooth and Custom-Diff, improve faithfulness over SD in some cases but show reduced performance in the cuisine and sport domains.
In terms of instruction-following score, Instruct-Imagen performs notably lower in the flower (31.3) and insect (21.9) domains, while Imagen-3 achieves significantly higher scores in the cuisine (93.8) and sport (96.9) domains.

\begin{table}[!htp]
    \centering
    \small
    \caption{Detailed breakdown of human evaluation.}
    \vspace{2mm}
    \tabcolsep 5pt
    { 
    \begin{tabular}{lccccccccc}
    \toprule
     Model & Aircraft & Vehicle & Cuisine & Flower & Insect & Landmark & Plant & Sport & Overall \\
    \midrule
    \multicolumn{10}{c}{Faithfulness to entity} \\
    \midrule
    \BackboneMarker{} SD & 2.07 & 3.57 & 3.42 & 2.36 & 1.61 & 1.95 & 2.09 & 3.02 & 2.51 \\
    \BackboneMarker{} Imagen & 3.19 & 3.62 & 3.55 & 3.21 & 1.54 & 2.11 & 2.04 & 3.3 & 2.82 \\
    \BackboneMarker{} Flux & 3.51 & 3.89 & 3.35 & 2.52 & 1.44 & 1.77 & 1.83 & 3.59 & 2.74 \\
    \BackboneMarker{} Imagen-3 & 3.74 & 4.04 & 4.26 & 3.37 & 1.63 & 2.31 & 1.98 & 4.04 & 3.17 \\
    \midrule
    \RetrievalMarker{} Custom-Diff & 2.59 & 3.90 & 2.52 & 3.82 & 2.37 & 3.02 & 2.85 & 2.10 & 2.90  \\
    \RetrievalMarker{} DreamBooth & 3.26 & 3.88 & 3.23 & 3.18 & 2.46 & 3.06 & 2.64 & 2.93 & 3.08 \\
    \RetrievalMarker{} Instruct-Imagen & 4.23 & 4.70 & 4.43 & 4.48 & 3.78 & 4.06 & 4.02 & 4.08 & \textbf{4.22} \\
    \midrule
    \multicolumn{10}{c}{Instruction-following} \\
    \midrule
    \BackboneMarker{} SD & 68.8 & 75.0 & 62.5 & 87.5 & 62.5 & 65.6 & 84.4 & 71.9 & 72.2 \\
    \BackboneMarker{} Imagen & 59.4 & 84.4 & 75.0 & 84.4 & 53.1 & 59.4 & 84.4 & 78.1 & 72.2  \\
    \BackboneMarker{} Flux & 56.3 & 87.5 & 96.9 & 78.1 & 59.4 & 71.9 & 84.4 & 81.3 & 76.9 \\
    \BackboneMarker{} Imagen-3 & 78.1 & 87.5 & 93.8 & 84.4 & 71.9 & 71.9 & 84.4 & 96.9 & 83.6  \\
     \midrule
    \RetrievalMarker{} Custom-Diff & 59.4 & 75.0 & 65.6 & 78.1 & 71.9 & 65.6 & 81.3 & 59.4 & 69.5 \\
    \RetrievalMarker{} DreamBooth & 68.8 & 71.9 & 71.9 & 81.3 & 68.8 & 75.0 & 78.1 & 75.0 & 73.8 \\
    \RetrievalMarker{} Instruct-Imagen & 46.9 & 56.3 & 56.3 & 31.3 & 21.9 & 50.0 & 59.4 & 50.0 & 46.5 \\
    \bottomrule
    \end{tabular}} \\
    (\BackboneMarker{}: Text-to-Image Models, \RetrievalMarker{}: Retrieval-augmented Models)
    \label{tab:instruction_he}
\end{table}

\section{Details of the Automatic Metric Scores Across Domains}
\label{app:auto_metrics}

We have presented the automatic metric evaluation for backbone and retrieval-augmented text-to-image models, including image-text alignment and image-entity alignment scores.
Here, we include a detailed breakdown of the automatic metrics across eight domains of the~\ourdata benchmark in~\cref{tab:instruction_metrics_all}. Specifically, we include a detailed ablation study of the image-entity alignment metric using cosine similarity scores based on two popular image features: the CLIP-Image and the DINO features.

The retrieval models---Custom-Diff, DreamBooth, and Instruct-Imagen---consistently outperform the backbone models in DINO score across the insect, landmark, and plant domains. In contrast, Custom-Diff shows notably worse DINO and CLIP-T scores in the cuisine and sport domains. These results are consistent with human evaluations.

\begin{table}[!htp]
    \centering
    \small
    \caption{Detailed breakdown of automatic metrics.}
    \vspace{2mm}
    \tabcolsep 5pt
    { 
    \begin{tabular}{lccccccccc}
    \toprule
     Model & Aircraft & Vehicle & Cuisine & Flower & Insect & Landmark & Plant & Sport & Overall \\
    \midrule
    \multicolumn{10}{c}{Image-Text Alignment: CLIP-T} \\
    \midrule
    \BackboneMarker{} SD & 0.341 & 0.354 & 0.343 & 0.336 & 0.322 & 0.330 & 0.332 & 0.342 & 0.338 \\
    \BackboneMarker{} Imagen & 0.329 & 0.344 & 0.338 & 0.337 & 0.323 & 0.312 & 0.316 & 0.335 & 0.329 \\
    \BackboneMarker{} Flux & 0.341 & 0.346 & 0.329 & 0.327 & 0.320 & 0.308 & 0.314 & 0.343 & 0.329 \\
    \BackboneMarker{} Imagen-3 & 0.345 & 0.351 & 0.352 & 0.343 & 0.325 & 0.315 & 0.324 & 0.352 & 0.338 \\
    \midrule
    \RetrievalMarker{} Custom-Diff & 0.333 & 0.349 & 0.314 & 0.339 & 0.321 & 0.315 & 0.326 & 0.298 & 0.324 \\
    \RetrievalMarker{} DreamBooth & 0.333 & 0.347 & 0.335 & 0.333 & 0.322 & 0.318 & 0.324 & 0.335 & 0.331 \\
    \RetrievalMarker{} Instruct-Imagen & 0.293 & 0.324 & 0.315 & 0.320 & 0.295 & 0.294 & 0.302 & 0.316 & 0.307 \\
    \midrule
    \multicolumn{10}{c}{Image-Entity Alignment: CLIP-I} \\
    \midrule
    \BackboneMarker{} SD & 0.640 & 0.655 & 0.673 & 0.719 & 0.619 & 0.616 & 0.628 & 0.618 & 0.646 \\
    \BackboneMarker{} Imagen & 0.678 & 0.650 & 0.671 & 0.733 & 0.630 & 0.632 & 0.630 & 0.545 & 0.646 \\
    \BackboneMarker{} Flux & 0.652 & 0.653 & 0.663 & 0.738 & 0.609 & 0.614 & 0.618 & 0.561 & 0.639 \\
    \BackboneMarker{} Imagen-3 & 0.677 & 0.656 & 0.665 & 0.726 & 0.623 & 0.638 & 0.632 & 0.582 & 0.650 \\
    \midrule
    \RetrievalMarker{} Custom-Diff & 0.688 & 0.678 & 0.593 & 0.759 & 0.619 & 0.655 & 0.662 & 0.491 & 0.643 \\
    \RetrievalMarker{} DreamBooth & 0.698 & 0.676 & 0.692 & 0.742 & 0.660 & 0.681 & 0.657 & 0.589 & 0.674 \\
    \RetrievalMarker{} Instruct-Imagen & 0.776 & 0.719 & 0.746 & 0.825 & 0.810 & 0.735 & 0.751 & 0.648 & 0.751 \\
    \midrule
    \multicolumn{10}{c}{Image-Entity Alignment: DINO} \\
    \midrule
    \BackboneMarker{} SD & 0.449 & 0.540 & 0.367 & 0.386 & 0.181 & 0.306 & 0.189 & 0.379 & 0.350 \\
    \BackboneMarker{} Imagen & 0.610 & 0.559 & 0.359 & 0.469 & 0.198 & 0.353 & 0.207 & 0.331 & 0.386 \\
    \BackboneMarker{} Flux & 0.524 & 0.576 & 0.382 & 0.477 & 0.153 & 0.343 & 0.218 & 0.369 & 0.380 \\
    \BackboneMarker{} Imagen-3 & 0.588 & 0.590 & 0.351 & 0.435 & 0.194 & 0.366 & 0.215 & 0.369 & 0.389 \\
    \midrule
    \RetrievalMarker{} Custom-Diff & 0.555 & 0.576 & 0.285 & 0.490 & 0.263 & 0.382 & 0.267 & 0.288 & 0.388 \\
    \RetrievalMarker{} DreamBooth & 0.580 & 0.567 & 0.406 & 0.435 & 0.280 & 0.407 & 0.250 & 0.371 & 0.412 \\
    \RetrievalMarker{} Instruct-Imagen & 0.758 & 0.711 & 0.539 & 0.646 & 0.526 & 0.545 & 0.450 & 0.482 & 0.582 \\
    \bottomrule
    \end{tabular}} \\
    (\BackboneMarker{}: Text-to-Image Models, \RetrievalMarker{}: Retrieval-augmented Models)
    \label{tab:instruction_metrics_all}
\end{table}

\section{Details of the MLLM Evaluation Scores Across Domains}

We report MLLM evaluation results for both the backbone and retrieval-augmented text-to-image models, including scores for text alignment and entity alignment.
\cref{tab:domain_mllm} presents a detailed breakdown of these scores across the eight domains of the~\ourdata benchmark.
Notably, retrieval-augmented models consistently outperform backbone models in entity alignment within the vehicle, insect, landmark, and plant domains. In contrast, Custom-Diff shows substantially lower entity alignment scores in the cuisine and sport domains.
\begin{table}[!htp]
\centering
    \small
    \caption{Detailed breakdown of MLLM evaluation.}
    \vspace{2mm}
    \tabcolsep 5pt
    { 
    \begin{tabular}{lccccccccc}
    \toprule
     Model & Aircraft & Vehicle & Cuisine & Flower & Insect & Landmark & Plant & Sport & Overall \\
    \midrule
    \multicolumn{10}{c}{MLLM Text Alignment} \\
    \midrule
    \BackboneMarker{} SD & 3.06 & 3.62 & 2.97 & 4.22 & 2.84 & 3.78 & 3.69 & 3.47 & 3.46 \\
    \BackboneMarker{} Imagen & 3.16 & 3.78 & 3.75 & 4.12 & 2.88 & 3.53 & 3.94 & 3.75 & 3.61 \\
    \BackboneMarker{} Flux & 3.31 & 4.31 & 4.28 & 4.12 & 3.5 & 3.94 & 4.44 & 3.75 & 3.96 \\
    \BackboneMarker{} Imagen-3 & 3.41 & 4.25 & 4.25 & 4.38 & 4.09 & 3.88 & 4.72 & 4.41 & 4.17 \\
    \midrule
    \RetrievalMarker{} Custom-Diff & 3.22 & 3.5 & 3.53 & 3.72 & 2.66 & 3.44 & 3.47 & 2.75 & 3.29 \\
    \RetrievalMarker{} DreamBooth & 3.16 & 3.66 & 3.31 & 3.88 & 2.94 & 3.53 & 3.41 & 3.09 & 3.37 \\
    \RetrievalMarker{} Instruct-Imagen & 2.53 & 3.25 & 2.75 & 2.41 & 1.91 & 2.94 & 3.0 & 2.28 & 2.63 \\
    \midrule
    \multicolumn{10}{c}{MLLM Entity Alignment} \\
    \midrule
    \BackboneMarker{} SD & 2.34 & 3.12 & 3.09 & 2.28 & 1.22 & 1.97 & 1.69 & 3.44 & 2.39 \\
    \BackboneMarker{} Imagen & 2.81 & 3.16 & 3.25 & 2.72 & 1.12 & 2.41 & 1.59 & 3.19 & 2.53 \\
    \BackboneMarker{} Flux & 2.41 & 2.97 & 2.47 & 1.75 & 1.12 & 1.72 & 1.22 & 2.69 & 2.04 \\
    \BackboneMarker{} Imagen-3 & 3.19 & 3.34 & 4.06 & 3.03 & 1.31 & 2.25 & 1.69 & 3.75 & 2.83 \\
    \midrule
    \RetrievalMarker{} Custom-Diff & 2.94 & 3.41 & 1.78 & 3.28 & 1.88 & 3.34 & 2.91 & 2.03 & 2.7 \\
    \RetrievalMarker{} DreamBooth & 3.12 & 3.5 & 2.66 & 2.59 & 1.91 & 3.53 & 2.59 & 2.66 & 2.82 \\
    \RetrievalMarker{} Instruct-Imagen & 3.88 & 4.06 & 3.91 & 4.03 & 3.22 & 4.09 & 3.31 & 3.25 & 3.72 \\
    \bottomrule
    \end{tabular}} \\
    (\BackboneMarker{}: Text-to-Image Models, \RetrievalMarker{}: Retrieval-augmented Models)
    \label{tab:domain_mllm}
\end{table}

\section{Details of the Human Evaluation Scores Across Prompts}
We provide a detailed breakdown of human evaluation results across five evaluation tasks in the~\ourdata benchmark, as shown in~\cref{tab:prompt_human}.
For the faithfulness score, the ranking across prompts remains consistent with the overall ranking. Instruct-Imagen achieves the highest score, followed by Imagen-3. The retrieval-based methods, DreamBooth and Custom-Diffusion, come next, while the other base models—Imagen, Flux, and Stable-Diffusion—score comparatively lower.

The instruction-following scores across prompts align with the average performance, with Imagen-3 emerging as the top performer. Notably, Flux and Imagen-3 excel in \textit{Location}, achieving scores of 90 and 87.8, respectively, whereas Imagen and other models fall below 80. Similarly, Imagen-3 demonstrates strong performance in \textit{Composition}, with a score of 94.6, highlighting its robust instruction-following capabilities in these aspects. In contrast, DreamBooth leads in \textit{Style} with the highest score of 90.2, followed by Custom-Diffusion (84.3) and Stable-Diffusion (80.4), showcasing the strong ability of Stable Diffusion and its retrieval-augmented variants to generate accurate styles. In the \textit{Material} category, all models struggle with instruction adherence, though Imagen-3 achieves a relatively higher score.

\begin{table}[!htp]
    \centering
    \small
    \caption{Detailed breakdown of human evaluation.}
    \vspace{2mm}
    \tabcolsep 5pt
    { 
    \begin{tabular}{lcccccc}
    \toprule
     Model & Basic & Location & Composition & Style & Material & Overall \\
    \midrule
    \multicolumn{7}{c}{Faithfulness to entity} \\
    \midrule
    \BackboneMarker{} SD & 3.08 & 2.87 & 2.03 & 2.16 & 2.65 & 2.51 \\
    \BackboneMarker{} Imagen & 3.96 & 3.16 & 2.38 & 2.53 & 2.76 & 2.82 \\
    \BackboneMarker{} Flux & 3.50 & 3.04 & 2.67 & 2.32 & 2.54 & 2.74 \\
    \BackboneMarker{} Imagen-3 & 4.08 & 3.55 & 2.78 & 2.79 & 3.13 & 3.17 \\
    \midrule
    \RetrievalMarker{} Custom-Diff &  4.34 & 2.77 & 2.73 & 3.12 & 2.80 & 2.90 \\
    \RetrievalMarker{} DreamBooth &  4.18 & 3.33 & 2.72 & 2.95 & 2.94 & 3.08 \\
    \RetrievalMarker{} Instruct-Imagen & 4.90 & 4.18 & 4.16 & 4.23 & 4.21 & 4.22 \\
    \midrule
    \multicolumn{7}{c}{Instruction-following} \\
    \midrule
    \BackboneMarker{} SD & 100.0 & 80.0 & 64.3 & 80.4 & 53.1 & 72.2 \\
    \BackboneMarker{} Imagen & 90.0 & 75.6 & 80.4 & 66.7 & 59.2 & 72.2  \\
    \BackboneMarker{} Flux & 90.0 & 90.0 & 83.9 & 58.8 & 61.2 & 76.9 \\
    \BackboneMarker{} Imagen-3 & 90.0 & 87.8 & 94.6 & 76.5 & 69.4 & 83.6\\
    \midrule
    \RetrievalMarker{} Custom-Diff &  100.0 & 72.2 & 66.1 & 84.3 & 46.9 & 69.5 \\
    \RetrievalMarker{} DreamBooth & 90.0 & 77.8 & 67.9 & 90.2 & 53.1 & 73.8 \\
    \RetrievalMarker{} Instruct-Imagen & 100.0 & 46.7 & 58.9 & 47.1 & 20.4 & 46.5 \\
    \bottomrule
    \end{tabular}} \\
    (\BackboneMarker{}: Text-to-Image Models, \RetrievalMarker{}: Retrieval-augmented Models)
    \label{tab:prompt_human}
\end{table}

\section{Details of the Automatic Metric Scores Across Prompts}
We provide a detailed breakdown of automatic metrics across five evaluation tasks in the~\ourdata benchmark, as shown in~\cref{tab:prompt_auto}.
We observe that \textit{Location} achieves the highest average image-text score (0.334), followed by \textit{Composition} (0.330), \textit{Style} (0.326), and \textit{Material} (0.319). This pattern suggests that generating entities within a given context is relatively easier, whereas accurately modifying materials remains more challenging.
In the categories of \textit{Location}, \textit{Composition}, and \textit{Material}, the base models tend to achieve higher image-text scores, with Imagen-3 attaining the highest score among them. Among retrieval-based models, DreamBooth outperforms Custom Diffusion and Instruct-Imagen. Notably, in \textit{Material}, Imagen scores lower (0.317) than DreamBooth (0.319) and Custom Diffusion (0.320), indicating greater difficulty for Imagen in this category.

For the image-entity score, we find that \textit{Location} again ranks highest with an average score of 0.431, followed by \textit{Material} (0.417), \textit{Style} (0.399), and \textit{Composition} (0.364). This ranking highlights the challenge of maintaining entity fidelity when prompts require complex compositions.
In the categories of \textit{Location}, \textit{Composition}, and \textit{Material}, retrieval-augmented models generally outperform base models, with Instruct-Imagen achieving the highest score, followed by DreamBooth and Custom Diffusion. Imagen-3 and Flux also surpass their respective predecessors, Imagen and Stable-Diffusion. However, in an unexpected result for \textit{Composition}, Imagen (0.333) outperforms Imagen-3 (0.317).

While most categories exhibit higher image-entity scores for \textit{Material} compared to \textit{Style}, Imagen-3 and Flux surprisingly perform better on styles. Additionally, other models tend to achieve higher image-entity scores on \textit{Style} than on \textit{Composition}, with DreamBooth being an exception. These results highlight the varying strengths of different models in preserving entity fidelity across diverse prompt types.

\begin{table}[!htp]
    \centering
    \small
    \caption{Detailed breakdown of automatic metrics.}
    \vspace{2mm}
    \tabcolsep 5pt
    { 
    \begin{tabular}{lcccccc}
    \toprule
     Model & Basic & Location & Composition & Style & Material & Overall \\
    \midrule
    \multicolumn{7}{c}{Image-Text Alignment: CLIP-T} \\
    \midrule
    \BackboneMarker{} SD &  0.308 & 0.342 & 0.333 & 0.346 & 0.332 & 0.338 \\
    \BackboneMarker{} Imagen &  0.304 & 0.338 & 0.329 & 0.328 & 0.317 & 0.329 \\
    \BackboneMarker{} Flux & 0.299 & 0.340 & 0.340 & 0.308 & 0.324 & 0.329 \\
    \BackboneMarker{} Imagen-3 & 0.303 & 0.347 & 0.345 & 0.322 & 0.333 & 0.338 \\
    \midrule
    \RetrievalMarker{} Custom-Diff & 0.308 & 0.326 & 0.324 & 0.335 & 0.320 & 0.324 \\
    \RetrievalMarker{} DreamBooth & 0.308 & 0.336 & 0.327 & 0.341 & 0.319 & 0.331 \\
    \RetrievalMarker{} Instruct-Imagen & 0.302 & 0.312 & 0.311 & 0.305 & 0.291 & 0.307 \\
    \midrule
    \multicolumn{7}{c}{Image-Entity Alignment: DINO} \\
    \midrule
    \BackboneMarker{} SD & 0.467 & 0.379 & 0.295 & 0.311 & 0.349 & 0.350 \\
    \BackboneMarker{} Imagen & 0.472 & 0.409 & 0.333 & 0.376 & 0.385 & 0.386 \\
    \BackboneMarker{} Flux & 0.464 & 0.394 & 0.316 & 0.383 & 0.368 & 0.380 \\
    \BackboneMarker{} Imagen-3 &  0.483 & 0.412 & 0.317 & 0.403 & 0.391 & 0.389 \\
    \midrule
    \RetrievalMarker{} Custom-Diff & 0.578 & 0.411 & 0.348 & 0.359 & 0.400 & 0.388 \\
    \RetrievalMarker{} DreamBooth &  0.597 & 0.440 & 0.376 & 0.371 & 0.417 & 0.412 \\
    \RetrievalMarker{} Instruct-Imagen & 0.626 & 0.573 & 0.566 & 0.593 & 0.606 & 0.582 \\
    \bottomrule
    \end{tabular}} \\
    (\BackboneMarker{}: Text-to-Image Models, \RetrievalMarker{}: Retrieval-augmented Models)
    \label{tab:prompt_auto}
\end{table}

\section{Details of the MLLM Evaluation Across Prompts}

We provide a detailed breakdown of MLLM evaluation results across five evaluation tasks in the~\ourdata benchmark, as shown in~\cref{tab:prompt_mllm}.
For MLLM text alignment, the ranking across prompt types largely mirrors the overall trend. Imagen-3 consistently achieves the highest overall score (4.17), followed by Flux (3.96) and Imagen (3.61). Among retrieval-augmented models, DreamBooth and Custom-Diffusion perform moderately (3.37 and 3.29, respectively), while Instruct-Imagen scores the lowest (2.63). Imagen-3 leads across most prompt types, particularly excelling in \textit{Composition} (4.59) and \textit{Material} (3.78). Flux also performs well, especially in \textit{Location} (4.31) and \textit{Composition} (4.23). In contrast, Custom-Diffusion and DreamBooth show weaker alignment, particularly in \textit{Composition} and \textit{Material}, with scores below 3.5.
The MLLM entity alignment scores follow a similar pattern, with Instruct-Imagen achieving the highest overall score (3.72), outperforming all other models across all prompt types. It particularly excels in \textit{Composition} (3.91), \textit{Style} (3.67), and \textit{Material} (3.65), demonstrating strong grounding of entity representations in the image. Among the base models, Imagen-3 performs best (2.83), while Flux consistently underperforms across all prompts. These results further underscore the effectiveness of retrieval augmentation for enhancing entity alignment, especially for complex instructions involving \textit{Style} and \textit{Material}.

\begin{table}[!htp]
    \centering
    \small
    \caption{Detailed breakdown of MLLM evaluation.}
    \vspace{2mm}
    \tabcolsep 5pt
    { 
    \begin{tabular}{lcccccc}
    \toprule
     Model & Basic & Location & Composition & Style & Material & Overall \\
    \midrule
    \multicolumn{7}{c}{MLLM Text Alignment} \\
    \midrule
    \BackboneMarker{} SD &  4.6 & 3.61 & 3.3 & 3.86 & 2.69 & 3.46\\
    \BackboneMarker{} Imagen &  5.0 & 3.7 & 3.86 & 3.45 & 3.06 & 3.61\\
    \BackboneMarker{} Flux & 5.0 & 4.31 & 4.23 & 3.35 & 3.41 & 3.96\\
    \BackboneMarker{} Imagen-3 & 5.0 & 4.28 & 4.59 & 3.75 & 3.78 & 4.17\\
    \midrule
    \RetrievalMarker{} Custom-Diff & 4.6 & 3.3 & 3.2 & 3.84 & 2.51 & 3.29\\
    \RetrievalMarker{} DreamBooth & 4.7 & 3.34 & 3.25 & 3.94 & 2.69 & 3.37\\
    \RetrievalMarker{} Instruct-Imagen & 4.6 & 2.6 & 2.86 & 2.82 & 1.84 & 2.63\\

    \midrule
    \multicolumn{7}{c}{MLLM Entity Alignment} \\
    \midrule
    \BackboneMarker{} SD & 2.9 & 2.87 & 2.16 & 1.84 & 2.27 & 2.39\\
    \BackboneMarker{} Imagen & 3.2 & 2.89 & 2.48 & 1.94 & 2.41 & 2.53\\
    \BackboneMarker{} Flux & 2.0 & 2.24 & 2.43 & 1.71 & 1.59 & 2.04 \\
    \BackboneMarker{} Imagen-3 &  3.82 & 3.18 & 2.63 & 2.41 & 2.63 & 2.83\\
    \midrule
    \RetrievalMarker{} Custom-Diff & 3.7 & 2.7 & 2.57 & 2.8 & 2.51 & 2.7\\
    \RetrievalMarker{} DreamBooth &  4.1 & 3.09 & 2.55 & 2.45 & 2.76 & 2.82\\
    \RetrievalMarker{} Instruct-Imagen & 4.2 & 3.61 & 3.91 & 3.67 & 3.65 & 3.72\\
    \bottomrule
    \end{tabular}} \\
    (\BackboneMarker{}: Text-to-Image Models, \RetrievalMarker{}: Retrieval-augmented Models)
    \label{tab:prompt_mllm}
\end{table}

\section{Details of the Correlation with Human Evaluation}
\label{app:correlation}

To quantify the consistency between automatic metrics and human evaluation results, we computed Pearson and Spearman correlations. CLIP-T and CLIP-I show moderate alignment with user evaluations.
Here, we provide a detailed breakdown of the correlation across eight domains in~\cref{tab:correlation_per_category}.
While these metrics show some alignment with human evaluations, discrepancies remain in certain categories. Notably, the correlation between CLIP-T and the instruction-following score in the landmark domain, as well as the correlation between CLIP-I and the faithfulness score in the plant domain, is negative. This underscores the limitations of automatic metrics in fully capturing human judgment.
Although DINO demonstrates stronger alignment with user evaluations of faithfulness, correlation variability persists across categories. For instance, correlations are lower in the vehicle and landmark domains, highlighting the need for more accurate automatic metrics to better reflect human evaluations and assess model performance.

Additionally, both MLLM entity alignment and text alignment scores show higher correlations with human evaluations in most domains, indicating stronger consistency as automatic proxies. However, exceptions arise in the aircraft and vehicle domains, where the correlation values are noticeably lower. This suggests that while MLLM-based scores generally align well with human judgments, their effectiveness may be reduced in domains characterized by complex or diverse visual features, such as aircraft and vehicles. Further refinement of these metrics could enhance their robustness and reliability across all categories.

\begin{table}[htp]
    \centering
    \small
    \caption{Per-category correlation with human evaluation.}
    \vspace{2mm}
    \tabcolsep 3pt
    \begin{tabular}{ll|cccccccc}
        \toprule
        Metric & Type & Aircraft & Vehicle & Cuisine & Flower & Insect & Landmark & Plant & Average \\
        \midrule
        \multirow{2}{*}{CLIP-T}  & Pearson  & \textbf{0.138} & 0.499 & 0.105 & 0.788 & 0.567 & \textbf{-0.299} & 0.564 & 0.337 \\
                                & Spearman & \textbf{0.168} & 0.454 & 0.286 & 0.836 & 0.554 & \textbf{-0.166} & 0.558 & 0.384 \\
        \midrule
        \multirow{2}{*}{CLIP-I} & Pearson  & 0.330 & 0.195 & \textbf{0.100} & 0.215 & 0.528 & 0.356 & \textbf{-0.051} & 0.239 \\
                                & Spearman & 0.548 & 0.323 & \textbf{0.238} & 0.287 & 0.503 & 0.623 & \textbf{-0.143} & 0.340 \\
        \midrule
        \multirow{2}{*}{DINO} &  Pearson  & 0.655 & \textbf{0.367} & 0.549 & 0.551 & \textbf{0.277} & 0.680 & 0.492 & 0.510 \\
                                & Spearman & 0.575 & \textbf{0.311} & 0.735 & 0.430 & \textbf{0.210} & 0.700 & 0.565 & 0.504 \\
        \midrule
        \multirow{2}{*}{MLLM-T} & Pearson  & 0.477 & 0.687 & 0.583 & 0.673 & 0.586 & 0.790 & 0.575 & 0.618 \\
                                & Spearman & 0.478 & 0.605 & 0.536 & 0.604 & 0.577 & 0.748 & 0.510 & 0.589 \\

        \midrule
        \multirow{2}{*}{MLLM-I} & Pearson  & 0.468 & 0.467 & 0.754 & 0.783 & 0.746 & 0.742 & 0.705 & 0.703 \\
                                & Spearman & 0.475 & 0.393 & 0.711 & 0.764 & 0.686 & 0.748 & 0.704 & 0.695\\
        \bottomrule
    \end{tabular}
    \label{tab:correlation_per_category}
\end{table}

\section{Dataset Statistics}
\label{app:data_stats}

\ourdata benchmark focuses on evaluating faithfulness to knowledge-grounded concepts.
To ensure diversity, we select entities from eight specialized domains and construct diverse prompts in each domain for evaluation. For each entity, we collect a set of support images as inputs to assess retrieval-augmented models, where support images are used to enhance the model's predictions.
In addition, we collect a set of evaluation images for conducting human evaluation. 
A detailed breakdown of the data statistics is provided in~\cref{tab:data_stats}.

Next, we present the distribution of different evaluation tasks in~\cref{tab:prompts}. The types of evaluation tasks include: 1) generating the knowledge entity, 2) the knowledge entity in context, 3) the composition of entities, 4) creation in different styles, and 5) creation in different materials.

\begin{table}[!htp]
    \centering
    \small
    \caption{Statistics of~\ourdata benchmark.}
    \vspace{2mm}
    \begin{tabular}{l|c|c|c|c|c}
    \toprule
      Domain   &  \#Entities & \#Prompts & \#Support Images & \#Eval Images & \# (Entitiy, Prompt) \\
    \midrule
       Aircraft  & 48 & 20 & 469 & 237 & 960 \\
       Vehicle  & 50 & 20 & 500 & 250 & 1000 \\
       Flower  &  18 & 20 & 180 & 90 & 360 \\
       Insect & 50 & 20 & 500 & 250 & 1000\\
       Plant & 48 & 20 & 480 & 240 & 960 \\
       Landmark & 50 & 20 & 500 & 250 & 1000 \\
       Cuisine & 31 & 20 & 310 & 155 & 620 \\
       Sport & 27 & 20 & 270 & 135 & 540 \\
    \bottomrule
    \end{tabular}
    \label{tab:data_stats}
\end{table}

\begin{table}[htp]
    \centering
    \small
    \caption{Statistics of~\ourdata evaluation tasks.}
    \vspace{2mm}
    \begin{tabular}{l|c|c}
    \toprule
      Evaluation task & \#Prompts & Percentage (\%) \\
    \midrule
Basic &	295	& 4.58 \\
Location &	1969 &	30.57 \\
Composition &	1467 &	22.78 \\
Style &	1365 & 21.20 \\
Material &	1344 & 20.87\\
    \bottomrule
    \end{tabular}
    \label{tab:prompts}
\end{table}

\section{Details on Human Evaluation}
We employ five annotators per image to ensure robust assessments. The raters are hired through Prolific.com, a third-party rating service. For the binary task ("Adherence to Prompt Beyond References"), we observe agreement among at least 4 out of 5 annotators in 75\% of cases and perfect agreement (5 out of 5) in 44\% of cases. For the Likert scale task ("Faithfulness to Reference Entity"), we calculate Krippendorff’s Alpha at 0.60, indicating a good agreement for a subjective task of this complexity. Additionally, we achieve an IoU-like score of 0.53, which penalizes outliers and demonstrates moderate consensus, and an average pairwise Cohen’s Kappa of 0.25, reflecting fair pairwise agreement. The average standard deviation of ratings is 0.74, reflecting moderate variability in annotator judgments. These metrics collectively demonstrate the reliability of our human evaluation.

\section{Details on Evaluated Models}
To facilitate the reproducibility of the \textsc{KITTEN} benchmark experiments, we provide detailed descriptions of the training and inference setups for all evaluated models. For retrieval-augmented methods (DreamBooth and Custom-Diffusion), we follow their original implementations and fine-tune each model per entity using 10 reference images that are disjoint from the evaluation set. We use the AdamW optimizer with a learning rate of $5 \times 10^{-6}$ and default $\beta$ values ($\beta_1 = 0.9$, $\beta_2 = 0.999$). Training is conducted for 1{,}000 steps with a batch size of 5. 
Instruct-Imagen is used solely in inference mode with the provided reference images and does not require fine-tuning. Backbone models (e.g., Imagen, Imagen-3, and Flux) are used as released, without additional training. All images are generated using evaluation text prompts, with no overlap in prompts or images between the test and support sets. Evaluation with GPT-4o-mini is performed via API calls.
Experiments are conducted on a cluster of eight NVIDIA A100 GPUs (each with 40GB of memory). Fine-tuning takes approximately 20 minutes per entity, and inference requires roughly 5 seconds per image. To support full reproducibility, we will release our benchmark data, evaluation code, and generation scripts.

\section{Human Annotation Instructions}
\label{app:instruction}
We provide the complete instructions given to human raters for evaluating the generated images in the \ourdata benchmark.

\section*{Rater Instructions}

In this task, you will be provided with a Prompt, Reference Images, and a Generated Image. Your job is to assess the factual accuracy of the generated image with respect to the prompt and the reference images. The goal is to ensure that the entity described in the prompt is factually correct and accurately represented. While the reference images offer a visual starting point, you may conduct your own research (e.g., Google Search) to clarify the appearance of the entity and ensure its accurate depiction in the generated image.

\textbf{Part 1: Reference Alignment}

\textbf{Faithfulness to Prompt Entity (Factuality)}

Your first task is to evaluate how faithfully the generated image represents the reference entity. Consider whether the reference entity's key features and overall appearance are accurately depicted.

\textit{Question:} How faithfully does the generated image represent the entity mentioned in the prompt?

\textit{Candidate Answers:}\\
1 (Not faithful at all): The generated image does not represent the reference entity at all. There are no discernible visual similarities to the reference entity.\\
2 (Barely faithful): The generated image faintly represents the reference entity, with significant effort needed to see any resemblance. Minor visual elements may be present, but crucial features or characteristics are missing or significantly misrepresented.\\
3 (Somewhat faithful): The generated image somewhat represents the reference entity, but it’s not prominent. There is a clear visual connection in terms of composition, style, or some key elements, but there are noticeable differences, omissions, or misinterpretations.\\
4 (Mostly faithful): The generated image mostly represents the reference entity and presents it. The generated image draws strong visual inspiration with a strong connection in terms of overall composition, style, key elements, and/or subject matter, despite some variations in details.\\
5 (Completely faithful): The generated image fully represents the reference entity accurately. It captures all key elements, composition, and style in a way that is almost identical to the reference entity.

\textbf{Open Questions for Reference Alignment}

\textit{Visual Similarities:} Describe any visual similarities between the generated image and the reference images, focusing on elements that enhance the recognizability of the entity. Be specific about shape, color, texture, composition, objects, or overall style.\\
\textit{Visual Differences:} Describe any differences in the generated image that negatively impact its faithfulness to the entity in the prompt and are not specified by the prompt. Focus on aspects that affect recognizability, and avoid mentioning changes that do not impact identification (e.g., angle or color for cars or aircraft).

\textbf{Part 2: Text-Image Adherence}

\textbf{Adherence to Prompt Beyond References}

Next, evaluate whether the generated image accurately and comprehensively depicts all aspects of the scene or entity described in the prompt that are not already reflected in the reference images. This involves checking for details in the prompt that go beyond what is shown in the reference images.

\textit{Question:} Does the generated image accurately and comprehensively depict any aspects of the scene or entity described in the prompt that are not already reflected in the reference images?

\textit{Candidate Answers:}

Yes: The generated image accurately and comprehensively depicts aspects of the scene or entity described in the prompt that are not already reflected in the reference images.\\
No: The generated image fails to accurately and comprehensively depict aspects of the scene or entity described in the prompt that are not already reflected in the reference images.

\textit{If the answer is No:}

\textit{Misalignments:} Explain the misalignments between the generated image and the prompt text. Focus on elements or concepts that are not present in the reference images. Be specific about which aspects are missing, inaccurate, or misrepresented.

\textbf{Optional: Open-Ended Feedback}

\textit{Question:} Do you have any other comments or observations about the generated image? (optional)

\section{MLLM Annotation Instructions}
\label{app:mllm}
We provide the complete instructions given to GPT-4o-mini for evaluating the generated images in the \ourdata benchmark.

\textbf{Rater Instructions}

Your job is to assess the factual accuracy of the generated image with respect to the prompt and the reference images. The goal is to ensure that the entity described in the prompt is factually correct and accurately represented.

\textbf{Part 1:}

In this task, you will be provided with a Prompt, Reference Images, and a Generated Image. Evaluate how faithfully the generated image represents the reference entity. Consider whether the key features and overall appearance of the reference entity are accurately depicted.

\textit{Question 1}: How faithfully does the generated image represent the entity mentioned in the prompt?

\textit{Candidate Answers}:\\
1 (Not faithful at all): The generated image does not represent the reference entity at all. There are no discernible visual similarities to the reference entity.
\\
2 (Barely faithful): The generated image faintly represents the reference entity, with significant effort needed to see any resemblance. Minor visual elements may be present, but crucial features or characteristics are missing or significantly misrepresented.
\\
3 (Somewhat faithful): The generated image somewhat represents the reference entity, but it’s not prominent. There is a clear visual connection in terms of composition, style, or some key elements, but there are noticeable differences, omissions, or misinterpretations.
\\
4 (Mostly faithful): The generated image mostly represents the reference entity and clearly presents it. The generated image draws strong visual inspiration with a strong connection in terms of overall composition, style, key elements, and/or subject matter, despite some variations in details.
\\
5 (Completely faithful): The generated image fully represents the reference entity accurately. It captures all key elements, composition, and style in a way that is almost identical to the reference entity.
\\

\textit{Answer in the exact format below}:\\
Question 1:\\
Answer: [1–5]\\
Reason: [Provide a clear explanation for your answer]\\

\textbf{Part 2:}

In this task, you will be provided with a Prompt and a Generated Image. Evaluate how well the generated image captures all aspects described in the prompt. Focus on background elements, contextual details, materials, styles, and other visual features.

\textit{Question 2}: How well does the generated image depict the details described in the prompt?

\textit{Candidate Answers}:\\
1 (Not at all): None of the described elements are present in the image. \\
2 (Slightly): A few minor elements are present, but most are missing or inaccurate.\\
3 (Moderately): Some elements are present and somewhat accurate, but others are missing or misrepresented.\\
4 (Mostly): Most of the described elements are clearly and accurately depicted.\\
5 (Completely): All relevant aspects of the prompt are thoroughly and accurately represented.\\

\textit{Answer in the exact format below}:\\
Question 2:\\
Answer: [1–5]\\
Reason: [Provide a clear explanation for your answer]\\

\section{Broader Impacts}
\ourdata, a benchmark for evaluating text-to-image models' ability to generate accurate depictions of real-world entities such as landmarks and animals, offers insights into the current strengths and limitations of generative models. Our findings suggest that, while these models have made notable progress, they continue to struggle with fine-grained visual accuracy. Moreover, retrieval-augmented approaches may over-rely on reference images, limiting their flexibility in handling creative or compositional prompts. 
These observations highlight opportunities to develop more robust AI systems capable of supporting cultural representation, educational applications, and inclusive visual generation. At the same time, they raise important concerns about the potential for misinformation, representational bias, and the evolving role of AI in creative workflows. By encouraging more rigorous evaluation and grounded generation, \ourdata~aims to promote more trustworthy and responsible use of generative models across domains such as education, communication, and media.

\newpage
{\small
\bibliographystyle{plain}
\bibliography{main}

\begin{thebibliography}{10}

\bibitem{bakr2023hrs}
Eslam~Mohamed Bakr, Pengzhan Sun, Xiaogian Shen, Faizan~Farooq Khan, Li~Erran Li, and Mohamed Elhoseiny.
\newblock {HRS-Bench}: Holistic, reliable and scalable benchmark for text-to-image models.
\newblock In {\em ICCV}, 2023.

\bibitem{blackforest2024flux}
{Black Forest Labs}.
\newblock Flux, 2024.

\bibitem{chen2022reimagen}
Wenhu Chen, Hexiang Hu, Chitwan Saharia, and William~W Cohen.
\newblock Re-imagen: Retrieval-augmented text-to-image generator.
\newblock {\em arXiv preprint arXiv:2209.14491}, 2022.

\bibitem{cho2023davidsonian}
Jaemin Cho, Yushi Hu, Roopal Garg, Peter Anderson, Ranjay Krishna, Jason Baldridge, Mohit Bansal, Jordi Pont-Tuset, and Su~Wang.
\newblock Davidsonian scene graph: Improving reliability in fine-grained evaluation for text-image generation.
\newblock {\em arXiv preprint arXiv:2310.18235}, 2023.

\bibitem{feng2023factkb}
Shangbin Feng, Vidhisha Balachandran, Yuyang Bai, and Yulia Tsvetkov.
\newblock Factkb: Generalizable factuality evaluation using language models enhanced with factual knowledge.
\newblock {\em arXiv preprint arXiv:2305.08281}, 2023.

\bibitem{gokhale2022spatial}
Tejas Gokhale, Hamid Palangi, Besmira Nushi, Vibhav Vineet, Eric Horvitz, Ece Kamar, Chitta Baral, and Yezhou Yang.
\newblock Benchmarking spatial relationships in text-to-image generation.
\newblock {\em arXiv preprint arXiv:2212.10015}, 2022.

\bibitem{gordon2023mismatch}
Brian Gordon, Yonatan Bitton, Yonatan Shafir, Roopal Garg, Xi~Chen, Dani Lischinski, Daniel Cohen-Or, and Idan Szpektor.
\newblock Mismatch quest: Visual and textual feedback for image-text misalignment.
\newblock {\em arXiv preprint arXiv:2312.03766}, 2023.

\bibitem{hessel2021clipscore}
Jack Hessel, Ari Holtzman, Maxwell Forbes, Ronan~Le Bras, and Yejin Choi.
\newblock {CLIPScore}: A reference-free evaluation metric for image captioning.
\newblock {\em arXiv preprint arXiv:2104.08718}, 2021.

\bibitem{heusel2017gans}
Martin Heusel, Hubert Ramsauer, Thomas Unterthiner, Bernhard Nessler, and Sepp Hochreiter.
\newblock {GANs} trained by a two time-scale update rule converge to a local nash equilibrium.
\newblock In {\em NeurIPS}, 2017.

\bibitem{hu2024instructimagen}
Hexiang Hu, Kelvin C.~K. Chan, Yu-Chuan Su, Wenhu Chen, Yandong Li, Kihyuk Sohn, Yang Zhao, Xue Ben, Boqing Gong, William Cohen, Ming-Wei Chang, and Xuhui Jia.
\newblock Instruct-imagen: Image generation with multi-modal instruction.
\newblock In {\em CVPR}, 2024.

\bibitem{Hu_2023_ICCV}
Hexiang Hu, Yi~Luan, Yang Chen, Urvashi Khandelwal, Mandar Joshi, Kenton Lee, Kristina Toutanova, and Ming-Wei Chang.
\newblock Open-domain visual entity recognition: Towards recognizing millions of wikipedia entities.
\newblock In {\em ICCV}, 2023.

\bibitem{hu2023tifa}
Yushi Hu, Benlin Liu, Jungo Kasai, Yizhong Wang, Mari Ostendorf, Ranjay Krishna, and Noah~A Smith.
\newblock {TIFA}: Accurate and interpretable text-to-image faithfulness evaluation with question answering.
\newblock In {\em ICCV}, 2023.

\bibitem{huang2023t2i}
Kaiyi Huang, Kaiyue Sun, Enze Xie, Zhenguo Li, and Xihui Liu.
\newblock {T2I-CompBench}: A comprehensive benchmark for open-world compositional text-to-image generation.
\newblock In {\em NeurIPS}, 2023.

\bibitem{baldridge2024imagen}
{Imagen 3 Team}.
\newblock Imagen 3.
\newblock {\em arXiv preprint arXiv:2408.07009}, 2024.

\bibitem{ku2024imagenhub}
Max Ku, Tianle Li, Kai Zhang, Yujie Lu, Xingyu Fu, Wenwen Zhuang, and Wenhu Chen.
\newblock Imagenhub: Standardizing the evaluation of conditional image generation models.
\newblock In {\em ICLR}, 2024.

\bibitem{kumari2022customdiffusion}
Nupur Kumari, Bingliang Zhang, Richard Zhang, Eli Shechtman, and Jun-Yan Zhu.
\newblock Multi-concept customization of text-to-image diffusion.
\newblock In {\em CVPR}, 2023.

\bibitem{lee2024holistic}
Tony Lee, Michihiro Yasunaga, Chenlin Meng, Yifan Mai, Joon~Sung Park, Agrim Gupta, Yunzhi Zhang, Deepak Narayanan, Hannah Teufel, Marco Bellagente, et~al.
\newblock Holistic evaluation of text-to-image models.
\newblock In {\em NeurIPS}, 2024.

\bibitem{li2024genaibench}
Baiqi Li, Zhiqiu Lin, Deepak Pathak, Jiayao Li, Yixin Fei, Kewen Wu, Tiffany Ling, Xide Xia, Pengchuan Zhang, Graham Neubig, and Deva Ramanan.
\newblock Genai-bench: Evaluating and improving compositional text-to-visual generation.
\newblock {\em arXiv preprint arXiv:2406.13743}, 2024.

\bibitem{lim2024addressing}
Youngsun Lim and Hyunjung Shim.
\newblock Addressing image hallucination in text-to-image generation through factual image retrieval.
\newblock In {\em IJCAI Workshop}, 2024.

\bibitem{lin2015microsoftcococommonobjects}
Tsung-Yi Lin, Michael Maire, Serge Belongie, Lubomir Bourdev, Ross Girshick, James Hays, Pietro Perona, Deva Ramanan, C.~Lawrence Zitnick, and Piotr Dollár.
\newblock Microsoft coco: Common objects in context.
\newblock In {\em ECCV}, 2015.

\bibitem{materzynska2023customizing}
Joanna Materzy\'nska, Josef Sivic, Eli Shechtman, Antonio Torralba, Richard Zhang, and Bryan Russell.
\newblock Customizing motion in text-to-video diffusion models.
\newblock {\em arXiv preprint arXiv:2312.04966}, 2023.

\bibitem{muhlgay2023generating}
Dor Muhlgay, Ori Ram, Inbal Magar, Yoav Levine, Nir Ratner, Yonatan Belinkov, Omri Abend, Kevin Leyton-Brown, Amnon Shashua, and Yoav Shoham.
\newblock Generating benchmarks for factuality evaluation of language models.
\newblock {\em arXiv preprint arXiv:2307.06908}, 2023.

\bibitem{gpt}
OpenAI et~al.
\newblock Gpt-4 technical report, 2023.

\bibitem{dinov2}
Maxime Oquab, Timothée Darcet, Théo Moutakanni, Huy Vo, Marc Szafraniec, Vasil Khalidov, Pierre Fernandez, Daniel Haziza, Francisco Massa, Alaaeldin El-Nouby, Mahmoud Assran, Nicolas Ballas, Wojciech Galuba, Russell Howes, Po-Yao Huang, Shang-Wen Li, Ishan Misra, Michael Rabbat, Vasu Sharma, Gabriel Synnaeve, Hu~Xu, Hervé Jegou, Julien Mairal, Patrick Labatut, Armand Joulin, and Piotr Bojanowski.
\newblock Dinov2: Learning robust visual features without supervision.
\newblock In {\em TMLR}, 2024.

\bibitem{radford2021clip}
Alec Radford, Jong~Wook Kim, Chris Hallacy, Aditya Ramesh, Gabriel Goh, Sandhini Agarwal, Girish Sastry, Amanda Askell, Pamela Mishkin, Jack Clark, Gretchen Krueger, and Ilya Sutskever.
\newblock Learning transferable visual models from natural language supervision.
\newblock In {\em ICML}, 2021.

\bibitem{ramesh2022dalle}
Aditya Ramesh, Prafulla Dhariwal, Alex Nichol, Casey Chu, and Mark Chen.
\newblock Hierarchical text-conditional image generation with clip latents.
\newblock {\em arXiv preprint arXiv:2204.06125}, 2022.

\bibitem{ramesh2021dalle}
Aditya Ramesh, Mikhail Pavlov, Gabriel Goh, Scott Gray, Chelsea Voss, Alec Radford, Mark Chen, and Ilya Sutskever.
\newblock Zero-shot text-to-image generation.
\newblock In {\em ICML}, 2021.

\bibitem{rombach2022stablediffusion}
Robin Rombach, Andreas Blattmann, Dominik Lorenz, Patrick Esser, and Björn Ommer.
\newblock High-resolution image synthesis with latent diffusion models.
\newblock In {\em CVPR}, 2022.

\bibitem{ruiz2023dreamboothfinetuningtexttoimage}
Nataniel Ruiz, Yuanzhen Li, Varun Jampani, Yael Pritch, Michael Rubinstein, and Kfir Aberman.
\newblock Dreambooth: Fine tuning text-to-image diffusion models for subject-driven generation.
\newblock In {\em CVPR}, 2023.

\bibitem{saharia2022photorealistic}
Chitwan Saharia, William Chan, Saurabh Saxena, Lala Li, Jay Whang, Emily~L Denton, Kamyar Ghasemipour, Raphael Gontijo~Lopes, Burcu Karagol~Ayan, Tim Salimans, et~al.
\newblock Photorealistic text-to-image diffusion models with deep language understanding.
\newblock In {\em NeurIPS}, 2022.

\bibitem{wiles2024revisiting}
Olivia Wiles, Chuhan Zhang, Isabela Albuquerque, Ivana Kaji{\'c}, Su~Wang, Emanuele Bugliarello, Yasumasa Onoe, Chris Knutsen, Cyrus Rashtchian, Jordi Pont-Tuset, et~al.
\newblock Revisiting text-to-image evaluation with gecko: On metrics, prompts, and human ratings.
\newblock {\em arXiv preprint arXiv:2404.16820}, 2024.

\bibitem{wu2024conceptmix}
Xindi Wu, Dingli Yu, Yangsibo Huang, Olga Russakovsky, and Sanjeev Arora.
\newblock Conceptmix: A compositional image generation benchmark with controllable difficulty.
\newblock {\em arXiv preprint arXiv:2408.14339}, 2024.

\bibitem{yarom2024you}
Michal Yarom, Yonatan Bitton, Soravit Changpinyo, Roee Aharoni, Jonathan Herzig, Oran Lang, Eran Ofek, and Idan Szpektor.
\newblock What you see is what you read? improving text-image alignment evaluation.
\newblock In {\em NeurIPS}, 2024.

\end{thebibliography}
}

\end{document}